%% file: main.tex
\pdfoutput=1
\documentclass{article}
\PassOptionsToPackage{dvipsnames}{xcolor}
\usepackage{iclr2026_conference,times}
\input{math_commands.tex}

\usepackage{hyperref}
\hypersetup{pdftitle={Verifiable, Articulable, and Tacit Components of Preference},
            pdfauthor={Alexander Spangher, Sheldon Huang, Andreas Haupt, Noah D. Goodman, Diyi Yang, Daniel E. Ho, Sanmi Koyejo}}
\usepackage{url}
\usepackage{amsmath}
\AtBeginDocument{\let\argmax\relax\DeclareMathOperator*{\argmax}{argmax}}   %

\usepackage{graphicx}
\usepackage{booktabs}
\usepackage[inline]{enumitem} 
\usepackage{colortbl}
\usepackage{array}
\usepackage{tikz}
\usepackage{wrapfig}
\usepackage{xcolor}   %
\usepackage{mathtools}
\usepackage{subcaption}
\usepackage{titletoc}

\usepackage{float}
\usepackage{pdflscape}
\usepackage{longtable}
\usepackage{caption} %
\usepackage{placeins} %
\usepackage[normalem]{ulem} %
\NewDocumentCommand{\ubrace}{O{black} m m}{{\color{#1}\underbrace{\color{black}#2}_{\color{#1}#3}}}
\NewDocumentCommand{\ubrack}{O{black} O{.3pt} O{.6ex} m m}{{\color{#1}\underbracket[#2][#3]{\color{black}#4}_{\color{#1}#5}}}
\newsavebox{\ubcb}\newsavebox{\ublb}\newlength{\ubcw}\newlength{\ublw}\newlength{\ubw}
\newcommand{\ubpass}[1]{#1}
\newcommand{\ubmeasure}[3]{\sbox{\ubcb}{$#2$}%
  \sbox{\ublb}{\begingroup\let\substack\ubpass\scriptsize\strut#3\endgroup}%
  \settowidth{\ubcw}{\usebox{\ubcb}}\settowidth{\ublw}{\usebox{\ublb}}%
  \setlength{\ubw}{\ubcw}\ifdim\ublw>\ubw\setlength{\ubw}{\ublw}\fi}
\newlength{\ubslot}
\NewDocumentCommand{\tbrace}{O{black} O{.4pt} O{.9mm} m m}{\ubmeasure{#1}{#4}{#5}%
  \mathop{\vtop{\offinterlineskip\hbox to\ubw{\hss\usebox{\ubcb}\hss}\kern1.2pt
  \hbox to\ubw{\begin{tikzpicture}[baseline=0pt]
    \useasboundingbox (0,-\ubslot) rectangle (\the\ubw,0);
    \draw[#1,line width=#2,decorate,
    decoration={brace,mirror,amplitude=#3,raise=0pt}](0,0)--(\the\ubw,0);\end{tikzpicture}}\kern2.4pt
  \hbox to\ubw{\hss\usebox{\ublb}\hss}}}}
\NewDocumentCommand{\ubline}{O{black} O{.3pt} m m}{\ubmeasure{#1}{#3}{#4}%
  \mathop{\vtop{\offinterlineskip\hbox to\ubw{\hss\usebox{\ubcb}\hss}\kern2.6pt
  \hbox to\ubw{\textcolor{#1}{\rule{\ubw}{#2}}}\kern2.6pt\hbox to\ubw{\hss\usebox{\ublb}\hss}}}}
\usepackage{epigraph}  %
\usetikzlibrary{arrows.meta,positioning,decorations.pathreplacing,decorations.pathmorphing,patterns,shapes.symbols,shapes.geometric,calc}

\definecolor{exmetricbg}{RGB}{238,248,224}
\input{exmetric_breakable.tex}

\definecolor{vgreen}{HTML}{369661}
\definecolor{aamber}{HTML}{417497}
\definecolor{torange}{HTML}{46237A}
\definecolor{tpink}{HTML}{D6C9E8}

\title{Verifiable, Articulable, and Tacit\\ Components of Preference}
\author{Alexander Spangher$^{1}$\thanks{\footnotesize Corresponding author: \texttt{alexspan@stanford.edu}.}, \ Sheldon S. Huang$^{1,2}$, \ Andreas Haupt$^{1}$, \ Noah D.~Goodman$^{1}$, \\
\textbf{Diyi Yang$^{1}$, \ Daniel E.~Ho$^{1}$ \ \& \ Sanmi Koyejo$^{1}$} \\[4pt]
$^{1}$Stanford University \qquad $^{2}$University of Toronto
}

\iclrfinalcopy %

\begin{document}
\maketitle
\lhead{Preprint.} %
\vspace{-3mm}
\begin{abstract}
\vspace{-2mm}
What makes a short story gripping; a news article newsworthy; or a math proof elegant? These constructs resist articulation or verification; their meaning is at least partially \textit{tacit}. %
However, modern AI models are improved primarily via  \textit{articulated} constitutions, rubrics and verifiers (i.e. in RLAIF and RLVR); \textit{tacit} components of preferences are typically understudied.
We introduce a large, labeled preference dataset CreativePreferences, containing 2.8M texts labeled by 317M human preference judgments across 7 creative domains, with 42 benchmark tasks for preference prediction. %
We model these labels with %
executable programs, rubric banks and densely trained models (\textcolor{ForestGreen}{\textit{$\mathcal{V}$: verifiability}}, \textcolor{blue}{$\mathcal{A}$: \textit{articulability}} and \textcolor{Fuchsia}{\textit{$\mathcal{VAT}$: full taste models}}, respectively). 
We observe robust \textit{articulability gaps}, $\mathcal{VAT}-\mathcal{VA}$; and \textit{verifiability gaps}, $\mathcal{VA}-\mathcal{V}$; we estimate upper and lower bounds for each gap with a novel measurement approach that discovers articulable and verifiable metrics, identifies spurious variables and estimates the value of undiscovered metrics using \textit{capture-recapture}. %
These gaps occur across \textit{all} domains, even in domains traditionally treated as fully verifiable: \textit{correctness-centered domains} (i.e. mathematics and software engineering) and \textit{claim- and novelty-centric} domains (i.e. news, patents, peer review). The \textit{size} of the gap varies based on domain (e.g. peer review and creative writing have the largest articulability gaps) and %
widens as more people take part in the judgment, consistent with \citet{collins2010tacit}'s \textit{collective tacit knowledge}. We show two consequences: (1) on human generations, the full model more closely matches human preferences, often in disagreement with articulated criteria, and (2) in an analogy to Goodhart's law, articulating preference shifts it \textit{away} from the tacit dimension.
Articulability and verifiability gaps are consequential; we give recommendations on when tasks can be prompted; how learning mechanisms might improve; and when to leave judgments with humans.%
\vspace{1.5mm}
\end{abstract}
\vspace{-4pt}

\begin{figure}[H]   %
\vspace{-14pt}\setlength{\abovecaptionskip}{5pt}   %
\centering
\includegraphics[width=1.0\textwidth,trim={10pt 42pt 90pt 29pt},clip]{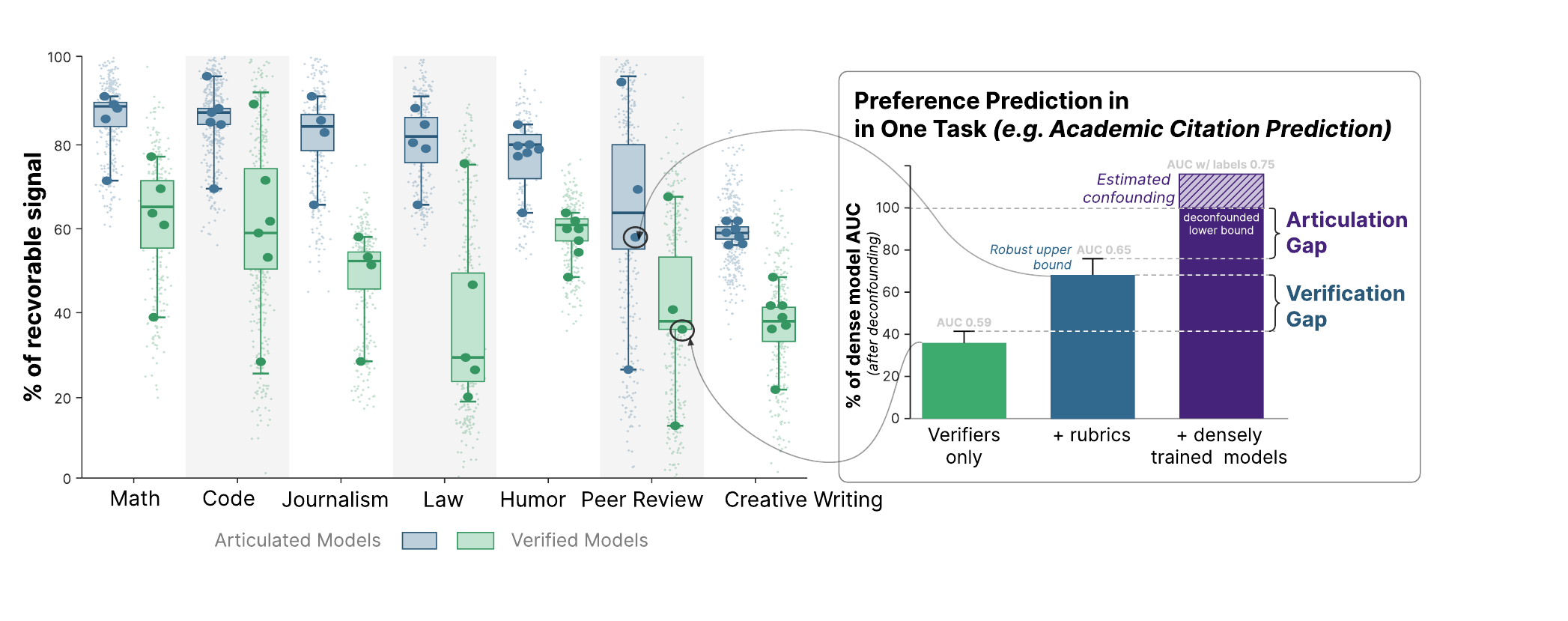}
\caption{\textbf{Verification and articulation gaps in preference prediction.} Each blue and green point in LHS boxplot is a single task of 42, example: \textit{Predict whether academic paper X receives a prize}). RHS barplot shows how we calculate: first, we measure an upper bound of task performance with \textcolor{Fuchsia}{dense full models}, estimated-optimal \textcolor{ForestGreen}{executable programs} and \textcolor{blue}{articulated rubrics}. 
Gaps are robustly estimated and comparable across tasks: we deconfound datasets, match preference labels and robustly estimate missing metrics. Light dots are subsample of $b{=}3{,}000$ bootstrap draws / task. %
}
\label{fig:concept-row}
\label{fig:explainer}
\label{fig:community-articulability}
\label{fig:domain-shares}
\label{fig:cover-photo}
\end{figure}

\section{Introduction}
\label{sec:intro}
\vspace{-1mm}
\epigraph{``[Mathematical beauty] is almost \textbf{impossible to state}; it is \textbf{felt} rather than formulated.'' {\tiny ---\,\citet{poincare1910mathematical}}}%
{}

\vspace{-8mm}
Tacit knowledge, where humans collectively agree on a construct but struggle to articulate it, drives the development and evaluation of high-quality creative work \citep{csikszentmihalyi1996creativity, collins2010tacit}. Concretely, this poses a challenge to AI models: %
agents learn via reward models, which are typically given by \textit{verifiers}, \textit{constitutions} and \textit{rubrics}. %
What impact does \textit{tacit knowledge} in these domains have? %
We find surprisingly little work evaluating this; despite reward modeling being well-studied \citep{lambert2024rewardbench}, prior work has focused on a limited set of domains \citep{zhou2024rmb} and has not focused on measuring tacitness \citep{wang2024helpsteer2,cui2024ultrafeedback} (\S\ref{sec:methods}). 

In this work, we study 7 creative domains\footnote{Defined in Section \ref{sec:dataset}, a creative domain is a domain where the merits of a work is judged and selected by a \textit{field}, or groups of humans rather than fixed, outside standards \citep{csikszentmihalyi1996creativity}. %
}: mathematics, coding, law, academia, journalism, creative writing, humor. We introduce a large, novel dataset, \textbf{CreativePreferences} to study tacit preferences, which contains 317 million preference judgments on 2.8 million texts across 42 benchmark prediction tasks; these tasks are split into: predict \textit{community voting preferences}, \textit{gatekeeper verdicts} and \textit{expert curations} (Section \ref{sec:dataset}). 
We model each preference judgment by:

\vspace{-4mm}
\begin{equation}
\tbrace[torange][.25pt][.6mm]{P(X)\vphantom{\gamma}}{\text{\strut preference}}\;=\;
\tbrace[ForestGreen][.25pt][1mm]{\textcolor{ForestGreen}{\mathcal{V}}(X)\vphantom{\gamma}}{\substack{\text{\strut\emph{\textcolor{ForestGreen}{verifiable metrics}}}}}\;+\;
\tbrace[blue][.25pt][.6mm]{\textcolor{blue}{\mathcal{A}}(X)\vphantom{\gamma}}{\substack{\text{\strut\emph{\textcolor{blue}{articulable rubrics}}}}}\;+\;
\tbrace[Fuchsia][.25pt][.6mm]{\textcolor{Fuchsia}{\mathcal{T}(X)}\vphantom{\gamma}}{\substack{\text{\strut\emph{\textcolor{Fuchsia}{articulability gap}}}}} \;+\;
\tbrace[torange][.25pt][.6mm]{\epsilon(X)\vphantom{\gamma}}{\substack{\text{\strut noise \& personal preference}}}
\label{eq:intro-pref-model}
\end{equation}
\vspace{-4mm}

For example, to model the preference signal \textit{academic paper X is well-cited}, we might consider metrics: \exmetric{\textit{experiments are reproducible}}, which we can check deterministically (\textcolor{ForestGreen}{$\mathcal{V}$}); \exmetric{\textit{compares to appropriate baselines}}, a construct we can describe in words (\textcolor{blue}{$\mathcal{A}$}); \exmetric{\textit{elegant presentation}} (\textcolor{Fuchsia}{$\mathcal{T}$}). The first two terms we measure with \textit{executable programs} and \textit{LLM-judged rubrics} via a novel metric discovery algorithm we introduce; the third we measure by training a dense model $\mathcal{VAT}$ on text and preferences. %
We highlight \emph{articulability gaps} ($\mathcal{VAT} - \mathcal{V}\mathcal{A}$): what we can predict beyond what we can articulate; and \emph{verifiability gaps} ($\mathcal{VA}-\mathcal{V}$): what we can articulate beyond what we can check
\footnote{\textit{Note:} we do not positively establish our measured quantities measure \textit{tacit knowledge} described by \citet{polanyi1966tacit, collins2010tacit} only negatively rule out competing explanations; to not overclaim, we refer to them as \textit{gaps}. Additionally, what lies above $T$ is not addressed: it contains idiosyncratic preference and noise.}.
We show these gaps are robust by showing our verifiers and rubrics are \textit{estimated-optimal} and that the datasets are cleaned and deconfounded; we describe our procedure in Section \ref{sec:res-frontier}.

Articulability and verifiability gaps exist in every field that we study. 
While ${\ge}85\%$ of preference in code, math, journalism and law can be articulated; only $61$--$85\%$ can in humor, creative writing and peer review (Figure \ref{fig:cover-photo}). Tasks with less articulability are also less verifiable ($\rho =.63$), except for humor: humor is highly articulable, less verifiable\footnote{It seems jokes truly \textit{can} be explained in words.}.
Even domains that are traditionally treated as verifiable have gaps: math preference has 8--26\% articulability gaps across tasks; software engineering preference has $8\%$-$28\%$ \citep{guo2025deepseekr1, lambert2024tulu3}.

Why do articulability and verifiability gaps arise across fields? Using human annotators, we judge that gaps exist when work is novel, clearly written and surprising; these are difficult constructs to specify. %
Gaps emerge, too, based on the social construction of preference: community voting yields the least articulable shared preference, while curated preferences yields the most and in fields with more judges, gaps grow larger ($\rho=.46$); this accords with \textit{collective tacit knowledge}, or knowledge that emerges when social groups form cultures at a scale that resists articulation \citet{collins2010tacit}. %

Finally, we show consequences for AI model development. Across tasks: (1) $\mathcal{VAT}$ models agree more with real human judgements; (2) in an analogue to \citet{goodhart1984problems}, the more we emphasize articulated metrics in preference models, the more inarticulated preference we \textit{lose}; the reverse is not true (Section~\ref{sec:res-optim}). As AI models are increasingly making consequential decisions in tasks which require tacit knowledge (e.g. preference judgement and reward modeling, rule-following and safety \citep{kripke1982wittgenstein, wittgenstein1953pi}), 
we argue for more robust research into where and why tacit knowledge emerges and the consequences if our models fail to acquire it. We close with recommendations: measurement focused on tacit knowledge can help identify more articulable tasks; these are supported by current learning mechanisms. Human-in-the-loop approaches should be considered for those tasks that are less. In tasks requiring even more tacit knowledge and preference is more intersubjective (e.g. creative writing), we should bound that AI may never surpass humans.

\begin{table}[!t]
\centering
\resizebox{\textwidth}{!}{\input{figures/tab_datagrid.tex}}
\caption{\textbf{CreativePreferences}: seven domains (columns) with three preference variable types (rows): \textit{gatekeeper verdicts} are when a text is evaluated by a decision-maker who accepts or rejects it. \textit{Expert curation} are preference signals transmitted by selecting top items from a pool. \textit{Community votes} preferences are aggregations of individual votes on items. Labels that list several corpora contain more than one task; we collect 42 tasks across 37 label-types. %
For details, see App.~\ref{app:inventory}.%
}
\label{tab:datasets}
\end{table}

\section{The CreativePreferences Dataset}
\label{sec:methods}

\label{sec:dataset}

To facilitate the study of tacit preferences, we introduce \textbf{CreativePreferences} (Table \ref{tab:datasets}), to our knowledge the largest, broadest dataset of preferences across creative domains. We base our definition of \textit{creative task} from Csikszentmihalyi's Systems Model of Creativity \citep{csikszentmihalyi1996creativity}: a domain is creative when its products' value is decided by the \emph{field}, or experts, rather than predetermined criteria (e.g. editors, juries and voters; %
instead of compilers and proof checkers). Math and coding, for instance, are creative tasks because \textit{humans} ultimately decide whether to merge code and give prizes. 
We choose 7 domains that are traditionally studied in reward-modeling (e.g. math and coding) and tasks colloquially considered creative (e.g. creative writing and humor).

CreativePreferences captures 317 million preference judgments across 2.8 million texts, 42 tasks and 7 creative domains  (Appendix~\ref{app:inventory}), allowing for large, cross-field comparative studies of preference. Existing datasets fall short: creative-domain benchmarks are typically within a single field (e.g. creative writing) \citep{litbench2025}; focus on correctness-based tasks, chat or safety \citep{patwardhan2025gdpval, lambert2024rewardbench}; or elicit labels \citep{wang2024helpsteer2, cui2024ultrafeedback}. 

We use \textit{found} rather than \textit{elicited} judgments: we do not ask raters to compare items, we observe which texts are rewarded in the real world. In doing so, we mimic how \textit{humans} learn their field's norms: e.g. an academic learns what their field is prioritizing not just by reading criteria or querying their weary advisor, but by observing and generalizing from paper acceptances and prizes \citep{lamont2009how, english2005economy}. We consider three ways these rewards are transmitted: \emph{expert curation} deliberately selects a few winners from a large pool (e.g. a paper prize) \citep{lamont2009how, english2005economy}; \emph{gatekeeper verdicts} are accept/reject decision on every submitted item (e.g. a paper acceptance); \emph{community votes} are votes cast by members (e.g. paper citations) \citep{surowiecki2004wisdom}.\footnote{CreativePreferences has additional criteria:
\begin{enumerate*}
    \item Datasets had to be textual; we did not consider multimodal datasets (e.g. \citealp{hessel2023androids}).
    \item Where possible, text corpora were labeled with multiple preferences: e.g. ICLR academic articles transmit \textit{verdicts}, %
     \textit{curation}, %
     and \textit{community} variables. %
\end{enumerate*}}.

\section{Methods}

We measure gaps in average reward between three sets of learners, each prompted with a different representation of the same training set: one is given the training data itself; one is given a compression of the data to text; one is given a compression of the data to a formula. We expect the three to be nested: text cannot carry more signal than the data, formulas cannot carry more than text, by the data processing inequality \citep{cover2006elements}. A formula is a finite string in a formal language and is a special case of text; however, not everything that can be said can be written as a formula \citep{searle1969speech}. We make no positive claim as to whether these gaps are caused by tacit knowledge, as defined by \citet{collins2010tacit}, although we test many competing explanations for what these gaps are (Section \ref{sec:res-frontier}). Nor, if dense models \textit{do} capture tacit knowledge, do we make claims as to whether they capture the \textit{full scope} of human tacit knowledge in these tasks.

\subsection{Measuring Articulability and Verifiability Gaps} \label{sec:measuring}

Eq. \ref{eq:intro-pref-model}, defines the verifiability gap as articulable preference that is not verifiable, $\mathcal{VA}-\mathcal{V}$, and the articulability gap as shared preference that is not articulable, $\mathcal{VAT}-\mathcal{VA}$. Now, we describe how to measure this. Both $\mathcal{A}$ and $\mathcal{V}$ models are implemented as a set of metrics $M(x) = m_1(x), m_2(x),...$\footnote{E.g. for creative writing, $m^i=$\exmetric{\textit{Uses verbal economy}})}
, aggregated by a dense model, $f$. Each $m^i$ is implemented as a \textit{program} executed by a compiler; or a \textit{prompt} executed by an LLM (described shortly). If $M$ consists only of programs\footnote{A program is a deterministic procedure run with the text as input; programs include executable checks (e.g. \exmetric{\textit{can the code compile}}) and surface checks (e.g. \exmetric{\textit{how long is the text}}), see Appendix~\ref{app:vexec}.}, $M_v$, then $f \circ M_v$ is the verifiable model $\mathcal{V}$%
; if $M$ is a mixture of programs \textit{and} prompts, $M_{av}$, then $f \circ M_{av}$ is the articulated model, 
$\mathcal{VA}$
. Finally, we implement the \textit{full} taste model, $\mathcal{VAT} = T(X, M(X))$\footnote{The full model is reported as: $\mathcal{VAT}=\max(T, \mathcal{VAT})$.}, a large, transformer model $T$ trained on text, preferences and $M$. %
Practically, this helps us compare different AI post-training regimes: Reinforcement Learning with Verifiable Rewards (RLVR) \citep{lambert2024tulu3}, RL with AI Feedback (RLAIF) \citep{lee2023rlaif} and RL with Human Feedback (RLHF) \citep{ouyang2022training, christiano2017deep}. Theoretically, this distinction tests differences between formally specifiable constructs; constructs that can be articulated to a LLM reader; and constructs that can only be directly learned.
For gaps %
to be meaningful and robust, we estimate the \textit{upper bound} of $\mathcal{V}$ and $\mathcal{VA}$ performance \textit{and} we ensure $\mathcal{VAT}$ is not spuriously high. We discuss this now.%

\begin{figure}[t]
	\centering
	\resizebox{.98\textwidth}{!}{\input{figures/fig_mining_loop.tex}}
	\caption{\textbf{How we mine criteria, discover spurious variables, and estimate effects of undiscovered metrics.} %
    (1) First, we discover metrics using metric discovery algorithms. (2) Each metric is tested for whether it is quality or spurious: we test if each metric (a) captures signal external to text itself (\textit{channel test}, e.g. the ``time of day'' of post); (b) is correlated with other metrics (\textit{path test}, e.g. ``first-person'' language correlates to favored genre); or (c) is qualitatively not valuable (\textit{LLM-audit}). (3) We estimate the upper bound of performance from metrics not yet found yet using \textit{capture-recapture} \citep{good1953population, chao1984nonparametric, efron1976estimating}.
    }
	\label{fig:mining_loop}
\end{figure}

\paragraph{Discovering Articulated and Verifiable Metrics}\label{sct:mining}

First, we discover a set of metrics by running four metric discovery methods: AutoMetrics \citep{ryan2026autometrics}, ExperiGen \citep{sengupta2026experigen}, HypoGeniC \citep{zhou2024hypogenic} and HypotheSAEs \citep{movva2025hypothesaes}. Each algorithm generates metrics to predict outcomes, $P(X)$. For example, AutoMetrics shows an LLM examples of text and labels and asks the LLM to propose metrics that predict the labels; HypotheSAEs trains a sparse autoencoder, keeps latent features whose activations best predict labels, and asks an LLM to identify similarities in texts with those features. We combine metrics from these algorithms to obtain a large pool of metrics\footnote{Most algorithms are implemented as originally described. We adjust AutoMetrics as follows: we iteratively show it misclassified examples, generate metrics, rescore, and generate more metrics (see Appendix \ref{app:metric-discovery}).}. For example, in software engineering, we generate metrics for Github pull requests $m^i_v=$\exmetric{\# unit tests moving fail$\to$pass} and $m^i_v=$\exmetric{\# unit tests moving pass$\to$fail}. To see additional examples of discovered articulated and verifiable metrics, see Table \ref{tab:topfeatures}. %

\paragraph{Implementing Articulated and Verifiable Metrics}
\label{sct:articulated_and_verifiable_metrics}

After our discovery algorithms propose metrics, we implement each metric as \textit{both} a program, $m_v^i$ and a rubric, $m_a^i$. Our approach to coding and articulating both sets of metrics is similar. We use a joint \textit{fidelity} objective
\footnote{\textit{Fidelity} combines: label consistency (i.e. how often do independent runs of an LLM agree), a reconstruction bottleneck (i.e. can labels applied by a prompt carry enough signal that the prompt can be reconstructed) and synthetic measurements. A synthetic measurement is a constructed metric with known, checkable values (e.g. ``how many question marks in the sentence?'') A batch whose synthetic measurements are not correctly detected has its prompts re-scored \citep{companion_prompts}.}
, introduced in \citep{companion_prompts}, which provides an \textit{unsupervised} measure of metric quality. We then apply prompt-optimization algorithms: $\epsilon$-certification for prompted metrics, $m^i_a$ \citep{companion_prompts}; and GEPA for coded metrics, $m^i_v$ \citep{agrawal2025gepa}. In both algorithms, an LLM is asked to critique the prompt or code in order to improve an objective (here, our fidelity objective). The $\epsilon$-estimation algorithm explicitly determines when a prompt is \textit{optimal} for its LLM; in GEPA, we run each until performance plateaus. We test different judge sizes and families to confirm that we are not capacity constrained or limited by model deficiencies (models used at each step listed in Appendix~\ref{app:models}, \ref{app:family-robustness}). Finally, we select the top $k$ features in $M_v$ and $M_{av}$ using sparse regression and train transformer models $f$ to obtain $\mathcal{V}$ and $\mathcal{VA}$.
\footnote{For all learned models, we train Llama-3.1-8B with LoRA ($r{=}16$, $\alpha{=}32$), learning rate $5{\times}10^{-5}$, batch size 16, two epochs, on 80/10/10 train/eval/test splits. Splits are defined by context (prompt, question, etc.). $\mathcal{VAT}$ models are trained on the text followed by a block of criterion scores (i.e. \texttt{text: <text>, VA metrics: <criterion>: <score>}). See Appendix~\ref{app:models}.}. 
For each metric, the judge reads the full text (Appendix~\ref{app:text-budget}).

\paragraph{Estimating $\mathcal{V}$ and $\mathcal{A}$ Upper Bounds} Although our discovered metrics, $M_a$ and $M_v$, are found using several different metric discovery algorithms, \textit{undiscovered} metrics may still exist that impact performance of our models. Thus, we wish to estimate the value of metrics that we \emph{failed to find}. We do this using \textit{capture-recapture} \citep{boehme2018stads, chao1984nonparametric, companion_prompts}:
\label{sct:missing-mass}%
\begin{equation}
\newcommand{\whiskerph}{\vphantom{\textstyle\sum_{m \in U} v_m\, G \cdot \tfrac{\mu}{1-\mu}}}%
\tbrace[torange][.25pt][.6mm]{\textstyle\sum_{m \in U} v_m\whiskerph}{\text{\strut unmined value}}
\;\le\;
\tbrace[torange][.25pt][.6mm]{G \cdot \tfrac{\mu}{1-\mu}\whiskerph}{\text{\strut unmined value (at last round's rate)}}
\;+\;
\tbrace[torange][.25pt][.6mm]{s \cdot v_{\mathrm{cap}}\whiskerph}{\text{\strut slack for rare-events}},
\label{eq:whisker}
\end{equation}
\noindent Here, $U$ is the set of metrics never proposed in any metric discovery round and $v_m$ is the performance that metric $m$ would add to $\mathcal{V}$ or $\mathcal{VA}$; $G = f(M_{n+1}) - f(M_n)$ is the gain of metrics discovered in round $n+1$; $\mu$ is the probability that the next metric has never been seen, or the Good--Turing missing mass \citep{good1953population} 
\footnote{We estimate this with $\hat\mu = f_1/N$, the fraction of the final round's $N$ proposals that were first-time discoveries. Each sampling round conditions on earlier rounds, so $\hat\mu$ is the missing mass of the final round's proposal distribution (per-round series and uncertainty split in Appendix~\ref{app:whisker-backtest})}. We assume, in general, that metrics are discovered \textit{roughly} in order of salience (i.e. metrics mined at round $n$ are worth more than those mined at round $n+1$); to bound this assumption, $s \cdot v_{\mathrm{cap}}$ is a budget for rare, high-value criteria: $s$ is the number of new metrics allowed to break the salience ordering and $v_{\mathrm{cap}}$ the most any one of them would have to be worth to completely \textit{close} the articulability or verifiability gap\footnote{The bound we draw is the first term, $G\hat\mu/(1-\hat\mu)$, computed from proposal counts alone. The second term is a sensitivity budget rather than part of the bound: with $s{=}1$, $v_{\mathrm{cap}} = (\text{gap} - \text{whisker})/s$ is the value a single metric that escaped the salience ordering would have to carry to close the gap (Appendix~\ref{app:missing-mass-derivation}). In the truncation backtest the first term held at all $30$ truncation points of the seven multi-round campaigns (Appendix~\ref{app:whisker-backtest}).}. %
We test whether this leads to instability using a backtest: we truncate each metric discovery runs at every round and check whether the realized gain stayed inside the bound computed at that round. (Appendix~\ref{app:whisker-backtest}). %
See Appendix~\ref{app:missing-mass-derivation} for the full derivation, its extension to multiple decorrelated proposers, and the rare-event analysis.

\paragraph{Identifying and Deconfounding Spurious Variables: Deriving a Lower Bound for $\mathcal{VAT}$}\label{sct:deconfound}

A spurious feature is any feature that predicts labels  without capturing the underlying construct we wish to measure. Identifying spurious variables is crucial; our full model $\mathcal{VAT}$ can fit to these and thus show  large articulability and verifiability gaps while capturing signal that is irrelevant to preferences. %
We develop a set of tests: \emph{channel tests} (i.e. to identify metrics that capture signals external to the text, like time-of-day \citep{poliak2018hypothesis, kaushik2018reading, feng2018pathologies}), \emph{path tests} (i.e. to identify metrics that are \textit{correlated} with genuine or spurious metrics and not the construct \citep{peters2016invariant, veitch2021counterfactual}), and a final LLM adjudication (i.e. to identify metrics that qualitatively do not appear reasonable \citep{messick1995validity}).
\footnote{Examples of channel tests are: scrambling the text (i.e. whether the full model, with the feature, still predicts the outcome); top-n-gram probes (i.e. whether top n-grams are clearly spurious). Examples of path tests: stratifying on \textit{known} spurious variables (i.e. whether the suspected variable still survives; (examples of path test variable might be if horror prompts draw more first-person stories and also more upvotes). Our LLM adjudication asks the LLM: ``is \{feature\} spurious in this task, where spurious means  that the feature measures on the capture, selection, or context of the item rather than the quality of the work? Answer yes/no.'' \citep{poliak2018hypothesis, mccoy2019right}. Example of a planted artifact: a footnote mark appended to 55\% of positive and 15\% of negative jokes; a resample of jokes making era predictive. These are detected at 100\% accuracy.} 
(Table~\ref{tab:audit-bench}, see Appendix \ref{app:spurious-audit} for the full audit). 
For metrics we discover are spurious, we deconfound in several ways: if it is present in $>60\%$ of the data, we stratify the dataset by the spurious feature; if it is a surface feature that is easily removable, we strip it from the text using regular expressions; otherwise, we residualize the model's score on the remaining spurious variables
\footnote{\textit{Stratifying} fits models \textit{within} levels of the spurious feature (e.g.\ within a length decile). Stripping deletes the spans using regexes (bot notices, signatures, URLs) before training. \textit{Residualizing} regresses each model's score on the spurious variables to correct the prediction (i.e. Corrected = Actual Prediction - Score predicted based on spurious block), so the effect of spurious variables is removed \citep{snoek2019confounds, chyzhyk2022confounds, frisch1933partial, lovell1963seasonal}. We note that controlling for these variables changes our dataset and thus the measurement target: a model fit after stratifying out length, topic or genre predicts choices conditioned on these attributes. This is a classical problem in text analysis, and we take a common approach \citep{gururangan2018annotation, poliak2018hypothesis, mccoy2019right, landeiro2016robust, biber1988variation}.}. We give more detail, including spurious features we discovered, in Appendix \ref{app:dense-decay} and \ref{app:balancing}. $\mathcal{V}$, $\mathcal{VA}$ and $\mathcal{VAT}$ are trained on the same stratified and stripped datasets; residualization is applied to each model's score.

\begin{table}[t]
\centering
\setlength{\tabcolsep}{5pt}
\setlength{\fboxsep}{1.6pt}
\scriptsize
\renewcommand{\arraystretch}{1.28}
\newcommand{\vchip}[3]{\colorbox{vgreen!#1}{\strut\tiny\textcolor{black}{#3}}}
\newcommand{\achip}[3]{\colorbox{aamber!#1}{\strut\tiny\textcolor{black}{#3}}}
\begin{tabular}{@{}>{\scshape\raggedright\arraybackslash}p{1.8cm}p{5.75cm}p{5.85cm}@{}}
\toprule
 & {\color{vgreen!55!black}\textbf{Top verifiable checks $V$}} &
   {\color{aamber!60!black}\textbf{Top articulated criteria $A$}} \\
\midrule
software code \newline {\upshape\scriptsize $y$: PR merged} &
\vchip{58}{white}{no pass$\to$fail regressions} \newline
\vchip{30}{black!75}{tests flip fail$\to$pass} \newline
\vchip{22}{black!75}{patch applies; suite executes} &
\achip{32}{black!75}{message says what changed, why} \newline
\achip{30}{black!75}{structured subject/body, issue links} \newline
\achip{27}{black!75}{single-purpose, reviewable change} \\
\addlinespace[7pt]
creative writing \newline {\upshape\scriptsize $y$: story upvotes} &
\vchip{20}{black!70}{plot-arc shape proxy: paragraphing, rising action density} \newline
\vchip{16}{black!70}{form/medium-fit check} \newline
\vchip{13}{black!70}{worldbuilding-specificity check} &
\achip{64}{white}{deliberate contrasts of image and tone} \newline
\achip{47}{white}{no stumbles read aloud; rhythm holds} \newline
\achip{38}{black!80}{micro-tension keeps pages turning} \\
\addlinespace[7pt]
math \newline {\upshape\scriptsize $y$: answer accepted} &
\vchip{27}{black!75}{lint checks} \newline
\vchip{18}{black!70}{machine-checkable-proof check} \newline
\vchip{13}{black!70}{formula-correctness check} &
\achip{56}{white}{every inference warranted, no jumps} \newline
\achip{55}{white}{brief, inevitable, clear argument} \newline
\achip{50}{white}{airtight steps, no hidden assumptions} \\
\bottomrule
\end{tabular}
\caption{\textbf{What a program can check vs.\ what can be articulated to a judge:} top three features per task, highlighted proportional to explanatory strength. The $
\mathcal{V}$ checks: an LLM writes a deterministic program implementing the named construct (e.g.\ the plot-arc check scores paragraph structure and rising verb-density as in \citet{yufei_spangher}). Math's correctness checks fail to predict answer acceptance (AUC $\le .55$, against $.65$--$.66$ for its articulated criteria). For details, see Appendix~\ref{app:prompts}.}
\label{tab:topfeatures}
\end{table}

\paragraph{Additional Robustness Estimates and Training Details: Bootstrap, Influence and Sensitivity} Finally, we take other steps to challenge the robustness of our gaps. Every $\mathcal{V}$, $\mathcal{VA}$ and $\mathcal{VAT}$ AUC we report is a mean over $b=3,000$ bootstrapped test samples. For every task we report largest change when any single group of data is left out\footnote{Each dataset contains natural groups of data: e.g. all responses to the same Reddit story prompt, the Stack Exchange question, the rulemaking docket. Leaving a group out drops every item in this group.}, the change when half the data is left out, and the change under different modeling choices (e.g. fold assignment, regularization, imputation). %
Finally, we take the following steps to prevent overfitting: only the training and evaluation data splits are used for metric discovery, criteria are admitted by predictive value, prompts are selected on the evaluation split, and a campaign stops when a round adds less than the noise floor. The test split is not seen except when scoring our trained models Table~\ref{tab:master-results}; the upper and lower bound estimates discussed next use proposal counts, not test labels. See App. ~\ref{app:standard-cis}, \ref{app:battery} and \ref{app:whisker-backtest}.

\section{Results}
\label{sec:results}

\subsection{Articulability gaps occur across nearly all tasks and datasets}
\label{sec:res-gaps}

We find verifiability and articulability gaps exist across all fields we study (Figure \ref{fig:concept-row}). Creative Writing and Academic peer review are two domains with the largest gaps, representing $34$--$39\%$ of shared preference
\footnote{We report AUC-based discrimination  here because it is interpretable, in Appendix~\ref{app:info-view} we recompute every task in bits, where the decomposition is additive; model orderings match in $40/47$ populations and share rankings correlate at $\rho=.93$.}
, across different preference variables. %
However, even domains that are typically treated as verifiable, in RLVR, have substantial gaps: math preference, with up to $26\%$ %
and coding with up to $28\%$ of shared preference
(nevertheless, verifiable metrics explain between $13\%$ and $80\%$ of shared preference, a substantial share, in every domain; Table~\ref{tab:master-results}). One explanation for peer review's wide gap is that its standards are set inside small, insular communities and learned by participation \citep{lucy2023words, vilhena2014finding}. 

Our results in Humor were surprising, as we initially hypothesized it would be one of the hardest fields to articulate. We found that, while humor has relatively large \textit{verifiability gaps} ($34\%$ of \textit{articulable signal} is \textit{verifiable} across its preference signals; the lowest of our seven fields, Figure~\ref{fig:verifiability-boxplots}), it has relatively small \textit{articulability gaps}. Although researchers typically regard Humor as a tacit field (see Appendix \ref{app:quotes:humor}; prior attempts to model it computationally have failed \citep{shahaf2015inside, hessel2023androids}), we find articulated criteria carry most of shared preference. On r/Jokes crowd upvotes, articulated criteria carry ${\sim}85\%$ of recoverable preference signal (Table~\ref{tab:master-results}); and its strongest criteria are comedy mechanics: ``virality/shareability'' $.613$ and ``misdirection/reveal'' $.607$. Some of these metrics might seem difficult for \textit{humans} (e.g. virality might seem hard to predict), however, it is likely that LLMs contain enough situational knowledge in their pretraining to make such predictions, highlighting that articulability is judge-relative \citep{companion_codability}.%

\subsection{Articulability Gaps are Robust}
\label{sec:res-frontier}\label{sct:robustness}

Four reasons to doubt the validity of verifiability and articulability gaps are: (1) We may not have articulated \textit{enough} metrics (genuine and spurious) (2) We may not have articulated metrics \textit{well enough} and (3) We have not fit metrics with capable-enough models. (4) Model capacity, judge capacity, prompt expressiveness or dataset leaks could account for gaps.
We address these objections here; for a more specific task-by-task discussion, see App. \ref{app:whisker-backtest}, \ref{app:dense-decay} and \ref{app:negative}.

\paragraph{\textit{Q1: Could we have articulated more metrics?} A1: No, discovering more genuine and spurious metrics does not close the gaps.} As described in Section \ref{sec:methods}, we estimate effects of \textit{unseen} genuine and spurious metrics, and we find minimal effects across all tasks. For example, on Creative Writing's r/WritingPrompts upvotes, $\mathcal{VA}$ rises $AUC=.607 \to .665$ as $k: 8 \to 144$ with an asymptote of $.668$. The degree to which spurious features \textit{decay} performance also saturates: on the same task, $\mathcal{VAT}$  decays by $.05$, from $AUC=.79$ as $k:0 \to 70$ towards an asymptote of $.73$%
\footnote{Conversely, adding spurious metrics to $\mathcal{V}$ \& $\mathcal{VA}$ models, i.e. like genuine metrics, moves AUC $.67 \to .69$.}
. Overall, the $\mathcal{VAT}$ model retains $34$--$43\%$ of the task's shared preference. Mutual information between the $\mathcal{VA}(X)$ and $y$ is positive on $45$ of $47$ trials. 
For more details, see Appendix~\ref{app:whisker-backtest}, ~\ref{app:dense-decay}, ~\ref{app:info-view}.%

\paragraph{\textit{Q2: Could we have articulated our existing metrics better, with better judges?} A2: No, better prompted/coded metrics and judge capacity also do not close the gap.} \citet{companion_prompts} previously explored how prompting improves both through prompt-optimization and model scaling. We run a similar analysis: running prompt optimizations to saturation and then increasing model size, as in \citet{ruan2024observational}, we extrapolate each metric's fidelity 
\footnote{Recall, we defined our \textit{fidelity} objective in Section \ref{sct:articulated_and_verifiable_metrics}: consistency, reconstruction, and synthetic metric scoring. Fidelity only observes text, not labels, so optimizing a metric's fidelity can sharpen how well it measures its own construct but not fit the outcome.}
asymptotically. Fully optimizing all metrics' prompts for each judge adds at most $.05$, and estimating an optimal judge's performance vs. current judges moves the articulated model by at most $.031$ across tasks (Appendix~\ref{app:family-robustness}, ~\ref{app:alt-posttraining}). In other words, having an \textit{optimal} metric prompt on an \textit{optimal} model still leaves the $\mathcal{V}$ and $\mathcal{VA}$'s asymptote below $\mathcal{VAT}$ in all cases. One explanation for this: tracking each metric's item-level discrimination as prompts improve %
\citep{lord1968statistical, embretson2000item}, we find prompting metrics better appears to cannibalize the effects of adding new metrics. %

\paragraph{\textit{Q3: Can better model architectures aggregate metrics better?} A3: No, and metrics do not span the decision-space of $\mathcal{VAT}$.} A third hypothesis explaining observed gaps is: an underexpressive model architecture might miss nonlinearities and interactions between articulated metrics. We address this concern in two ways. First, we train different model architectures on articulated criteria (e.g. logistic, boosted trees, random forests, MLPs, degree-2 interactions, an RBF kernel machine); this does not affect held-out AUC (Appendix~\ref{app:negative}). Secondly, we regress the $\mathcal{VAT}$ \emph{score} on articulated criteria: if the full model's advantage were only better \emph{aggregation} of the criteria we already have, then $\mathcal{VAT}$ \textit{itself} would be reconstructible from the metrics. We find that a substantial part of the $\mathcal{VAT}$'s ranking is \textit{not} reconstructible from metrics ($\rho=.56$; Appendix~\ref{app:negative}). The unreconstructed residual, or the component of the full model \textit{orthogonal} to any metric is not noise: it explains $85\%$ of the articulability gap, a share that is specific to our full model\footnote{More concretely, in r/WritingPrompts upvotes, it adds $+.108$ AUC in a creative writing task, or AUC=$.665 \to.773$, against full $AUC(VAT)=.792$.}. %

\paragraph{\textit{Q4: Are gaps caused by other explanations (e.g. pretraining, corpora or judge differences)?} A4: The corpus- and population-side alternatives we can test do not explain gaps (Appendix~\ref{app:alternatives}).} We tested three further alternatives to gaps as tacit knowledge. \emph{Pretraining exposure}: if certain terms (e.g. ``mathematical elegance'') were simply seen more during pretraining, preferences in some fields might be understood better. To address this, we measure how often phrases
\footnote{Phrases for venues or genres, e.g.\ ``flash fiction'', ``math.stackexchange'', ``area chair''; Appendix~\ref{app:alt-density}.} 
appear in a large pretraining corpus, Dolma \citep{soldaini2024dolma, liu2024infinigram}; corpus density does not predict gaps ($\rho{=}.12$, $p{=}.73$). Nor does item-level exposure: on the five tasks with enough items dated after Llama-3.1's pretraining cutoff (December 2023), the gap is no larger on items the base model could have seen than on later ones (pooled difference $-.005$ $[-.033, +.023]$; Appendix~\ref{app:gap-battery}). We also split training by time: our $\mathcal{VAT}$ model is trained only on papers from before the Llama-3.1 pretraining cutoff and tested on papers from after it (Appendix~\ref{app:temporal}). \emph{Subfield mixture}: if, for instance, subfields had opposing \textit{articulated} preferences (e.g. in humor, \textit{dark comedy} might emphasize taboo while \textit{slapstick} does not), then a pooled set of metrics could not capture subfield variance while a full model enough capacity could. To investigate this, we built separate articulated models across identifiable subfields in 8 tasks and pooled them; this does not close the gap (e.g. $.032$ vs. $.021$ pooled vs. single-model for N\&C responses, $.126$ vs. $.124$ for r/WritingPrompts upvotes, Table~\ref{tab:alt-subfield}; Appendix~\ref{app:alt-subfield}). %
\emph{Post-training}: differences in post-training across fields might also have an effect on judge performance. To test this, we compared performance of base models at 8B and 12B with post-trained models by optimizing prompts for each judge separately; we do not find a meaningful difference in articulability gains across fields, the instruct-minus-base gain is $.06$, $.05$ and $.00$ for a Llama-3.1-8B pair and $.10$, $.06$ and $.07$ for a Gemma-3-12B pair on peer review, N\&C and creative writing (Appendix~\ref{app:alt-posttraining}). An untrained $\mathcal{VAT}$ model, a retrieval-augmented $\mathcal{VA}$ model and $\mathcal{T}$-only dense model reported in Appendices~\ref{app:gap-battery}, \ref{app:field-object} and \ref{app:master-results}. %

\section{Analysis}%
\label{sec:res-decisions}

Having measured articulability gaps in Section \ref{sec:res-gaps}, and having ruled out other explanations for these gaps in Section \ref{sec:res-frontier}, we now want to explore why these gaps exist, and what the consequences are.  %

\subsection{Why do articulability gaps occur?}

Examples of items with large articulability gaps, or high $\mathcal{VAT}$ scores and low $\mathcal{VA}$: a horror story, \textit{``The creatures were all smiles. A bunch of tall, pale, ghostly things with these dark smiles.''}; a news article, \textit{``Aphantasia: A life without mental images''}; a press release, \textit{``Target Addresses Firearms in Stores''}
\footnote{Horror story: \url{https://www.reddit.com/r/WritingPrompts/comments/15cm4uu/} (July 2023), full prompt: ``Humans are 1000x more sensitive to the smell of petrichor \ldots\ than sharks are to blood \ldots\ They smell like petrichor.'' News article: \url{https://www.bbc.co.uk/news/health-34039054} (August 2017), headline: ``Aphantasia: A life without mental images''. Press Release: \url{https://corporate.target.com/article/2014/07/target-addresses-firearms-in-stores} (July 2014), headline: ``Target Addresses Firearms in Stores''.}
Two human annotators described what 50 high-gap texts per task had in common compared with low-gap texts (Appendix~\ref{app:item-insights}). 62\% of texts had within-text indicators: clear writing (51\% of such texts, for example: \textit{stated its point up front}, and \textit{was ``reasonable'' in claims}) and whether the text contained a twist (44\%, for example: \textit{bringing the response to a bigger picture}, or \textit{containing a surprise at the end}). 22\% of texts had corpus-relative indicators: novelty (79\% of such texts, including: novelty of \textit{concept, approach or writing style}), conversely, overusing idioms appeared in negative examples (65\%, \textit{obvious puns}, \textit{academic articles using cliched titles}). These properties were possible for annotators to spot and name but often difficult to further describe in words and thus challenging for LLMs to detect; some corpus-relative properties, such as novelty, might be known to the full model during training, as it has read the corpus; we consider this as part of what enculturation transmits \citep{csikszentmihalyi1996creativity}. 
\begin{wrapfigure}{r}{0.44\textwidth} %
\vspace{-10pt}
\centering
\includegraphics[width=0.44\textwidth]{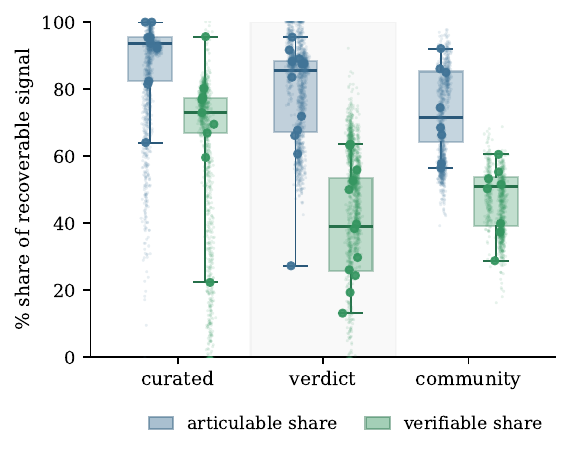}
\caption{\textbf{Shares by preference channel.} Recoverable preferences by preference channel; one point per task, faint points are bootstrap draws.}
\label{fig:articulated-share}
\label{fig:verifiability-boxplots}
\vspace{-8pt}
\end{wrapfigure}
We find scant evidence of other textual cues we hypothesized would be present: e.g. jargon \citep{lucy2023words}, reading grade \citep{fleshkincaid}, and median educational level of human judges do not predict gaps (Appendix~\ref{app:field-jargon}, ~\ref{app:field-language}, ~\ref{app:field-deciders}) Since such attributes are measurable, they may already be proxied in $\mathcal{VA}$.

Social features surrounding the creation of text contain more measurable signal. When we group tasks by the \textit{preference channel} that each task falls into, we see significant differences (Kruskal--Wallis $p{=}.011$ over the 28 tasks; App. ~\ref{app:gap-battery}). Recall, in Section \ref{sec:dataset}, we defined three types of preference channel: \textit{gatekeeper verdict}, where a decision-maker accepts or rejects a work (e.g. paper acceptance); \textit{expert curation}, where top-$k$ works are selected from a pool (e.g. paper prizes); and \textit{community voting}, where preference labels are aggregated across individuals (e.g. citations). In Figure \ref{fig:verifiability-boxplots}, we show that expert curated preferences are the most articulable while community voting preferences are the least
\footnote{Does lower predictability, or more preference disagreement, impact gaps? While curated preferences often have the lowest \textit{absolute} gaps \textit{and} the lowest predictability, the \textit{relative} gaps, we find, do \textit{not} correlate with predictability ($\rho{=}.09$); eleven tasks matched on performance still have relative gaps ranging from $0$--$44\%$ (Appendix Fig.~\ref{fig:auc-ceilings}). We define: recoverable signal = $AUC(\mathcal{VAT}) -.5$. Articulability gap = $AUC(\mathcal{VAT}) - AUC(\mathcal{VA})$. Articulated share = $\frac{AUC(\mathcal{VA}) -.5}{AUC(\mathcal{VAT}) -.5}$. Relative gap = 1 - articulated share}. 
In academic peer review, for example, $64\%$ of paper prize predictability is articulable, %
$61\%$ paper acceptance, %
and only $58\%$ on citation percentile
\footnote{This holds after controlling for between $43$--$57$ spurious variables for each label; including topic, venue and year variables. Creative writing gives another example: best-of lists (i.e. expert curated) show no articulability gap at all, while only $57\%$ of shared preference in crowd-rated stories (i.e. community voting) is articulable.}.
Prior work has found that popularity and critical judgment decouple in creative domains \citep{salganik2006experimental} %
for example, within academia: the metrics that predict citations agree only weakly, we observe, with those that predict prizes, with rank agreement $\rho{=}.47$ (see Appendix~\ref{app:subfields}). \textit{Within} preference channel, we also observed that the number of evaluators per decision correlated with gaps ($\rho{=}.46$; $n{=}28$, $p{=}.016$; the deconfounded gap is flat in training-set size, Appendix~\ref{app:scaling}, ~\ref{app:gap-battery}). \citet{collins2010tacit}'s taxonomy of origins and types of tacit knowledge defines \textit{collective tacit knowledge} as knowledge that arises when groups of people jointly hold social norms that change too often to be put into words and are \textit{felt} rather than \textit{stated}. Our findings offer evidence of this effect.

\subsection{Consequences of Articulability Gaps}

\paragraph{Consequence \#1: Articulation gaps change judgements on creative outputs.}
\label{sec:res-selection}
How do $\mathcal{VA}$ and $\mathcal{VAT}$ models differ in their rankings on creative output? We measure this with \textit{selection regret}, or how far down a model's ranked list the human's preferred choice is: we look at tasks where natural text groupings exist, within which humans choose single item among many (e.g. StackOverflow answers to one question, or Reddit comments on a post)
\footnote{Formally, regret$=(1/|G|) * \sum_{g \in G} ( y[i_T(g)] - y[i_A(g)] )$, where $G$ is all groups for the corpus, $y[i]$ whether or not humans chose item i, and $i_T(g)$, $i_A(g)$ is the top item chosen by $\mathcal{VAT}$ and $\mathcal{VA}$, respectively.}
and observe how our models rank these texts. In modern AI systems, regret has consequences: if a reward model ranks outputs with ``poor taste'' higher, post-trained systems will drift in this direction. Or, beyond model training, a decision-model in a human-AI hybrid system
\footnote{In journalism, for example, the NYTimes has up to 8 editing layers before publication \citep{clayman1998gatekeeping, gans1979deciding}; in software engineering, every code change at Google must be reviewed and approved before submission \citep{sadowski2018modern}.} 
might promote actions that diverge from human desires. Across tasks, we observe that the $\mathcal{VAT}$ model's preferred item is at least as preferred by humans as the $\mathcal{VA}$ model's; in every task where the two rank items differently, VAT's is more preferred more times than not (Table~\ref{tab:regret}). 
We also observe \textit{dose response} effects: we induce greater gaps by refitting $\mathcal{VA}$ on random $k$-subsets of each metric bank and find that regret tracks the induced gap ($\rho = +.65$, Table~\ref{tab:dose}). Interestingly, across tasks, the gap is not calibrated: a small articulability gap does not imply a $\mathcal{VA}$ is safe because decision margins between tasks differ: regret is decided within a group, by how close the candidates score, so the same gap in AUC costs more where choices are close\footnote{Math.SE accepted answers has a smaller gap than Math.SE votes but a larger regret ($+.068$ against $+.014$)}. The dose-response experiment shows regret also does not depend on \textit{which} metrics define the gap: removing the most predictive, the least predictive, or random metrics moves regret in step with  AUC (Appendix~\ref{app:regret-all}); regret follows the size of the gap and not the identity of the missing criteria: articulability and verifiability gaps are therefore a measure of selection risk within a task.

\begin{table}[t]
\centering\small
\begin{tabular}{lrrr}
\toprule
task & regret (full model $-$ articulated) & 95\% CI & disagreement odds \\
\midrule
peer review: citations & $+.171$ & $[.125, .216]$ & $4.4:1$ \\
peer review: accept/reject & $+.109$ & $[.078, .140]$ & $2.1:1$ \\
math.SE: accepted answers & $+.067$ & $[.029, .103]$ & $1.4:1$ \\
peer review: spotlight & $+.065$ & $[.020, .109]$ & $1.3:1$ \\
creative writing: upvotes & $+.061$ & $[.013, .107]$ & $1.5:1$ \\
\bottomrule
\end{tabular}
\caption{Selection regret on a sample of tasks where groups of human text responses (e.g. StackOverflow questions) containing both a chosen and an unchosen item %
(Appendix~\ref{app:regret-all}). Regret as defined in \S\ref{sec:res-selection}: the full model's hit rate minus the articulated model's, averaged over groups. Intervals are 3,000-sample bootstraps; regret is positive in 13 of 16 tasks (sign test one-sided $p{=}.011$; Appendix~\ref{app:regret-all}). Disagreement odds is when the instruments pick different items, how often the community's actual winner is $\mathcal{VAT}$'s pick vs.\ $\mathcal{VA}$'s. When the two instruments disagree, for example, in peer review:citations, $\mathcal{VAT}$'s pick matches humans' recorded choice 4.4:1.}
\label{tab:regret}
\end{table}

\paragraph{Consequence \#2: Goodhart's Analogue: Selecting by Articulable Metrics Sheds Unstated Preference.}
\label{sec:res-optim}

Having shown that the full model is a better proxy for human preference on \textit{real} text, we now study preference measurements on \textit{generated} text. We return to an idea expressed in Section \ref{sec:intro}:  as rubric- and constitution-based post-training optimizes models \textit{towards} articulated metrics, do they drift \textit{away} from full judgement? To test this, we start with tasks where prompts exist naturally (e.g. human questions on Stack Overflow)%
. For each prompt, we generate a pool of candidate LLM-generated responses (18 generated across three model families
\footnote{We use Llama-3.1-8B-Instruct, Qwen2.5-14B-Instruct, and phi-4; the judge's model family (Gemma) was deliberately excluded as a generator, as models are known to prefer generations from their own families \citep{panickssery2024llm}. Generator identity does not separate the instruments (real-vs-generated AUC $.46$--$.53$ for every family) on some tasks (e.g. creative writing); on others (e.g. Math.SE) the full model rates its own family's answers higher, so Llama candidates are dropped from every readout}. 
and six instruction conditions
\footnote{Our prompts here, abridged: (a) plain: ``write like a human''; ``very human'' (idiosyncratic voice, small imperfections); (b) casual (a redditor replying on their phone); (c) literary (polished and ambitious); and (d) a high-temperature condition ($T{=}1.3$). This helps generators be less homogeneous in their quality (Appendix~\ref{app:pools}).}
, plus real human stories written in response to that prompt). As in the previous experiment, the articulated model and the full model rank their favorites; however, unlike before, we lack ground truth about human preference. So we compare the two models' rankings of the same pool with each other. %
The \textit{cross-model percentile rank} is where on the \textit{first} model's ranking the \textit{other} model's top pick sits \citep{gao2023scaling, wen2025rethinking}. We observe an asymmetry: $\mathcal{VAT}$'s \textit{top pick} is, on average, in the $.7$ percentile of $\mathcal{VA}$'s rankings, while $\mathcal{VA}$'s top pick is only in the $.6$ on $\mathcal{VAT}$'s. This means, concretely, $\mathcal{VA}$ values $\mathcal{VAT}$'s selection \textit{higher} than $\mathcal{VAT}$ values $\mathcal{VA}$'s. We leave a reinforcement-learning run to future work. Cross-model rankings are relative; they might both be wrong. However, the direction fits what each model is exposed to: the articulated criteria are largely \textit{visible} to $\mathcal{VAT}$, so $\mathcal{VAT}$'s top pick satisfies them and thus ranks highly in $\mathcal{VA}$'s ranking; conversely, the articulation gap is \textit{invisible} to $\mathcal{VA}$, so $\mathcal{VA}$'s top pick ranks less favorably in $\mathcal{VAT}$'s ranking. %
\footnote{An example of the disagreement. For the prompt: ``You are a necromancer\ldots you use your powers for a much\ldots'', the articulated model's top story (of 22) is a master-and-apprentice scene, half dialogue, scored $1.0$ on every escalation and payoff criterion and $0$ on both voice criteria; the full model's top story is a first-person account with no dialogue (``I was never one for grand gestures\ldots My first was little Janet, murdered by her own mother for a thrall spell that went wrong''), scored the reverse. Each model ranks the other's pick around 10th. Across all $7{,}236$ generated stories the same split holds: the articulated model emphasizes escalation, payoff and dialogue, the full model on rhythm, subtext, ambiguity and voice.}

\section{Discussion}

Thus far, we have shown that articulability and verifiability gaps exist across 7 creative domains and 42 tasks; we have developed new methods to estimate them robustly, with optimal rubrics and programs (i.e. estimated-upper bound $\mathcal{VA}$); and deconfounded dense models (i.e. estimated lower-bound $\mathcal{VAT}$). We have explained reasons for why these gaps might arise, specifically, that they arise when text is novel, helpful, clear, and not-overused idioms; they also result from scenarios where social construction of preference involves more judges. We have discussed consequences of these gaps: because $\mathcal{VA}$ and $\mathcal{VAT}$ models' judgments diverge, $\mathcal{VA}$ becomes a poorer reflector of human preference and $\mathcal{V}$ amplifies these discrepancies: coded programs capture as low as $13\%$ of shared preference across our tasks, and even mathematics keeps a $20\%$ or larger unverifiable share. Using either as a reward model optimizes models away from inarticulated preference. The consequences likely extend beyond what we have shown: poor writing patterns (i.e. ``AI-slop'') \citep{padmakumar2024does, doshi2024generative}; misalignment \citep{gao2023scaling}; safety failures\footnote{
Whether or not the \textit{substance} of a rule is obeyed depends not on the text of the rule itself (\S 202, \citealp{wittgenstein1953pi}), but how the community interprets the rule and what tacit knowledge is layered onto it \citep{kripke1982wittgenstein, winch1958idea, daston2022rules}.
} \citep{debraji} and functionality failures \citep{raji2022fallacy} can all result from underspecified reward models.

Humans accommodate for underspecification by reasoning and by using background and tacit knowledge \citep{polanyi1966tacit}. We have avoided the claim that articulation and verification gaps in CreativePreferences tasks \textit{are} measurements of tacit knowledge that $\mathcal{VAT}$ models are able to learn. Although we have ruled out, by negation, other explanations for the gaps (e.g. too few criteria, criteria stated badly, weak judges, weak aggregators, etc., App.~\ref{app:alternatives}), it is unclear whether LLMs can acquire all forms of human tacit knowledge with current learning mechanisms. To understand why, consider tacit knowledge's sources: \citet{collins2010tacit} defines tacit knowledge as being somatic or collective in origin. Somatic tacit knowledge (STK) refers to knowledge we gain from interacting with the natural world and internalized in our body \citep{polanyi1958personal, dreyfus1986mind}. Collective tacit knowledge (CTK) refers to knowledge we gain from interacting in social systems and is held by these systems \citep{wittgenstein1953pi, collins2010tacit, collins1974tea}. On the one hand, language models do not exist in physical space \citep{harnad1990symbol, brooks1991intelligence, bender2020climbing, bisk2020experience, mahowald2024dissociating} it is questionable whether they can engage in the empathic mechanisms required to understand of human minds \citep{merleauponty1962phenomenology, tomasello2005understanding, yejin_theory_of_mind, polanyi1958personal}. On the \textit{other} hand, \textit{some} forms of tacit knowledge clearly \textit{have} been able to enter AI systems: 
the ability to understand natural language at all, or play the ``language game'' \citep{wittgenstein1953pi}, depends on some degree of tacit knowledge acquisition; this knowledge is continually demonstrated when models: e.g. translate underspecified prompts into responses \citep{ouyang2022training}, decode qualitative constructs \citep{companion_prompts, spangher_newsinterview}, and adjust to cultural contexts \citep{hershcovich2022challenges, adilazuarda2024towards}, among other feats.

\subsection{Recommendations and Future work}

With these possibilities and limitations in mind, we offer a roadmap for future work and a framework for thinking about tacit knowledge in our processes: 
acquiring tacit knowledge, we argue, is crucial for continued progress on frontier tasks. Thus, we make the following recommendations:%

\paragraph{Recommendation and Future Work \#1 (Measurement): Gaps should be \textit{measured} where possible; for tasks with small gaps, prompt-instruction and rubrics likely suffice; tasks with large gaps may present underspecification risks.} 
AI can deliver socially beneficial outcomes in many tasks (e.g. science, law, coding); however, we show that these tasks have substantial articulation and verification gaps. The larger the gap, the more preference an articulated objective leaves out, and thus more cautious we should be: \citet{zhuang2020consequences} show that optimizing a proxy that omits part of the true objective can make the outcome worse than not optimizing at all. We present our methods for finding gaps as \textit{one} approach to measurement that allows us to analyze taste at scale, across time and from experts who are less likely to participate in scientific trials. Further research aimed at developing additional measurement approaches, like \textit{in-situ} user designs, might allow us to discover gaps with less outcome data, measure different forms of tacit knowledge (e.g. STK vs. CTK) within gaps, and obtain richer \textit{descriptive} insights into human tacit processes. Our work focused on preferences and is directly useful for a set of social goals, like ``\textit{adapt AI better into human processes}'', ``\textit{reduce AI slop}''. Other goals may not be as preference-focused (e.g. \textit{``emulate human intuition in scientific discoveries''}) but might, according to \citet{polanyi1958personal}, still involve substantial tacit knowledge. New data collection approaches can help us measure tacit knowledge in these tasks, like recording workflows \citep{gum_omar} or novel approaches to interviews \citep{klein1989critical, militello1998applied} (i.e. interviews need to be adjusted to capture the fact that tacit knowledge can be difficult for humans to describe in words). Finally, better measurement of tacit knowledge within tasks might help us identify if \textit{subtasks} within a task that can be fully articulated, yielding hybrid articulable and tacit designs\footnote{In support of hybrid articulable and tacit systems, \citet{daston2022rules} identifies tasks as combinations of \textit{thin} vs. \textit{thick} rules. A thin rule is fully specified, like a cooking recipe with exact measurements and timings; a thick rule relies on experience and interpretation.}. %

\paragraph{Recommendation and Future Work \#2 (Learning): A deeper understanding of how humans acquire and use tacit knowledge can help us devise new learning mechanisms.}

Let us consider \citet{polanyi1958personal}'s idea that scientific discovery requires substantial human intuition and tacit knowledge: Einstein, he says, imagined he was ``traveling the lightbeam'' to develop a hypothesis on relativity. How might we design systems that acquire this degree of tacit intuition? Current learning mechanisms \textit{do} capture tacit knowledge during \textit{pretraining} (i.e. \citet{companion_prompts} observes that AI can interpret rich, social concepts), \textit{fine-tuning} (i.e. we observe that further training can predict beyond what a model can express) and likely \textit{behavioral learning}. Is tacit knowledge gained through these mechanisms enough to enable an AI system to ``travel the lightbeam''? Is additional data required? Additional pretraining data or extensive rewards might expose a model to examples of human to enable these kinds of subtle processes to emerge. Preference-pair elicitation \citep{christiano2017deep, stiennon2020learning} might more directly tune competing approaches, although elicitation is a more artificial representation of how judgment happens in the field \citep{csikszentmihalyi1996creativity}, is not as scalable and, under observation tends to return articulated preferences \citep{nisbett1977telling}. And what if such approaches are simply too data inefficient to enable such rich capabilities to emerge as \citet{polanyi1958personal} depicts? There are other areas of tacit knowledge acquisition that are poorly understood even in humans, and might yield novel approaches. For example, humans regularly emulate \citep{Grush, lydia_hopper}, imitate \citep{tomasello1999cultural}, and use empathy \citep{decety2004functional, dewaal2008putting, clark1996using, tomasello2005understanding} to both infer states of mind as well as make predictions; \citet{polanyi1958personal}'s conception of \textit{indwelling}, for instance, is predicated on an ability to robustly project the self into another. Learning mechanisms for machines that are more directly tuned towards tacit knowledge acquisition, like emulation learning \citep{spangher_thesis}, proactive social feedback \citep{lave1991situated, collins1974tea}, and apprenticeship \citep{apprenticeship_learning_abbeel} could help us directly address tacit deficits that, we predict, will stymie frontier tasks. %

\paragraph{Recommendation and Future Work \#3 (Deferral): When tacit knowledge is not recoverable by learning methods, consider delegating to humans.}

Finally, for some tasks, machines may simply never acquire enough tacit knowledge to be able to perform some tasks at their frontier, in which case we speculatively advocate more human-in-the-loop approaches or, more radically, to bound expectations that AI will simply never perform well in these tasks.
To conceptualize this family of tasks, let us consider an extreme case of tacit knowledge that not only \textit{can not} but \textit{must not} be articulated: it \textit{must remain} tacit because articulation would destroy the knowledge itself. In fields like creative writing, humor and music, one \textit{goal} is to establish shared experience through tacit meaning \citep{susan_langer}. The knowledge conveyed is not in the articulated content itself, but like a non-verbal Turing test, a code to establish that the speaker and listener \textit{share an understanding}; the \textit{speaker} behind the words is more important than the words themselves. 
\citet{colins_the_shape_of_actions} broaden this view to describe why certain tasks \textit{resist} articulation: some tasks resist because they are \textit{polimorphic}: a social group keeps revising the social context \textit{precisely so that} no fixed rule can be written for it and \textit{continued engagement} with the group is necessary (their example: ``\textit{writing a genuine love letter}''); unlike a \textit{mimeomorphic} action, which is done the same way and can be mechanized. %
Measurements determining the bounds of tacit knowledge, especially beyond the bounds of dense model training, might be useful to help model developers re-prioritize and to plan for human-in-the-loop where it is most needed.

\section{Related work}
\label{sec:related}

\paragraph{Tacit knowledge, tacit rules and learning methods.} Tacit knowledge has long been studied in the context of philosophy, starting with descriptions of \textit{knowing} (e.g. ``we can know more than we can tell'' \citep{polanyi1966tacit, polanyi1958personal}, knowing \textit{how} and knowing \textit{that} \citep{ryle1949concept}). \citet{schon1983reflective} and \citet{argyris1985action, nisbett1977telling} expanded tacit knowledge to contain professional judgment and the gap between the rules people espouse and the rules they act on; these intersect with rule-science \citep{daston2022rules} more broadly, where the distinction between \textit{thin} and \textit{thick} rules (\textit{thin rules} can be fully specified, like cooking recipes; \textit{thick rules} are harder to \textit{articulate}, and require apprenticeship to interpret). Like many aspects of modern philosophy, tacit \textit{knowledge} and tacit \textit{rule-following} converge earlier, in Wittgenstein's philosophy: neither language nor rules contain their own meaning, and what determines correctness is the community interpretation \citep{wittgenstein1953pi, kripke1982wittgenstein, winch1958idea}. For example, food-safety inspectors applying one written code to identical conditions disagreed 60\% of the time; what reduced the disagreement was peer review among inspectors, not more rules \citep{ho2017does, ho2017managing}. Psychologists and sociologists have studied how tacit knowledge is learned: norms become internalized implicitly \citet{dreyfus1986mind}; processes of inference over candidate rules \citep{goodman2008rational, goodman2015concepts} and  unstated intent \citep{goodman2016pragmatic} occur  from examples. \citet{collins2010tacit, collins1974tea, collins2007rethinking} identified the origins of tacit knowledge: somatic tacit knowledge (STK), which originates in the body, and collective tacit knowledge (CTK), which originates in society. CTK is acquired by taking part, by imitation and shared intention \citep{lave1991situated, tomasello1999cultural, tomasello2005understanding}, rehearsal and feedback \citep{yang2024social}. Although researchers are increasingly discussing tacit knowledge in the context of machines \citep{dreyfus1972what, autor2014polanyis, collins2018artifictional, fjelland2020why, budding2025what, lu2025tacit, zou2026creative, li2026what}, the most direct discussion in recent years has been in the context of embodiment \citep{harnad1990symbol, brooks1991intelligence, bender2020climbing, bisk2020experience, mahowald2024dissociating}, which is a partial representation of STK. Although social norms have previously been \textit{written down} at scale \citep{ziems2022moral, ziems2023normbank, shi2024culturebank} including from a group's behavior \citep{franken2023social}, we find work in \textit{tacit knowledge} and preferences to be understudied. Our work is the first broad and computational study to start to parse tacit knowledge in preferences and, to our knowledge, is also the first to systematically begin to measure and address different types of tacit knowledge across tasks.

\paragraph{Reward modeling: Rubrics, judges and reward signals.} Post-training techniques, generally centered around reinforcement learning -- with either verifiable rewards, RLVR \citep{guo2025deepseekr1, lambert2024tulu3}, which in practice also include rule-based format and length checks, as in DeepSeek-R1; or LLM-judge rewards \citep{bai2022constitutional, zheng2023judging, liu2023geval, kim2024prometheus, gunjal2025rubrics, viswanathan2025checklists, lee2023rlaif, franken2024self} -- have become a major driver of progress in AI. Thus, reward models, rubrics, and verifiers have come under renewed scrutiny as a learning mechanism \citep{stegeman2016designing, pan2022effects, skalse2022defining, casper2023open, frick2025evaluate, truong2025machine}. Any fixed criterion set under-represents the construct it stands for \citep{messick1994interplay}, an argument since carried to AI evaluation \citep{schaeffer2023emergent, salaudeen2025measurement, desai2026what}, and this becomes pronounced in AI, where use-cases extend beyond tasks the evaluator anticipated \citep{zou2026creative}; a metric ceases to be a valid metric once it is optimized against \citep{goodhart1984problems, manheim2018categorizing, gao2023scaling, wen2025rethinking}. Preference models densely trained on \textit{elicited} decisions \citep{christiano2017deep, stiennon2020learning, ouyang2022training, bai2022training, rafailov2023direct} had been dominant reward-modeling approach that may have addressed these concerns, but the cost and scale of collecting preference data for agentic tasks grows exponentially, especially when such tasks become longer, more varied and more complicated \citep{lambert2024rewardbench, zhou2024rmb}. CreativePreferences uses massive amounts of \textit{found} preferences, rather than \textit{elicited} preferences; these capture decisions originating from the field, in natural settings. Although other work has examined parts of our analysis (e.g. peer review \citep{kang2018peerread, cortes2021inconsistency, si2025can}; code review \citep{bacchelli2013expectations, rigby2013convergent, bosu2015characteristics, sadowski2018modern}; humor \citep{shahaf2015inside, hessel2023androids}; creative writing \citep{fan2018hierarchical, chakrabarty2024art}; journalism \citep{spangher2021modeling, spangher2024tracking, spangher2024do, spangher2025spatial, spangher2023identifying, spangher2024explaining, spangher2025novel, spangher2022sequentially, spangher2022newsedits, spangher2024newsedits2, spangher_newsinterview}; mathematics question quality \citep{ravi2014great}; community moderation \citep{chandrasekharan2018internet, fiesler2018reddit}; and legal and administrative judgment \citep{ho2017does, ho2020improving, guha2023legalbench}) ours is the first to systematically combine all of these fields and compare them together with the same methods.

\paragraph{Discovering natural language criteria and program synthesis.} Our $\mathcal{V}$ and $\mathcal{VA}$ models rely on automated methods for building rubrics and verifiers given natural text data. As such, we utilize and build on several strains of work: HypoGeniC \citep{zhou2024hypogenic}, NLParam \citep{zhong2024nlparam}, HypotheSAEs \citep{movva2025hypothesaes}, ExperiGen \citep{sengupta2026experigen} and AutoMetrics \citep{ryan2026autometrics}, with GEPA and $\epsilon$-estimation as prompt-evolution engines we use \citep{agrawal2025gepa, companion_prompts}. \citet{companion_codability} mines 53,413 human-written criteria from online rubrics, professional guidance and community rules and studies how reliably judges apply them; our $\mathcal{V}$ and $\mathcal{VA}$ banks draw on those criteria. The programs in $\mathcal{V}$ are written by a language model and revised after failures, which places them in the line of program synthesis with language models \citep{chen2021evaluating, austin2021program} and, more closely, of methods that state a hypothesis or concept in words and then implement it as a program \citep{ellis2021dreamcoder, wang2024hypothesis, zelikman2023parsel, li2024automated, goodman2015concepts}. However, our approach differs in several ways. Each individual metric and program that we develop is optimized for an unsupervised metric, fidelity \citep{companion_prompts}, because we want to capture the construct each metric represents rather than proxy the preference variable. We are the first rubric generation paper, to our knowledge, to use capture-recapture \citep{good1953population, chao1984nonparametric, efron1976estimating} to estimate how much a longer search could add. Finally, our spurious variable estimation proposes a novel use-case for discovered metrics. In sum, our contribution is a novel procedure around discovery, the bound on undiscovered criteria and the spurious-feature audit, applied to existing discovery methods. Closest to our measurements of articulation and verification gaps are systems that learn a person's or a field's standards from behavior rather than from statements: general user models built from recorded computer use \citep{gum_omar}, apprenticeship learning from demonstrations \citep{apprenticeship_learning_abbeel}, and metric elicitation, which recovers the performance metric an expert is implicitly using \citep{hiranandani2019performance, hiranandani2020fair}. Our emphasis on measuring of what people cannot state; there are related methods in the critical decision method and applied cognitive task analysis \citep{klein1989critical, militello1998applied}, which we address in the recommendations.

\paragraph{Creative domains, expertise and agreement without definition.} We take the definition of a creative domain from the systems model, in which a field of gatekeepers rather than a fixed standard confers value \citep{csikszentmihalyi1996creativity, becker1982art, bourdieu1993field, english2005economy, lamont2009how}. Research across the domains we analyzed has repeatedly found that expert evaluators can agree without being able to articulate shared criteria, see Appendix \ref{app:quotes}. Mathematicians agree on which proofs are beautiful while disagreeing under absolute rating \citep{wells1990beautiful, sa2024agree, thurston1994proof, geist2010peer, greiffenhagen2024checking}, and disagreeing on what their own words for intuition denote \citep{burton1999intuition}. Practitioners decline to formalize what a good refactor entails \citep{buse2010learning, fowler1999smells}. Scientific reviewers apply criteria they were never given \citep{guetzkow2004originality, brunet2022making}, use explicit criteria after the fact to justify holistic judgments \citep{bloxham2011mark, roumbanis2021oracles, hug2024referees}; judges describe deciding by hunch and reasoning afterwards \citep{hutcheson1929judgment, cardozo1921nature}; editors and humorists decline to state their standards when asked \citep{lish_parisreview, white1941humor, makkai_prizes}. Art connoisseurship has attempted to build explicit methods to audit judgment which rarely work \citep{morelli1892italianpainters, friedlander1942onart}; the forecasting prediction literature found that experts have poor insight into their own cue use and that unstated channels can dominate stated ones \citep{hoffman1960paramorphic, goldberg1970man, highhouse2008stubborn, tsay2013sight}. Cultural-market experiments showed that the same works get different verdicts under different social exposure \citep{salganik2006experimental, surowiecki2004wisdom}.

\section{Conclusion}
\label{sec:conclusion}

Preferences in complex fields decompose, and the decomposition is measurable. We have shown that CreativePreferences, a novel dataset of 2.8M texts and 317M human preference judgments across 42 tasks and seven creative domains, contains articulability and verifiability gaps. These  gaps exist in every domain, including the ones treated as verifiable in RLVR; they survive more criteria, better prompts, stronger judges and better aggregation; and they widen as more people take part in the judgment, consistent with collective tacit knowledge. They are consequential: the full model agrees more with humans' actual choices, and optimizing generations towards articulated criteria shifts them away from tacit preference. While we have not shown that these gaps equate directly to tacit knowledge in humans, we see evidence for the remaining gap as collective tacit preference. We close with recommendations: (1) Measure articulation and verification gaps where possible (2) Study how humans acquire tacit knowledge to devise new learning mechanisms, using found data, emulation, correction by social groups and apprenticeship rather than a return to preference-pair elicitation. And (3) where tacit knowledge is not recoverable by learning methods, delegate to humans, or bound expectations. Tacit knowledge is a core component of what language models already do; identifying it and studying it directly is, we argue, crucial for continued progress on frontier tasks.

\section{Limitations}
\label{sec:limitations}

\paragraph{Single-seed training runs.} The full model is one full-data run per task, not a seed ensemble, and the clean-eval AUCs are single-seed point estimates. Where an eval split is small (528 items for caption finalists, 223 for the peer revealed task), the sampling error of $\Delta$ is several AUC points so significance is lower; on all other tasks, gaps are significant.

\paragraph{Whether this is tacit knowledge.} The gap is defined by an estimator failure: what a full model learns from the labels that metrics alone do not not carry. Everything we did to identify this gap as tacit knowledge is by negating of some other hypothesis. To state again: 
\begin{itemize}
    \item Against the objection that too few metrics were discovered and they were badly stated, weakly judged or badly aggregated: we mined metrics to saturation, optimized prompts to saturation, extrapolated judge capacity, swept aggregators, and regressed the full model's score on the bank (Section~\ref{sec:res-frontier})
    \item Against the objection that differences in corpus composition led to training differences: we tested pretraining density, subfield mixture and post-training (Appendix~\ref{app:alternatives}). 
    \item Against the objection that the residual is something other than preference we tested one decider's memorised taste, timing and exposure, the coarse criterion readout, judge noise, a single holistic judgment, label noise, relative labels, subpopulations and drift by item age (Appendix~\ref{app:gap-battery}); we tried to name the residual from the high-gap items themselves, from the full model's own features with a sparse autoencoder, and from what two human annotators saw in those items, and none of it closed more than $.025$ of the gap (Appendices~\ref{app:negative} and~\ref{app:item-insights}); and we trained on early years and read late ones (Appendix~\ref{app:temporal}). 
\end{itemize}
What survives these tests are gaps that grow with data while the metrics saturate, and that grows with the number of people who take part in the judgment. 

However, to state explicitly, this gap is \textit{not} proven to be the following. 
\begin{itemize}
    \item It is not shown to be tacit knowledge as a person holds it. 
    \item It is not separated into Collins's somatic and collective kinds; although the fact that more people passing judgement correlates with a bigger gap points to CTK, nothing explicitly tests this.
\end{itemize}

\paragraph{Mined and automatically constructed rubrics and verifiers.} Every articulated or programmed metric was either mined from 53,413 metrics, introduced by \citet{companion_codability}; or proposed by a language model and refined against an unsupervised objective. No rubric was written by a domain expert for this study and no verifier was built by hand. A rubric written by someone who knows the field's practice, or a checker built by hand, might carry what the mined criteria did not. Consequently, the estimated upper bounds for articulation and verification are for automatic and mined articulation. The whisker of Section~\ref{sct:missing-mass} bounds what further proposals could add; a hand-written rubric would have to exceed the estimated upper bound to change a reading.

\paragraph{Possible judge and label deficiencies.} Every articulated score comes from a language-model judge. We implement tests to detect drift within our algorithms, but the articulated model cannot know more than its judge does, and different judges might give different measurements. We try to address this by testing many different judges. Additionally, as labels are \textit{found} proxies for preference, they are poor proxies sometimes and may introduce considerable noise. For example: in notice \& comments, labels are derived by matching comments to judgements, a merged pull request reflects the maintainer's priorities as well as the patch, and citation counts measure attention. What we call taste is the part of the full model's performance that the criteria do not reproduce, so it is defined relative to that trained model. We do not separate shared but unstated preference from the variation between individual raters; that needs rater-level data, which we have only for peer review, see  Appendix~\ref{app:standard-cis}.

\paragraph{Estimator sensitivity.} Our estimate might depend on how we split the data. For the agency-agreement label on notice-and-comment data, the articulated model scores $.612$ when the grouped folds are stratified by agency and $.524$ under the frozen cross-domain \texttt{GroupKFold} design. The outcome label on the same comments gives the same score under both designs, so the instability belongs to the agreement label, whose positives cluster within dockets. We report both numbers whenever we quote this task. Its full model has the same problem: $.566$ on the evaluation split and $.639$ on the test split, where every other task has evaluation and test within noise of each other.

\bibliographystyle{iclr2026_conference}
\bibliography{refs-shared,references}

\clearpage
\appendix
\begin{center}{\Large\bf Appendix}\end{center}
\startcontents[appendix]
\printcontents[appendix]{}{1}{\setcounter{tocdepth}{2}}
\vspace{1em}

\section{The CreativePreferences dataset: inventory and audit}
\label{app:part-dataset}

This appendix documents the corpus: an inventory of every released dataset and an audit table that records, for each task, how the label was produced and what supports each comparison. The construction and balancing of each dataset follows in Appendix~\ref{app:balancing}.

\subsection{Full dataset inventory}
\label{app:inventory}

We release the corpus collection behind this paper as
\textsc{CreativePreferences}:  2.8 million preference-labeled texts carrying 317 million recorded preference acts (reviews, editorial decisions, votes, follows, citations, likes; 16 million of these are individually recorded decisions, reviews and votes, the rest are vote and engagement totals; two rows carry labels derived by an LLM judge and are counted as zero acts, eight rows record one decision per text, and the tweet-engagement and Reddit-score rows are lower bounds) (rows marked $^{\dagger}$, e.g. the New Yorker caption-contest tasks, are not included in any main-text table or average. For New Yorker, specifically, the task is multimodal, and the cartoon each caption answers is unavailable to every instrument)
across 44 datasets in seven fields, each carrying one of three preference
channels: an accountable \emph{gatekeeper verdict}, an editorial \emph{expert curation},
or an aggregate \emph{community/crowd} signal. Table~\ref{tab:datasets} in
the main text condenses Tables~\ref{tab:inventory}--\ref{tab:inventory-community}; this section gives the authoritative per-dataset
listing behind each task.

Every dataset obeys five construction disciplines. (i)~\emph{Institutional
labels}: the preference signal is the community's own recorded trace
(votes, acceptances, removals, prize lists, homepage placements), never a
judgment we commissioned; the benchmark measures preferences communities
actually enacted. (ii)~\emph{Matched rendering}: both label classes pass one
shared normalization pipeline, and a character-$n$-gram probe (grouped
out-of-fold) is run as a detector of pipeline artifacts before any instrument
reads the text; tasks whose classes were fetched by different systems
(e.g., research corpora paired with API recollection) must pass this check or
are rebuilt. (iii)~\emph{Anchored judging}: every scored batch of the articulated-criteria
judge carries blinded known-label anchor items, so the judge's validity
is certified per batch rather than assumed. (iv)~\emph{Stable splits}:
train/evaluation/test membership is a content hash, never a seeded shuffle,
so an item's split does not change as the corpus grows. (v)~\emph{Declared confounds}: each task
lists its known nuisance channels (length, era, venue, genre mix), which
feed the deconfounding analyses of \S\ref{sec:res-frontier} and
Appendix~\ref{app:dense-decay}.

Two dataset families require a more detailed description. The \emph{moderation}
family pairs research corpora of removed posts with our own era-matched
recollection of surviving posts from the same communities and windows: the
r/Jokes removal task contrasts moderator-removed jokes with kept jokes under
matched normalization, and the macro-norm tasks build on the Chandrasekharan
et al.\ removal corpus across nine subreddits, five humor-adjacent
(r/funny, r/Showerthoughts, r/nottheonion, r/me\_irl, r/tifu) and four
long-form writing communities (r/nosleep, r/books, r/asoiaf,
r/gameofthrones), with kept-side posts sampled uniformly over the
corpus's eleven-month window and bot/moderation notices stripped from both
classes. The moderation tasks are the benchmark's largest verdict tasks and the setting for
the per-community analyses of Appendix~\ref{app:subfields}. The
\emph{production-verdict} family captures decisions made about creative work
by institutions (a publisher's acceptance, a television show's cut-for-time
decision): small, costly tasks that test whether an item's text predicts a decision
made partly on information outside the text (Appendix~\ref{app:new-tasks}).

\begin{table}[!htbp]
\centering
\scriptsize
\setlength{\tabcolsep}{3.5pt}
\resizebox{\textwidth}{!}{%
\begin{tabular}{l>{\raggedright\arraybackslash}p{4.1cm}lr>{\raggedright\arraybackslash}p{5.6cm}}
\toprule
Field & Dataset & Status & $n$ & Notes (1s/0s; sub-corpora) \\
\midrule
Coding & GitHub PR merged (enriched) & inst. & 44{,}751 & 594 repositories; within-repository balance \\
 & Stack Overflow accepted answer & inst. & 16{,}001 & 5{,}972 / 10{,}029 \\
Math & mathlib PR merged & inst. & 8{,}494 & 7{,}509 / 985; canonical clean slice 7{,}956 \\
 & Math.SE accepted answer & inst. & 13{,}001 & 4{,}960 / 8{,}041; 4{,}960 questions \\
Law \& policy & Patent application granted & inst. & 579{,}084 & 290{,}198 / 288{,}886; nested claim-level corpus 59{,}937 \\
 & Comment received a response & inst. & 9{,}521 & 7{,}439 / 2{,}082; 1{,}814 dockets \\
Academia & ICLR accept/reject & inst. & 28{,}402 & 11{,}193 / 17{,}209; 6{,}030-row scored task \\
Journalism & Editorial pickup (press releases) & inst. & 72{,}315 & 1{,}478 / 70{,}837; 2{,}956-row scored task \\
Fiction & Book deals (RoyalRoad market pickup) & inst. & 1{,}274 & Kindle-Unlimited pickup 637/637; at chance on all instruments (App.~\ref{app:new-tasks}) \\
 & Kindle Scout publisher acceptance & inst. & 726 & 240 / 486 of 1{,}024 archived campaigns; excerpt-only context \\
 & Macro-norm removals (long-form subs) & inst. & 64{,}788 & r/nosleep $\cdot$ r/books $\cdot$ r/asoiaf $\cdot$ r/gameofthrones; 1:1 within-sub; era-uniform kept side \\
 & WritingPrompts mod-removals & live & 333{,}834 & removed-comment universe; 414 pre-removal texts recovered; 23{,}057 kept contrast; prospective collector running \\
Humor & r/Jokes mod-removals & inst. & 106{,}937 & removed-joke universe (184{,}774 kept); 18{,}879-row matched-normalized scored task \\
 & Macro-norm removals (humor subs) & inst. & 73{,}268 & r/funny $\cdot$ r/Showerthoughts $\cdot$ r/nottheonion $\cdot$ r/me\_irl $\cdot$ r/tifu; 1:1 within-sub; era-uniform kept side \\
 & SNL cut-for-time & inst. & 72 & 36 aired / 36 cut, ASR-matched pilot of 2{,}549 aired / 87 cut catalog \\
\bottomrule
\end{tabular}}
\caption{\textsc{CreativePreferences} inventory, \emph{gatekeeper verdict}
channel. Status: \emph{inst.} = V/A instruments scored; \emph{coll.} =
collected but not yet instrumented; \emph{live} = acquisition still running.
Counts are verified against the primary artifacts (ledger files, build
manifests, raw stores). Construction, filters, and confound handling per
dataset are in Appendix~\ref{app:balancing}.}
\label{tab:inventory}
\end{table}

\begin{table}[!htbp]
\centering
\scriptsize
\setlength{\tabcolsep}{3.5pt}
\resizebox{\textwidth}{!}{%
\begin{tabular}{l>{\raggedright\arraybackslash}p{4.1cm}lr>{\raggedright\arraybackslash}p{5.6cm}}
\toprule
Field & Dataset & Status & $n$ & Notes (1s/0s; sub-corpora) \\
\midrule
Coding & Competition editorial-approach match & inst. & 6{,}353 & AtCoder 2{,}495 $\cdot$ LeetCode 1{,}995 $\cdot$ CodeChef 995 $\cdot$ CF 868 \\
 & Stack Overflow bounty award & inst. & 23{,}498 & 7{,}103 / 16{,}395; within-question \\
 & Release-highlight picks & coll. & 4{,}654 & Rust 4{,}031-PR pool $\cdot$ mathlib 623 picks \\
Math & AoPS editorial approach & inst. & 25{,}454 & 17{,}560 / 7{,}894; 5{,}202-row scored task \\
 & Math.SE bounty award & inst. & 23{,}972 & 8{,}503 / 15{,}469; within-question \\
Law \& policy & Comment: change made & inst. & 7{,}084 & 2{,}824 / 4{,}260; second label agency-agreement 5{,}046 \\
Academia & Best-paper awards & inst. & 11{,}613 & 960 / 10{,}653 \\
 & Spotlight/oral pick & inst. & 7{,}941 & 2{,}016 / 5{,}925 \\
Journalism & Homepage placement & inst. & 183{,}708 & 91{,}854 / 91{,}854; 12{,}998-row scored task \\
Fiction & Curated-venue pick (Wigleaf Top 50) & inst. & 1{,}568 & 404 / 1{,}164 \\
 & Reedsy winner/shortlist & live & 115{,}670 & scrape at 66{,}189; badges so far 64 winner / 130 shortlist; 367 contests \\
 & RoyalRoad Community Magazine & inst. & 2{,}103 & 2{,}012 labeled / 26 ranked winners across 8 editions; pilot power \\
 & Reddit awards / best-of (WP) & coll. & 14{,}714 & gilded$+$awarded of 2{,}457{,}997 comments; r/bestofWP 822 picks \\
Humor & \#HashtagWars staff picks & inst. & 12{,}734 & 1{,}109 top-10/winner / 11{,}625; 4{,}228-row scored task \\
 & Style Invitational tiers & inst. & 9{,}637 & top-tier 1{,}549 (322 winner $+$ 1{,}227 runner-up); 316 weeks \\
 & Caption-contest finalists$^{\dagger}$ & inst. & 5{,}218 & 678 / 4{,}540; 227 contests \\
 & McSweeney's archive & coll. & 17{,}483 & pieces scraped; ``Best of'' anthology picks pending transcription \\
\bottomrule
\end{tabular}}
\caption{\textsc{CreativePreferences} inventory, \emph{expert curation}
channel (conventions as in Table~\ref{tab:inventory}).}
\label{tab:inventory-curation}
\end{table}

\begin{table}[!htbp]
\centering
\scriptsize
\setlength{\tabcolsep}{3.5pt}
\resizebox{\textwidth}{!}{%
\begin{tabular}{l>{\raggedright\arraybackslash}p{4.1cm}lr>{\raggedright\arraybackslash}p{5.6cm}}
\toprule
Field & Dataset & Status & $n$ & Notes (1s/0s; sub-corpora) \\
\midrule
Coding & Stack Overflow answer votes & inst. & 12{,}202 & 6{,}391 / 5{,}811; 137{,}994 ranked pairs collected \\
 & PR linked-issue reactions & coll. & 154{,}633 & 13 repositories \\
Math & Math.SE answer votes & inst. & 100{,}000 & 50{,}000 / 50{,}000; 11{,}629-row scored task \\
Law \& policy & Comment co-signing & inst. & 9{,}520 & 342 positive \\
 & Law SE answer votes & coll. & 4{,}192 & balanced; 72{,}369-post pool \\
Academia & Citation percentile & inst. & 2{,}387 & 1{,}176 / 1{,}211; from 27{,}233-paper collected corpus \\
Journalism & Tweet engagement & inst. & 46{,}456 & within outlet$\times$day median split; 784 groups; 31{,}129-row scored task \\
 & BBC most-read & inst. & 50{,}761 & 22{,}357 / 28{,}404; 2{,}251 days; criteria mining run to convergence \\
Fiction & Story upvotes (WritingPrompts) & inst. & \,96{,}080 & 48{,}040 / 48{,}040; 7{,}008-row dense-held-out scored population \\
 & RoyalRoad follower counts & inst. & 135{,}598 & listing universe; 3{,}604-row scored task from 10{,}611 random deep pages, within genre$\times$year strata \\
Humor & Joke upvotes (r/Jokes) & inst. & 383{,}786 & 191{,}893 / 191{,}893; 16{,}000-row scored task \\
 & Crowd-loved captions & inst. & 10{,}893 & within-contest vote split; 223 contests \\
\bottomrule
\end{tabular}}
\caption{\textsc{CreativePreferences} inventory, \emph{community/crowd}
channel (conventions as in Table~\ref{tab:inventory}).}
\label{tab:inventory-community}
\end{table}

One corpus is excluded from these tables by design: the twelve-domain
district-court outcome collection (facts-only opinions with win/loss labels).
A litigation outcome is a ruling on a dispute, not a preference expressed
about a creative product, so the district-court corpus does not fit the benchmark's
preference mapping; we retain it as an appendix-only resource and document
its construction alongside the others in Appendix~\ref{app:balancing}.

\begin{landscape}
\subsection{Audit table: what each task's label is, and what supports each comparison}
\label{app:audit}

Table~\ref{tab:audit} records, for each task of Table~\ref{tab:master-results}, the decision-maker, the candidate set, the information available at decision time, how the label was obtained, and the size of the scored population, so that the AUCs of Table~\ref{tab:master-results} can be audited. Provenance: O = the decision or event is observed directly; X = extracted from records; M = matched across corpora by our pipeline; A = model-assisted. $n$, positives and groups are the scored population behind the row's articulated model (the frame footnotes of Table~\ref{tab:master-results} apply). The gap interval is the paired grouped bootstrap of $VA{-}V$ (3,000 draws), or of $VAT{-}VA$ (2,000 draws) where that is the row's bakeoff, from Appendix~\ref{app:standard-cis}; n/r = no per-item scores on disk. Maturity: mature bank, floor harness (the $^{i}$ rows of Table~\ref{tab:master-results}), or withdrawn.

\scriptsize
\setlength{\tabcolsep}{3pt}
\begin{longtable}{p{2.6cm}p{2.3cm}p{2.5cm}p{2.7cm}cp{1cm}p{1cm}p{0.9cm}p{1.9cm}p{3.6cm}p{1.2cm}}
\caption{Audit table, one row per task of Table~\ref{tab:master-results}. Columns are defined in the text above.}\label{tab:audit}\\
\toprule
task & decision-maker & candidate set & information at decision time & prov. & $n$ & positives & groups & gap 95\% CI & label uncertainty / selection effects & maturity \\
\midrule
\endfirsthead
\toprule
task & decision-maker & candidate set & information at decision time & prov. & $n$ & positives & groups & gap 95\% CI & label uncertainty / selection effects & maturity \\
\midrule
\endhead
\bottomrule
\endfoot
GitHub PR merged & repository maintainers & pull requests opened on 594 repositories & diff, description, CI results, author identity, review thread & O & 11{,}452 & 9{,}205 & 255 & [+.020, +.073] & merge reflects maintainer priorities and timing as well as the patch & mature \\
Stack Overflow accepted answer & the question's asker & answers to the same question & all answers, votes at the time, asker's own problem & O & 16{,}001 & 5{,}972 & 5{,}972 & [+.070, +.085] & one person's choice; asker may accept early or never & mature \\
Competition editorial-approach & contest editorial (official solution) & accepted solutions per problem & problem statement, editorial & A & 999 & n/r & n/r & n/r & label = similarity of the solution to the editorial approach; model-assisted, not a gatekeeper verdict on the item & mature \\
Stack Overflow bounty award & the bounty poster & answers to the bountied question & all answers, votes, poster's own problem & O & 23{,}498 & 7{,}103 & 7{,}036 & [+.036, +.044] & one person's choice under a deadline & mature \\
Stack Overflow answer votes & site voters (aggregate) & answers to the same question & answers in vote order, accepted mark & O & 12{,}202 & 6{,}391 & 5{,}972 & n/r & vote order and acceptance bias exposure; threshold on score & mature \\
mathlib PR merged & mathlib maintainers & pull requests to mathlib & diff, CI build, reviewer comments & O & 7{,}956 & 7{,}502 & 31 & n/r & 94\% positives; full model unstable at this balance (u/q) & mature \\
Math.SE accepted answer & the question's asker & answers to the same question & all answers, votes, asker's problem & O & 2{,}600 & 992 & 992 & [+.023, +.062] & accepted $\wedge$ score$\ge$3 vs. rest; asker's choice & mature \\
AoPS editorial approach & AoPS wiki editors (reference solutions) & forum solutions per problem & problem, reference solutions & A & 5{,}202 & 3{,}503 & 606 & [+.006, +.027] & label = similarity to the wiki solution; model-assisted & mature \\
Math.SE bounty award & the bounty poster & answers to the bountied question & all answers, votes & O & 23{,}972 & 8{,}503 & 8{,}278 & [+.028, +.035] & one person's choice & mature \\
Math.SE answer votes & site voters (aggregate) & answers to the same question & answers in vote order & O & 2{,}326 & 1{,}166 & 1{,}140 & [+.026, +.061] & score threshold; exposure effects & mature \\
N\&C: received response & the rulemaking agency & public comments on one docket & comment text, docket, all other comments & M & 1{,}904 & 1{,}454 & 1{,}010 & [+.022, +.060] & response matched to comment by our pipeline; unmatched = not found, not ignored & mature \\
N\&C: rule change made & the rulemaking agency & comments the agency responded to & comment, docket, final-rule text & X & 1{,}417 & 638 & 692 & [+.000, +.043] & majority-vote label over matched responses & mature \\
N\&C: cosigner agreement & other commenters (co-signing organizations) & comments on one docket & comment text & X & 1{,}009 & 656 & 498 & [+.008, +.062] & cosigner counts extracted from letters & mature \\
Patent application granted & USPTO examiner & claims of one application & claims, specification, prior art & O & 59{,}937 & 36{,}022 & n/r & n/r & withdrawn: label-conditional candidate set fed every model & withdrawn \\
ICLR accept/reject & area chairs, from reviewer recommendations & submissions to one venue-year & full paper, reviews, rebuttal & O & 1{,}244 & 569 & 1{,}239 & [+.043, +.096] & reviewer noise: half the reviewers predict the decision at AUC.881 (split-half ceiling) & mature \\
Spotlight/oral pick & area chairs / program chairs & accepted papers of one venue-year & paper, reviews & O & 1{,}571 & 392 & 1{,}571 & [+.003, +.064] & quota-driven; venue conventions differ & mature \\
Citation percentile & the field (citing authors) & accepted papers of one venue-year & the paper, over time & X & 478 & 253 & 478 & [+.156, +.254] & citation counts measure attention; venue and year normalized & mature \\
Editorial pickup (press releases) & news editors at tracked outlets & press releases in the same period & release text; outlet's own beat & M & 605 & 303 & 45 & [+.051, +.115] & release$\leftrightarrow$article matching recall; k$\ge$3 outlets & mature \\
Homepage placement & outlet editors & articles live on the outlet's homepage snapshot & article, competing stories, news cycle & X & 12{,}998 & 6{,}507 & 1{,}229 & n/r & spatial position parsed from snapshots; outlet-held-out dense fails, story-grouped used & mature \\
BBC most-read & BBC readers (aggregate) & articles published the same day & headline and article in site context & X & 10{,}147 & n/r & 455 & n/r & list membership; exposure and placement confound & mature \\
Tweet engagement & readers (aggregate) & an outlet's tweets on one day & tweet text, outlet, timing & O & 31{,}129 & n/r & 508 & n/r & engagement threshold within outlet-day; follower base not controlled & mature \\
Kindle Scout publisher accept & Kindle Scout editors & nominated manuscripts in one campaign wave & excerpt, nominations & O & 726 & n/r & 3 & [+.105, +.142] & 60-row test leg; three campaign waves & mature \\
RoyalRoad market pickup & publishers (Amazon KU pickup, inferred from stubs) & serials on RoyalRoad & the serial's early chapters & M & 1{,}274 & 637 & 1{,}274 & n/r & pickup inferred from chapter removal; noise in $y{=}0$ & mature \\
Wigleaf Top 50 & a guest editor, from the series editor's longlist & longlisted flash fiction & the story & O & 1{,}568 & 404 & 1{,}568 & n/r & both classes expert-selected; 26 positives per year & mature \\
RoyalRoad Magazine pick & RoyalRoad staff & serials on the site & the serial & O & 2{,}012 & 26 & 8 & n/r & 1.3\% positives, eight editions & mature \\
RoyalRoad follower count & readers (aggregate) & serials within genre and year & the serial as listed & O & 3{,}604 & 1{,}888 & 149 & n/r & follower threshold within stratum & mature \\
Macro-norm removals (4 long-form subs) & subreddit moderators & posts to the same subreddit & post, subreddit rules & O & 64{,}788 & 32{,}394 & 4 & [$-$.024, +.151] & pooled over four communities; within-community fits differ (App.\ subfields) & mature \\
Story upvotes (WritingPrompts) & readers (aggregate) & stories written to the same prompt & story in thread order & O & 7{,}008 & 3{,}516 & 5{,}136 & [+.112, +.135] & vote threshold; posting-time exposure & mature \\
r/Jokes moderator removal & subreddit moderators & posts to r/Jokes & post, rules & O & 18{,}879 & n/r & 55 & [+.043, +.111] & enforcement inconsistency; 55 temporal strata & mature \\
Macro-norm removals (5 humor subs) & subreddit moderators & posts to the same subreddit & post, rules & O & 73{,}268 & 36{,}634 & 5 & [+.135, +.185] & pooled over five communities & mature \\
\#HashtagWars staff picks & the show's staff & submissions to one hashtag game & the tweet & O & 924 & 80 & 8 & [+.046, +.169] & 8 games; 9\% positives & mature \\
Joke upvotes & readers (aggregate) & jokes posted in the same period & the joke & O & 3{,}163 & 1{,}547 & 10 & [+.030, +.064] & vote threshold; topic-grouped & mature \\
Style Invitational tiers & the column's editor & entries to one week's contest & the entry & O & 8{,}063 & 1{,}518 & 316 & n/r & tiers: winner / runner-up / honorable mention & mature \\
SNL cut-for-time & the show's producers & sketches rehearsed for one episode & the sketch & O & 72 & 36 & 6 & n/r & n=72; underpowered & mature \\
\end{longtable}
\end{landscape}

\input{appendix_balancing.tex}

\section{Metric implementation and its checks}
\label{app:part-metrics}

This appendix describes how the criteria were implemented and checked: what a deterministic program can certify, the judge prompts and scoring code, the text each model reads, and whether the gaps hold when the judge family or the dense-model family changes.

\subsection{Executable verifiability: what a procedure can certify}
\label{app:vexec}

Two reviewers pointed out that ``checkable via code'' is not the same as verifiable. We therefore audited every $V$ column in the paper against a stricter definition: a feature is \textit{executable} if a deterministic procedure certifies a stated property of the item (i.e. the item compiles, passes tests, resolves to a reference, matches a reference answer, satisfies an arithmetic identity, or is confirmed by a lookup the item itself points to), and \textit{surface} if it is a statistic of the text (lengths, structure, lexical or readability counts). Of the $940$ $V$ columns behind Table~\ref{tab:master-results}, $17$ are executable, all in the GitHub pull-request task (tests run against the patch); $918$ are surface; $5$ are metadata or learned scores. We then built executable checks for every task where one was feasible with data on disk, and fit them under the same protocol as $V$ (grouped five-fold out-of-fold AUC, logistic head, no tuning). Table~\ref{tab:vexec} reports, per task, the share of items each check applies to and the AUC of the executable columns alone, of the existing $V$ bank, and of the two combined.

\begin{table}[!htbp]
\centering\scriptsize\setlength{\tabcolsep}{3pt}\renewcommand{\arraystretch}{1.15}
\begin{tabular}{@{}p{2.1cm}p{4.4cm}p{2.0cm}p{1.8cm}p{1.05cm}p{1.05cm}@{}}
\toprule
task & executable check & share of items the check applies to & AUC, executable columns alone & AUC, existing $V$ bank & $V$ + executable \\
\midrule
Code (GitHub PR merged) & tests pass after patch (17 cols); patch well-formed; added Python/JS/Go code parses (``compiles''); lint errors & 3.7\% (tests); 99\% (patch); 94\% (compiles) &.553 /.548 within-repository (tests);.528 (compiles, univariate) & -- & no change \\
Code (GitHub PR merged; deep) & added dependencies exist on PyPI/npm; bandit findings on added Python & 5.3\% (deps); 3\% high findings &.518 /.494 & -- & -- \\
Math (AoPS editorial approach) & final answer matches the reference key & 46.2\% &.572 (univariate.649) &.706 &.706 \\
Law (N\&C, three labels) & cited CFR/USC/FR sections exist; cited part matches the rule; docket id matches & 11--13\% cite anything &.44--.50 &.59--.60 &.59--.60 \\
Journalism (editorial pickup) & internal arithmetic identities; URLs resolve; dates and weekdays agree & 0.1\% / 20\% / 50\% &.462 &.611 &.607 \\
Journalism (editorial pickup) & stated revenue, income and EPS match the filer's SEC XBRL facts & 0.6\% &.458 &.611 &.610 \\
Law (N\&C, three labels; deep) & quoted NPRM text verbatim/fuzzy; requested change within the rule's scope; cosigner-count identity; period-length claim vs.\ the FR window & 18--19\% quote anything; 5\% cite CFR &.45--.50 &.59--.60 &.59--.60 \\
Journalism (editorial pickup; deep) & within-issuer restated facts; BLS/Census figures; event-calendar dates; ratio identities; Wikidata entity attributes & 21.5\% (union) &.465 &.611 &.603 \\
Math (mathlib PR merged; deep) & Lean syntax well-formed; no \texttt{sorry}/\texttt{admit}/debug commands; new files registered; deprecations dated & 95--100\% (syntax); 0.3\% (\texttt{sorry}) &.539 &.668 &.665 \\
Peer review (ICLR verdict; deep, full texts) & statcheck/GRIM; table identities; DOI, arXiv and repository links resolve; abstract numbers reappear & 0\% (statcheck), 8\% (tables), 30--90\% (links) &.508 &.600 (surface) &.597 \\
Math (Math.SE accepted / votes; deep) & displayed equalities hold under SymPy; LaTeX balanced; numeric claims consistent & see note &.512 /.511 &.588 /.572 &.588 /.570 \\
Patents (claim fell) & dependency references resolve and precede; no multiple-on-multiple dependency; antecedent basis; numbering contiguous & 82\% dependent claims &.702 (confounded; see text) & withdrawn & -- \\
Creative writing (r/WritingPrompts) & prompt constraints satisfied (word/sentence limits, required or forbidden tokens, lipograms, form, acrostics) & 0.4\% &.490 &.612 &.612 \\
Humor (\#HashtagWars) & referenced title exists (Wikidata) & 2.9\% &.507 &.555 &.559 \\
Humor (r/Jokes, captions) & exact duplicates (none found); caption rules (none stated) & 0\% & -- &.584 /.632 & -- \\
\bottomrule
\end{tabular}
\caption{\textbf{Executable checks per task.} Coverage is the share of items to which the check applies; AUCs are grouped five-fold out-of-fold under the $V$ protocol. ``Deep'' rows are the second, more thorough pass. The Math.SE row's coverage varies by check (see the project notes).}
\label{tab:vexec}
\end{table}

Table~\ref{tab:vexec} shows three things. (The deeper law and journalism checks, built after the first pass, change none of the three: coverage stays below a quarter of items and the AUC of the executable columns stays at chance.) First, executable checks exist in every field, but the share of items they apply to is small outside code and math: most public comments cite no legal authority, most press releases state no figure that any filing can confirm, and the figures that can be confirmed are segment or trailing-year numbers that a consolidated fact does not match. Second, where a check does apply and does carry signal (the AoPS final answer, univariate AUC $.649$), it is redundant with the surface bank for the label we study: whether an answer is right is already legible in how the solution is written. Third, the one field where structural checks scored well on their own, patents ($.702$), is confounded: the number of an application's claims sampled into our file predicts the label at $.721$ by itself, and the residual after controlling for that count is about $+.03$. We therefore read the verifiable share of Figure~\ref{fig:concept-row} as a surface share everywhere except code, and we do not claim that the executable tier would change any ordering in the paper. For published-article tasks (headline picks, most-read) a further selection effect applies: every published article has already passed editorial verification, so an executable check would find almost no variance to explain. The full audit, the per-check definitions.

\subsection{Judge prompts, rubric texts, and scoring code for Table~\ref{tab:topfeatures}}
\label{app:prompts}
\textbf{Scoring prompt (all articulated criteria).} Every $A$ feature is scored with
one prompt per (item, criterion), zero temperature, single-token answer
(\texttt{google/gemma-4-31b-it} for every task except the two departures listed in Appendix~\ref{app:models}):
\begin{quote}\small\ttfamily
You are an expert [domain] judge. You are given [the item] and ONE quality
criterion. Decide how strongly the item, on its own evidence, satisfies that
criterion. Answer with EXACTLY ONE token:\\
1.0 = clearly satisfies \quad 0.5 = partially / borderline \quad
0.0 = fails / cuts against \quad NA = no evidence bearing on it\\
CRITERION: [name] \quad DESCRIPTION: [text below]
\end{quote}
Scorer implementations: \texttt{datasets/va\_gemma\_banks/\allowbreak score\_va\_gemma\_banks.py}
(protocol source; every bank scorer imports its judge settings from it), \texttt{datasets/peer-review/vat\_3y/\allowbreak score\_va\_gemma\_3y.py}
(peer review), \texttt{datasets/humor/caption\_multiy/\allowbreak score\_va\_gemma\_captions.py}
(captions), and the per-task scorers under \texttt{datasets/*/} and \texttt{methods/taste\_decomposition/closure/*/} (Appendix~\ref{app:models}).

\textbf{Full rubric texts.} Table~\ref{tab:topfeatures} in the main text
lists, for each task, the three strongest verifiable checks (green) and the
three strongest articulated criteria (amber), each abbreviated to a line.
The judge of record never sees those abbreviations: it scores each item against the full rubric
text of each criterion. The complete rubric texts behind the articulated
entries of Table~\ref{tab:topfeatures} are:

\begin{itemize}\small
\item \emph{Change/commit/PR communication quality}: \texttt{Titles and messages clearly
summarize what and why, provide sufficient context/rationale, avoid narrating the
diff, and scale detail to the change; large CLs are annotated to aid review.}
\item \emph{Commit message structure and content}: \texttt{Accurate, descriptive commit
messages with conventional subject/body separation and structured footers (issue
references, explicit BREAKING CHANGE notes).}
\item \emph{Small, focused, reviewable changes}: \texttt{Keep changes well-scoped and
single-purpose; prefer small, incremental CLs that ease review, merging, and
rollback.}
\item \emph{Virality/shareability drivers}: \texttt{Features that drive sharing:
surprise/interestingness, high arousal, interactivity/variability, collective
dynamics, and visual salience.}
\item \emph{Laugh yield and impact}: \texttt{Frequency, consistency, and strength of
laughs; ability to elicit vivid, positive audience reactions (beyond tepid
amusement).}
\item \emph{Misdirection and reveal design}: \texttt{Primes an initial reading with fair
cues and conceals an alternate interpretation, then flips expectations with a
justified reveal.}
\item \emph{Contrast and juxtaposition to heighten effect}: \texttt{Create intensity and
meaning by deliberate contrasts (image, tone, circumstance); select striking
particulars and counterpoints to sharpen focus and feeling.}
\item \emph{Read-aloud test for flow and pacing}: \texttt{Reading the work aloud surfaces
stumbles, breathless lines, awkward rhythm, or dull patches.}
\item \emph{Reader momentum and page-turnability}: \texttt{Maintain forward motion toward
tangible goals; use micro-tension and curiosity devices; omit or compress
skippable material.}
\item \emph{Line-by-line justifiability and explicitness (gap-aware)}: \texttt{Each
inference is warranted and checkable; nontrivial facts are cited or proved;
assumptions used are explicit; any omitted steps are minor and clearly signaled
(no disguised jumps).}
\item \emph{Elegance and beauty of proofs}: \texttt{Value brevity with simplicity and
inevitability, aesthetic clarity, and generality --- arguments that capture core
ideas beautifully.}
\item \emph{Logical correctness, rigor, and completeness}: \texttt{Airtight logic with
every step justified, case splits exhaustive/disjoint, no hidden assumptions,
field-appropriate standards of rigor.}
\end{itemize}

\textbf{Verifiable checks.} The execution checks for code come from the
certified test-execution fleet (\texttt{datasets/code-review/pr\_test\_execution/}):
each patch is applied in a containerized environment and the repository's
held-out test suite executed before/after, yielding fail$\to$pass flips (996
certified), clean pass$\to$fail rows (6{,}557), and patch-application status;
the merge odds ratio is Mantel--Haenszel across repos. The generated checks
for creative writing and math are Python programs written by an LLM, one per criterion,
executed deterministically over the raw text
(\texttt{runs/validity\_full/v2/<task>/codegen\_exec\_results.jsonl}). Humor's
surface checks are the 16 deterministic caption features defined in
\texttt{score\_va\_gemma\_captions.py}.

The listing below gives the
complete program behind the creative-writing ``plot-arc shape proxy'' entry
(criterion: \emph{classical plot structure and causality}), as an example of what a generated check can and cannot measure. The program
operationalizes the construct with three surface proxies: length adequacy,
paragraph structure, and whether verb density (\texttt{-ed}/\texttt{-ing}
tokens) rises after the opening third. The example shows the defining property of the $V$
tier: the check is exactly replayable, and it captures only as much of the
construct it names as a program can compute. The program does not compute causality; the
$V\to A$ step measures the difference between what this program computes and
what the criterion means.

\begin{quote}\scriptsize\ttfamily
def score(text: str) -> float:\\
\hspace*{1em}n = max(1, len(re.findall(r"\textbackslash w+", t)))\\
\hspace*{1em}len\_score = math.tanh(n/500)~~\# length adequacy for plot\\
\hspace*{1em}paras = [p for p in t.split("\textbackslash n\textbackslash n") if p.strip]\\
\hspace*{1em}para\_score = math.tanh(len(paras)/8)~~\# paragraphing signals structure\\
\hspace*{1em}\# split into thirds; verb (-ed/-ing) density per third\\
\hspace*{1em}vcounts = [density(third) for third in thirds(t)]\\
\hspace*{1em}\# escalation: middle and end should carry the action\\
\hspace*{1em}esc = 0.8 if vcounts[1], vcounts[2] >= 0.9*vcounts[0] else 0.5\\
\hspace*{1em}return clip(0.4*len\_score + 0.3*para\_score + 0.3*esc, 0, 1)
\end{quote}

\subsection{Models used at every step}
\label{app:models}

This section lists every pipeline step that involves a language model, with the exact model, its serving configuration and the tasks it was applied to, as recorded in the run configurations and result ledgers.

\paragraph{Judge of record for the articulated criteria.} \texttt{google/\allowbreak gemma-4-31b-it}, bfloat16, served offline with vLLM, temperature $0$, at most six output tokens, one prompt per (item, criterion), answer read as one of the four tokens \texttt{1.0}, \texttt{0.5}, \texttt{0.0}, \texttt{NA} (prompt in Appendix~\ref{app:prompts}). Two weight snapshots were used: \texttt{3548789868c5} on the machine that scored most banks, and \texttt{518276fb130d} on a second machine for the RoyalRoad banks and the notice-and-comment prompt-optimization rounds. This judge scored the articulated bank of every row of Table~\ref{tab:master-results} and every ``$+$mined'' column; two rows were first scored by this judge on 2026-09-29 (next paragraph).

\paragraph{Two rows rescored to the same protocol.} Until 2026-09-29 two rows departed from this protocol. \emph{Editorial pickup (press releases)}: the base bank of $40$ criteria had been scored by \texttt{nvidia/\allowbreak Llama-3.3-70B-Instruct-FP8} with one joint JSON prompt covering all $40$ criteria (a score in $[0,1]$ plus an applicability flag, text capped at $3{,}600$ characters, $65\%$ of readings not applicable). We rescored the base bank with the judge of record, one prompt per (item, criterion), $4{,}000$ characters of text (the mined rounds' cap and about the full model's $1{,}024$-token budget): the not-applicable rate fell to $21\%$, four of $40$ criteria collapsed to a constant, and the scrambled-text anchor AUC was $.96$. Refit under the frozen protocol the row moved from $V$ .653 / $VA$ .681 / $+$mined .744 / $T$ .774 / $\mathcal{VAT}$ .771 to $V$ .653 / $VA$ .658 / $+$mined .749 / $T$ .774 / $\mathcal{VAT}$ .750; the full model $T$ is unchanged and the gap between $T$ and $VA$ widened. \emph{Competition editorial-approach}: this row had no language-model judge at all; its ``articulated'' bank was $139$ deterministic coded metrics. We built an articulated bank for this row by the procedure used for the AoPS and Stack Overflow banks: three label-blind proposer sessions (Claude Opus) each read $25$ accepted solutions, code only, from disjoint problem sets and proposed $20$ criteria; the $60$ proposals were merged and audited to $35$ (no criterion may refer to an editorial or reference solution, to the approach of others, or to length). Scored with the judge of record (code read as $4{,}000$ head $+$ $2{,}000$ tail characters; not-applicable rate $.23$, one collapsed criterion, scrambled-text anchor AUC $1.00$) and refit under the row's certified protocol (stratified group folds by problem, seeds 0--2), the row moved from $V$ .743 / $VA$ .754 / $T$ .724 / $\mathcal{VAT}$ .754 to $V$ .743 / $VA$ .734 [.687, .784] / $T$ .724 / $\mathcal{VAT}$ .737. The judged criteria add nothing beyond the programs on this row, and the bank's ensemble score anti-correlates with code length ($r = -.59$ with log characters): the criteria the judge could apply reward clean, short files. The competition row's label was produced by Claude Sonnet and its full model is ModernBERT-base (marked $^{m}$ in Table~\ref{tab:master-results}).

\paragraph{Labels produced by a model.} The AoPS editorial-approach label was assigned by \texttt{Qwen/\allowbreak Qwen3.5-122B-A10B-FP8} (snapshot \texttt{fb53b9f3}, thinking disabled). The same model, under the tag \texttt{qwen\_thinking\_fp8}, scored an earlier harness of nine tasks and the first Math.SE and mathlib banks; those scores are superseded, appear nowhere in Table~\ref{tab:master-results}, and were never pooled with Gemma scores.

\paragraph{Judge-family sweep (Appendix~\ref{app:family-robustness}).} Six judges, all with the frozen prompt and token scheme above: Llama-3.1-70B in bfloat16 (the run tag does not record whether the base or the instruct checkpoint was served); \texttt{Qwen2.5-14B-Instruct}; \texttt{phi-4}; \texttt{gemma-2-9b-it}; \texttt{Llama-3.1-8B-Instruct}; and \texttt{Mistral-7B-Instruct-v0.3}. Reasoning-tuned models were excluded because their thinking templates are incompatible with the one-token readout.

\paragraph{Aggregators for $\mathcal{V}$, $\mathcal{A}$ and $\mathcal{VA}$.} The aggregator $f$ of Section~\ref{sct:articulated_and_verifiable_metrics} is a transformer model trained over the matrix of criterion scores, fit out of fold within groups and selected on the evaluation split; Table~\ref{tab:master-results} reports its held-out AUC.

\paragraph{Dense and fused models.} The dense model $T$ is Llama-3.1-8B with LoRA (rank $16$, $\alpha{=}32$, learning rate $5{\times}10^{-5}$, batch $16$, $1{,}024$ tokens, two epochs, checkpoint chosen on the evaluation split) trained on the text alone. The fused model $\mathcal{VAT}$ candidates are: the same recipe trained on the text followed by a block of criterion scores (\texttt{VA metrics: <criterion>: <score>}); a rank mean of $T$ and $\mathcal{VA}$; an evaluation-weighted blend; the better parent; and a logistic stack fit on the evaluation split. The candidate is chosen on the evaluation split and reported on the test split, which is why the $\mathcal{VAT}$ column can be lower than $T$ on a task where the chosen candidate lost on test. Family checks: Qwen2.5-7B replaces the Llama base on two tasks; ModernBERT-base is the full model of the competition row.

\paragraph{Spurious-feature audit (Appendix~\ref{app:spurious-audit}).} The blind routing auditor that reads each mined criterion in a fresh session was Claude Sonnet for every campaign except three rounds (GLM-5.2 for one BBC round; \texttt{gpt-5.6} for one BBC round, one code round and one caption round). The arbiter that resolves disagreements with provenance visible was Claude Opus, with the same \texttt{gpt-5.6} exceptions. The planted-artifact validation (Table~\ref{tab:audit-bench}) used \texttt{gpt-5.6} through the Codex command-line client as a fresh blind adjudicator.

\paragraph{Criteria proposers in the mining campaigns.} Each mining round presented the same misclassified items to a fleet of independent proposers from up to three model families, whose proposals were pooled, merged and audited: two Claude Opus sessions (one with a fresh salt) and one Claude Sonnet session, run as sealed subagents; two or three \texttt{gpt-5.6} sessions through the Codex command-line client (reasoning effort high, read-only); and two GLM-5.2 sessions (thinking on) through the z.ai API. Table~\ref{tab:proposer-fleet} gives the composition per campaign. The exact Claude model versions were not recorded by the fleet runner, which logged them only as ``opus'' and ``sonnet''.

\begin{table}[!htbp]
\centering\footnotesize
\resizebox{\linewidth}{!}{%
\begin{tabular}{llll}
\toprule
campaign & proposing rounds & proposers per round & families \\
\midrule
joke upvotes & 5 & 8 (rounds 1--5); 5 in round 6, Claude slots missing & Claude, Codex, GLM \\
N\&C received a response & 4 & 4 (Opus, Sonnet, two Codex); GLM in round 2 only & Claude, Codex (GLM once) \\
editorial pickup (press) & 2 & 6 & Claude, Codex, GLM \\
BBC most-read & 6 & 8 (four Codex, four GLM) & Codex, GLM \\
spotlight/oral pick & 5 & 4 in rounds 1--2; 6 from round 3 & Claude, Codex, GLM \\
citation percentile & 6 & 4 in rounds 1--2; 6 from round 3 & Claude, Codex, GLM \\
\#HashtagWars staff picks & 4 & 4 & Claude, Codex \\
Style Invitational tiers & 2 & 4 & Claude, Codex \\
story upvotes & 7 & 4 & Claude, Codex \\
AoPS editorial approach & 1 & 8 & Claude, Codex, GLM \\
Math.SE accepted answer & 2 & 8 & Claude, Codex, GLM \\
Math.SE answer votes & 2 & 8 & Claude, Codex, GLM \\
RoyalRoad follower count & 2 & 6 (three Claude, three Codex) & Claude, Codex \\
N\&C rule change & 2 & 4 & Claude, Codex \\
GitHub PR merged & 0 & one decomposition round only (\texttt{gpt-5.6}) & Codex \\
\bottomrule
\end{tabular}}
\caption{Proposer fleet per mining campaign, from the header of each round's proposal file. A slot is counted when it returned proposals. Campaigns absent from the table ran no mining rounds (``n/c'' in Table~\ref{tab:master-results}).}
\label{tab:proposer-fleet}
\end{table}

\paragraph{Program writers for the verifiable checks.} The codegen programs behind the creative-writing and math checks (three programs per criterion: keyword, structure and holistic variants) were written by Claude Sonnet-class subagents from a fixed specification; the run recorded the output directory but not the model version. The GEPA refinement of coded programs used the Codex command-line client as its reflection model without pinning a model variant, so it ran on the client's default \texttt{gpt-5.6} model at the time. The execution-fleet checks for GitHub pull requests involve no language model.

\paragraph{Prompt optimization of the articulated criteria.} The fidelity objective and $\epsilon$-certification are the companion paper's \citep{companion_prompts}; that paper's configuration of record uses \texttt{Llama-3.1-8B-Instruct} as the judge under optimization, \texttt{Llama-3.3-70B-Instruct} as reviser, reconstructor, generator and grader, \texttt{Qwen2.5-72B-Instruct} for the acceptance check and Claude Sonnet 4.5 as the oracle, all through OpenRouter; the reconstruction channel uses Llama-3.1-8B-Instruct as executor and GLM-5 as reconstructor. The notice-and-comment bank was further optimized in five GEPA rounds with Claude Sonnet as proposer and fidelity reference against the Gemma-4-31B judge. The judge-specific optimization of Appendix~\ref{app:alt-posttraining} used \texttt{gpt-5.6} through Codex as proposer and decoder, with the eight judges named there.

\paragraph{Metric-discovery head-to-head (Appendix~\ref{app:metric-discovery}).} The paper's own loop, the adjusted AutoMetrics of Section~\ref{sct:mining}, is the proposer fleet above. The three published methods were rerun with one shared model, \texttt{Qwen/\allowbreak Qwen3-32B}, served with vLLM on one GPU (16{,}384-token context): thinking on for criterion generation and neuron naming (temperature $.7$), off for prediction and annotation (temperature $0$). HypotheSAEs used \texttt{BAAI/\allowbreak bge-large-en-v1.5} embeddings and top-$k$ sparse autoencoders of $64{\times}4$ and $256{\times}8$, both in our reimplementation and in the authors' package (version 2.0.1, default training settings). Every resulting bank was scored with the Gemma-4-31B judge of record. The RL-XAR-style rubric meta-optimization arm run on joke upvotes (Appendix~\ref{app:metric-discovery}) used the same \texttt{Qwen/\allowbreak Qwen3-32B} server for its rubric writer and meta-optimizer (thinking on) and its $0$--$10$ judge (thinking off).

\paragraph{Analysis-time readers (Appendices~\ref{app:subfields} and \ref{app:item-insights}).} The model that states what separates two item sets, writes the per-item notes and synthesizes them is \texttt{gpt-5.6} through the Codex command-line client (reasoning effort medium, web search disabled). The second reader is Claude Opus; its reads were produced in agent sessions whose model version was not recorded. Every candidate description was then scored on held-out items by the Gemma-4-31B judge of record, so the readers propose descriptions and the judge of record measures them.

\paragraph{Generators for the generated pools (Appendix~\ref{app:pools}).} \texttt{meta-llama/\allowbreak Llama-3.1-8B-Instruct}, \texttt{Qwen/\allowbreak Qwen2.5-14B-Instruct} and \texttt{microsoft/\allowbreak phi-4}, one sample per model per condition under six prompting conditions (five at temperature $1.0$, one at $1.3$; top-$p$ $.95$), with per-item seeds fixed in advance. The judge's family (Gemma) was excluded as a generator, and the readouts that drop the Llama family exclude the dense model's family as well. The articulated selector is the frozen bank recipe with criteria scored by Gemma-4-31B; the dense selector is the task's Llama-3.1-8B LoRA reward model.

\paragraph{Post-training appendix (Appendix~\ref{app:alt-posttraining}).} Two base-versus-instruct pairs, read at temperature $0$ from first-token log-probabilities: \texttt{Llama-3.1-8B} with \texttt{Llama-3.1-8B-Instruct}, and \texttt{gemma-3-12b-pt} with \texttt{gemma-3-12b-it}. The judge-specific optimization additionally used \texttt{gemma-4-31b-it}, \texttt{Qwen2.5-32B-Instruct}, \texttt{phi-4} and \texttt{Mistral-Small-24B-Instruct-2501}.

\paragraph{What was not recorded.} No run pinned an exact version for the Claude models (proposers, auditor, arbiter, program writers, reader), the Codex client's default model at the time of the GEPA program refinement, or the checkpoint variant of the 70B judge in the family sweep. We report these models as the runs logged them.

\subsection{What each model was allowed to read}
\label{app:text-budget}

If the articulated judge read less of each item than the full model, part of the gap could reflect information access rather than tacit preference. We audited the text budgets of both models from the run configurations for the tasks below. In every audited task the judge reads at least as much of the item as the full model, and in the two long-text tasks the judge reads more. The remaining tasks use short items (abstracts, comments, answers) that fit within both budgets. Where the judge reads a head-and-tail window and the full model a right-truncated prefix, the two models see different windows of the same item; we have not measured the effect of this difference.

\begin{table}[!htbp]
\centering\footnotesize
\resizebox{\textwidth}{!}{%
\begin{tabular}{p{.30\textwidth}p{.32\textwidth}p{.30\textwidth}}
\toprule
task & articulated judge (Gemma-4-31B) & full model (Llama-3.1-8B LoRA) \\
\midrule
creative writing, story upvotes (and other story tasks) & $3{,}600$ head $+$ $2{,}400$ tail characters ($\approx 1{,}400$ tokens) & first $1{,}024$ tokens ($\approx 4{,}000$ characters) \\
code, GitHub PR merged (v3, enriched record) & full record (median $6{,}801$ tokens) & first $4{,}096$ tokens \\
peer review (abstract tasks) & full abstract and metadata & full abstract and metadata \\
press releases, editorial pickup & first $4{,}000$ characters ($\approx 1{,}000$ tokens; was $3{,}600$ under the retired Llama scoring) & first $1{,}024$ tokens ($\approx 4{,}000$ characters) \\
\bottomrule
\end{tabular}}
\caption{Text budget per model for the audited tasks, from the run configurations. Columns: what the articulated judge (Gemma-4-31B) and the full model (Llama-3.1-8B LoRA) read of each item.}
\label{tab:text-budget}
\end{table}

\subsection{Robustness of the gap estimates across model families}
\label{app:family-robustness}

The instruments behind our gap estimates each involve a model choice: the
articulated bank is scored by a judge model, and the dense bound $T$ is a
trained reward model with a particular base. We test whether the gaps
hold when the model \emph{family} on each side is swapped, with everything
else (rows, banks, prompts, splits, estimators, and folds) held byte-for-byte
identical.

\paragraph{Judge-family sweep.} We rescored two complete tasks (math.SE bounty
and Stack Overflow bounty; ${\sim}1.7$M judgments per judge) with six
additional open-weight judges spanning five model families beyond the
Gemma-4-31B judge of record, then refit the identical out-of-fold ladder on
each judge's matrix. The coded $V$ models never consult the judge and reproduce
identically under every judge ($.691$ math / $.738$ SO), which certifies that the
$A$ and $VA$ columns of Table~\ref{tab:judge-sweep} differ only in the judge.

\begin{table}[!htbp]
\centering
\begin{tabular}{lcccc}
\toprule
 & \multicolumn{2}{c}{math.SE bounty} & \multicolumn{2}{c}{SO bounty} \\
Judge & $A$ & $VA$ & $A$ & $VA$ \\
\midrule
Gemma-4-31B (record) &.696 & \textbf{.723} &.753 & \textbf{.779} \\
Llama-3.1-70B &.686 &.711 &.745 &.772 \\
Qwen2.5-14B &.680 &.708 &.736 &.766 \\
phi-4 (14B) &.672 &.708 &.726 &.766 \\
gemma-2-9b &.678 &.708 &.738 &.765 \\
Llama-3.1-8B &.659 &.705 &.722 &.757 \\
Mistral-7B-v0.3 &.626 &.700 &.675 &.748 \\
\bottomrule
\end{tabular}
\caption{Judge-family sweep on two bounty tasks. Rows: judge model; columns: held-out AUC of the articulated bank alone ($A$) and with the verifiable features ($VA$), on math.SE bounty and Stack Overflow bounty, with the same criteria and prompts for every judge.}
\label{tab:judge-sweep}
\end{table}

Table~\ref{tab:judge-sweep-cw} reports the same sweep on a third task, creative-writing story upvotes ($7{,}008$ items in the dense-held-out population; the same $45$ criteria and prompts for every judge). Five judges from five families completed the run (the weights of a sixth, Gemma-2-9B, were not available on the scoring machine). $VA$ ranges from $.619$ (Llama-3.1-8B) to $.660$ (Gemma-4-31B), a spread of $.041$, and the order follows judge capability; the Gemma-4 rerun reproduces the paper's $VA$ for this task ($.650$) within $.01$. The verifiable features alone give $.612$ for every judge, so the weakest judge adds almost nothing beyond $V$ and the strongest adds $.048$. All five judges order the anchors correctly (known positives above known negatives above scrambled text; coherent-vs-scrambled AUC $.97$--$1.00$), and no criterion collapses to a constant.

\begin{table}[!htbp]
\centering\footnotesize
\resizebox{\textwidth}{!}{%
\begin{tabular}{lcccc}
\toprule
Judge & family & $A$ & $VA$ & anchors (pos vs.\ neg / coherent vs.\ scrambled) \\
\midrule
Gemma-4-31B (record rerun) & Gemma &.648 &.660 &.57 / 1.00 \\
phi-4 (14B) & Phi &.634 &.647 &.66 /.98 \\
Qwen-2.5-14B-Instruct & Qwen &.632 &.640 &.71 / 1.00 \\
Mistral-7B-Instruct-v0.3 & Mistral &.592 &.630 &.66 /.97 \\
Llama-3.1-8B-Instruct & Llama &.560 &.619 &.57 /.99 \\
\bottomrule
\end{tabular}}
\caption{Judge-family sweep on creative-writing story upvotes ($7{,}008$ items, $45$ criteria, identical prompts). Columns: judge; model family; grouped out-of-fold AUC of the articulated bank alone ($A$, best of linear and gradient-boosted heads) and with the verifiable features ($VA$; $V$ alone is $.612$); anchor-battery AUCs (50 known positives vs.\ 50 known negatives, and coherent vs.\ scrambled text).}
\label{tab:judge-sweep-cw}
\end{table}

Two further tasks extend the sweep to full banks that include every mined round, scored under each judge (Table~\ref{tab:judge-sweep-jokes-nc}). On joke upvotes the three alternative judges lie within $.020$ of one another and $.05$--$.07$ below the judge of record; the bank was mined with that judge, so part of this margin may be adaptation of the mined criteria to the judge of record rather than judge capability, and we report the margin as an upper limit on the judge-capability effect. On notice-and-comment responses the judges lie within $.015$ of one another and of the record, and the ordering does not follow capability; the verifiable features carry most of the signal on this task, so the choice of judge changes $VA$ little.

\begin{table}[!htbp]
\centering\footnotesize
\resizebox{\textwidth}{!}{%
\begin{tabular}{llcccc}
\toprule
 & & \multicolumn{2}{c}{joke upvotes ($16{,}000$)} & \multicolumn{2}{c}{N\&C responded ($9{,}521$)} \\
Judge & family & $A$ & $VA$ & $A$ & $VA$ \\
\midrule
Gemma-4-31B (record) & Gemma &.748 &.752 &.672 &.733 \\
Gemma-4-31B (re-run, control) & Gemma &.749 &.752 &.670 &.736 \\
Qwen-2.5-14B-Instruct & Qwen &.687 &.699 &.631 &.718 \\
phi-4 (14B) & Phi &.671 &.687 &.644 &.719 \\
Llama-3.1-8B-Instruct & Llama &.662 &.679 &.669 &.732 \\
\bottomrule
\end{tabular}}
\caption{Judge-family sweep on two further tasks, full banks (base criteria plus every mined round), scored under each judge with identical prompts. The control row re-scores both full banks with the judge of record on a second machine and a second run; it reproduces the record within $.003$, so the drops below it are judge effects, not run effects. Columns: grouped out-of-fold AUC of the articulated bank alone ($A$) and with the verifiable features ($VA$), gradient-boosted head, on the full population of each task (joke upvotes: 50 topic groups; notice-and-comment responses: 1,814 dockets).}
\label{tab:judge-sweep-jokes-nc}
\end{table}

\paragraph{Quantization of the judge's answer.} Each criterion is read as one of four tokens ($1$, $.5$, $0$, not applicable). A reviewer asked whether this four-token readout discards information the judge has. We rescored math.SE bounty with the same judge, rows, criteria and prompts, and replaced the argmax token with the expected score under the judge's top-20 log-probabilities over the three scored options (not applicable when those options carry under $5\%$ of the mass). The two readouts agree per criterion at a median Spearman $\rho$ of $.81$ across $31$ criteria, and the soft readout changes the grouped out-of-fold AUC by $+.003$ for the articulated bank alone ($.692 \to.696$) and by $+.004$ for $VA$ ($.720 \to.724$). The not-applicable rate is the same under both ($.22$ vs.\ $.20$). Quantizing the judge's answer to four tokens therefore costs at most $.004$ AUC on this task.

The sweep shows three regularities. \emph{First, the articulated recovery varies little across judge families}:
the total $VA$ spread across six families is $.023$ AUC (math) and $.031$
(SO), and every alternative judge lands \emph{below} the record, so the
judge of record is the strongest available articulated bound and every swap
\emph{widens} the reported gap $T-(VA)$; the gaps are not an artifact of a
weak judge of record. \emph{Second, the judges order by capability, not
by family}: within the Llama family alone $VA$ climbs
$.705\to.711$ / $.757\to.772$ from 8B to 70B, the two 14B-tier judges
from different families (Qwen2.5, phi-4) tie exactly, and the weakest
recovery belongs to the smallest, oldest judge (Mistral-7B). \emph{Third,
criterion-level implementations differ by judge even where the bank-level AUC
agrees}: per-criterion cross-judge rank correlations against the record have
median $\rho=.16$--$.45$ (math) and $.18$--$.59$ (SO), rising monotonically
with judge capability (Llama-3.1-70B tops both, with bank-mean agreement
$\rho=.81$ on SO), while the same criteria draw ``no evidence'' verdicts
under all judges. The sweep reproduces, on this paper's own tasks, the
companion paper's judge-ladder finding: \emph{which} words carry the signal
depends on the judge; \emph{how much} signal the bank recovers does not, and judges
converge toward the record's criterion profile as they grow larger. Translated to
Figure~\ref{fig:articulated-share}, the articulable-share estimates for these
tasks shift by four to seven points under matched-capability judges and by at
most roughly ten under the weakest (7B) judge; cross-task orderings are
unchanged. Reasoning-tuned models (Qwen3, R1 distills, GLM-Z1) are excluded
by design: their thinking templates are incompatible with the frozen
single-token readout.

A possible objection is that all judges above used the
\emph{same} frozen scoring template, which was developed with the
record judge in the loop, so the record's top position could reflect
the template's fit to the record judge rather than judge quality. We tested this objection directly. Holding
the criterion texts frozen (they are the construct, and were authored
independently of any judge), we let the strongest challenger
(Llama-3.1-70B) and the weakest (Mistral-7B) each \emph{select its own}
scoring template from five candidates by label-blind fidelity (anchor
separation and answer rate on a small probe), then rescored both
tasks in full under each judge's chosen template. On math, both judges
independently preferred our default template; on SO, both preferred a
definition-first variant that reduced their ``no evidence'' rates.
Yet $VA$ changed by at most $.003$ under the chosen templates, never
upward (Llama-70B $.711\to.710$ / $.772\to.769$; Mistral-7B $.700\to.700$ /
$.748\to.746$). The bank-level $VA$ is insensitive to the scoring
template, so the record judge's advantage reflects judge capability, not
template fit.

\paragraph{An empirical bound on the gain from judge choice: the seven-family panel.}
Rather than extrapolating to an ``ideal judge,'' we measure the
ceiling of judge choice directly: an ensemble that stacks \emph{all seven} judges'
criterion-score matrices into one articulated model (three to four times the
features of any single judge) is a strictly larger instrument than any judge
choice. The seven-judge ensemble beats the record judge by $+.004$ (math, $.722\to.726$) and
$+.005$ (SO, $.779\to.784$); the per-criterion \emph{mean} across judges is
slightly worse than the record alone. The remaining gain from judge choice for the
articulated model is therefore about half a point of AUC on these tasks,
an order of magnitude smaller than the gaps the paper reports.

\paragraph{Dense-family swap.} On the other side of the gap, every dense bound
$T$ in this paper is a Llama-3.1-8B LoRA reward model. We retrained the dense
model from a Qwen2.5-7B base (the closest capacity-matched base in a second
family) on two tasks (math.SE bounty and Stack Overflow accepted answer),
with the frozen recipe, identical rows and splits, and three seeds per task,
selecting on the eval split and reporting test exactly as everywhere else.
The dense bound $T$ barely moves: math.SE bounty $T =.733$ (Llama) vs.\ $.726$
(Qwen, $\Delta{=}{-}.007$); SO accepted $T =.743$ vs.\ $.738$
($\Delta{=}{-}.005$), with per-seed test spreads of ${\sim}.01$ in each
family. The gap numerator is therefore not an artifact of the Llama
base: a second family recovers the same dense signal to within seed
noise, and since the Llama model is the (slightly) stronger bound on both
tasks, keeping it as the model of record is the conservative choice: a
weaker $T$ would only shrink the reported gaps. Together, the two swaps
test both instruments behind the gap estimate on the same tasks: $VA$
moves by at most $.031$ across judges and $T$ by at most
$.007$ across dense families, both within the confidence intervals already reported in
Table~\ref{tab:master-results}.

\section{The missing-value bound}
\label{app:part-bound}

This appendix derives the bound of Eq.~\ref{eq:whisker}, backtests it on the sealed mining campaigns, and compares our mined banks with published criterion-discovery methods.

\subsection{Derivation and extensions}
\label{app:missing-mass-derivation}

This section derives the missing-value bound of Eq.~\ref{eq:whisker}. We build the bound in three steps: the base construction under a single proposal distribution, its extension to multiple decorrelated proposers, and the rare-event analysis that supplies the bound's second term.

\paragraph{Setup:} Let $c \in \mathcal{C}$ be the countable universe of criteria our mining process discovers, treated as the \emph{species} \citep{chao1984nonparametric, bunge1993estimating}, and let $q$ be the proposal distribution over $\mathcal{C}$ induced by the mining loop (\S\ref{sct:mining}). Each round draws $N$ proposals from $q$, and after the final round $D \subseteq \mathcal{C}$ are the discovered species and $U = \mathcal{C} \setminus D$ are the undiscovered ones; the \emph{missing mass} is $\mu = q(U)$. The Good--Turing estimate of $\mu$ is
\begin{equation}
\hat\mu \;=\; f_1 / N,
\label{eq:goodturing}
\end{equation}
where $f_1$ is the number of the final round's proposals that were
first-time discoveries \citep{good1953population, efron1976estimating}.

To estimate missing \emph{value} from missing mass, we estimate each $c$'s marginal value $v_c \ge 0$ (i.e. the AUC it would add to the articulated model). Because bank criteria can carry redundant signal, the realized gain of any species \emph{set} is subadditive, so the sum $\sum_{c \in S} v_c$ upper-bounds the realized gain of $S$. $G = VA_{new} - VA$ is the campaign's actual mining gain, from the discovered mass $q(D)= 1 - \mu$.

\textbf{Salience-ordered mining} We assume that our mining process surfaces salient, high-value criteria first, so the value carried per unit of proposal mass is non-increasing in discovery order:
\begin{equation}
\frac{1}{q(U)}\sum_{c \in U} v_c \;\le\; \frac{G}{q(D)}.
\label{eq:salience}
\end{equation}

\textbf{Bound.} Rearranging Eq.~\ref{eq:salience} and substituting
Eq.~\ref{eq:goodturing},
\begin{equation}
\underbrace{\textstyle\sum_{c \in U} v_c}_{\text{unmined value}}
\;\le\; G \cdot \frac{\mu}{1-\mu}
\;\approx\; \big(VA_{new} - VA\big)\cdot
\frac{\hat\mu}{1 - \hat\mu},
\label{eq:whisker-base}
\end{equation}
This base bound is the first term of Eq.~\ref{eq:whisker}, and it has two properties. (1) The bound is \emph{wide where mining has not converged} and lower bounds $\gamma(X)$. (2) New criteria \emph{names} never run out ($\hat\mu$ stays at $.4$--$.7$ even in terminal rounds), but \textit{informative} criteria do, so Eq.~\ref{eq:whisker-base} overstates the unmined value where new names outpace new value. We validate the
construction by holding out each multi-round campaign at round $k$, computing the bound from rounds ${\le}k$, and comparing it with the gain actually realized afterward; the bound exceeds the realized gain, with substantial slack, from mid-campaign onward (Appendix~\ref{app:whisker-backtest}).

\paragraph{Frozen sampler or evolving process.} Good--Turing assumes the proposals are draws from one distribution. Ours are not: each mining round is shown what earlier rounds found and is directed at the residual, so the proposal distribution changes across rounds. We therefore interpret $\hat\mu$ as the missing mass of the \textit{final} round's proposal distribution (i.e. the process that would generate the next round), and make no claim about a stationary population of criteria. The per-round series for BBC most-read is $.45,.41,.49,.51,.49,.61$ over six rounds; the series rises at the end because the residual-directed proposer begins producing criteria the bank does not absorb. The cross-proposer recapture below is the complementary check: a second, decorrelated proposer that rediscovers our species is evidence that the missing mass is small regardless of how the sampler evolved.

\paragraph{Extension: multiple proposers.} The bound above conditions on a single proposal distribution $q$, and its salience assumption fails exactly for a criterion that is valuable but rarely proposed \emph{by our sampler}. The construction extends to $J$ deliberately decorrelated proposer families $q_1, \dots, q_J$ (different base models, personas, taxonomy-guided enumeration, prompts that ban the existing bank). Discoveries pool into $D = \bigcup_j D_j$, and the missing mass of the mixture, $\mu_{\cup}$, is estimated from the overlap pattern between families in the capture--recapture manner \citep{chao1984nonparametric, boehme2018stads}: how often family $j$ rediscovers what family $j'$ already found measures how much of the space each family misses. Substituting $\mu_{\cup}$ for $\mu$ in Eq.~\ref{eq:whisker-base} tightens the whisker, but the more important change is to the salience assumption: salience ordering now has to fail for every family at once, so an escaping criterion must lie in the intersection of the $J$ regions that the families miss, and the overlap data estimates the size of that intersection rather than assuming it is empty. The design also adds a check the single-sampler construction cannot run: each family's bound, computed alone, must cover the value the other families in fact discovered.

\paragraph{Rare events.} No sample-based bound can detect a criterion whose proposal mass is arbitrarily small: an atom with mass $\varepsilon$ and large value $v^*$ appears in no finite campaign, contributes nothing to $\hat\mu$, and violates Eq.~\ref{eq:salience} directly. Two statements bound this failure mode. On the mass axis, a \emph{detection-resolution} bound: a criterion with proposal mass at least $\varepsilon$ survives $NRJ$ draws unproposed with probability at most $e^{-\varepsilon NRJ}$ (for $R$ rounds of $N$ proposals across $J$ families), so at confidence $1-\delta$ every criterion still undiscovered has mass below $\varepsilon_{res} = \ln(1/\delta)/(NRJ)$. Our loop re-weights $q$ toward unexplained examples each round, which accelerates discovery in practice but makes this rate a calibration rather than a theorem. On the value axis, the budget $s \cdot v_{\mathrm{cap}}$ of Eq.~\ref{eq:whisker}: at most $s$ below-resolution criteria violate salience ordering, each worth at most $v_{\mathrm{cap}}$. We do not choose $s$ and $v_{\mathrm{cap}}$; we invert them. Setting $s = 1$ and solving for the value that closes a task's remaining gap gives
\begin{equation}
k^* \;=\; \frac{\big(VAT - VA_{new}\big) \;-\; G \cdot \hat\mu/(1-\hat\mu)}{\max_{c \in D} v_c},
\label{eq:kstar}
\end{equation}
the multiple of the best criterion ever discovered that a single never-proposed criterion would need to reach (while also sitting below $\varepsilon_{res}$) for the task's tacitness reading to be wrong. We cannot measure $\max_{c \in D} v_c$ directly, since criteria enter the bank in rounds; a round's gain is at least the marginal value of any one criterion in it, so putting the largest single-round gain in the denominator understates $k^*$. Under that substitution, with the full model of Table~\ref{tab:master-results} and the committed whisker of Table~\ref{tab:whisker-coverage}, four campaigns need no undiscovered criterion at all: on press pickup, BBC most-read, spotlight picks and \#HashtagWars the bank plus its whisker already reaches or exceeds the full model, which is why those rows carry the $h$ mark in Table~\ref{tab:master-results} and are not read as tacit. On the four whose gap survives the whisker, $k^*$ is at least $1.0$ on joke upvotes, $5.4$ on N\&C responses, $6.4$ on citations and $22$ on story upvotes: a single undiscovered criterion would have to outweigh the best round of discovery by that factor, while also having escaped every proposal round. The bound is not assumption-free. Each assumption has an empirical check: the truncation backtest for salience ordering (Appendix~\ref{app:whisker-backtest}), the cross-family overlap check for regions the sampler misses, and $k^*$ for the rare-event tail.

\subsection{Backtesting the missing-value bound}
\label{app:whisker-backtest}

The whiskers on the $VA$ bars (Appendix Figure~\ref{fig:vat-community})
are \emph{upper bounds} on the articulated value still unmined (the salience-ordered-tail term of Eq.~\ref{eq:whisker}):
$\mathrm{whisker} = (VA_{new}-VA) \times \hat\mu/(1-\hat\mu)$,
where $\hat\mu$ is the Good--Turing missing mass of real-criteria species in the
final mining round (fraction of that round's proposals that were first-time
discoveries). The construction assumes miners surface salient, high-value
criteria first, so value per unit of discovered mass is non-increasing; the
linear extrapolation then bounds the remaining value from above.

\begin{figure}[!htbp]
 \centering
 \includegraphics[width=\textwidth]{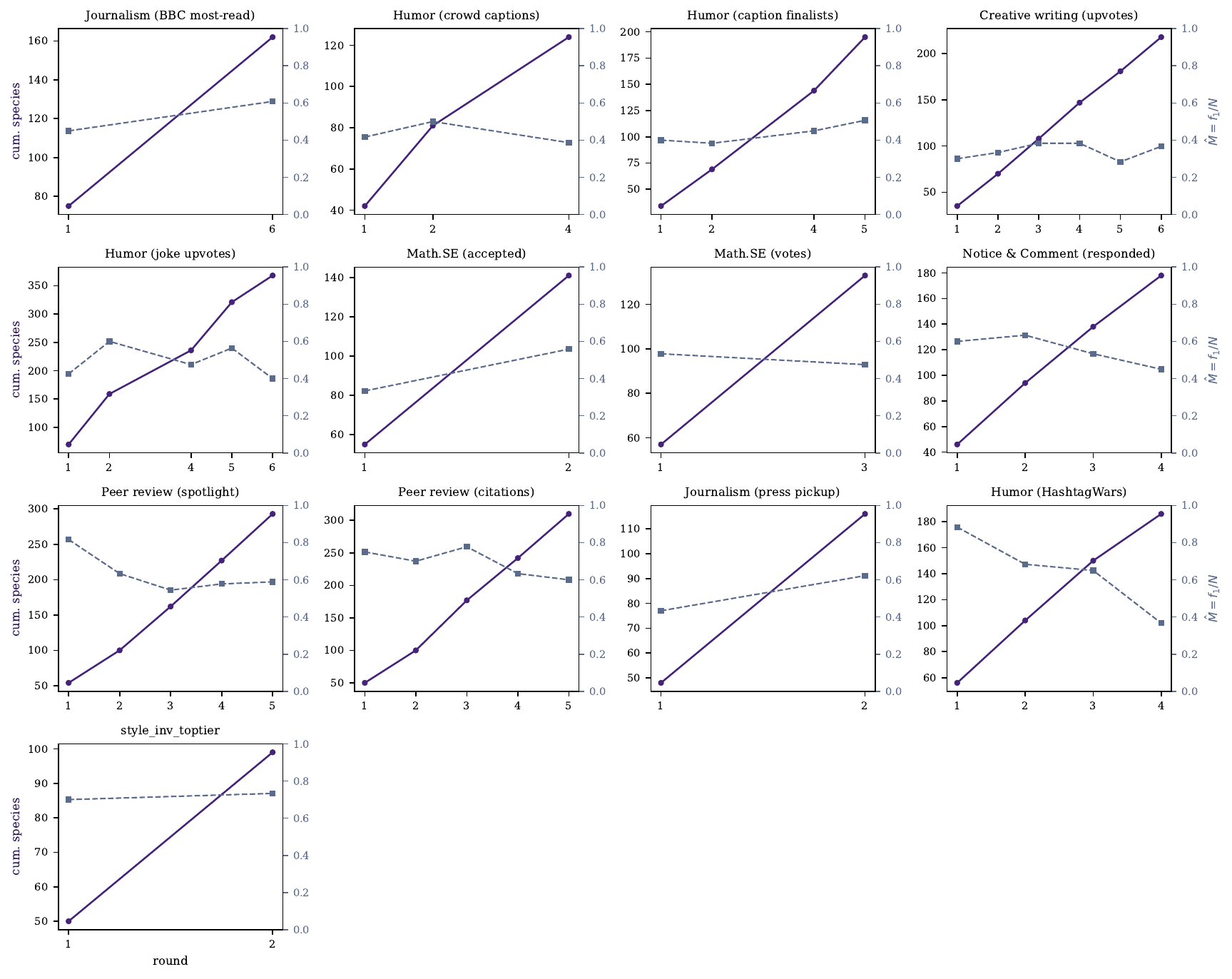}
 \caption{\textbf{Species accumulation and Good--Turing missing mass by
 mining round}, Track-A (quality) criteria, for every multi-round closure
 campaign. Orange: cumulative distinct criteria species discovered (each
 round's two-judge-merged vocabulary; cross-round deduplication happens at
 routing, so the accumulation slightly overcounts). Slate, dashed: the
 per-round missing mass $\hat\mu = f_1/N$ (fraction of that round's proposals
 that were first-time species). The figure shows the contrast that motivates
 the value-weighted whisker construction below: the species vocabulary keeps
 growing ($\hat\mu$ stays at $.4$--$.7$ even in final rounds; new
 \emph{names} never run out) while the held-out value of each round's
 additions saturates (the stopping rule that ended each campaign).
 }
 \label{fig:missing-mass-curves}
\end{figure}

We validated the two halves of this construction separately. The \emph{mass}
estimate was checked by a fresh proposal fleet whose new-species rate matched
the predicted missing mass, by a round-ahead backtest, and by
leave-one-proposer-family-out recovery. We backtest the \emph{value} translation
directly by truncation holdout: cut each multi-round campaign at
round $k$, compute the bound from rounds $\le k$ only, and compare it with the
gain actually realized in the remaining rounds.

\begin{table}[!htbp]
\centering\footnotesize
\begin{tabular}{llrrrl}
\toprule
campaign & truncated after & bound & realized after & used & holds \\
\midrule
BBC most-read & round 1 &.007 &.018 & 255\% & \textbf{no} \\
BBC most-read & round 2 &.010 &.011 & 104\% & borderline \\
BBC most-read & round 3 &.020 &.005 & 24\% & yes \\
BBC most-read & round 4 &.029 & $-$.002 & 0\% & yes \\
r/Jokes upvotes & round 1 &.012 &.010 & 86\% & yes \\
r/Jokes upvotes & round 2 &.039 &.001 & 2\% & yes \\
r/Jokes upvotes & round 4 &.025 & $-$.001 & 0\% & yes \\
\bottomrule
\end{tabular}
\caption{Truncation backtest of the missing-value bound on the two campaigns
with full per-round records (Table~\ref{tab:whisker-backtest-expanded} extends it to seven campaigns) (each evaluated in its own campaign frame). The
bound fails only at the earliest truncations, where the observed gain is a
single noisy round; from mid-campaign onward the bound holds at every truncation with
substantial slack. Figure whiskers are computed at the \emph{final} round of
terminal campaigns, the regime in which the backtest validates the bound.}
\label{tab:whisker-backtest}
\end{table}

Quality-criteria ($M_{quality}$) missing mass does
\emph{not} fall across rounds (BBC: $.45,.41,.49,.51,.49$). The space of
plausible criteria has a long tail of rare species, so each new proposal
round yields first-time discoveries at a roughly constant rate even as the
\emph{value} of those discoveries decays to zero; additional mining rounds
therefore tighten the bound by extending the flat-gain record rather than by
exhausting the mass. A calibrated point estimate (the realized future gain in
the terminal regime, roughly $0$--$.005$ in the backtest) is far smaller than
the bound itself; we nonetheless report the bound because the bound, not the point estimate, has
holdout support, and the width of the bound indicates how far a mining
campaign has converged.

\paragraph{Full coverage: the bound on every eligible task.} The
whisker requires a sealed mining campaign (the Good--Turing input) and a
mined increment in the unified frame; of the twelve master-table tasks with
a mined column, ten now carry the frozen bound, computed from each
campaign's latest sealed $M_{quality}$ species artifact. In Table~\ref{tab:whisker-coverage}, $\hat\mu$ is the Good--Turing missing mass at the final round, the mined gain is $G = VA_{new}-VA$, the whisker is $G\hat\mu/(1-\hat\mu)$ (Eq.~\ref{eq:whisker-base}, the first term of Eq.~\ref{eq:whisker}), and $VA$ + whisker is the bounded articulated ceiling that the appendix ladder figures (Figures~\ref{fig:vat-verdict-bars}, \ref{fig:vat-curation}, \ref{fig:vat-community}) draw:

\begin{table}[!htbp]
\centering\scriptsize\setlength{\tabcolsep}{4pt}
\resizebox{\textwidth}{!}{%
\begin{tabular}{lccccl cc}
\toprule
Task & $\hat\mu$ & mined gain & whisker & $VA$ + whisker & status & $G_{c}$ & whisker$_{c}$ \\
\midrule
Editorial pickup (press) &.589 &.091 & $+.130$ &.879 & bound &.022 & $+$.032 \\
Citation percentile &.456 &.069 & $+.058$ &.782 & bound &.048 & $+$.040 \\
Math.SE votes &.283 &.029 & $+.011$ &.651 & bound &.016 & $+$.006 \\
Joke upvotes &.400 & flat$\times$3 & $+.022$ &.743 & bound+tick &.028 & $+$.019 \\
BBC most-read &.608 & flat$\times$2 & $+.065$ &.843 & bound+tick &.041 & $+$.064 \\
Spotlight/oral pick &.589 &.060 & $+.086$ &.675 & bound & $-$.003 & $+$.000 \\
\#HashtagWars picks &.367 &.174 & $+.101$ &.804 & bound &.029 & $+$.017 \\
AoPS editorial approach &.492 &.001 & $+.001$ &.773 & bound &.001 & $+$.001 \\
Story upvotes (r/WritingPrompts) &.367 &.015 & $+.009$ &.674 & bound &.015 & $+$.009 \\
N\&C: received response &.450 &.016 & $+.013$ &.804 & bound &.016 & $+$.013 \\
GitHub PR merged & \multicolumn{4}{c}{no sealed proposal fleet} & n/e & n/e & n/e \\
ICLR accept/reject & \multicolumn{4}{c}{campaign exploratory, not sealed} & n/e & n/e & n/e \\
\bottomrule
\end{tabular}}
\caption{The missing-value bound on every eligible task, at the final sealed round of each mining campaign. Columns: $\hat\mu$, the Good--Turing missing mass ($f_1/N$ at the final round); mined gain, $G = VA_{new}-VA$, the increase in $VA$ that the campaign realized (flat$\times k$: the last $k$ rounds realized no gain); whisker, $G\hat\mu/(1-\hat\mu)$, the bound of Eq.~\ref{eq:whisker-base}; $VA$ + whisker, the bounded articulated ceiling; status, whether the task reports the bound alone or the bound with the calibrated terminal tick ($+.005$, the realized future gain in the backtest's terminal regime). n/e = not eligible (no sealed campaign). The two right-hand columns re-express the bound in the campaign's own frame: $G_c$ is the gain the campaign realized on its honest population between its first and final round (paired grouped bootstrap, Table~\ref{tab:whisker-uncertainty}), and whisker$_c = \max(0,G_c)\hat\mu/(1-\hat\mu)$. The mined-gain column is the master-table difference (the $+$mined column minus $VA$), which mixes the campaign frame with the bakeoff frame; for spotlight, press and HashtagWars the two frames disagree, and the campaign-frame bound is the smaller one.
}
\label{tab:whisker-coverage}
\end{table}

On the tasks whose gaps the main text emphasizes, the bounded
articulated ceiling stays below the full model (story upvotes
$.674 <.792$; citation percentile $.782 <.884$), so the whisker does not close those gaps. The three widest
whiskers (editorial pickup $+.130$, HashtagWars $+.101$, spotlight $+.086$) flag exactly the
\emph{under-mined} campaigns (large realized gains with high residual
mass). On those three tasks we treat the gap as unresolved rather than
tacit, because further mining could still narrow it. Only campaigns with two or more consecutive flat rounds receive the calibrated $.005$ tick; all other whiskers are drawn as
solid bounds. Per-task inputs and artifacts:
\texttt{results/u13\_extension\_harvest.json}.

\paragraph{Expanded backtest (seven campaigns).} The two-campaign
backtest above was the only one with persisted per-round records at the time;
five further multi-round campaigns have both per-round species and per-round
articulated AUC in a frozen frame, so we extend the truncation test to all
seven campaigns with persisted per-round records (seven of the twelve master-table campaigns with a mined column) ($30$ truncations for which a frozen articulated AUC exists). Each row cuts a campaign at round $k$
of $K$, computes the bound from rounds $\le k$, and compares it with the gain
realized in rounds $k{+}1..K$:

\begin{table}[!htbp]
\centering
\scriptsize
\resizebox{\textwidth}{!}{%
\begin{tabular}{llrrrrl}
\toprule
campaign & $k/K$ & $\hat\mu_k$ & bound & realized & used & holds \\
\midrule
BBC most-read & 1/6 & 0.45 & +.007 & +.020 & 297\% & \textbf{no} \\
BBC most-read & 2/6 & 0.41$^{*}$ & +.010 & +.014 & 130\% & \textbf{no} \\
BBC most-read & 3/6 & 0.49$^{*}$ & +.020 & +.008 & 39\% & yes \\
BBC most-read & 4/6 & 0.51$^{*}$ & +.029 & +.001 & 4\% & yes \\
BBC most-read & 5/6 & 0.49$^{*}$ & +.025 & +.003 & 12\% & yes \\
r/Jokes upvotes & 1/6 & 0.42 & +.012 & +.011 & 94\% & yes \\
r/Jokes upvotes & 2/6 & 0.60 & +.039 & +.002 & 5\% & yes \\
r/Jokes upvotes & 4/6 & 0.47 & +.025 & $-$.000 & -0\% & yes \\
r/Jokes upvotes & 5/6 & 0.56 & +.034 & +.001 & 3\% & yes \\
Spotlight/oral pick & 1/5 & 0.82 & +.000 & +.020 & n/d & \textbf{no} \\
Spotlight/oral pick & 2/5 & 0.63 & +.000 & +.015 & n/d & \textbf{no} \\
Spotlight/oral pick & 3/5 & 0.54 & +.000 & +.009 & n/d & \textbf{no} \\
Spotlight/oral pick & 4/5 & 0.58 & +.012 & $-$.000 & -2\% & yes \\
Citation percentile & 1/5 & 0.75 & +.020 & +.040 & 201\% & \textbf{no} \\
Citation percentile & 2/5 & 0.70 & +.049 & +.025 & 52\% & yes \\
Citation percentile & 3/5 & 0.78 & +.125 & +.010 & 8\% & yes \\
Citation percentile & 4/5 & 0.63 & +.091 & $-$.006 & -7\% & yes \\
\#HashtagWars picks & 1/4 & 0.88 & +.000 & +.009 & n/d & \textbf{no} \\
\#HashtagWars picks & 2/4 & 0.68 & +.017 & $-$.003 & -20\% & yes \\
\#HashtagWars picks & 3/4 & 0.65 & +.019 & $-$.006 & -30\% & yes \\
Story upvotes (r/WritingPrompts) & 1/7 & 0.30 & +.000 & +.015 & n/d & \textbf{no} \\
Story upvotes (r/WritingPrompts) & 2/7 & 0.33 & +.002 & +.011 & 614\% & \textbf{no} \\
Story upvotes (r/WritingPrompts) & 3/7 & 0.38 & +.005 & +.006 & 111\% & borderline \\
Story upvotes (r/WritingPrompts) & 4/7 & 0.38 & +.009 & +.000 & 5\% & yes \\
Story upvotes (r/WritingPrompts) & 5/7 & 0.28 & +.005 & +.001 & 26\% & yes \\
Story upvotes (r/WritingPrompts) & 6/7 & 0.37 & +.007 & +.002 & 30\% & yes \\
N\&C: received response & 1/5 & 0.60 & +.008 & +.010 & 132\% & borderline \\
N\&C: received response & 2/5 & 0.63 & +.013 & +.007 & 54\% & yes \\
N\&C: received response & 3/5 & 0.53 & +.005 & +.010 & 202\% & \textbf{no} \\
N\&C: received response & 4/5 & 0.45 & +.008 & +.005 & 66\% & yes \\
\bottomrule
\end{tabular}}
\caption{Truncation backtest of the missing-value bound on seven campaigns. Rows: a campaign cut at round $k$ of $K$; columns: $\hat\mu_k$, the missing mass at round $k$; bound, the whisker computed from rounds $\le k$; realized, the gain actually obtained in rounds $k{+}1..K$; used, the mined gain entering the bound at round $k$; holds, whether realized $\le$ bound. n/d: the bound at that truncation was zero because no gain had yet been realized, so the share used is undefined and the bound fails by construction.}
\label{tab:whisker-backtest-expanded}
\end{table}

As in the two-campaign backtest, the failures concentrate at early truncations: the bound holds in
$18$ of $30$ truncations ($2$
borderline), and $5$ of the $10$ failures are
rows where the cumulative gain up to $k$ is zero or negative. The bound is
then trivially zero, a degenerate early-campaign case rather than a
miscalibration of the mass-to-value scaling. Restricting to $k \ge 3$, the
bound holds in $13$ of $16$ rows ($1$
borderline). Two caveats: campaigns stop when observed
gains flatten, so late-truncation coverage is partly expected by construction
(the test is informative about the mass-to-value \emph{scaling}, not about
stopping-rule optimality); and the N\&C campaign's frozen readout frame
(monitor-full) has near-zero cumulative gain, so the N\&C rows are shown in the
honest frame and the N\&C bound is the least stable of the seven. $^{*}$BBC
rounds 2--5 use the masses published in the original table; their species
files were not persisted and are being regenerated from the saved
two-judge verdicts. Artifact: \texttt{results/u13\_backtest\_extended.json}.

\paragraph{Species definition and sensitivity to the merge rule.} A
species is a set of proposals judged to name the same criterion. Two
partitions are computed per round: a similarity partition (single-linkage
on sentence embeddings at cosine $.79$) and a \emph{strict} two-judge
partition (two independent judges must both call a pair the same criterion;
pairs are shortlisted at cosine $\ge.55$). The two partitions are not nested (the
strict pass rebuilds from singletons rather than refining the similarity
clusters), so $\hat\mu$ can move in either direction between them, and in
Table~\ref{tab:species-sensitivity} it moves in both directions:

\begin{table}[!htbp]
\centering
\scriptsize
\resizebox{\textwidth}{!}{%
\begin{tabular}{lcccc}
\toprule
campaign / round & $\hat\mu$ similarity-only & $\hat\mu$ merged & $\hat\mu$ strict two-judge & leave-one-proposer-out range \\
\midrule
HashtagWars r4 & 0.73 & 0.73 & 0.37 & [0.73, 0.91] \\
Spotlight r5 & 0.78 & 0.78 & 0.59 & [0.75, 0.84] \\
Citations r5 & 0.57 & 0.57 & 0.60 & [0.52, 0.60] \\
AoPS r1 & 0.50 & 0.56 & 0.49 & [0.53, 0.64] \\
BBC r1 & 0.55 & 0.45 & 0.45 & [0.38, 0.53] \\
BBC r6 & 0.54 & 0.61 & 0.61 & [0.58, 0.67] \\
Press r2 & 0.59 & 0.59 & 0.62 & [0.53, 0.67] \\
math.SE votes r3 & 0.28 & 0.28 & 0.47 & [0.29, 0.32] \\
math.SE accepted r2 & 0.33 & 0.33 & 0.56 & [0.28, 0.36] \\
r/Jokes r1 & 0.28 & 0.42 & 0.42 & [0.39, 0.50] \\
r/Jokes r2 & 0.32 & 0.60 & 0.60 & [0.60, 0.65] \\
r/Jokes r4 & 0.38 & 0.47 & 0.47 & [0.44, 0.53] \\
r/Jokes r5 & 0.32 & 0.56 & 0.56 & [0.56, 0.64] \\
r/Jokes r6 & 0.49 & 0.40 & 0.40 & [0.42, 0.50] \\
\bottomrule
\end{tabular}}
\caption{Sensitivity of the missing mass $\hat\mu$ to the species-merge rule. Rows: campaign and round; columns: $\hat\mu$ under the similarity-only partition, under the default two-judge merge, under the strict two-judge partition, and its range across leave-one-proposer-out recomputations. Values in red were recomputed from the saved verdicts and re-embedded proposals (bge-large, cosine $.79$); where a round has only one two-judge partition, the strict and merged columns coincide by construction, and for the HashtagWars, spotlight, citations and press rows the merged column is the similarity-only partition.
}
\label{tab:species-sensitivity}
\end{table}

Differences of $.1$--$.4$ in $\hat\mu$ across merge rules are common and change the
whisker multiplier $\hat\mu/(1-\hat\mu)$ by up to a factor of two. This sensitivity has two
consequences. First, where a strict two-judge partition exists,
Table~\ref{tab:whisker-coverage} uses the strict partition, because it gives the more conservative estimate of
distinct criteria; elsewhere the table uses the merged partition. Second,
a whisker is better reported as a \emph{range} of bounds across admissible merge rules than
as a single value, and the leave-one-proposer-out ranges (last column of Table~\ref{tab:species-sensitivity})
show that proposer identity contributes a further $\pm.05$--$.10$. Neither the
merge rule nor proposer identity changes any qualitative conclusion in the paper, because the
tasks whose gaps the main text emphasizes have small realized mining gains,
which cap the whisker regardless of $\hat\mu$. Artifact:
\texttt{results/u13\_species\_sensitivity.json}.

\paragraph{Three sources of uncertainty in the whisker.} The whisker $G\cdot\hat\mu/(1-\hat\mu)$ has three uncertain inputs. (i) $\hat\mu$ is a Good--Turing estimate from $N$ proposals; we use the standard error $\sqrt{f_1+2f_2}/N$ \citep{good1953population} rather than a bootstrap over proposals, because resampling proposals removes singletons and biases $f_1/N$ downward. (ii) $G$ is the realized mining gain in the campaign's own frame; its standard error is the bootstrap spread stored in the campaign's round artifacts for creative writing and notice-and-comment, and a paired grouped bootstrap of the campaign's reproduced heads for the other tasks. (iii) The $VA$ level itself has the grouped-bootstrap interval of Appendix~\ref{app:standard-cis}; the $VA$ interval shifts the bar the whisker sits on but does not enter the whisker. Table~\ref{tab:whisker-uncertainty} propagates (i) and (ii) through the whisker by Monte Carlo. The $\hat\mu$ interval is the larger of the two propagated sources on four tasks (citation percentile, story upvotes, BBC most-read, joke upvotes) and the $G$ interval on the other six; where the whisker exceeds $.005$, its upper limit is two to four times its point value.

\begin{table}[!htbp]
\centering\footnotesize
\resizebox{\textwidth}{!}{%
\begin{tabular}{lllllll}
\toprule
task & $\hat\mu$ & 95\% CI & $G$ & se$(G)$ & whisker & 95\% CI \\
\midrule
Editorial pickup (press) &.589 & [.405,.772] &.022 &.015 & $+$.032 & [$+$.000, $+$.108] \\
Citation percentile &.456 & [.288,.623] &.048 &.015 & $+$.040 & [$+$.012, $+$.093] \\
Math.SE votes &.283 & [.166,.401] &.016 &.007 & $+$.006 & [$+$.001, $+$.015] \\
Joke upvotes &.400 & [.219,.581] &.028 &.003 & $+$.019 & [$+$.008, $+$.040] \\
BBC most-read &.608 & [.458,.759] &.041 &.003 & $+$.064 & [$+$.034, $+$.133] \\
Spotlight/oral pick &.589 & [.411,.767] & $-$.003 &.013 & $+$.000 & [$+$.000, $+$.037] \\
\#HashtagWars picks &.367 & [.171,.563] &.029 &.026 & $+$.017 & [$+$.000, $+$.061] \\
AoPS editorial approach &.492 & [.331,.653] &.001 &.003 & $+$.001 & [$+$.000, $+$.008] \\
Story upvotes (r/WritingPrompts) &.367 & [.160,.573] &.015 &.004 & $+$.009 & [$+$.002, $+$.021] \\
N\&C: received response &.450 & [.236,.664] &.016 &.010 & $+$.013 & [$+$.000, $+$.044] \\
\bottomrule
\end{tabular}}
\caption{Inputs to the missing-value whisker and their uncertainty, at the final round of each campaign. Columns: $\hat\mu$ (Good--Turing missing mass) with its 95\% interval; the mined gain $G$ in the campaign's own frame and its standard error from a 2,000-draw paired grouped bootstrap of the campaign's reproduced heads; the whisker $G\hat\mu/(1-\hat\mu)$ with its 95\% Monte-Carlo interval (3,000 draws). $\hat\mu$ and $G$ are the values of Table~\ref{tab:whisker-coverage}; whiskers may differ in the third decimal by rounding. The citation-percentile $\hat\mu$ is the pooled-partition value of that table; its standard error uses the final round's $N{=}90$ and the strict partition's doubleton count.}
\label{tab:whisker-uncertainty}
\end{table}

\paragraph{Four further campaigns.} After the seven backtested campaigns we ran the sealed protocol on four more tasks under the frozen stopping rule: two consecutive proposing rounds with monitor gain below $\varepsilon=.005$, five rounds at most. RoyalRoad followers: two rounds, $G=+.0030$, $\hat\mu=.356$, whisker $+.0017$. Stack Overflow accepted: three rounds, $G=+.0036$, $\hat\mu=.685$, whisker $+.0078$. Stack Overflow bounty: four rounds, $G=+.0237$, $\hat\mu=.511$, whisker $+.0248$. Math.SE bounty: five rounds, $G=+.0371$, $\hat\mu=.663$, whisker $+.0730$, the gap to the full model moving $.085 \to .048$; rounds 2 and 4 carry a degraded-audit flag (both auditors missed one planted probe, and the readmission rule was applied). On ICLR accept/reject the campaign is held at round 0: the paired refit-seed spread of $\mathcal{VA}$ on the monitor split is $.012$, above $\varepsilon$, so no round can be resolved; the ICLR campaign's earlier pilot used a single self-proposer and has no valid Good--Turing estimate. None of the four tasks has an evaluation frame for the head-to-head, so the four tasks do not enter Table~\ref{tab:metric-discovery}.

\subsubsection{\texorpdfstring{Metric discovery head-to-head, and the articulated ceiling}{Metric discovery head-to-head, and the articulated ceiling}}
\label{app:metric-discovery}

We ran three published criterion-discovery methods on the same inputs as our mining campaigns (the 2,000 labeled training rows of each task, whole groups, held-out rows excluded), with the same LLM (Qwen3-32B) wherever the method needs one, and scored every resulting bank with the paper's judge protocol (Gemma-4-31B, one criterion per prompt, the same token scheme) on the task's evaluation frame. \textbf{HypoGeniC} \citep{zhou2024hypogenic} keeps an online bank of at most 20 hypotheses that predict the training labels, replacing hypotheses whenever ten misclassified examples accumulate; we cap the loop at 300 rows and obtain the predictions of the top-$k$ hypotheses in one call. \textbf{HypotheSAEs} \citep{movva2025hypothesaes} trains sparse autoencoders on text embeddings, selects the 20 neurons that best separate the labels and names each from its top-activating examples; we report both our reimplementation and the authors' package (their selection statistic, training schedule and fidelity-scored naming). \textbf{ExperiGen} \citep{sengupta2026experigen} runs four rounds in which a generator proposes 30 hypotheses and an experimenter tests each by annotating 300 training rows and fitting a multivariate logistic regression; we keep every hypothesis (about 120 per task). Each bank is fit by the frozen stack of Section~\ref{sct:articulated_and_verifiable_metrics}; \emph{union} stacks the four external banks, \emph{ours+union} stacks them on ours. The articulated ceiling of Section~\ref{sct:missing-mass} is $\max(\text{ours}, \text{ours+union})$ plus the whisker. Table~\ref{tab:metric-discovery} reports all values; the whisker is the salience-ordered term of Equation~\ref{eq:whisker} from the task's campaign (zero where the campaign realized no gain; absent where no campaign was run).

\paragraph{A fourth method on one task: rubric meta-optimization.} After the head-to-head was sealed, \citet{weston2026unslopping} described RL-XAR, in which a \emph{meta-prompt} instructs a model to write a scoring rubric, a judge scores texts with the rubric, and a meta-optimizer is shown the meta-prompt, the score gap the rubric achieves between expert human text and model text on a small training set, and the failure cases, and rewrites the meta-prompt to ``widen the gap for genuine-quality reasons, not superficial tells''; the best meta-prompt by held-out gap is kept and the writer is then trained by reinforcement learning against the rubric. We reimplemented the rubric-discovery loop from that description as a fourth discovery arm and ran it on joke upvotes, with the gap defined as the community's positive minus negative outcome (the quantity every arm here optimizes) and without the RL stage: eight criteria per rubric, $24$ training and $16$ held-out items per class, four failure cases per side, six iterations, Qwen3-32B as rubric writer (thinking on), judge (thinking off) and meta-optimizer (thinking on), 492 model calls in 4 minutes. The training gap widened from $+1.44$ to $+2.10$ on the $0$--$10$ scale while the held-out gap did not improve, so keep-best returned the first-iteration rubric (held-out gap $+1.89$). Scored by the judge of record on the same $3{,}163$ evaluation rows and refit with the frozen stack, the rubric's eight criteria reach $VA$ .678 ($\mathcal{VAT}$ .743, tacit share 0.28), against .689 for our $74$-criterion bank, .691 for ExperiGen ($101$ criteria), .686 for HypoGeniC ($20$) and .604 for HypotheSAEs ($20$). Adding the RL-XAR rubric to the union of the published methods moves the union from .707 to .717 and the union stacked on ours from .716 to .721. Per criterion it is the most efficient method we ran; as a bank it does not close the gap on this task, and the meta-optimizer's revisions (toward ``earned'' surprise and ``emotional truth'', away from puns and in-jokes) fit the $48$ training items rather than the held-out ones. The RL-XAR criteria, meta-prompts and per-iteration gaps are in the released artifacts.

\begin{table}[h]
\centering
\caption{Metric-discovery head-to-head on the fourteen tasks with a shared evaluation frame: held-out AUC of the full model $T$, our bank, each external bank, their union, and the union stacked on ours; the campaign whisker; the committed articulated ceiling $\max(\text{ours},\text{ours+union})+\text{whisker}$; the full model $\max(T, \mathcal{VAT})$ with $\mathcal{VAT}$ refit on ours+union; and the articulability gap, full model minus ceiling. ExperiGen is the strongest external method (best on ten tasks, ahead of our bank on six); the package-faithful HypotheSAEs is last on every task. The gap survives on twelve of the thirteen bounded tasks; on BBC most-read the whisker covers it ($-.005$), and on N\&C rule change the two are at parity ($+.001$).}
\label{tab:metric-discovery}
{\scriptsize
\setlength{\tabcolsep}{3pt}
\resizebox{\textwidth}{!}{\begin{tabular}{lrrrrrrrrrrrrr}
\toprule
task & $n$ & $T$ & ours & HypotheSAEs & HypoGeniC & HypotheSAEs pkg & ExperiGen & union & ours+union & whisker & ceiling & full & gap \\
\midrule
Joke upvotes & 3,163 & .747 & .689 & .604 & .686 & .556 & .691 & .707 & .716 & $+$.019 & .734 & .747 & $+$.013 \\
Math.SE accepted & 2,600 & .644 & .574 & .532 & .554 & .526 & .582 & .579 & .575 & $+$.001 & .576 & .644 & $+$.068 \\
Math.SE votes & 2,326 & .654 & .603 & .535 & .611 & .569 & .628 & .632 & .631 & $+$.006 & .637 & .660 & $+$.023 \\
N\&C rule change & 1,417 & .624 & .612 & .545 & .607 & .487 & .632 & .644 & .633 & 0 & .633 & .634 & $+$.001 \\
N\&C responded & 1,904 & .817 & .791 & .501 & .675 & .470 & .723 & .736 & .802 & $+$.013 & .815 & .841 & $+$.026 \\
ICLR spotlight/oral & 1,571 & .594 & .529 & .517 & .557 & .455 & .514 & .556 & .573 & 0 & .573 & .594 & $+$.021 \\
ICLR citation pct. & 478 & .884 & .655 & .672 & .598 & .637 & .728 & .731 & .726 & $+$.040 & .766 & .884 & $+$.118 \\
ICLR accept & 1,244 & .777 & .668 & .501 & .609 & .538 & .637 & .639 & .675 & --- & --- & --- & --- \\
N\&C agreement & 1,009 & .603 & .563 & .544 & .591 & .482 & .573 & .578 & .583 & 0 & .583 & .603 & $+$.020 \\
WP story upvotes & 7,008 & .792 & .665 & .572 & .616 & .602 & .689 & .695 & .703 & $+$.009 & .712 & .792 & $+$.080 \\
HashtagWars picks & 924 & .732 & .529 & .486 & .633 & .512 & .589 & .619 & .562 & $+$.017 & .579 & .732 & $+$.153 \\
Press pickup & 605 &.774 &.660 &.464 &.493 &.540 &.664 &.644 &.655 & $+$.032 &.692 &.774 & $+$.083 \\
AoPS approach & 5,202 & .781 & .771 & .729 & .718 & .624 & .763 & .760 & .770 & $+$.001 & .772 & .782 & $+$.010 \\
BBC most-read & 10,147 & .822 & .719 & .688 & .694 & .688 & .741 & .759 & .763 & $+$.064 & .827 & .822 & $-$.005 \\
\bottomrule
\end{tabular}}}
\end{table}

Two results bear on the main text. First, no articulated model, ours or any other or all of them stacked, reaches the full model on any task; the whisker covers the remaining distance on BBC most-read and N\&C rule change is at parity. The mean tacit share over the thirteen bounded tasks is $.38$ against our bank alone and $.21$ against the committed ceiling, with the full model in both cases the better of $T$ and $\mathcal{VAT}$ refit on the bank in question. Second, the union's realized increment over our bank exceeds our campaign's whisker on ten of thirteen tasks, which is why the committed ceiling stacks the union on our bank rather than relying on the whisker alone. Our contribution is therefore the certification (species collapse, missing mass, whisker, and the spurious-variable audit applied to every bank), not the discovery algorithm; the certification can be applied to any proposer.

\textbf{The bound against the baselines.} The whisker predicts what our own campaign left unmined, so the whisker can be checked against what the external methods found. Over the 13 tasks with a bound, ours plus whisker beats the best single external bank on 5 and loses on 8; against ours$\cup$ext it wins 4 and loses 9. Four of the losses in both columns are tasks where the campaign gain was zero or negative (AoPS, spotlight/oral, N\&C rule change, N\&C agreement): a campaign whose gain was zero or negative yields a zero whisker even when $\hat\mu$ indicates that most of the mass was unseen (spotlight/oral: $G=-.003$, $\hat\mu=.59$, whisker $0$, union $+.040$). The committed ceiling therefore takes the maximum of ours and ours$\cup$ext before adding the whisker. The external banks themselves are not degenerate: of the 400 criteria in the first 20 finished HypoGeniC and HypotheSAEs banks there are no exact duplicates, no near-duplicates at $.85$ string similarity and no label restatements, against the 54--68\% degeneracy found in our own mined banks. The normalizer rewrites HypotheSAEs' ``a text is in Group A if'' frame into a criterion a judge can score on one item, and merges semantic duplicates that string matching misses.

\section{Deconfounding the full model}
\label{app:part-deconfound}

This appendix gives the full procedure that Section~\ref{sct:deconfound} summarizes: the spurious-feature audit, what we stratify, strip and residualize once a feature is declared spurious, and how the full model's performance decays as declared channels are removed.

\subsection{\texorpdfstring{The spurious-feature audit}{The spurious-feature audit}}
\label{app:spurious-audit}

The main text (Section~\ref{sct:deconfound}) names three tests: channel tests, path tests and an LLM adjudication. This section gives the full procedure in five stages, which map onto those three: stage 1 is the fidelity gate of the metric-discovery loop, stages 2 and 3 are the channel and path tests, stage 4 is a path test carried out by intervening on the capture layer, and stage 5 is the LLM adjudication.

``Spurious'' is not a purely statistical predicate. A feature is
spurious when its association with the label flows through a path that is not
the quality construct (a property of the capture, the selection, or the
context rather than of the work). With the construct latent and one
observational corpus, no statistic computed from (text, feature, label) can
decide which path a given association takes. Nor would full causal-graph
discovery settle it: the graph supplies the paths, but calling a path
illegitimate is a construct-validity commitment, an argument rather than a
statistic \citep{cronbach1955construct, messick1994interplay}. Our audit is
therefore built as a set of \emph{local} tests, each of which severs or
controls one suspected path under one stated assumption
\citep{pearl2009causality}, and its output for each candidate feature is
an evidence table: one row per test, with the result and
the assumption that result depends on. A decision to route a feature to $M_{spurious}$ cites that feature's evidence
table. The stages are ordered by what they can and cannot conclude.

\begin{figure}[t]
\centering
\resizebox{.98\textwidth}{!}{\input{figures/fig_spurious_audit.tex}}
\caption{\textbf{The spurious-feature audit.} A candidate metric goes through three tests. \emph{Channel tests} ask whether it captures signals external to the text, by scrambling the text and by probing its top n-grams. \emph{Path tests} ask whether its association with the label is reducible to a known spurious variable, by stratifying on that variable and checking whether the metric survives. The \emph{LLM adjudication} asks whether the metric measures the capture, selection or context of the item rather than the quality of the work. Each test writes one row of an evidence table (result plus the assumption it depends on), and the adjudication routes the metric, citing those rows, either to the criterion bank as a genuine metric or to the spurious-variable ledger, where it is deconfounded: stratified on if present in more than $60\%$ of the data, stripped by regular expression if it is a removable surface feature, and otherwise residualized out of the models' scores.}
\label{fig:spurious-audit}
\end{figure}

\textbf{Stage 1: instrument checks.} (1)~A \emph{degeneracy check}
drops criteria whose scores collapse to one modal value. (2)~An \emph{anchor
battery} scores blinded known-label anchor items in every batch and requires
the positive $>$ negative $>$ scrambled ordering. These two checks certify that a
criterion measures something; they say nothing about spuriousness.

\textbf{Stage 2: channel tests, which identify what a criterion
reads.} (3)~The \emph{scramble probe} scores each criterion on word-scrambled
versions of the items, which destroy everything order-dependent (argument,
misdirection, pacing) while preserving order-free statistics (length,
vocabulary register, topic words, punctuation). A quality-named criterion
whose scrambled scores still separate the classes is reading bag-of-words
surface, regardless of its name; scrambled AUC near $.5$ is consistent with
the named construct. The probe detects \emph{mislabeled channels}, not
illegitimate paths. For a feature declared as surface (word count) the probe is
vacuous, and a legitimately order-free criterion (``cites specific
statutes'') can fail the probe falsely, which stage~5 must catch. This is
negative-control logic \citep{lipsitch2010negative}: quality is absent from
scrambled text by construction, so separation that survives scrambling shows that the criterion reads a channel other than quality.
(4)~A \emph{surface-proxy regression} reports the fraction of each
criterion's variance explained by the declared length/format family.
(5)~\emph{Paraphrase invariance} rescores a sample under meaning-preserving
paraphrase; content criteria should be invariant, and style-signature criteria
should change. (6)~The \emph{capture-fingerprint probe} trains a character-$n$-gram
classifier on the class split; a criterion correlated with the classifier's score is
suspected of reading the capture pipeline rather than the text.

\textbf{Stage 3: path tests of whether the criterion's association
reduces to a suspect channel.} Path tests use the same arithmetic as
deconfounding, but they answer a different question. Deconfounding subtracts
a channel already ruled illegitimate from a readout; here the subtraction is
a diagnosis: it factors the candidate's label association into the part that
flows through the suspect channel and the part beyond it. A criterion whose
signal disappears once the channel is held fixed has been shown
 to measure that channel under a quality-sounding name. Reducibility to a channel
is the testable half of spuriousness; whether the channel itself is a
legitimate part of what the community rewards is the normative half, and it
is deferred to stage~5. (7)~\emph{Within-stratum survival} recomputes the
criterion's AUC inside strata of the suspected confound (era bins, venues,
topic clusters, length deciles); signal that vanishes within strata was
carried by the stratum variable. (8)~\emph{Matched-pair survival} recomputes
the association inside caliper-matched positive/negative pairs on the
suspect channel's percentile rank. (9)~The \emph{group-shuffle null}
permutes labels within versus across the task's natural groups (prompt,
docket, contest); a criterion that predicts the label across groups but not within them is
reading group identity. (10)~The \emph{transport test} fits the
criterion--label relation on one era or source slice and evaluates it on
another; collapse or sign reversal marks the association as slice-local
(selection or drift), though construct stability across slices is itself a
commitment, since genuine taste drift is the rival explanation. (11)~The
\emph{capture-route decomposition} is a multitrait-multimethod design
\citep{campbell1959convergent}: for items captured through two routes (fan
transcripts versus shared ASR; a research corpus versus our own
recollection), criterion variance is split into an item share and a route
share, and a high route share marks the criterion as a property of the
capture.

\textbf{Stage 4: interventions, restricted to the capture layer.}
Free-text rewriting is excluded from the audit because tacit features
are by definition unnamed, so no edit to content can be verified to preserve
them. The remaining interventions are (12)~\emph{shared re-derivations}, where one
class-blind pipeline re-derives both classes from the source artifacts
(re-transcribing aired and cut sketches with a single ASR system dropped the
transcriber-signature separability from $.73$ to $.55$ without changing a
word of content), and (13)~\emph{enumerable-span ablation}, stripping
declared content-free spans (bot notices, signatures, URLs) from both
classes.

\textbf{Stage 5: the declared commitment.} (14)~A \emph{construct
audit} (an LLM adjudication with arbitration on disagreement) assigns
the final label, quality of the work versus property of its capture,
selection, or context, citing the evidence rows above. This stage is not a
test and cannot be replaced by one: whether length is part of what a
community rewards or an artifact of how we sampled it is a decision about
the construct, and the audit makes that decision explicit,
recorded, and open to challenge rather than leaving it implicit in a threshold.

Of the fourteen components, this paper's tasks run (1), (2), (3) as
part of the anchor battery, (6), (8), and (14) everywhere, and (12) and (13)
where a shared source artifact exists; (9), (10), and (11) are specified
here and are the audit's next increment, computable from existing artifacts.
The audit itself is validated two ways. First, by \emph{planted artifacts}:
features with known ground truth (an injected class-correlated formatting
token, a deliberately era-skewed resample) are routed to
$M_{spurious}$, and held-out genuine criteria are not. Second, by the
\emph{historical leak benchmark}: the corpus-construction record
(Appendix~\ref{app:balancing}) contains leaks we originally found by hand
(transcriber signatures in the sketch task, bot-merged pull requests
mislabeled as rejections, author-status flags in titles), and each is a
labeled test case that the battery recovers unaided. The
literature grounds the logic of the tests; the planted and historical cases
test our implementation of them.

\paragraph{Validation run.} \emph{Planted artifact.} On joke upvotes we appended a footnote mark to $55\%$ of positive and $15\%$ of negative jokes (alone AUC $.70$), wrote two criteria that read it, one disguised as craft (``typographic closure after the punchline'') and one literal, and scored both plus two genuine base criteria with the judge of record on the altered texts, together with $300$ word-scrambled items. Table~\ref{tab:audit-bench} gives the evidence table. Both planted criteria receive three flags (surface-proxy $R^2$ $.89$ and $1.00$, capture-fingerprint correlation $.60$ and $.62$, AUC $.70$ collapsing to $.50$ inside token strata); the two controls, the $47$ base criteria and the $65$ mined quality criteria receive none; six of the $68$ nuisance criteria are flagged by the statistical stages alone. The scramble probe misses the token, because shuffling moves it off the end of the text; a position-dependent artifact requires the fingerprint or stratum tests. \emph{Era-skewed resample.} Redrawing the population so that positives come from later eras (era alone AUC $.75$ against $.59$) flags the era covariate (within-era AUC $.53$) and none of the $180$ record criteria, none of which correlates with era ($|\rho|\le.07$). \emph{Historical leaks.} On the pre-fix code-review titles the fingerprint probe surfaces the \texttt{[WIP]} markers ($1{,}848$ titles) among its top $n$-grams but not the bot-merge notice, which sits in $85$ of $141{,}644$ titles; the enumerable-span test catches it (merge rate $.19$ inside the notice against $.86$ outside). On the sketch task the fingerprint AUC is $.62$ on the fan transcripts and $.56$ on the shared re-transcription ($n{=}40$ and $72$). \emph{Construct audit.} A fresh blind adjudicator routes all nine test criteria correctly, including the disguised planted one; across the $39$ mining rounds it also routed all $156$ planted probe pairs correctly.

\begin{table}[!htbp]
\centering\scriptsize
\resizebox{\textwidth}{!}{%
\begin{tabular}{llcccccl}
\toprule
case & truth & scramble AUC & surface $R^2$ & fingerprint $\rho$ & AUC pooled $\to$ within stratum & stage 5 & routed \\
\midrule
planted token, disguised criterion & spurious &.51 &.89 &.60 &.70 $\to$.50 & incidental & yes \\
planted token, literal criterion & spurious &.51 & 1.00 &.62 &.71 $\to$.50 & incidental & yes \\
control: reveal is present and locatable & quality &.49 &.05 &.22 &.59 $\to$.59 & quality & no \\
control: point survives as a quotable formulation & quality &.49 &.03 &.13 &.62 $\to$.63 & quality & no \\
112 record quality criteria (47 base, 65 mined) & quality & & \multicolumn{3}{c}{0 of 112 flagged} & & no \\
68 record nuisance criteria & spurious & & \multicolumn{3}{c}{6 of 68 flagged by statistics} & incidental & yes (stage 5) \\
era covariate, skewed resample & spurious & & & &.75 $\to$.53 & & yes \\
code titles: \texttt{[WIP]} marker & leak & \multicolumn{4}{l}{surfaced by the fingerprint probe's top $n$-grams} & incidental & yes \\
code titles: bot-merge notice & leak & \multicolumn{4}{l}{missed by top $n$-grams (85 titles); span test: merge rate.19 vs.\.86} & incidental & yes (span test) \\
sketch transcripts: capture route & leak & \multicolumn{4}{l}{fingerprint AUC.62 on fan transcripts,.56 on shared ASR} & incidental & yes \\
\bottomrule
\end{tabular}}
\caption{Validation of the spurious-feature audit on planted artifacts and historical leaks. Columns are the appendix's tests: label AUC on word-scrambled items; variance explained by the length/format family (plus the token indicator); Spearman correlation with a character-$n$-gram capture classifier; AUC pooled and inside strata of the suspect channel; the blind construct audit's label; and whether the battery routes the case to $M_{spurious}$. Flag thresholds: surface $R^2>.5$, $|\rho|>.5$, within-stratum loss $>.08$ AUC, scrambled AUC more than $.10$ from $.5$.}
\label{tab:audit-bench}
\end{table}

\subsection{Deconfounding procedure: what we stratify, strip and residualize}
\label{app:deconfounding}

A spurious feature is any feature that predicts labels without capturing the underlying construct we wish to measure. The full model $\mathcal{VAT}$ can fit to such features, so a gap between $\mathcal{VAT}$ and $\mathcal{VA}$ could reflect them rather than the construct. This appendix gives the procedure that Section~\ref{sct:deconfound} summarizes: how we decide that a feature is spurious, what we do with it once we have, which grouping units and training recipe the deconfounded models share, and what we declared spurious in each task. The decay curves that measure how much of the full model's performance rested on the declared features are in Appendix~\ref{app:dense-decay}; the full test battery and its validation are in Appendix~\ref{app:spurious-audit}; the per-corpus filters, matching variables and leak fixes that the grouping units below rest on are in Appendix~\ref{app:balancing}.

\paragraph{Deciding that a feature is spurious.} We run three tests on every candidate metric. \emph{Channel tests} identify variables that are external to the text, such as time of day, or that read the text only at its surface: we score the metric on word-scrambled copies of the items (a metric whose scrambled scores still separate the labels is reading length, vocabulary or formatting, regardless of its name), and we inspect the top $n$-grams of a character-level classifier trained on the label to see whether they are clearly spurious (bot notices, datelines, markup residue). These are stages 1--2 of Appendix~\ref{app:spurious-audit}. \emph{Path tests} identify variables \textit{within} the text that are correlated with external variables and not the construct: we stratify on \textit{known} spurious variables (era bins, venues, topic clusters, length deciles, the task's grouping unit) and ask whether the suspected metric still predicts the label inside the strata; a metric whose signal vanishes within strata was carried by the stratum variable. These are stage~3 of Appendix~\ref{app:spurious-audit}. The final \emph{LLM adjudication} asks, for every metric that passes the first two tests, ``is \{feature\} spurious in this task, where spurious means that the feature measures the capture, selection, or context of the item rather than the quality of the work? Answer yes/no.'' A second model arbitrates disagreements. This is stage~5 of Appendix~\ref{app:spurious-audit}; it is a judgment about the construct rather than a statistic. Metrics declared spurious go to the nuisance ledger $M_{spurious}$ and are never used as features; every readout is discounted for them as described below.

\paragraph{Validation.} We validate the tests with planted artifacts and with spurious variables we had found by hand. On joke upvotes we appended a footnote mark to $55\%$ of positive and $15\%$ of negative jokes and resampled the jokes so that posting era predicts the label; both planted features, including one disguised as a craft criterion, are routed to the ledger, and none of the $112$ genuine criteria are. The human-identified cases are bot-merge notices and \texttt{[WIP]} markers in GitHub pull-request titles and transcriber signatures in the SNL sketches; the battery recovers each unaided (Table~\ref{tab:audit-bench}).

\paragraph{What we do with a spurious feature.} We deconfound in one of three ways, chosen by how the feature is distributed in the data. If it is present in more than $60\%$ of the items, we \emph{stratify}: models are fit and evaluated within levels of the feature (e.g.\ within a length decile), so the comparison the model learns is between items that share it. If it is a surface feature that is easily removable, we \emph{strip} it: regular expressions delete the spans (bot notices, signatures, URLs) from every item before training, and where two capture routes exist we re-derive both classes with one class-blind pipeline (the aired and cut sketches were re-transcribed with a single ASR system). Otherwise we \emph{residualize}: we regress each model's score on the block of remaining spurious variables and correct the prediction, Corrected $=$ Actual Prediction $-$ Score predicted from the spurious block. The regression is cross-fitted under grouped five-fold splits on the task's grouping unit, so the partial-out never sees its evaluation rows, and we check the residual with nonlinear probes (a linear probe, an MLP, a boosted tree and a $k$-NN probe), because a linear regression cannot remove a nonlinear function of the same features. Linear partial-out also removes quality variance that happens to be collinear with a spurious feature (a longer story can be better \emph{and} longer), so the residualized full-model values are lower bounds.

\paragraph{Grouping units and training recipe.} Every held-out number in the paper is computed under the task's grouping unit, so that an item is only ever compared with items that share its context: the question on Stack Exchange, the story prompt on WritingPrompts, the docket in Notice \& Comment, the repository on GitHub, the contest for captions and hashtag games, the homepage snapshot for placement, the subreddit-era stratum for removals, the paper for peer review, the company for press releases and a topic cluster for jokes. Train/eval/test splits are 80/10/10 and are defined by the same units, so no group is split across folds. All learned models use one recipe: Llama-3.1-8B with LoRA ($r=16$, $\alpha=32$), learning rate $5\times10^{-5}$, batch $16$, two epochs, trained on the binary target.

\paragraph{What we declared in each task.} Table~\ref{tab:ledgers} lists, for the twelve tasks with a full-model decay ledger, the grouping unit, the number of declared spurious variables and the three with the largest label association on their own. The declared features are those the definition predicts: circulated-joke signatures and posting time for jokes, datelines and wire-service boilerplate for press releases, answer position for Stack Exchange, fashionable-subfield vocabulary and non-native grammar for peer review, citation apparatus and disclosed organizational standing for public comments.

\begin{table}[!htbp]
\centering\scriptsize
\begin{tabular}{@{}l l r p{7.6cm}@{}}
\toprule
Task & Grouping unit & Declared & Largest declared spurious features (by label AUC on their own) \\
\midrule
Creative writing (crowd) & story prompt & 70 & Fragmented staccato paragraphing  /  line-break repetition; Blank-line and short-paragraph density; Dialogue punctuation mark count \\
N\&C responded & docket & 56 & First-hand local exposure; First-person credential or stake self-disclosure; Identified accountable author with stated standing \\
Peer review (citation) & paper & 57 & Extent of currently fashionable subfield vocabulary; Trend-Aligned Vocabulary; Count of enumerated evaluation settings and resource markers \\
Press releases (pickup) & company & 23 & Extent of departure from templated corporate register; Extent of newswire dateline, ticker and boilerplate formatti; Formal wire-service dateline / distribution convention \\
Jokes (crowd) & topic cluster & 58 & Extent of overlap with stock circulated joke material; Pre-owned chestnut: the pun arrives already familiar; Canonical-chestnut retelling fingerprint (position in retell \\
Math SE (accepted) & question & 36 & number of answers; answer position; is the first answer \\
Math SE (vote) & question & 36 & answer position (percentile); answer position; is the first answer \\
N\&C rule change & docket & 22 & Technical identifier density; Legal / regulatory citation apparatus density; Professional legal drafting \\
N\&C agreement & docket & 25 & Contested consequential question versus ministerial housekee; Technical identifier density; Concrete organizational standing disclosed \\
Peer review (accept/reject) & paper & 43 & Non-idiomatic English and grammatical slips; A public repository URL appears in the abstract; Passive-voice density \\
Hashtag wars (verdict) & hashtag game & 44 & Contest boilerplate load; Boilerplate tag placed before vs. after the punchline; Tweet opens with the @midnight mention as first token \\
Peer review (spotlight) & paper & 52 & Memorable title framing; Density of currently fashionable technical terms and paradig; Trend-term concentration \\
\bottomrule
\end{tabular}
\caption{Declared spurious variables per task. ``Declared'' is the size of the task's nuisance ledger ($M_{spurious}$ criteria plus structural columns such as posting time and answer position); the last column names the three ledger entries with the largest label AUC on their own, as recorded in the deconfounding run. The full ledgers are in the run artifacts.}
\label{tab:ledgers}
\end{table}

Channels found later, by reading the items with the largest gaps, are declared and tested in Appendix~\ref{app:gap-battery} (Table~\ref{tab:item-spurious}).

\paragraph{What deconfounding changes.} Controlling for these variables changes the dataset and thus the measurement target: a model fit after stratifying out length, topic or genre predicts choices conditioned on these attributes. We accept the change of target because the alternative is to report a gap that a shortcut could explain. The residualized values are conservative because linear partial-out also removes quality variance collinear with a spurious feature; the stacked increment of Appendix~\ref{app:dense-decay} remains the primary deconfounding statistic.

\subsection{Spurious-feature decay curves for the full model}
\label{app:dense-decay}

The saturation curve in \S\ref{sec:res-frontier} asks how far \emph{adding}
articulated criteria can raise the bank; the decay curve here asks the opposite
question of the full model: how much of its performance rests on the spurious
channels our mining process declared. For each task we residualize the dense
readout on random subsets of $k$ declared nuisance channels ($M_{spurious}$ criteria plus declared structural columns) and track the held-out AUC of the
residual as $k$ grows. The residualization is cross-fitted (linear
regression under grouped five-fold splits on the task's grouping unit, so the
partial-out never sees its evaluation rows), with twenty random subsets per
$k$ and the full channel set as the terminal point. Where the curve has
flattened we also report the fitted exponential asymptote
$A + B e^{-k/\tau}$; where it has not (three tasks), the full-$K$ residual is
the bound and no asymptote is quoted.

\begin{table}[!htbp]
\centering
\footnotesize
\resizebox{\textwidth}{!}{%
\begin{tabular}{lrcccc}
\toprule
Task & $K$ & Full-model AUC & Residual AUC (all $K$) & Asymptote & placebo drop (share) \\
\midrule
Creative writing (crowd) & 70 &.792 &.737 &.738 & +.008 (0.15) \\
N\&C responded & 56 &.817 &.733 &.733 & +.035 (.42) \\
Peer review (citation) & 57 &.884 &.773 &.777 & +.029 (0.26) \\
Press releases (pickup) & 23 &.774 &.729 &.726 & +.034 (0.73) \\
Jokes (crowd) & 58 &.747 &.641 &.641 & +.004 (0.04) \\
Math SE (accepted) & 36 &.644 &.579 &.560$^{e}$ & +.016 (0.24) \\
Math SE (vote) & 36 &.654 &.578 &.555$^{e}$ & $-$.003 (-0.04) \\
N\&C rule change & 22 &.624 &.569 &.571 & +.011 (.20) \\
N\&C agreement & 25 &.603 &.568 &.569 & +.018 (.50) \\
Peer review (accept/reject) & 43 &.777 &.654 & n/i & +.008 (0.06) \\
Hashtag wars (verdict) & 44 &.732 &.591 &.564$^{e}$ & +.017 (0.12) \\
Peer review (spotlight) & 52 &.594 &.569 & n/i & +.004 (0.17) \\
Caption contest (finalist)$^{\dagger}$ & 38 &.612 &.521 &.518 & +.008 (0.09) \\
\bottomrule
\end{tabular}}
\caption{Full-model performance under progressive nuisance partial-out. $K$ = number
of declared nuisance channels; ``Residual AUC'' = held-out AUC of the full model's
readout after cross-fitted linear residualization on all $K$ channels;
``Asymptote'' = fitted floor $A$, quoted only where the curve plateaus
($|A - \text{full-}K| <.01$); ``---'' marks tasks still declining at full
$K$, for which the full-$K$ residual is an upper bound on the floor. One task
(BBC most-read) is omitted: its dense ledger is not yet built. $^{e}$: extrapolated from a curve still declining at full $K$, so read as an upper estimate; n/i: no finite asymptote is identifiable from the fitted curve (the full-$K$ residual is the quotable bound). ``Placebo drop'' = the drop in AUC produced by residualizing on a row-permuted nuisance block with the same $K$ and folds (mean of the placebo runs), i.e. the part of the drop that is first-stage estimation noise rather than shared variance, with its share of the archived drop in parentheses. $^{\dagger}$: New Yorker caption-contest task; not included in any main-text table or average (multimodal).}
\label{tab:dense-decay}
\end{table}

\paragraph{Nonlinear residue after linear residualization.} The residual AUC in Table~\ref{tab:dense-decay} is a linear guarantee: within grouped five-fold splits the dense score is regressed on the $K$ nuisance channels on the training folds and the held-out residual is read out, so the residual is orthogonal, out of fold, to the linear span of the channels. Like LEACE \citep{belrose2023leace}, which certifies that no linear classifier can recover an erased concept, the linear residualization says nothing about nonlinear routes. We therefore tested residual nuisance predictability directly. On the same grouped folds, with probe training rows drawn from inner-fold residuals so that no probe sees its test fold's first-stage fit, we trained a linear probe, a two-layer 256-unit MLP with early stopping, a depth-4 gradient-boosted tree and a $k$-nearest-neighbour probe to recover each nuisance channel, and the whole block, from the residual, against a within-group permutation null. The linear probe is at its null in every task. A nonlinear probe recovers the joint block above its null in seven of thirteen tasks, explaining between $2\%$ and $11\%$ of the residual's variance (median $2.4\%$), against the $20$--$57\%$ of the dense score's variance that the same block explained before residualization; $14\%$ of individual channels remain nonlinearly readable, mostly length, volume and template channels, at $R^2 \le.09$. Closing the nonlinear route by subtracting the nonlinear out-of-fold prediction of the residual lowers the residual AUC by at most $.021$ net of a matched placebo (jokes), by $.009$--$.015$ in four tasks, and by at most $.005$ in the remaining eight; the creative-writing floor moves from $.737$ to $.735$ and the peer-review-citation floor from $.773$ to $.758$, both still above their banks, while N\&C agreement ($.568$ against a bank of $.563$ on this ledger's frame; the same label reads $A=.612$ and $.524$ under the two designs of \S\ref{sec:limitations}) sits at parity within the noise of the partial-out before and after the correction. The placebo column of Table~\ref{tab:dense-decay} bounds the residualization itself: a row-permuted nuisance block with the same $K$ and folds reproduces $4$--$26\%$ of the archived drop in ten tasks and $42$--$73\%$ in three (N\&C responded, N\&C agreement, press releases), so those three floors are lowered by first-stage estimation noise as well as by shared variance. We therefore read each linear residual together with these two bounds, the nonlinear correction and the placebo drop. The nonlinear residue of the declared block is small on a scalar residual but not on a representation: in the one task where we erased a 4096-dimensional hidden state with LEACE, an MLP still read the erased concept at AUC $.96$ against a linear $.47$.

In twelve of thirteen tasks the deconfounded dense model remains
well above chance, but in only \emph{three} does the residual also remain
above the corresponding articulated bank (creative writing, peer-review
citations, N\&C agreement); in the other ten the linear partial-out pulls
the full model to or below the bank.
The linear partial-out's deliberate conservatism (see below) makes those ten
uninformative about the gap rather than evidence against it. On
creative writing the
floor of $.738$ still exceeds the bank's own asymptote of $.668$, matching
the stacked-increment result ($+.099$ beyond bank-plus-nuisance) from the
opposite direction. Two
exceptions are informative. On the caption-contest
finalist task the dense floor is $.518$, essentially chance, so there
the dense advantage is mostly collinear with declared nuisance, and we
treat that task's gap accordingly. On ICLR accept/reject the
residual ($.654$) falls below the task's own mined bank ($.668$), and the
declared nuisance channels alone reach $.690$ nonlinearly. The dense
advantage on that row is likewise largely collinear with declared nuisance,
and we mark its gap accordingly in Table~\ref{tab:master-results}. Throughout, the linear-residual readout is deliberately
conservative against the full model: linear partial-out removes every
direction of dense variance collinear with a nuisance channel, including
genuinely shared quality variance (a longer story can be better \emph{and}
longer), so the floors are lower bounds and the stacked increment remains the
primary deconfounding statistic.

\section{Full Results}
\label{app:part-results}

This appendix reports complete results behind the main-text figures and robustness checks on them.

\subsection{All results}
\label{app:master-results}

The paper counts its material at several levels, and the counts differ because each counts a different unit. The inventory (Appendix~\ref{app:inventory}) lists $44$ datasets in seven fields. Where the main text quotes a single full-model AUC it is the larger of the $T$ and $VAT$ columns of Table~\ref{tab:master-results} (e.g. $.792$ for story upvotes). Table~\ref{tab:master-results} reports the $32$ task--label rows that carry at least an articulated model (a 33rd, patents, is withdrawn). The main-text figures use the $28$ of those rows that carry a full model (RoyalRoad magazine picks have none). Sixteen tasks admit a selection test (Table~\ref{tab:regret-all}); thirteen have a spurious-feature decay ledger (Appendix~\ref{app:dense-decay}); twelve ran a sealed mining campaign, ten of which carry the missing-value bound (Table~\ref{tab:whisker-coverage}); and the information view (Appendix~\ref{app:info-view}) counts $47$ populations because it includes the per-community refits. Each count in the text refers to one of these sets.

Table~\ref{tab:master-results} collects every instrumented task
of the benchmark and every instrument we ran on it: the verifiable model $V$
(programmatic checks only), the articulated model $VA$ (verifiable checks plus
the LLM-judged criteria bank, with the criteria-mining increment where a mining
campaign ran on that task), the text-only preference model $T$ (a preference model trained
end-to-end on text), and the full model $VAT$ (the fusion of $T$ with the metric models; the main text's full model is the better of the $T$ and $VAT$ columns). All values are
held-out AUC under the task's grouping unit (question, prompt, docket,
repository, contest, snapshot, subreddit-era stratum; Appendix~\ref{app:balancing}).
Fused values follow one selection protocol throughout: the fusion variant is
chosen on the evaluation split and reported on the disjoint test split; where a
group-bootstrap $95\%$ confidence interval was run, it is quoted. The
main-text figures are views of this table: Figure~\ref{fig:concept-row} (and, regrouped by channel, Figure~\ref{fig:articulated-share})
plots the gap between the best instrument and the best fully-articulated
instrument as a discrimination share; the channel ladders of
Appendix~\ref{app:companion} plot one task per field and channel.

\begin{table}[!htbp]
\centering
\scriptsize
\setlength{\tabcolsep}{4pt}
\renewcommand{\arraystretch}{1.12}
\resizebox{\textwidth}{!}{%
\begin{tabular}{ll l ccccc}
\toprule
Field & task & Channel & $V$ & $VA$ & $+$mined & $T$ & $VAT$ \\
\midrule
Coding & GitHub PR merged & verdict &.550 &.631 &.638 & u/q$^{m}$ &.692 \\
 & Stack Overflow accepted answer & verdict &.639$^{f}$ &.733 & n/c &.743 &.764 {\tiny[.736,\,.791]} \\
 & Competition editorial-approach & curated &.743 &.734 & n/c &.724$^{m}$ &.737$^{m}$ \\
 & Stack Overflow bounty award & curated &.739$^{f}$ &.787 & n/c &.804 &.811 {\tiny[.791,\,.829]} \\
 & Stack Overflow answer votes & community &.638 &.710 & n/c &.723 &.728 \\
\midrule
Math & Math.SE accepted answer & verdict &.591 &.632 &.618$^{f}$ &.644 &.630 \\
 & AoPS editorial approach & curated &.708 &.771 &.772 &.781$^{t}$ &.785 \\
 & Math.SE bounty award & curated &.691$^{f}$ &.714 & n/c &.733 &.738 {\tiny[.717,\,.759]} \\
 & Math.SE answer votes & community &.575 &.611 &.640 &.654 &.688 \\
\midrule
Law \& policy & N\&C: received a response & verdict &.711 &.775 &.791 &.817 &.832 \\
 & Patent application granted$^{w}$ & verdict & & & w/d & w/d & \\
 & N\&C: rule change made & curated &.591 &.612 &.616 &.624 &.636 \\
\midrule
Academia & ICLR accept/reject$^{f}$ & verdict & &.654 &.668 &.777 & \,.747 \\
 & Spotlight/oral pick & curated &.531 &.529 &.589$^{h}$ &.594 &.639 \\
 & Citation percentile & community &.695 &.655 &.724$^{h}$ &.884 &.888 \\
\midrule
Journalism & Editorial pickup (press releases) & verdict &.653 &.658 &.749$^{h}$ &.774 &.750 \\
 & Homepage placement & curated &.643 &.729 & n/c &.740 &.740 \\
 & BBC most-read & community &.672 &.737 &.778$^{h}$ &.823 &.817 \\
 & Tweet engagement & community &.542$^{f}$ &.600 & n/c &.643 &.646 {\tiny[.627,\,.665]} \\
\midrule
Creative writing & Kindle Scout publisher accept$^{p}$ & verdict &.612$^{f}$ &.744 & n/c &.744 &.792 {\tiny[.627,\,.935]} \\
 & RoyalRoad market pickup$^{0}$ & verdict & .554 &.566 & n/c &.486 & --- \\
 & Wigleaf Top 50 & curated &.573 &.605 & n/c &.559 &.593$^{f}$ {\tiny[.543,\,.640]} \\
 & RoyalRoad Magazine pick$^{p}$ & curated &.510 &.624 {\tiny[.550,\,.686]} & n/c & n/r & n/r \\
 & Story upvotes (WritingPrompts) & community & \,.609 &.650 &.665 &.792 &.790 \\
 & RoyalRoad follower count$^{b}$ & community &.610 &.632 & n/c &.699 &.698 {\tiny[.639,\,.752]} \\
 & Macro-norm removals (4 long-form subs)$^{s}$ & verdict &.551 &.606 & n/c &.889 &.889 \\
\midrule
Humor & r/Jokes moderator removal & verdict &.603$^{f}$ &.904 & n/c &.922 &.923 {\tiny[.908,\,.939]} \\
 & Macro-norm removals (5 humor subs)$^{s}$ & verdict &.561 &.714 & n/c &.816 &.816 \\
 & SNL cut-for-time$^{0}$ & verdict & .55 &.49 & n/c & --- & --- \\
 & \#HashtagWars staff picks & curated &.569 &.529 &.703$^{h}$ &.732 &.706 \\
 & Style Invitational tiers & curated &.596 &.601 & u/q$^{u}$ &.624$^{v}$ & n/r \\
 & Joke upvotes (r/Jokes) & community &.634 &.689 &.721 &.747 &.760 \\
\bottomrule
\end{tabular}}
\captionsetup{singlelinecheck=off}
\caption{Per-task provenance, sizes and paired intervals are in Table~\ref{tab:audit}. \textbf{Every task, every instrument} (held-out AUC;
grouping units and construction in Appendix~\ref{app:balancing}). ``$+$mined''
is the articulated bank after criteria-mining rounds, where a campaign ran;
``n/c'' means no mining campaign ran, ``n/r'' that the model was not run, ``u/q'' that it is unquoted, and ``---'' that it was not built for that task, not that it scored at
chance (the two tasks marked $^{0}$ are at chance; see the notes and Appendix~\ref{app:new-tasks}). Bracketed intervals are
group-bootstrap $95\%$ CIs on the fused test-split value. The $\mathcal{VAT}$ column is the fused candidate chosen on the evaluation split and scored on test; it is below $T$ on 7 of the 26 rows that have both and below the articulated model on 3, and $V$ exceeds $VA$ on 5 rows; the full model used for every gap and share is $\max(T,\mathcal{VAT})$. Of the $32$ rows, the $28$ with a full model are the points of Figures~\ref{fig:cover-photo} and~\ref{fig:articulated-share} and the units of every cross-task correlation and test; rows with no full model (n/r or --- in both full-model columns) are assigned no gap.}
\label{tab:master-results}
\end{table}

{\footnotesize\noindent\textbf{Notes to Table~\ref{tab:master-results}.} $^{0}$~At chance. RoyalRoad market pickup is weak but real on its full $1{,}274$-row population ($V$ $.554$ [.524,\,.586], $VA$ $.566$ [.535,\,.598]); the full model's $.486$ is a 142-row test split, and an equivalence test excludes AUC $\ge.60$ but not $\ge.55$. SNL cut-for-time has $n{=}72$ (anchors AUC $1.0$, $A$ alone $.44$); the interval on any AUC spans about $\pm.13$, so the task cannot exclude AUC $.60$. $^{b}$~The articulated model's anchor battery is weak on this population, so the full-model-over-articulated margin could measure an instrument limit, not tacitness.
$^{m}$~The code full model is withheld: its only trained model used a
non-standard backbone and truncation budget, not the paper's uniform
Llama-3.1-8B LoRA recipe (Appendix~\ref{app:balancing}); the fused value for
this row inherits the same provisional status. A standard Llama-LoRA retrain
is queued.
$^{v}$~Best instrument for the task; its per-instrument ladder is not part of
the unified frame. $^{f}$~Matched-rows frame: all models scored on the
same $1{,}244$ dense-held-out judge-scored rows (class-balanced draw, pos
rate $.457$); the dense model reads $.753/.751$ on its own full eval/test
splits (pos $.394$). The mined column's campaign was exploratory
(pre-freeze), and the deconfounded residual for this row falls below the
bank (Table~\ref{tab:dense-decay}), so we do not treat this row's gap as
resolved tacitness. VAT re-quoted under the declared eval-select/test-report
protocol (was $.754$, a pooled test-selected value). $^{w}$~Withdrawn after the verifiability audit: every model of this task read a candidate set of eight prior-art passages to which the examiner's gold-cited reference had been appended for positive rows only, so every number is conditioned on the label (Appendix~\ref{app:negative}). $^{h}$~Remaining discovery gain unresolved: the missing-value bound on the articulated model (Table~\ref{tab:whisker-coverage}) exceeds $.05$, so the gap in this row could still narrow with further mining. $^{m}$~Dense model trained with a non-standard recipe (ModernBERT-base rather than Llama-LoRA); a fused value in the same row inherits that parent. Reported provisionally; the Llama-LoRA retrain is in progress. $^{t}$~Reused single-seed model whose checkpoint was selected on the test split, so the value is upward-biased; printed for completeness. $^{f}$~Value from the pooled out-of-fold frame, while the row's other columns are the bakeoff test leg; the frames differ, so the column, not the within-row difference, is the quotable quantity. $^{u}$~Campaign run on a population later found contaminated; not quotable. Codes: n/c = no sealed mining campaign was run for this task; n/r = model not built; u/q = run but not quotable (see note); w/d = withdrawn. $^{s}$~Pooled nine-community fits (era-uniform
rebuilt populations); fusion adds nothing beyond the full model on these
tasks, and the pooled articulated value substantially understates
within-community recovery (Appendix~\ref{app:subfields}). Tasks
reported in their own frames, mathlib pull
requests (topic-residualized: $V$ $.635$, $+$tactics $.680$; the previously
quoted full-model $.736$ was a TF-IDF proxy, not the standard Llama-LoRA
full model. No dense number is quotable for this task: twelve seeds across four training regimes, the fourth a fresh Llama-3.1-8B LoRA run on area-grouped splits (held-out AUC $.41$--$.51$ across three seeds; the eval-selected seed lands at $.446$), produced unstable, at-or-below-chance held-out
models (455 negatives at a $94\%$ base rate; a class-weighted lexical
baseline on identical splits scores far higher), so the task's readable
ceiling is lexical and its gap is not computable),
macro-norm moderation removals
(Appendix~\ref{app:subfields}), and the twelve-domain legal-outcome corpus
(appendix-only), appear in Appendix~\ref{app:balancing} rather than here.
The two New Yorker caption-contest tasks are excluded from all cross-field
share readouts and from every main-text table and average; they appear only in appendix tables, marked $^{\dagger}$ because the cartoon, which the caption answers, is unavailable to
every instrument.\par}

The table shows three regularities. Within nearly every task the ordering
$V < VA \le \max(T, VAT)$ holds: verifiable checks recover a real but limited share, articulated criteria extend it, and a full model extends it again; the last step is the articulability gap of the main text. Fusion never falls meaningfully below its better parent and helps most where the two parents are close (Stack Overflow, the bounty tasks); where one parent dominates, the fused score tracks that parent. The two chance-level rows are the exception, and neither is a certified null: RoyalRoad market pickup is a weak but significant signal on its full $1{,}274$-row population, and SNL cut-for-time is an underpowered null at $n{=}72$ (Table~\ref{tab:master-results}, footnote $^{0}$).

\subsection{Cross-field charts: full instrument ladders per channel}
\label{app:companion}

This section gives the full per-channel views behind the main-text summary
figures. The main text compresses each task to two numbers (best instrument,
best articulated instrument; Figure~\ref{fig:articulated-share}); the charts
here show the whole ladder ($V$, $VA$, the mined increment, and the
fused stack), one chart per preference channel, plus the ceiling chart
behind the normalization argument and the schematic that defines the two
gaps. The numbers are those of
Table~\ref{tab:master-results}.

\begin{figure}[!htbp]
 \centering
 \includegraphics[width=.82\textwidth]{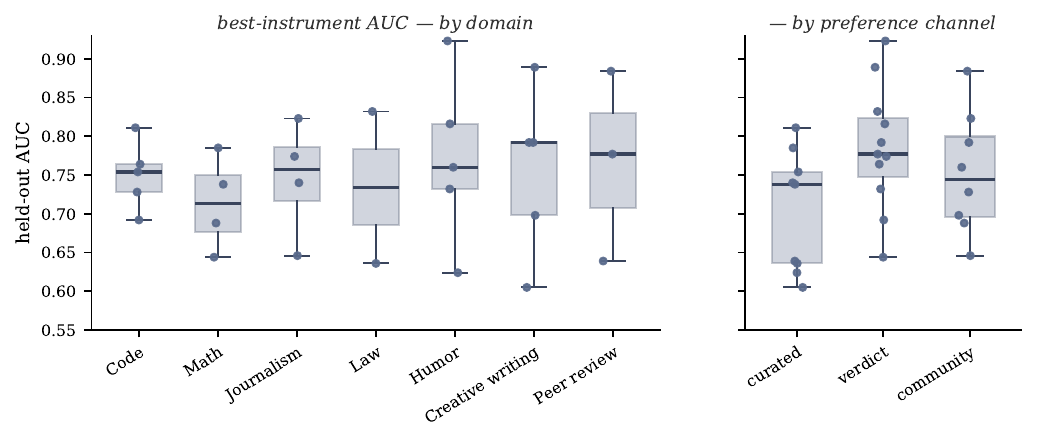}
 \caption{\textbf{Best-instrument AUC (the ceiling) for the tasks behind
 Figure~\ref{fig:concept-row}}, grouped by domain and by preference
 channel. The raw gap co-moves with this ceiling ($r{=}.49$, $p{=}.008$);
 the normalized share does not ($\rho{=}-.09$, $p{=}.65$; $n{=}28$).}
 \label{fig:auc-ceilings}
\end{figure}

\paragraph{The ceiling chart (Figure~\ref{fig:auc-ceilings}).} For each of
the tasks behind Figure~\ref{fig:concept-row} the chart plots the
best-instrument AUC (the total shared preference, regardless of
instrument). The ceiling chart supports the normalization choice in the main text.
Across tasks, the \emph{raw} articulability gap co-moves with this ceiling
($r{=}.49$, $p{=}.008$, $n{=}28$): part of any raw gap is mechanical, since a
task with more shared preference has more room for instruments to differ.
Figure~\ref{fig:concept-row} therefore plots the gap as a share of each
task's shared preference. The
normalized share is uncorrelated with the ceiling ($\rho{=}-.09$, $p{=}.65$, $n{=}28$, 95\% CI $[-.49,.35]$, bootstrap over tasks),
and twelve tasks matched on ceiling (AUC $.72$--$.80$) still span normalized
gaps of $0$--$44\%$ in the same domain order. The chart also records a fact
about channels used in \S\ref{sec:res-gaps}: curation has the lowest ceilings
of the three channels (prize juries are hard to predict for \emph{any}
instrument), which is why its small gaps must be read against its small
shared preference.

\begin{figure}[!htbp]
\centering
\includegraphics[width=\textwidth]{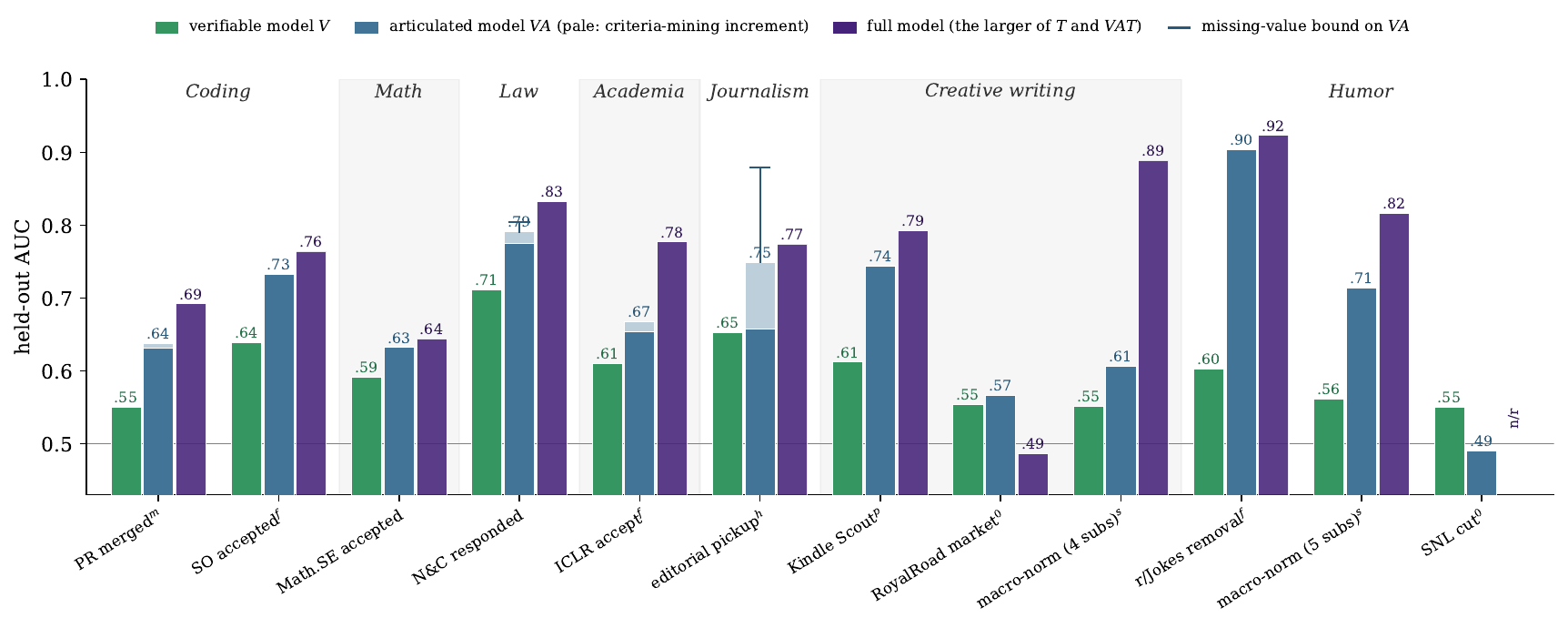}
\caption{Decision-maker \emph{verdicts}, unified-frame ladder. Green bars are
the verifiable model $V$; blue bars the articulated model $VA$, with
pale stacks marking criteria-mining increments where a mining campaign
ran; purple bars the full model, the larger of the $T$ and $VAT$ columns of Table~\ref{tab:master-results}. Every verdict task of that table is drawn, with its markers; n/r = no full model on record. Thin whiskers on the amber bars are the Good--Turing missing-value bounds of Appendix~\ref{app:whisker-backtest}, drawn under one standard on every task with a sealed mining campaign (editorial pickup $+.130$, notice-and-comment $+.013$); the peer-review campaign was exploratory and carries none.}
\label{fig:vat-verdict-bars}
\end{figure}

\paragraph{The verdict ladder (Figure~\ref{fig:vat-verdict-bars}).} Every verdict task of Table~\ref{tab:master-results}, twelve grouped by field, accountable single-gatekeeper decisions: merge a pull request,
accept an answer, respond to a comment, accept a paper, pick
up a press release, remove a post, accept a manuscript. Verdicts sit between the other two
channels in articulability, and the verdict ladder shows the two ways a verdict task's gap
can close: notice-and-comment closes mostly through the articulated bank
($VA$ $.775$, mined to $.791$, full model $.832$), while peer review leaves the
widest verdict band between the mined bank ($.668$) and the full model
($.777$, the text-only model $T$). The wide peer-review band is the accountable-decider signature discussed in
\S\ref{sec:res-gaps}. Humor's verdicts are the r/Jokes moderator removals ($.923$), the five-subreddit macro-norm removals ($.816$) and the underpowered SNL null; the \#HashtagWars staff pick, a curated task, is drawn in the curation ladder. Kindle Scout ($.792$), Stack Overflow accepted answers ($.764$) and the long-form macro-norm removals ($.889$) are drawn with the rest.

\begin{figure}[!htbp]
\centering
\includegraphics[width=\textwidth]{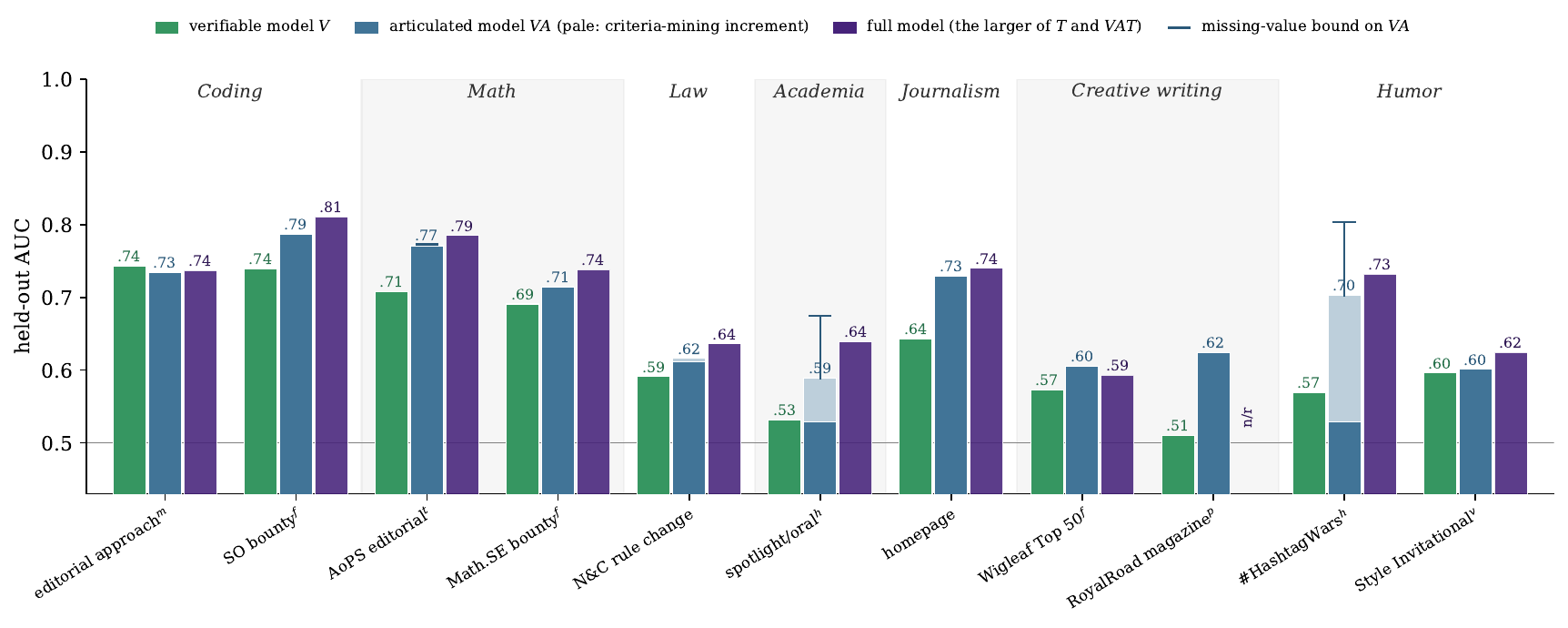}
\caption{\emph{Expert curation}, unified-frame ladder (same encoding as
Figure~\ref{fig:vat-verdict-bars}). Every curated task of Table~\ref{tab:master-results} is drawn; on Wigleaf the full-model bar is the fused value ($.593$, marker $f$), since the text-only model ($.559$) sits below the articulated model. Whiskers are the missing-value bounds (spotlight $+.086$, AoPS $+.001$, HashtagWars $+.101$); the spotlight bound exceeds the task's full model, so its gap is treated as unresolved (Table~\ref{tab:master-results}).}
\label{fig:vat-curation}
\end{figure}

\paragraph{The curation ladder (Figure~\ref{fig:vat-curation}).} The curation ladder covers every curated task of Table~\ref{tab:master-results}: eleven tasks grouped by field, each a deliberate few-from-many selection (editorial-approach picks, bounty
awards, oral/spotlight promotions, homepage placement, guest-editor lists).
Curation is the most articulable channel (smallest bands between blue and
purple) but also the lowest-ceilinged one, consistent with the accountability
reading: curators must defend selections, which favors criteria that can be
stated, while the scarcity of winners caps how much of the preference any
instrument can recover.  On competition editorial-approach match the articulated bank ($.734$, scored by the judge of record since 2026-09-29) sits below the verifiable bank ($.743$) and the fused model ($.737$) adds $.013$ over the text model ($.724$). This is one of the two tasks with almost no articulability gap, and one where the articulated criteria add nothing beyond what the verifiable programs measure. On Wigleaf the text-only model ($.559$) sits below the articulated model ($.605$); the bar drawn is the fused value ($.593$, pooled out-of-fold frame, marker $f$), and because the fused value also sits below the articulated model, Wigleaf is the other task with no articulability gap. The \#HashtagWars staff pick is drawn here as humor's curated task: its articulated model starts below its verifiable model ($.529$ vs $.569$) and reaches $.703$ after mining, the strongest case in the benchmark for criteria mining.

\begin{figure}[!htbp]
\centering
\includegraphics[width=\textwidth]{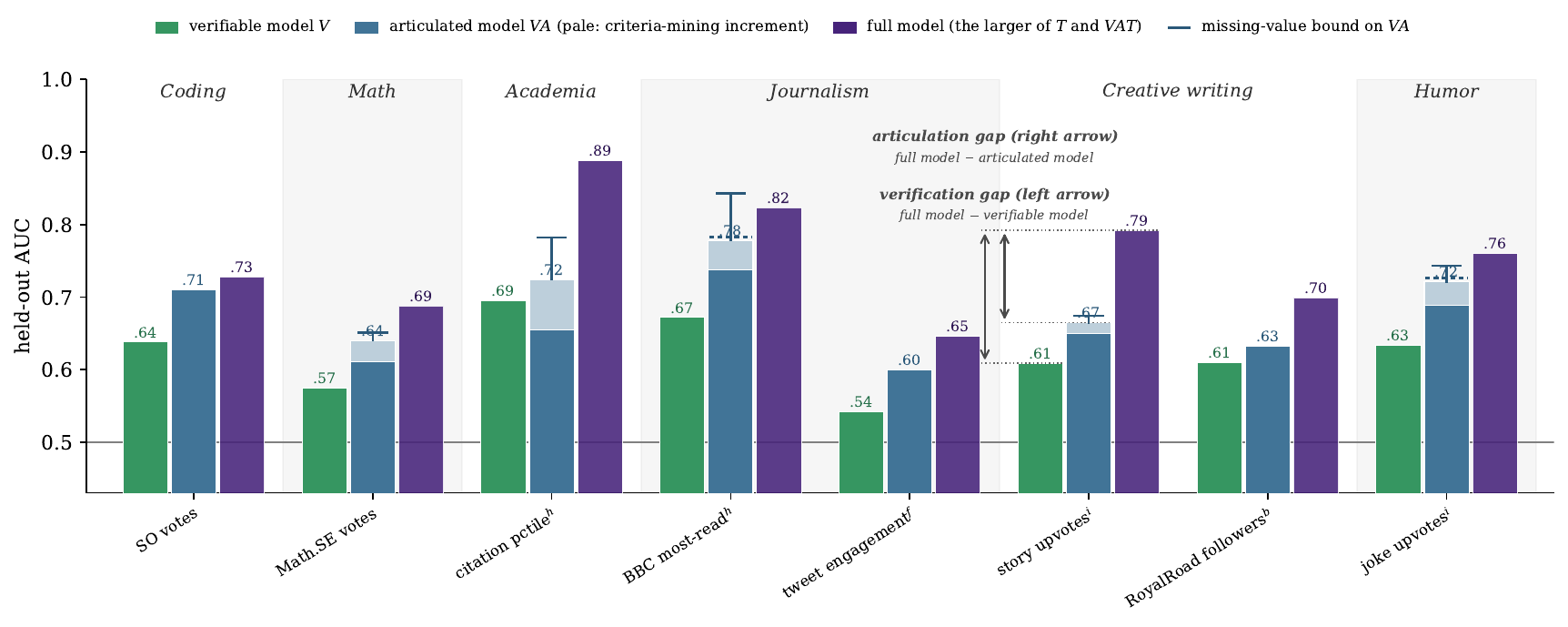}
\caption{\textbf{Community preference across seven fields (eight tasks), unified-frame
instruments.} Per task: verifiers $V$ (green), the articulated model $VA$
(blue) with its criteria-mining increment stacked on top (pale), and
the full model, the larger of $T$ and $VAT$ (purple). Thin solid whiskers are missing-value
\emph{upper bounds}: observed mining gain scaled by the ratio of unmined to
mined criteria mass (Good--Turing, last mining round); bars without whiskers
lack a campaign or a mined column, not evidence of zero remaining gain. Humor and journalism show their bounds like every other task ($.022$ and
$.065$) and \emph{additionally} a dashed tick at the calibrated residual
($\approx.005$): both campaigns closed $\geq 2$ consecutive proposing rounds
with held-out gain below $\varepsilon{=}.005$, the regime in which the
backtest  we report in Appendix~\ref{app:whisker-backtest} validates the realized-gain
estimate. The tick is a point estimate and the whisker a bound. The two campaigns' Good--Turing bounds at closure are $.022$ (humor, mass
$.564{\to}.400$ after strict cross-proposer species merging) and $.065$
(journalism, mass $.492{\to}.608$; the bound \emph{widens} because
unseen-species mass rose even as realized gain stayed at zero, the
mass$\neq$value pattern the backtest anticipates). Tasks grouped by field; the annotated creative-writing spans mark the
verification and articulation gaps.}
\label{fig:vat-community}
\end{figure}

\paragraph{The community ladder (Figure~\ref{fig:vat-community}).} The community ladder covers every community task of Table~\ref{tab:master-results}: eight tasks grouped by field, each an aggregate of anonymous votes (answer votes, joke and story upvotes,
most-read lists, citation percentiles). The community channel has the widest
articulability gaps ($15$--$44\%$ of shared preference), and the figure
also draws the paper's uncertainty accounting on top of the bars: thin whiskers
are \emph{upper bounds} on articulated value still unmined, computed from the
Good--Turing missing mass of criteria species in the final mining round
(Appendix~\ref{app:whisker-backtest}). Every campaign task carries this bound under one
standard; the two campaigns that mined to convergence (humor, journalism)
\emph{additionally} carry a dashed tick at the calibrated terminal residual
($\approx.005$), a point estimate of realized future gain, shown beside the
bound rather than in place of it, because the tick estimates the gain further mining would realize while the whisker bounds the articulated value still unmined. The annotated creative-writing column defines both
gaps geometrically: the step from the verifiable bar to the full-model bar
is the verification gap (what can be said but not checked), and the step from
the articulated bar to the full-model bar is the articulation gap (what is shared
but resists statement).

\paragraph{The gap-anatomy schematic (Figure~\ref{fig:gap-anatomy}).} The
schematic shows the structure behind the main text's decomposition. Panel (a) draws the two
measurable channels from a written input to a binary judgement (verifiable
metrics a program can check and articulable metrics an LLM judge can score),
with the full model bounding total shared preference; panel (b) names
the two differences between adjacent rungs. Every quantitative result in this
paper measures one of these two steps on one task.

\begin{figure}[!htbp]
\centering
\resizebox{.92\textwidth}{!}{\input{figures/fig_gap_anatomy.tex}}
\caption{\textbf{The articulability gap: channels and anatomy.} \textbf{(a)} A
written input $x$ reaches a binary judgement $y$ through two channels we can
measure: verifiable metrics $m_{\mathrm{ver}}(x)$ (checkable by a program) and
articulable metrics $m_{\mathrm{art}}(x)$ (statable in words, scored by an LLM
judge); the full model trained end-to-end on $(x,y)$ bounds the total
shared preference. \textbf{(b)} On a given task, the span from $V$ to $VAT$
is the \emph{verifiability gap} (what can be said but not checked), and the
step from $VA$ to $VAT$ is the \emph{articulability gap} (what is
shared but resists statement), which is our operationalization of tacit preference.
The training-signal spectrum this implies is discussed in \S\ref{sec:related}.}
\label{fig:gap-anatomy}
\end{figure}

\subsection{Standard robustness checks: bootstrap, influence, subsampling and fitting choices}
\label{app:battery}

This section reports the standard checks that a single interval does not show: how much each estimate moves under resampling, under dropping any one group, under less data, and under the fitting choices not swept elsewhere. Table~\ref{tab:battery} reports the first three for the articulated model ($VA$, gradient-boosted head where fit) on every task with per-item held-out scores on disk: the 3,000-draw grouped bootstrap interval; the largest change in AUC when any single group is left out; the standard deviation of the AUC over 30 random halves of the groups; and the within-group pairwise AUC as an alternative readout. Two tasks are sensitive to these checks. On r/Jokes moderator removal one of the 55 temporal strata accounts for much of the articulated model's AUC (leaving that stratum out moves the AUC by $.187$, and the interval spans $.650$--$.903$), so we treat the r/Jokes articulated AUC as unstable. On Wigleaf the articulated gain over $V$ has an interval spanning zero ($[-.028, +.027]$) and its sign flips when any of 112 of the 1,568 stories is left out, so we do not claim an articulated gain on Wigleaf. Everywhere else leave-one-group-out moves the AUC by at most $.003$, and halving the groups changes it by less than $.01$; the within-group pairwise AUC preserves every ordering.

For the three tasks with feature matrices on disk (math.SE bounty, Stack Overflow bounty, creative-writing story upvotes) we also varied what the frozen recipe holds fixed: five random group-to-fold assignments move $V$ and $VA$ by a standard deviation of at most $.0025$; logistic regularization ($C \in \{.1, 1, 10\}$) and gradient-boosting depth and learning rate (depth 2--4, rate.03--.06) move $VA$ within $.003$ on the bounty tasks and $.015$ on creative writing, where the deepest trees are worst; mean, median and zero imputation of missing judge scores differ by at most $.003$. None of these choices moves any gap by more than the intervals already reported.

\begin{table}[!htbp]
\centering\scriptsize
\setlength{\tabcolsep}{2.5pt}
\resizebox{\textwidth}{!}{%
\begin{tabular}{p{3.4cm}rrcccc c}
\toprule
task & $n$ & groups & AUC & boot.\ 95\% CI & LOGO $|\Delta|_{\max}$ & half sd & pairwise \\
\midrule
GitHub PR merged (transition frame) & 7{,}563 & 184 &.630 & [.587,.677] &.012 &.035 &.646 \\
Homepage placement & 12{,}998 & 1{,}229 &.734 & [.722,.746] &.001 &.006 &.708 \\
Kindle Scout & 726 & 3 &.724 & [.714,.790] &.059 &.028 &.715 \\
Math.SE bounty & 23{,}972 & 8{,}278 &.723 & [.716,.729] & n/r &.003 &.711 \\
RoyalRoad followers & 3{,}604 & 149 &.630 & [.611,.648] &.003 &.011 &.655 \\
RoyalRoad market pickup & 1{,}274 & 1{,}274 &.562 & [.530,.593] &.001 &.021 & n/r \\
Stack Overflow accepted & 16{,}001 & 5{,}972 &.717 & [.709,.724] & n/r &.004 &.723 \\
Stack Overflow bounty & 23{,}498 & 7{,}036 &.779 & [.772,.785] & n/r &.004 &.779 \\
Tweet engagement & 31{,}129 & 508 &.600 & [.593,.607] &.001 &.003 &.602 \\
Wigleaf Top 50 & 1{,}568 & 1{,}568 &.613 & [.581,.645] &.002 &.017 & n/r \\
r/Jokes moderator removal & 18{,}879 & 55 &.701 & [.650,.903] &.187 &.108 &.897 \\
r/Jokes removal (v1, retired) & 20{,}543 & 55 &.992 & [.966,.997] &.021 &.012 &.974 \\
\bottomrule
\end{tabular}}
\caption{Robustness of the articulated model's AUC on every task with per-item held-out scores. Columns: items; bootstrap groups; the AUC; its 3,000-draw grouped-bootstrap 95\% interval; the largest absolute change in AUC when one group is left out (n/r where groups exceed 3,000 and the bootstrap covers it); the standard deviation of the AUC over 30 random halves of the groups; and the pairwise AUC computed only within groups (n/r where groups are singletons). The retired v1 moderation task is shown because its interval illustrates the leak the information view flagged.}
\label{tab:battery}
\end{table}

\subsection{Standardized uncertainty for the instrument ladders}
\label{app:standard-cis}

Every task with per-row out-of-fold predictions (47 populations: all
unified-frame tasks built since the current machinery, every judge-sweep
rescore, and the per-community moderation refits) gets the same uncertainty
protocol: $95\%$ percentile confidence intervals from $3{,}000$ bootstrap
draws that resample the task's declared \emph{grouping unit} (question,
outlet-day, subreddit stratum, contest) rather than rows, for the $V$, $A$,
and $VA$ out-of-fold AUCs; the increments $A{-}V$ and $VA{-}V$ are
computed \emph{within each draw}, so their intervals respect row pairing.
Fused ($VAT$) intervals use the same grouped bootstrap on the fused test
split and appear in Table~\ref{tab:master-results} where available.
Table~\ref{tab:standard-cis-table} reports representative tasks.

\begin{table}[!htbp]
\centering
\footnotesize
\resizebox{\textwidth}{!}{%
\begin{tabular}{lrr ccc c}
\toprule
Task & $n$ & groups & $V$ & $VA$ & $\Delta(VA - V)$ \\
\midrule
math.SE bounty & 23{,}972 & 8{,}278 &.691 [.684,.699] &.723 [.716,.730] & [+.028,+.035] \\
SO bounty & 23{,}498 & 7{,}036 &.739 [.732,.746] &.779 [.772,.785] & [+.036,+.044] \\
SO accepted & 16{,}001 & 5{,}972 &.639 [.631,.648] &.717 [.708,.725] & [+.070,+.085] \\
r/Jokes removal & 18{,}879 & 55 &.603 [.542,.857] &.701 [.651,.903] & [+.043,+.111] \\
Kindle Scout & 726 & 3 &.612 [.608,.685] &.724 [.714,.790] & [+.105,+.142] \\
Macro-norm humor (pooled) & 73{,}268 & 5 &.561 [.548,.572] &.714 [.690,.742] & [+.135,+.185] \\
Macro-norm long-form (pooled) & 64{,}788 & 4 &.551 [.528,.586] &.606 [.504,.736] & [$-$.024,+.151] \\
\quad r/books (within) & 14{,}598 & 10 &.651 [.645,.657] &.837 [.832,.842] & [+.178,+.193] \\
\bottomrule
\end{tabular}}
\caption{Standardized uncertainty for the articulated ladder. Rows: tasks; columns: number of items and of bootstrap groups, the 3,000-draw grouped-bootstrap 95\% intervals for $V$ and $VA$, and the interval of their paired within-draw difference.}
\label{tab:standard-cis-table}
\end{table}

First, on every task whose grouping unit is plentiful,
the articulated increment's paired interval excludes zero, so the increments
are not resampling noise. Second, the one interval that \emph{crosses} zero
comes from the pooled long-form moderation task, which resamples only
four communities, and at that grain the pooled articulated increment is
uncertain. This interval is the confidence-interval expression of the
heterogeneity finding of Appendix~\ref{app:subfields}, where all nine
\emph{within}-community increments are positive
($[+.12,+.21]$). Wide intervals on few-group tasks (r/Jokes resamples $55$
temporal strata; Kindle Scout $3$ campaign waves) are the cost of
grouping at the unit where rows are exchangeable; we report these intervals rather
than resampling rows. Full per-task intervals for all $47$
populations are in \texttt{results/standard\_cis.json}.

\paragraph{Predictability of the label.} A gap between two instruments is only interpretable against how much of the label any instrument could recover. For most tasks we have one aggregate outcome per item (a vote total, a merge decision, a pick) and no repeated judgments, so the label's own reliability cannot be measured and $VAT$, the best text instrument, is the operative ceiling: $VAT$ reports how much of the aggregate preference a model that reads the item can predict. Peer review is the exception, because each paper carries several reviewer recommendations. We split each ICLR paper's non-meta reviews at random into two halves ($32{,}765$ papers, $3.8$ reviews per paper, 200 splits) and measured how well the mean recommendation of one half predicts the decision: AUC $.881$ [.879,.884]; all reviewers together reach $.953$, and the two halves agree at Spearman $\rho =.54$ ($.70$ after the Spearman--Brown correction to full length). Our text-only instruments on this task ($VA$ $.654$, $T$ $.777$) sit well below what half the reviewers already know, so the label is not the binding constraint there; the constraint is what can be read from the paper. Across all venues in the review corpus the same quantities are $.777$ (half), $.833$ (all) and $\rho =.56$. The caption-contest crowd ratings, the one other source of repeated judgments we hold, are not used as a reliability ceiling here; the caption tasks appear only in the appendix regret table (Table~\ref{tab:regret-all}), marked $^{\dagger}$.

\paragraph{Beyond sampling: the systematic sensitivity budget.} Sampling
intervals answer ``how much could this number move under resampling'';
the design-choice objections require a different accounting: how much each
\emph{measured} sensitivity could move the two sides of the gap. Table~\ref{tab:systematic-budget}
collects the bounds established in this section and in the companion paper.

\begin{table}[!htbp]
\centering
\footnotesize
\resizebox{\textwidth}{!}{%
\begin{tabular}{llll}
\toprule
Source & Side & Measured bound & Where \\
\midrule
Judge family (7 judges, 6 families) & $A$ $\downarrow$ & $\le.031$ & App.~\ref{app:family-robustness} \\
Judge panel (7-family ensemble) & $A$ $\uparrow$ & $+.005$ & App.~\ref{app:family-robustness} \\
Per-judge template optimization & $A$ $\uparrow$ & $\le.003$ & App.~\ref{app:family-robustness} \\
Per-judge prompt optimization & $A$ $\uparrow$ & $\le.05$ (peer review) & App.~\ref{app:alt-posttraining} \\
Prompt/implementation headroom & $A$ $\uparrow$ & ${\sim}.006$ & parallel forms \citep{companion_prompts} \\
Criteria-mining exhaustion & $A$ $\uparrow$ & asymptote $\pm.006$ + whisker & \S\ref{sec:results}, App.~\ref{app:whisker-backtest} \\
Dense base family & $T$ $\updownarrow$ & $\le.007$ & App.~\ref{app:family-robustness} \\
Dense training data & $T$ $\uparrow$ & $+.006$--$.013$ residual, growing & App.~\ref{app:scaling} \\
Spurious features & $T$ $\downarrow$ & decay to fitted asymptote & App.~\ref{app:dense-decay} \\
\bottomrule
\end{tabular}}
\caption{Systematic (non-sampling) sensitivity budget for the articulated side. Rows: sources of variation we measured; columns: which side of the comparison they move, the measured bound on their effect, and the appendix that measures it.}
\label{tab:systematic-budget}
\end{table}

The articulated side's total measured room for \emph{upward} movement (better judge
panel, better templates, better prompts, more mining) sums to roughly
$.01$--$.02$ AUC on the tasks swept here, rising to $.05$ on peer review under prompts optimized per judge (as we show in App.~\ref{app:alt-posttraining}); the dense side moves by ${\le}.01$ downward under family
swap and spurious discounting and is still growing slowly with data. Every
gap the main text emphasizes exceeds this combined budget; gaps smaller
than ${\sim}.02$ (e.g.\ math.SE bounty) fall within this budget, so we describe them as small
and do not rely on them.

\subsection{Full-model data scaling}
\label{app:scaling}

\begin{figure}[!htbp]
\centering
\includegraphics[width=.96\textwidth]{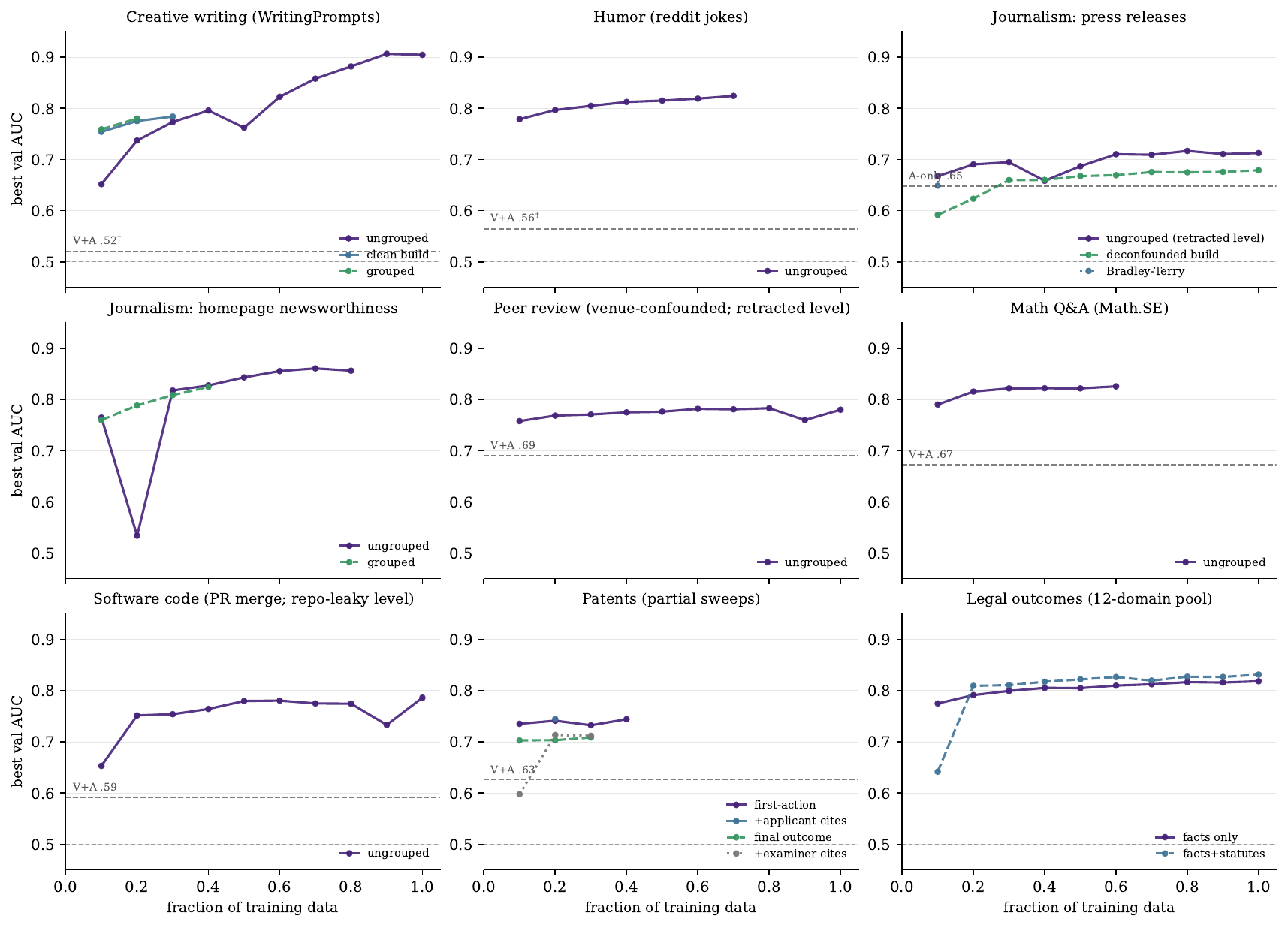}
\caption{Data-scaling of the Llama-3.1-8B LoRA full model (best validation AUC
vs.\ fraction of training data; one panel per sweep family). Most sweeps use
ungrouped splits, so \emph{levels} can be inflated by group identity (dashed
grouped variants show the level without group leakage where run, e.g.\ homepage
newsworthiness $.86\to.82$; the code-review and peer-review ungrouped levels are
known leak artifacts); the \emph{shapes} are the informative part: creative
writing is still climbing at full data while math Q\&A, legal outcomes, and
press releases saturate by 30--60\%.}
\label{fig:scaling}
\end{figure}

\begin{figure}[!htbp]
\centering
\includegraphics[width=.92\textwidth]{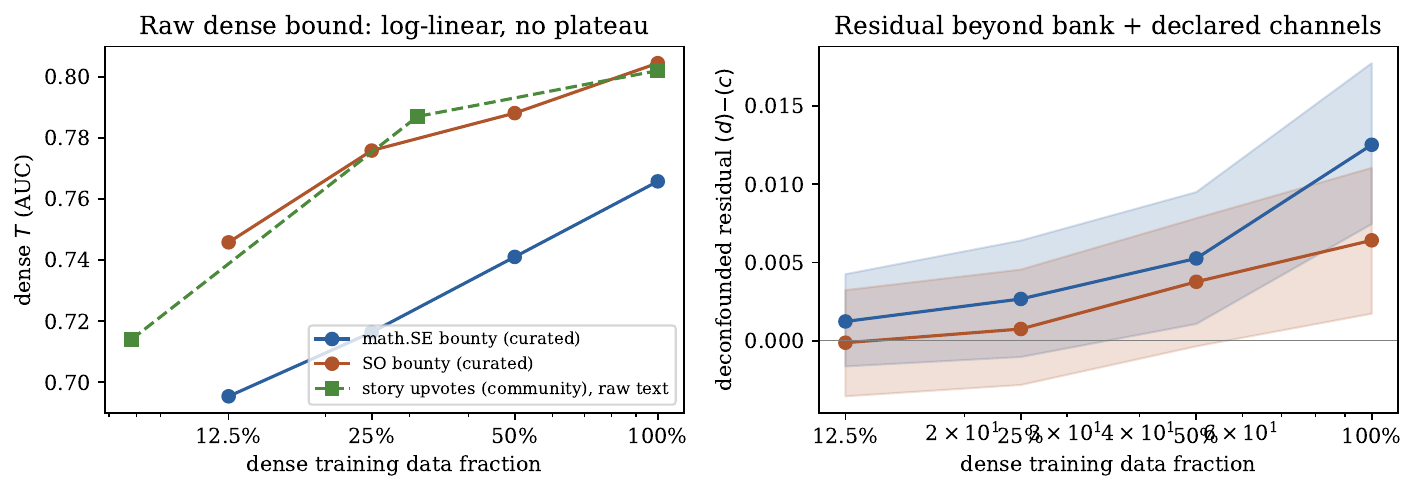}
\caption{\textbf{Residual scaling:} full models ($VAT$) retrained on nested fractions of two curated tasks' training data (identical eval/test rows throughout). Left: the raw dense bound grows log-linearly with data ($\sim$.03 AUC per doubling) with no plateau in range, and the later story-upvotes scale-up (raw text, three training sizes up to the full 76,867 rows, dashed) follows the same slope; a saturating power-law fit leaves the asymptote unidentified. Right: the \emph{deconfounded} residual (the full model's margin over the articulated bank plus declared channels, grouped bootstrap) is indistinguishable from zero at 12.5\% data and grows to $+.013$ [.007,.018] (math.SE bounty) and $+.006$ [.002,.011] (Stack Overflow bounty) at full data: the articulated bank absorbs roughly $90\%$ of what added data gives the dense model, and the unabsorbed part is still growing. Articulated-share estimates are therefore upper bounds that decline slowly with full-model data. The Llama-8B sweep above predates the current corpora and is retained for provenance only.}
\label{fig:residual-scaling}
\end{figure}

The articulability gap is measured against a dense bound that could, in
principle, still be rising with training data, so we measure how both the bound
and the gap move as the full model's training data grows. Figure~\ref{fig:scaling}
shows the historical Llama-8B sweeps (retained for provenance; most predate
the current grouped-split standard), and Figure~\ref{fig:residual-scaling}
the grouped-split rerun on two curated tasks, which separates what added data gives the
full model from what remains after the articulated bank and declared
channels are accounted for.

\subsection{Measuring Articulability Gaps in Terms of Information}
\label{app:info-view}

The main text reports AUC-based quantities because they are familiar and
threshold-free, but AUC differences are not a
decomposition of information. This section checks that the paper's ``signal''
language is backed by a genuine information decomposition. For binary outcomes, mutual information
is exactly the expected log-loss reduction of calibrated probabilities,
$I(S;y) = H(y) - H(y \mid S)$, and unlike AUC it decomposes additively by
the chain rule: $I(M, X; y) = I(M; y) + I(X; y \mid M)$, so ``the
articulated increment beyond the verifiable tier'' and ``the residual beyond
the measurements'' are well-defined information quantities.

\paragraph{Method.} For every task with per-row out-of-fold predictions
(the 47 populations of Appendix~\ref{app:standard-cis}), each model's
predictions are calibrated by isotonic regression fit on held-out
\emph{groups} (grouped cross-validation, so calibration never sees the rows
it converts), and information is computed as entropy minus cross-entropy,
debiased by a within-group permutation null. Where score distributions
shift across a task's grouping unit (detectable as negative debiased
bits, the calibration analogue of the era shifts our grouped AUCs already
condition on), calibration is instead fit \emph{within} each stratum and
we report the conditional information $I(S; y \mid G)$, the bits analogue
of the paper's stratified AUC readouts. Two small tasks fail both calibration routes
(three campaign waves; sparse repo groups) and report ranks only.

\paragraph{Results.} The information view gives the same orderings as the AUC view.
Within tasks, the model ordering ($V$ vs.\ $A$ vs.\ $VA$) agrees in $40$
of $47$ populations; of the seven exceptions, four are exact ties
(models within $.005$ bits), one is the \emph{retired} leaky v1 moderation
task (where the information view flags the leak at $1.25$ bits on the
$V$ model), and two are the small-task calibration
failures above. Across tasks, the ranking of verifiable shares under bits
agrees with the ranking under AUC at Spearman $\rho =.93$
($p < 10^{-4}$, $n{=}43$ populations with a verifiable model and positive articulated bits; 95\% CI $[.81,.98]$, bootstrap over populations; $[.72,.98]$ when the ten source tasks are resampled instead). The paper's substantive patterns replicate on
the bits scale: the pooled-versus-within-community heterogeneity result
(pooled moderation $VA$ $.07$--$.12$ bits vs.\ within-community
$.13$--$.28$), and the judge-capability ordering of the family sweep
(Mistral-7B $.037 \to$ phi-4 $.056 \to$ Qwen-14B $.065 \to$ Llama-70B
$.070 \to$ record $.078$ articulated bits on math.SE bounty, identical
order to the AUC ladder). Estimates carry the usual caveats: plug-in MI
bias is handled by the permutation null but small strata make absolute
levels less reliable than orderings, and all quantities are lower bounds
(imperfect models, imperfect calibration). Artifacts:
\texttt{results/mi\_bits\_all.json},
\texttt{results/mi\_bits\_jokes\_conditional.json}.

\subsection{Verifiability tracks articulability across tasks}
\label{app:artic-verif}

\begin{figure}[!htbp]
\centering
\includegraphics[width=.62\textwidth]{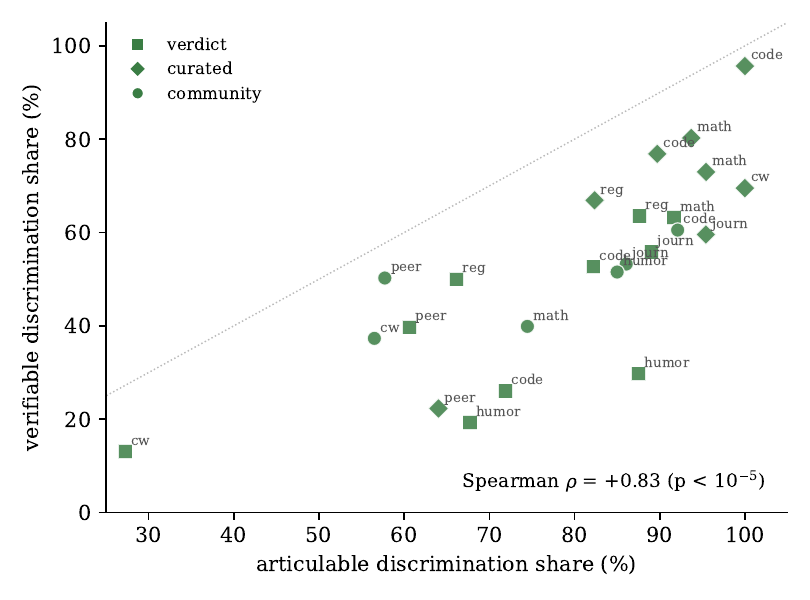}
\caption{Verifiable vs.\ articulable discrimination share, one point
per task of the earlier 24-task set ($n{=}24$, including the withdrawn patents task; marker = preference channel). The dotted diagonal is
verifiable $=$ articulable (checks carry everything statable). Spearman
$\rho =.83$ ($p < 10^{-5}$) on the plotted points; on the 28 tasks of Figure~\ref{fig:concept-row}, $\rho =.63$ (text). Humor sits far below the trend: highly
articulable, weakly verifiable. }
\label{fig:artic-verif}
\end{figure}

The two nested shares of the decomposition are strongly related across tasks:
tasks with larger articulation gaps also support fewer verifiable checks
(Figure~\ref{fig:artic-verif}). With both shares on the common
denominator of each task's shared preference (the convention of
Figure~\ref{fig:concept-row}), the articulable and
verifiable shares correlate at Spearman $\rho =.63$ ($p<.001$,
$n{=}28$ tasks, 95\% CI $[.29,.85]$, bootstrap over tasks with 3,000 resamples); on the 23 tasks that carried all three models before the five tasks added last (Kindle Scout, RoyalRoad followers, r/Jokes removals, tweet engagement, Style Invitational), the correlation is $\rho =.84$. Equivalently, the articulation gap ($100{-}$articulable
share) anti-correlates with verifiability at $\rho = -.63$. Part of this
association is structural: the shares are nested, so a task's articulable
share caps its verifiable share. When that coupling is removed by measuring instead
\emph{how much of what a field can state it can also check} (the ratio
$(V-.5)/(VA-.5)$), the association weakens to $\rho =.23$ ($p =.24$, $n{=}28$, 95\% CI $[-.18,.60]$; $\rho=.44$, $p=.03$, on the earlier 23 tasks).
The weaker association is suggestive but not significant at this task count: more-articulable fields
may verify a somewhat
larger fraction of what they articulate. The two quantities are correlated
but not redundant.
 What a field
cannot state, it \emph{a fortiori} cannot reduce to checks. The
converse fails in one place: humor articulates $85\%$ of its
shared preference (median) while verifying only $30\%$, the largest
stated-but-uncheckable band in the benchmark. For reward design,
the on-diagonal fields (code, math, law) are where
verifiable-reward training has most room; the below-diagonal band is
where rubric- and constitution-based training can reach signal that
verifiable rewards cannot; and the low-articulability corner (peer
review) is where neither articulated form captures what the dense model
finds.

We run two further checks on this correlation. First, the 28 tasks are not independent (several share texts, judges or banks), so we recomputed $\rho$ leaving out each domain in turn: it ranges from $.56$ (without code) to $.83$ (without humor); on the earlier 23 tasks the range was $.80$ (without code) to $.86$ (without law). Second, the two shares share a denominator and are nested ($V \le VA$), which induces a correlation on its own. On the earlier 23 tasks, we simulated the nested null by keeping each task's $V$ and denominator and shuffling the articulated increment $VA-V$ across tasks: the null $\rho$ is $.67$ (95\% range $.48$--$.82$), and the observed $.84$ exceeds it ($p =.012$). Shuffling $V$ across tasks instead gives $\rho =.24$. The correlation therefore does not reduce to coupling, although coupling alone produces about $.67$ of it, so the observed $\rho$ is to be compared with $.67$ rather than with zero. The second correlation (articulable share against verifiable-over-articulated) is fragile under the same check: on the 28 tasks, leaving out code or math drops it to $.07$ and $.12$, leaving out humor raises it to $.37$.

\section{\texorpdfstring{Why some fields carry more tacit preference: a wide screen}{Why some fields carry more tacit preference: a wide screen}}
\label{app:field-tacitness}

Section~\ref{sec:res-gaps} reports that tacit shares differ by field and by channel. This appendix examines which properties of a field make its preferences hard to state and screens four families of explanation: the language of the texts, how many people decide and from how many institutions, what the deciders were trained to do, and what the judgment is about. The outcome throughout is the tacit share of Table~\ref{tab:master-results}, over the nineteen tasks whose texts we hold locally. Two further subsections relax the pooling of the estimates, by subfield and by organization, and stress-test the pipeline on tasks where it should return chance.

Two facts constrain the analyses below. First, the outcome has two clusters: peer review ($.39$ over three tasks) and creative writing ($.39$ over two) sit far above humor ($.15$), code ($.14$), notice-and-comment ($.14$), math ($.12$) and press releases ($.11$). Any property that separates those two fields from the other five will predict the tacit share at $n{=}19$; distinguishing among such properties requires new fields or within-field contrasts. Second, this is a screen and not a test: roughly forty covariates were tried, the fields cluster, and no single number here would survive a correction for multiple comparisons. We report the screen because the properties that fail to predict the tacit share are as informative as those that do.

\subsection{Jargon does not predict tacit preference}
\label{app:field-jargon}

The most natural explanation is that specialised fields use specialised vocabulary, and that a rubric written in plain language cannot capture what that vocabulary expresses. We measure jargon as \citet{lucy2023words} do. Their type NPMI scores how specific a word is to a field,
\[
\mathcal{T}_f(t) \;=\; \frac{\log\!\big(P(t \mid f)/P(t)\big)}{-\log P(t,f)},
\]
where $f$ is a field's corpus and $P(t)$ is the word's probability in a background of all our corpora together with general English. Following them, we score only words appearing at least twenty times in the field, count a token as discipline-specific when $\mathcal{T}_f(t) > 0.1$ (their cutoff, with $0.2$ as their robustness check), and give each item the fraction of its tokens that are discipline-specific. Their sense-level measure, which induces word senses with a masked language model before scoring them, is not reproduced here, because it costs roughly ninety GPU hours in the original and refines the type-level measure, which already shows no relation to the tacit share.

The type NPMI picks out field vocabulary. Its highest-scoring words are notation and procedure: \emph{frac}, \emph{sqrt}, \emph{mathbb} in mathematics; \emph{import}, \emph{def}, \emph{github} in code; \emph{docket}, \emph{NPRM}, \emph{airworthiness} in notice-and-comment; \emph{adversarial}, \emph{generative}, \emph{outperforms} in peer review. Fiction's most distinctive words are contractions and the verbs of narration, \emph{sighed}, \emph{nodded}, \emph{stared}.

The jargon fraction does not predict the tacit share. Across the nineteen tasks the rank correlation is $-.13$ ($p{=}.60$); across the seven fields the jargon fraction gives $+.11$ and mean word distinctiveness $+.39$, neither significant at $n{=}7$. The pattern is not monotone but crossed: mathematics is the most jargon-dense field and has nearly the smallest gaps, creative writing is the least jargon-dense and has the largest, and peer review is dense \emph{and} large. Within tasks the jargon share is flat as well, which is the most direct form of the test because no field confound can enter: splitting each task's items into terciles by their own jargon share does not move the gap in a consistent direction (pull requests $.040/.045/-.032$, ICLR $.160/.091/.135$, Math.SE accepted answers $.095/.084/.108$), and an item's jargon share does not predict its label in any task (AUC $.40$--$.56$).

\begin{table}[!htbp]
\centering\footnotesize
\begingroup
\resizebox{\textwidth}{!}{%
\begin{tabular}{lrrrl}
\toprule
field & tacit share & jargon fraction & distinctiveness & most discipline-specific words \\
\midrule
peer review & $.392$ & $.406$ & $.118$ & llms, neural, adversarial, outperforms \\
creative writing & $.386$ & $.179$ & $.052$ & wp, didn, sighed, nodded, stared \\
humor & $.153$ & $.359$ & $.099$ & contest, prompt, entry, hashtag \\
code & $.139$ & $.408$ & $.074$ & import, com, self, github, def \\
notice and comment & $.135$ & $.259$ & $.089$ & docket, FAA, NPRM, airworthiness \\
mathematics & $.121$ & $.486$ & $.062$ & frac, math, sqrt, mathbb, infty \\
press releases & $.109$ & $.175$ & $.073$ & (company names), inauthentic \\
\bottomrule
\end{tabular}}
\caption{Scholarly jargon, measured as in \citet{lucy2023words}, against the tacit share. Jargon fraction is the mean share of an item's tokens with field-level type NPMI above $0.1$; distinctiveness is the mean of $\max(\mathcal{T}_f(t),0)$ over the field's vocabulary. Rank correlation with the tacit share is $+.11$ for the fraction and $+.39$ for distinctiveness over seven fields, and $-.13$ ($p{=}.60$) over the nineteen tasks. The two largest-gap fields sit at opposite ends of the jargon scale.}
\label{tab:field-jargon}
\endgroup
\end{table}

\subsection{The rest of the language of the texts}
\label{app:field-language}

Seven further surface properties were computed on 1{,}500 sampled items per task: the share of tokens rarer than about one per million in general English, the share absent from the general-English list, mean word length, lexical diversity as a moving-average type--token ratio, gzip compressibility, Flesch--Kincaid grade, and a novelty score, namely the AUC with which an item's distance from its task's centroid predicts the label.

\begin{table}[!htbp]
\centering\footnotesize
\begingroup
\resizebox{\textwidth}{!}{%
\begin{tabular}{lrrrrl}
\toprule
property & all ($n{=}19$) & community (6) & verdict (6) & curated (7) & with the object control \\
\midrule
rare-word share & $-.39$ & $-.26$ & $+.09$ & $-.75$ & $+.015$ ($t{=}0.5$) \\
out-of-vocabulary share & $-.43$ & $-.37$ & $+.20$ & $-.68$ & $-.003$ ($-0.1$) \\
mean word length & $+.35$ & $+.03$ & $+.94$ & $+.82$ & $+.027$ ($1.1$) \\
lexical diversity & $+.53$ & $+.14$ & $+.26$ & $+.71$ & $-.088$ ($-3.1$), sign flips \\
compressibility & $-.07$ & $-.94$ & $-.31$ & $+.21$ & $-.092$ ($-6.2$), length proxy \\
reading grade & $+.21$ & $+.09$ & $+.83$ & $+.46$ & $+.044$ ($2.0$) \\
novelty AUC & $+.23$ & $+.31$ & $+.66$ & $+.07$ & $-.021$ ($-0.9$) \\
median length in words & $+.27$ & $+.83$ & $+.31$ & $+.14$ & $+.051$ ($2.4$) \\
\bottomrule
\end{tabular}}
\caption{Surface properties of the texts against the tacit share: Spearman $\rho$ over all nineteen tasks and within each channel, and in the last column the coefficient per standard deviation once the composition-versus-answer indicator of \S\ref{app:field-object} is in the model. Only length survives that control with its sign intact.}
\label{tab:field-language}
\endgroup
\end{table}

Length is the one property that survives the composition-versus-answer control, and within the community channel length is strong on its own ($\rho{=}.83$ over six tasks, from fifteen-word jokes to six-hundred-word stories): the longer the text, the larger the tacit share. Reading grade tracks the gap within verdict labels, from ICLR abstracts at grade seventeen down to Stack Overflow answers at ten, but fiction is written at grade five and has the largest gap of all, so grade is a genre signature and not a measure of what the deciders were taught. Lexical diversity reverses sign under the control, and compressibility is a length proxy on short texts. No label rewards unusualness within its own corpus: the novelty AUC is between $.37$ and $.56$ everywhere, and its maximum is on a small-gap task.

\subsection{How many people decide, and from how many institutions}
\label{app:field-deciders}

Table~\ref{tab:field-participants} reports the participants behind every task, counted from the raw corpora; for each count we record the source file and column. Two cautions apply before any correlation is read off the table. First, on the community tasks the number of deciders per item \emph{is} the label: upvotes, followers and citation counts are what the binary outcome thresholds, so correlating them with the gap would bin the outcome by itself. We therefore report them as label-derived and exclude them from any inference, the same rule that forced a retraction of an earlier within-score-band analysis. Second, where a count is independent of the label, the count mostly restates the channel: one asker accepting an answer, one editor picking a contest winner, four reviewers scoring a paper, a crowd voting. The cross-task analysis already found that the number of people a label aggregates predicts the gap at $\rho{=}.46$, and already concluded that this correlation is the channel ordering in another form.

The institutional question cannot be answered with these corpora. Affiliation exists in the data for exactly the fields with small gaps: code review carries $8{,}982$ distinct GitHub organisations over $15{,}298$ repositories, with a top-ten share of $10.6\%$ and a Herfindahl index of $.002$, so code review is institutionally very diffuse; notice-and-comment carries $1{,}205$ named commenting organisations against twenty-two deciding agencies, the most concentrated decider side we measure (the top ten agencies take $82.8\%$ of the true comment volume); press releases carry $11{,}075$ issuing companies against seventeen tracked outlets. Mathematics competitions reduce to two administering bodies. But the two fields with the largest gaps have no affiliation field at all: no peer-review file we hold, including the OpenAlex citation join, carries an author institution, and Reddit and RoyalRoad have no institutions by construction. The fields where institutional tightness would matter most are therefore exactly the ones where tightness is unmeasured, and the four fields where it is measured span institutional structures from two bodies to nine thousand while their tacit shares sit within $.03$ of each other. Extracting affiliations for the peer-review tasks is the one direct way to test the institutional question; we have not done so.

\begin{table}[!htbp]
\centering\footnotesize
\begingroup
\resizebox{\textwidth}{!}{%
\begin{tabular}{llrlll}
\toprule
task & tacit & who decides & deciders per item & producers & institutions \\
\midrule
story upvotes & $.435$ & anonymous crowd & $219$ mean / $47$ median upvotes$^{\ell}$ & not in data & none by construction \\
citation percentile & $.423$ & the citing community & $59.7$ / $10$ citing works$^{\ell}$ & not in data & not in data \\
ICLR accept/reject & $.394$ & reviewers + area chair & $3.81$ / $4$ reviews & $62{,}442$ authors, $4.78$/paper & not in data \\
spotlight/oral & $.360$ & programme committee & committee absent from data & $60{,}689$ authors, $5.05$/paper & not in data \\
RoyalRoad followers & $.337$ & anonymous readers & $21.4$ / $2$ followers$^{\ell}$ & not in data & none by construction \\
pull request merged & $.281$ & repository insiders & $37.7$ / $29$ review comments & not in data & $8{,}982$ orgs, HHI $.002$ \\
Math.SE votes & $.255$ & anonymous voters & net score only$^{\ell}$ & not in data & not in data \\
Style Invitational & $.185$ & one columnist & $1$ (the editor) & $1{,}962$ entrants (heuristic) & none in data \\
joke upvotes & $.150$ & anonymous crowd & score only$^{\ell}$ & $322{,}501$ authors (raw scrape) & none by construction \\
N\&C rule change & $.147$ & a federal agency & $3{,}751$ / $28$ comments per docket & $1{,}205$ orgs named & $22$ agencies; top-10 $82.8\%$ \\
\#HashtagWars & $.125$ & show writing staff & $114$ / $124$ tweets per hashtag & not in data & none in data \\
N\&C response & $.123$ & a federal agency & as above & $994$ orgs named & $22$ agencies \\
SO accepted & $.117$ & the asker & $1$ & $10{,}064$ answerers & not in data \\
editorial pickup & $.109$ & news outlets & $1.21$ outlets$^{\ell}$ & $11{,}075$ companies & $17$ outlets; top-10 $92.1\%$ \\
Math.SE bounty & $.101$ & the bounty offerer & $1$ & not in data & not in data \\
Math.SE accepted & $.083$ & the asker & $1$ & not in data & not in data \\
SO votes & $.079$ & anonymous voters & net score only$^{\ell}$ & $10{,}064$ answerers & not in data \\
SO bounty & $.077$ & the bounty offerer & $1$ & not in data & not in data \\
AoPS editorial & $.046$ & an editorial board & canonical solution per problem & $4{,}369$ posters (upstream) & $2$ administering bodies \\
\bottomrule
\end{tabular}}
\caption{Participants behind each task, counted from the raw corpora. $^{\ell}$ marks a count that is label-derived: the binary outcome is a threshold on it, so it must not be correlated with the gap. Producer and institution counts are floors where identity is sparsely recorded (notice-and-comment names an organisation on $13$--$16\%$ of comments; the Style Invitational entrant count comes from a name-extraction heuristic with a $73\%$ match rate). Affiliation is absent from every peer-review file we hold and does not exist for the Reddit and RoyalRoad tasks, which is why the institutional-tightness question is left open rather than answered.}
\label{tab:field-participants}
\endgroup
\end{table}

Five properties of the fields were coded by hand, after the gaps were known, so they describe this sample rather than test a hypothesis. Years of formal training on the \emph{producer} side do not track the tacit share at all ($\rho{=}.00$). Years of training on the \emph{decider} side are not monotone but U-shaped: deciders with no formal training average $.26$, deciders with four to seven years average $.11$--$.16$, and the most trained deciders, ICLR reviewers and the citing community, average $.39$. Whether the decider works in what \citet{kahneman2009conditions} call a high-validity environment, one that returns quick and unambiguous feedback, correlates at $-.69$, and the presence of a formal correctness criterion such as a proof or a test at $-.60$; but humor, press releases and notice-and-comment have no such criterion and small gaps, so a formal criterion is sufficient for a small gap and not necessary. Decider type reproduces the channel ordering: audience $.31$, peer $.19$, gatekeeper $.18$, institution $.14$.

\subsection{What the judgment is about, and a direct test}
\label{app:field-object}

The clearest split is a property of the judgment, not of the texts or the deciders: whether the label evaluates a whole composition for its own sake, such as a paper, a story or a joke, or an answer to a posed problem, such as a Stack Exchange answer, a comment on a proposed rule, a pull request or a press pitch. Compositions average a tacit share of $.30$ over eight tasks and answers $.13$ over eleven, a one-sided Mann--Whitney $p$ of $.0006$ that is post hoc and should be read as a description. The only compositions with small gaps are the three humor tasks, all under thirty-one words. The code field contains the contrast within itself: a pull request, judged as a contribution to an active codebase, sits at $.28$, while Stack Overflow answers to a stated question sit between $.08$ and $.12$.

We hypothesize that item-intrinsic judgments carry their criteria with them, while corpus-relative judgments require having read the corpus, which the full model has done and a criteria bank has not. The hypothesis requires no new judging to test. We gave the articulated model a \emph{memory of the corpus} and no new criteria: for each item, the mean label of its twenty-five and five nearest training-fold neighbours under tf-idf cosine, plus its mean similarity to positives minus negatives, all computed inside the same grouped five-fold split the articulated model uses, with neighbours from an item's own group excluded on the training side, then fitted with the same gradient-boosted recipe. Table~\ref{tab:corpus-memory} reports the share of each task's gap that the corpus-memory features close.

\begin{table}[!htbp]
\centering\footnotesize
\begingroup
\resizebox{\textwidth}{!}{%
\begin{tabular}{lrrrrrrrr}
\toprule
task & $n$ & full model & articulated & memory alone & articulated + memory & gap & share closed \\
\midrule
citation percentile & $478$ & $.884$ & $.627$ & $\mathbf{.739}$ & $.746$ & $+.257$ & $\mathbf{+46\%}$ \\
\#HashtagWars & $924$ & $.732$ & $.553$ & $.546$ & $.542$ & $+.179$ & $-6\%$ \\
story upvotes & $7{,}008$ & $.792$ & $.658$ & $.587$ & $.675$ & $+.134$ & $+12\%$ \\
ICLR accept/reject & $1{,}244$ & $.777$ & $.647$ & $.574$ & $.657$ & $+.130$ & $+7\%$ \\
Math.SE accepted & $2{,}600$ & $.644$ & $.549$ & $.497$ & $.552$ & $+.095$ & $+4\%$ \\
Math.SE votes & $2{,}326$ & $.654$ & $.576$ & $.511$ & $.572$ & $+.077$ & $-6\%$ \\
press pickup & $605$ & $.774$ & $.699$ & $.445$ & $.671$ & $+.076$ & $-37\%$ \\
joke upvotes & $3{,}163$ & $.747$ & $.677$ & $.496$ & $.673$ & $+.070$ & $-6\%$ \\
spotlight/oral & $1{,}571$ & $.594$ & $.529$ & $.501$ & $.527$ & $+.065$ & $-3\%$ \\
N\&C cosigner agreement & $1{,}009$ & $.603$ & $.549$ & $.598$ & $.577$ & $+.054$ & $+50\%$ \\
pull request merged & $11{,}452$ & $.727$ & $.684$ & $.493$ & $.679$ & $+.043$ & $-12\%$ \\
N\&C responded & $1{,}904$ & $.833$ & $.793$ & $.735$ & $\mathbf{.852}$ & $+.040$ & $+150\%$ \\
AoPS editorial & $5{,}202$ & $.781$ & $.750$ & $.605$ & $.747$ & $+.031$ & $-10\%$ \\
N\&C rule change & $1{,}417$ & $.624$ & $.593$ & $.588$ & $.605$ & $+.031$ & $+39\%$ \\
\bottomrule
\end{tabular}}
\caption{The corpus-memory ablation on every task whose text we hold, ordered by gap size. The articulated model is refitted with three added features that encode only the labels of an item's nearest training-fold neighbours, so nothing new is articulated; the articulated column is this ablation's own refit and differs from Table~\ref{tab:master-results} by a few points. Share closed is $(\mathrm{AUC}_{VA+\mathrm{mem}}-\mathrm{AUC}_{VA})$ over the gap, and it is unstable wherever the gap is small: the three notice-and-comment rows are ratios on denominators of $.03$--$.05$, and on notice-and-comment responses the augmented articulated model passes the full model outright. Across the fourteen live tasks the share closed does not correlate with gap size ($\rho=+.02$), nor does the memory-alone AUC ($\rho=-.07$).  On citation percentile a memory of the corpus alone ($.739$) beats the entire criteria bank ($.627$) and closes $46\%$ of the gap; on joke upvotes it closes nothing. Remaining tasks were still running when this table was made.}
\label{tab:corpus-memory}
\endgroup
\end{table}

The corpus-memory ablation gives one strong case and no general law. On citation percentile, where the label is by construction a paper's standing relative to everything else published in its venue and year, a memory of the corpus does more than the entire bank of stated criteria ($.739$ against $.627$) and closes $46\%$ of the largest gap we measure. Nothing else in our battery closes that much of that gap. Joke upvotes is the opposite case: a punchline either works or it does not, neighbour labels are at chance, and the corpus-memory features add nothing. The pattern does not, however, generalise across tasks. \#HashtagWars has the second-largest gap and gains nothing, and the three notice-and-comment tasks have the smallest gaps and the largest proportional closures, including one where the augmented articulated model passes the full model. Across tasks the share closed is uncorrelated with the size of the gap. Two factors weaken the small-gap rows in any case: a ratio on a $.04$ denominator is noise, and neighbour labels can encode a group-level base rate rather than corpus-relative taste, which is the likely explanation on notice-and-comment, where neighbours share an agency and its response habits even though the fold structure holds out the docket. Corpus memory is therefore one mechanism that dominates on at least one task, not a general account of the gap, and because term-frequency neighbours are a crude memory, every share closed here is a lower bound on what a retrieval model could close.

The citation row sits alongside one within-task finding reported in Appendix~\ref{app:gap-battery}: the citation gap is three times larger on decade-old papers, whose labels have drifted furthest from anything their abstract states. Read together with the null on jargon and the U-shape on decider training, the two citation results suggest that where enculturation matters it is immersion in a field's output rather than formal training, which produces explicit criteria and acts in the opposite direction. We treat this suggestion as a screen result and not a test.

Five designs would turn this screen into a test: within-field composition-versus-answer contrasts in peer review and creative writing, matching the one code already provides; a retrieval-augmented articulated model on every task in the grid, rather than the five reported here; truncation curves, training the full model on the first $k$ words, to separate reading bandwidth from gestalt; new fields with highly trained deciders and instrumental labels, such as patents, grants and legal outcomes, where the U-shape above predicts small gaps despite high education; and a blind re-coding of the field properties by someone who has not seen the gaps.

\subsection{The social structure of articulability: subfields and organizations}
\label{app:subfields}
\label{sec:res-subfields}
\label{app:orgsize}

The gap estimates of the main text are pooled fits, and the subfield checks of \S\ref{sec:res-frontier} (Q4) hold the criteria vocabulary fixed. This section examines the two ways pooling can be relaxed.
 Horizontally, a field decomposes into
\emph{subcommunities} (subreddits within a platform, labels within a
rulemaking docket, subfields within a discipline), and part of what a pooled
fit reports as tacitness is un-modeled heterogeneity between
their criteria. Vertically, the deciding unit itself has \emph{organizational
structure} (headcount, decision throughput), and on the one field where that
structure is observable, the two layers of the verifiable model track different
parts of it. Both analyses show that articulability depends on the
social unit doing the judging: who shares criteria with
whom, and how much practice the deciding body has had at defending its
decisions.

\paragraph{Shared cores and subfield-specific criteria
(Figure~\ref{fig:infilling}).} Preference vocabularies are heavily shared across
related decisions within a field and much less so across subfields:
per-criterion rank agreement is $\rho =.76$ between the regulatory outcome and
agreement labels, against $\rho =.47$ between academic curation and citations ($n{=}171$ columns, 17 surface features and 154 criteria, each scored by its univariate AUC on the 2,202 papers that carry both labels; 95\% CI $[.28,.62]$, bootstrap over columns with 3,000 resamples).
What differs is the weighting and a small set of subfield-specific criteria: the
regulatory outcome label favours an evidence-and-actionability cluster while the
agreement label favours a burden-and-flexibility cluster. Figure~\ref{fig:infilling}
sketches the discovery procedure this suggests: mining on a
subfield's residual (using labels only on the training and monitoring splits; the test split is never touched by any discovery decision), with a candidate kept only if it adds held-out lift within
that subfield beyond the shared core. We present the procedure as a schematic, since the
empirical work so far establishes the signatures rather than the procedure.

\begin{figure}[!htbp]
\centering
\resizebox{.8\textwidth}{!}{\input{figures/fig_infilling.tex}}
\caption{\textbf{Metric infilling: fields share an articulated core; subfields introduce their own criteria.} Schematic of the discovery procedure: a field carries a shared metric set ($m_1$--$m_3$); mining on a subfield's residual (using labels only on the training and monitoring splits; the test split is never touched by any discovery decision) proposes candidate criteria, and a candidate is kept only if it increases scores on a held-out set. Preference vocabularies are heavily shared across related decisions (per-metric rank agreement $\rho{=}.76$ between regulatory outcome and agreement; $\rho{=}.47$ between academic curation and citations) while the \emph{weights} and a small set of subfield-specific criteria differ (metric names schematic).}
\label{fig:infilling}
\end{figure}

\paragraph{Why the metric tree failed.} The metric tree of Appendix~\ref{app:negative} assumed that subfields differ in their weights over a shared criteria vocabulary, and it scored below the flat bank; the infilling signatures above put the subfield structure in the vocabulary instead. We discuss the metric tree and the infilling signatures together in Appendix~\ref{app:negative}.

\paragraph{Organizational scale and the verifiable model.}
Where subcommunities decompose a field horizontally, organizational scale
stratifies its deciders vertically, and the regulatory field is the one place
we observe both structure and outcome at the level of the deciding unit.

\begin{figure}[!htbp]
\centering
\includegraphics[width=.9\textwidth]{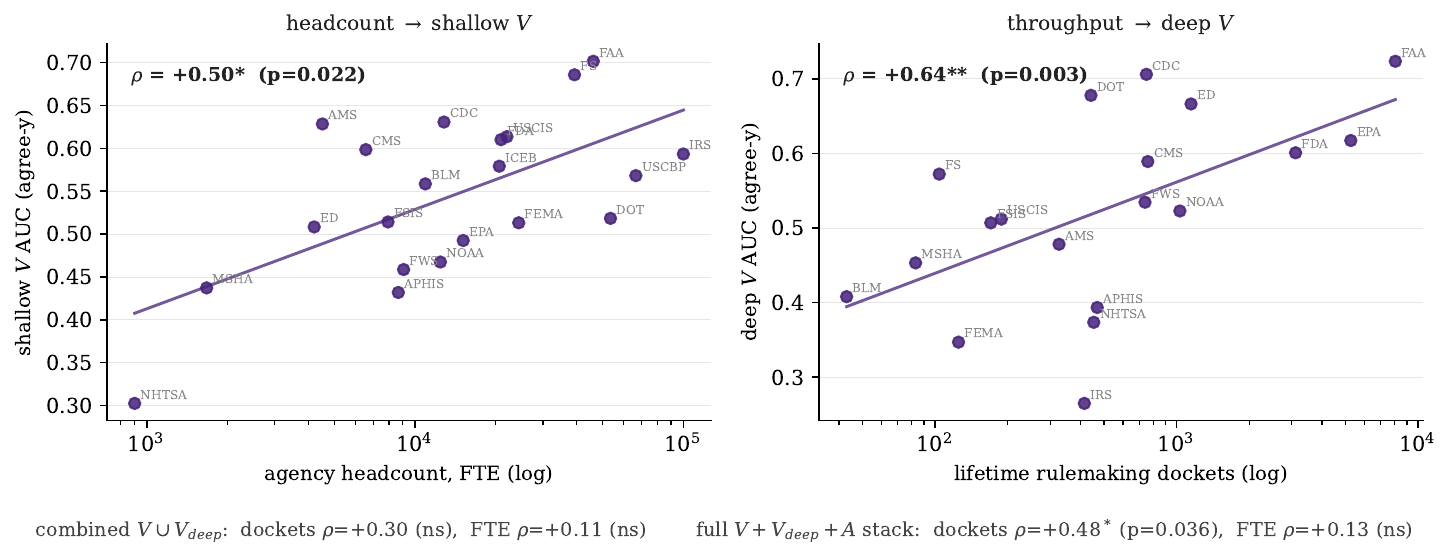}
\caption{\textbf{Organizational scale and the two layers of verifiability
(regulatory rulemaking, agency-agreement label).} Each point is a federal
agency. \emph{Headcount} predicts how well shallow surface verifiers work
($\rho{=}.50$, $p{=}.022$) while \emph{practice} (lifetime rulemaking-docket
throughput) predicts deep verifier signal ($\rho{=}.64$, $p{=}.003$; robust to
excluding FAA and to partialling sampled-docket count); the two cross-pairings
are null. Merging the layers dilutes the throughput signal ($\rho{=}.30$, ns)
while the full $V{+}V_{deep}{+}A$ stack retains it ($\rho{=}.48$, $p{=}.036$).
The GitHub counterpart, where an earlier analysis found the \emph{opposite}
sign (repos with $\le$5 maintainers more internally consistent, $p{=}.035$,
Mann--Whitney), rests on artifacts that were
since lost and a chance-floor reconstruction; its rebuild is queued and it is
not shown.}
\label{fig:orgsize}
\end{figure}

On the one field where organizational structure is observable at the level of
the deciding unit, federal agencies, the two layers of the verifiable model
dissociate (Figure~\ref{fig:orgsize}). Agency headcount predicts how well
shallow surface verifiers work ($\rho =.50$, $p =.022$) but not deep verifier
signal; lifetime rulemaking-docket throughput predicts deep verifier signal
($\rho =.64$, $p =.003$, $n = 19$) but not shallow, and survives excluding the
largest agency and partialling out sampled-docket count. Merging the two layers
dilutes the throughput effect to non-significance ($\rho =.30$) while the full
$V{+}V_{\mathrm{deep}}{+}A$ stack retains it ($\rho =.48$, $p =.036$). We had
intended a GitHub counterpart, where an earlier analysis found the opposite sign
(repositories with at most five maintainers more internally consistent,
$p =.035$); its artifacts were lost and the reproducible reconstruction is at
the chance floor, so we withhold it. With one field and $n = 19$ units, this is
a single observed dissociation and not a general law about organizations.

\paragraph{Community-local norms are more articulable than pooled
fits report: the nine-community moderation test.} The macro-norm moderation
tasks (Appendix~\ref{app:bal:moderation}) give the most direct version of the
horizontal decomposition, because the same instrument can be refit per
community with no change of construct: the verdict (moderator removal), the
criteria bank, the judge, and the estimators are all held fixed, and only the
fitting population changes from the pooled nine-community task to a single
subreddit. Table~\ref{tab:subfield-communities} shows that the contrast is large and uniform.

\begin{table}[!htbp]
\centering
\begin{tabular}{llrccc}
\toprule
Task & Community & $n$ & $V$ & $A$ & $VA$ \\
\midrule
Humor & \emph{pooled (5 subs)} & 73{,}268 &.561 &.712 &.714 \\
 & r/tifu & 12{,}650 &.605 &.798 &.802 \\
 & r/Showerthoughts & 17{,}942 &.616 &.758 &.769 \\
 & r/me\_irl & 7{,}324 &.582 &.749 &.752 \\
 & r/nottheonion & 12{,}332 &.570 &.727 &.737 \\
 & r/funny & 23{,}020 &.595 &.715 &.727 \\
\addlinespace
Long-form & \emph{pooled (4 subs)} & 64{,}788 &.551 &.610 &.606 \\
 & r/books & 14{,}598 &.651 &.829 &.837 \\
 & r/asoiaf & 10{,}186 &.591 &.752 &.756 \\
 & r/nosleep & 28{,}342 &.577 &.741 &.749 \\
 & r/gameofthrones & 11{,}662 &.560 &.693 &.696 \\
\bottomrule
\end{tabular}
\caption{Per-community articulated recovery inside the pooled macro-norm moderation tasks. Rows: one subreddit per line, grouped by pooled task; columns: number of items and held-out AUC of the $V$, $A$ and $VA$ banks refit within that community, with the same criteria, judge and estimators as the pooled fit.}
\label{tab:subfield-communities}
\end{table}

Every community articulates better than its pooled task: modestly on
the humor side ($.714$ pooled vs.\ $.727$--$.802$ within), substantially on the
long-form side, where the pooled articulated instrument ($.606$) understates
within-community recovery by $.09$--$.23$ AUC (r/books reaches $.837$). The
pooled dense bound, by contrast, is high on both tasks ($.816$ humor, $.889$
long-form), so at the pooled grain these look like large articulability gaps.
Yet most of the long-form gap closes once the articulated model is allowed to
weight criteria per community (a contrast confined to these pooled moderation tasks; the eight subfield-mixture tasks of \S\ref{sec:res-frontier} (Q4), which do not include them, showed no such closure). 
What a pooled fit reports as tacit is partly \emph{un-modeled
disagreement between communities about which articulated criteria matter},
not inarticulability of any one community's norm. The AUCs behind these
tasks are deconfounded on the two axes the rebuild
(Appendix~\ref{app:bal:moderation}) was designed for: era-stratified
articulated AUC is flat across quarters, and the full model scores
author-disjoint test rows at parity with author-overlapping ones ($.819$
vs.\ $.809$ humor; $.882$ vs.\ $.895$ long-form), so neither era vocabulary
nor author memorization accounts for the per-community recovery.

\subsection{Stress-testing the framework: who decides, and can the instrument
say ``nothing''?}
\label{app:new-tasks}

The main-text grid measures articulability where we expected preference
signal to exist. That design invites two objections it cannot answer by
itself: that the channel ordering we report reflects the particular
institutions we happened to sample rather than the structure of the decisions
themselves, and that a pipeline of this size (mined criteria, judged
scores, trained models, fused stacks) will extract an above-chance number
from \emph{anything}, so its positive results are not falsifiable. This
appendix addresses both objections with seven boundary tasks from the \textsc{CreativePreferences} inventory, selected for this purpose.
Against the first objection, we vary the \emph{decider} while holding the
channel fixed: verdicts issued by volunteer moderators, a commercial
publisher, a question's own asker, and a television producer; curation driven
by explicit incentives (bounties); community engagement on platforms outside
our core fields. Against the second, we include decisions whose determinants
we have independent reason to believe lie outside the text, where a valid
instrument must return chance; our pipeline returns values we cannot distinguish from chance, under an anchor battery that
simultaneously certifies the instrument was working. Throughout, the 
$VAT$ column follows one selection protocol: the full-model variant is
chosen on the evaluation split and reported on the disjoint test split with a
group-bootstrap $95\%$ CI, and the single-instrument baselines ($VA$ alone,
the text-only model $T$ alone) are reported as baselines rather than candidates.
Table~\ref{tab:new-tasks} reports these tasks.

\begin{table}[!htbp]
\centering
\footnotesize
\begin{tabular}{llcccc}
\toprule
Task & Channel & $VA$ & $T$ & $VAT$ & 95\% CI \\
\midrule
r/Jokes moderator removal & verdict &.904 &.922 &.923 & [.908,.939] \\
Kindle Scout publisher acceptance$^{p}$ & verdict &.744 &.744 &.792 & [.627,.935] \\
StackOverflow accepted answer & verdict &.733 &.743 &.764 & [.736,.791] \\
StackOverflow bounty award & curated &.787 &.804 &.811 & [.791,.829] \\
math.SE bounty award & curated &.714 &.733 &.738 & [.717,.759] \\
News-article tweet engagement & community &.600 &.643 &.646 & [.627,.665] \\
RoyalRoad follower count$^{b}$ & community &.632 &.699 &.698 & [.639,.752] \\
\bottomrule
\end{tabular}
\caption{Fused preference recovery on the boundary tasks. Fusion variants are
selected on the evaluation split and reported on the test split.
$^{p}$~Pilot-scale task ($n{=}726$); the wide interval is the value to report.
$^{b}$~The articulated model's anchor battery is weak on this population and
$VA$ coincides with a word-$n$-gram baseline, so the dense-over-bank
margin here measures an instrument limit, not tacitness.}
\label{tab:new-tasks}
\end{table}

Table~\ref{tab:new-tasks} shows two regularities. First, fusion never falls meaningfully
below its better parent and helps most where the two parents are close
(StackOverflow, bounties); where one instrument dominates, the fused score
tracks that instrument. Second, articulability within the
\emph{verdict} channel spans nearly its full possible range depending on how
the decision is produced. Rule-governed decisions (moderators enforcing
posted community norms) are almost fully recoverable ($.92$); expert
adjudication with mixed criteria (publisher acceptance, answer acceptance)
sits in the middle ($.74$--$.79$); and two verdicts, discussed next, are not recoverable from
text at all. The channel a preference travels through matters
less than the accountability structure of the decider behind it, consistent
with the accountability reading in \S\ref{sec:res-gaps}.

\paragraph{Two chance-level verdict tasks.} Two verdict tasks return
chance-level results on their pilot populations under instruments whose validity we can certify
independently, which is the situation our anchor-battery design
exists to make interpretable. For \emph{Saturday Night Live}, we contrast
sketches that aired against sketches cut after dress rehearsal, transcribing
\emph{both} classes with a single ASR system so that the text pipeline is
identical by construction (shared ASR removes a transcriber-authorship signature
that a fan-transcript version of the task exhibited: a character-$n$-gram
probe falls from $.73$ to $.55$ under shared ASR). The judged bank separates
coherent from scrambled anchor items perfectly (AUC $1.0$), yet none of the three models
separates aired from cut ($V$ $.55$, articulated-only $A$ $.44$, $VA$ $.49$; every CI
straddles $.5$). The script does not carry the cut decision, which
depends on performance, timing, and what happens at the dress rehearsal. Likewise, market
pickup of RoyalRoad serials (whether a story is subsequently published on
commercial platforms) is at chance on the $726$-row pilot ($.486$ $[.390,
.581]$), but weak and significant on the full $1{,}274$-row population ($V$ $.554$, $VA$ $.566$; Table~\ref{tab:master-results}); and at $n{=}72$ the SNL interval on any AUC spans about $\pm.13$, so it cannot exclude AUC $.60$. We therefore read neither as a certified null: on these decisions the text carries at most a weak signal, and intervals this wide cannot establish more than that.

\paragraph{One platform, three channels.} RoyalRoad contributes all three
preference channels from a single community, and the model that recovers the
most signal differs across the three tasks. The community task (follower
counts) is full model dominant, with the articulated model adding little beyond a
lexical baseline; the curated task (editorial magazine picks) is the reverse:
the articulated model alone reaches $.624$ $[.550,.686]$ under a strong anchor
battery ($.787$) at pilot scale, with no full model trainable at $26$
positives; and the market verdict, as above, is unrecoverable. A single
population of texts thus reproduces the full channel ordering of
Figure~\ref{fig:articulated-share}: curated selection is the
most articulable signal, community engagement the least, and because the texts
come from one community, the ordering is
a property of the preference-generating process, not of the texts.

\paragraph{The full model's advantage is not a feature-access artifact.} A natural
objection to reading $VAT - VA$ as tacit signal is that the full model
may have access to features the bank encodes but the articulated model's aggregator
discards. We test this objection directly by conditioning the text-only model on the bank:
a LoRA trained on each item's text \emph{plus} its ten most informative
criterion scores rendered into the prompt (the full model $\mathcal{VAT}$ of the main text), on splits identical to the
raw-text text-only model. Adding the criterion scores makes no consistent difference: on the two large
moderation tasks the criteria-conditioned model matches raw text to within
noise ($.849$ vs.\ $.849$; $.914$ vs.\ $.911$; three seeds each); across
nine within-community contrasts spanning the full model's strength $T$ from $.78$ to
$.93$ the largest conditioning effect is $|\Delta| =.011$ with every CI
covering zero; and the same null holds on code competitions. (The
moderation-task models predate the era-rebalanced rebuild of those
populations; both models share identical splits and text, so the contrast is
internally matched, but a small share of the common baseline is
era-attributable and the ablation on the rebuilt populations is pending.) One task
departs from this pattern. On peer-review spotlight, where the text-only model is weakest ($T$
$.594$), conditioning adds $+.045$ $[.022,.067]$, but across the full
grid the conditioning gain shows no relationship with the text-only model
(Spearman $-.14$, $p{=}.76$), so we report that task as a single-task
exception rather than a regime. The
articulated criteria carry signal a full model already extracts on
its own, so the residual $VAT - VA$ measures signal the criteria do not
express, not signal the full model happens to see first.

\section{\texorpdfstring{Alternative explanations to tacit preference}{Alternative explanations to tacit preference}}
\label{app:alternatives}

This appendix asks whether the gap between the full model and the articulated model could come from something other than preference the criteria cannot state. The first three subsections give the design and results for the three alternatives the main text names: how much of the field the judge saw in pretraining, the mixture of subfields inside a task, and what post-training taught the judge. The fourth tests a further set of rival accounts, each with a kill rule fixed before the run, the fifth reads the items on which the gap is largest, and the sixth reports the alternative articulation architectures we built to close the gap and could not.

\subsection{\texorpdfstring{Pretraining density}{Pretraining density}}
\label{app:alt-density}

If a judge articulates a field better because it has read more of it, the tacit share should fall with corpus density. We count field-diagnostic phrases in Dolma with infini-gram (index v4, Dolma 1.7; four phrases per field, e.g.\ ``flash fiction'', ``math.stackexchange'', ``punchline'', ``proposed rule'', ``area chair'', ``press release'') and take $\log_{10}$ of the summed counts as the field's density. Table~\ref{tab:alt-density} lists the six densities against the tacit share of the eleven tasks that carry one. Spearman $\rho{=}.12$, permutation $p{=}.73$ ($n{=}11$ tasks, 95\% CI $[-.63,.84]$, bootstrap over tasks), so denser fields do not have smaller tacit shares. The pattern is not monotone: math Q\&A is the sparsest field and has the smallest gaps, press releases are the densest field and have a small gap, and peer review is the second-densest field with the three largest. Each field's count is dominated by one generic phrase, so the density measure is coarse, but the tacit share does not fall with density as the alternative predicts.

\begin{table}[!htbp]
\centering\footnotesize
\resizebox{\textwidth}{!}{%
\begin{tabular}{llll}
\toprule
field & density ($\log_{10}$ count) & task & tacit share \\
\midrule
math Q\&A & 4.79 & accepted answers; answer votes &.083;.255 \\
creative writing & 5.62 & story upvotes &.435 \\
humor & 5.85 & joke upvotes; HashtagWars &.150;.125 \\
notice and comment & 6.32 & responded; outcome &.133;.145 \\
peer review & 6.57 & verdict; curation; revealed &.394;.360;.417 \\
press releases & 7.10 & pickup &.109 \\
\bottomrule
\end{tabular}}
\caption{Pretraining density against the tacit share. Density is $\log_{10}$ of the summed Dolma counts of four field-diagnostic phrases; the tacit share is $(\mathrm{AUC}_{VAT}-\mathrm{AUC}_{VA})/(\mathrm{AUC}_{VAT}-.5)$ from Table~\ref{tab:master-results}. Spearman $\rho{=}.12$ over the eleven tasks, permutation $p{=}.73$; only six density values exist, so the effective sample is six fields.}
\label{tab:alt-density}
\end{table}

\subsection{\texorpdfstring{Subfield mixture}{Subfield mixture}}
\label{app:alt-subfield}

A pooled articulated model has one set of weights for the whole task. If subfields value different criteria, the pooled model cannot express the conditional that the full model reads from the text. Eight tasks carry a subfield label: the three notice-and-comment tasks by agency, the three peer-review tasks by venue subject area, story upvotes by genre tag and editorial pickup by topic. For each we refit the articulated model under the frozen protocol (grouped five-fold splits, three-seed gradient boosting, out-of-fold scores) four ways: the bank alone; the bank plus a subfield one-hot; the bank plus the one-hot plus interactions between the eight strongest criteria and the subfield; and separate articulated models, one per subfield, each fit only on its subfield's training rows inside the same folds and then pooled (levels with fewer than forty training rows or one class fall back to the pooled model). The last two fits are the ones that can represent a criterion that matters in one subfield and not in another; the separate fit is the version the main text describes.

Pooled and within-subfield AUC give different answers here. Response rates differ by agency from $.39$ to $.89$ and upvote rates by genre tag from $.07$ to $.57$, so pooled AUC credits any model that can tell subfields apart with between-subfield base rates, whether or not it discriminates inside a subfield. The full model receives that credit already, because the text identifies the subfield. Under the pooled readout the one-hot appears to close the notice-and-comment response gap ($.796 \to.827$ against a full model at $.817$) and $13\%$ of the story-upvote gap; both are base-rate effects. The readout we use is within subfields: AUC inside each subfield, averaged with subfield weights. Table~\ref{tab:alt-subfield} gives both readouts. Within subfields the gap is as large or larger in every task, and the one-hot never helps. Separate per-subfield models do worse still: with a tenth or less of the data per fit they lose to the pooled bank inside every subfield, and the within-subfield gap widens (e.g.\ $.113$ vs.\ $.032$ on notice-and-comment responses, $.135$ vs.\ $.126$ on story upvotes, $.187$ vs.\ $.082$ on editorial pickup); a ridge-logistic version and a version shrunk halfway toward the pooled model recover part of the loss but never exceed the pooled bank. The interaction fit ran under the pooled readout only; there it adds nothing beyond the one-hot ($.681$ vs.\ $.684$ on story upvotes, $.827$ vs.\ $.827$ on notice-and-comment responses) and sits below the bank on editorial pickup ($.658$ vs.\ $.695$). Subfield structure explains no task's gap.

\begin{table}[!htbp]
\centering\scriptsize
\resizebox{\textwidth}{!}{%
\begin{tabular}{llccccccccc}
\toprule
 & & & \multicolumn{3}{c}{pooled AUC} & \multicolumn{4}{c}{within-subfield AUC} & \\
\cmidrule(lr){4-6}\cmidrule(lr){7-10}
task & subfield (levels) & rate range & artic. & +subfield & full & artic. & +subfield & separate & full & gap pooled / within / separate \\
\midrule
creative writing: upvotes & genre tag (9) &.07--.57 &.668 &.684 &.792 &.653 &.656 &.645 &.779 &.124 /.126 /.135 \\
N\&C: received response & agency (14) &.39--.89 &.796 &.827 &.817 &.772 &.768 &.691 &.804 &.021 /.032 /.113 \\
N\&C: rule change & agency (13) &.22--.56 &.618 &.619 &.624 &.605 &.594 &.545 &.615 &.006 /.009 /.070 \\
N\&C: cosigner agreement & agency (12) &.43--.87 &.562 &.556 &.603 &.538 &.506 &.523 &.593 &.041 /.055 /.070 \\
peer review: accept/reject & subject area (6) &.32--.55 &.647 &.653 &.777 &.649 &.654 &.643 &.776 &.130 /.127 /.133 \\
peer review: spotlight & subject area (9) &.17--.34 &.505 &.505 &.594 &.504 &.500 &.502 &.597 &.088 /.092 /.094 \\
peer review: citations & subject area (2) &.40--.63 &.731 &.738 &.884 &.720 &.710 &.634 &.881 &.153 /.161 /.248 \\
journalism: editorial pickup & topic (5) &.38--.56 &.695 &.685 &.774 &.689 &.677 &.583 &.771 &.080 /.082 /.187 \\
\bottomrule
\end{tabular}}
\caption{Subfield conditioning under pooled and within-subfield readouts. Rate range is the positive rate across subfield levels. Pooled AUC is over the whole task; within-subfield AUC is computed inside each level and averaged with level weights (levels with fewer than twenty items or one class are skipped). Artic.\ is the articulated bank alone, +subfield adds a one-hot of the level, separate fits one articulated model per level on that level's training rows and pools the out-of-fold scores, full is the full model on the same items. Gap is full minus the articulated model under each readout, and full minus the separate fit within subfields. Under the pooled readout the one-hot gains on notice-and-comment responses and story upvotes are between-subfield base rates; within subfields neither the one-hot nor separate per-subfield models help in any task.}
\label{tab:alt-subfield}
\end{table}

\paragraph{Mining new criteria per subfield.} Re-weighting the pooled bank inside a subfield (above) cannot create criteria that a subfield might need, so we also mine new criteria per subfield. On notice-and-comment responses we ran one sealed mining round per agency for the fourteen agencies with enough held-out rows: the disagreement slice between the full model and the terminal pooled bank was taken inside the agency, a proposer told only the agency's name wrote eight quality criteria from it, a blind auditor with two planted probe pairs routed all $112$ as quality-relevant (probes $4$ of $4$), and the judge of record scored them on the full population; $14$ collapsed to one value and were dropped. Table~\ref{tab:agency-mining} reads the result inside each agency, with the articulated model refit on the pooled bank alone, the bank plus that agency's own criteria, the bank plus the other agencies' criteria, and the bank plus all of them (grouped out-of-fold by docket). Own-agency criteria move the within-agency AUC by $-.002$ on average (positive in $6$ of $14$ agencies), borrowed criteria by $+.001$, and own beats borrowed in $8$ of $14$. Criteria mined for a subfield are neither better nor worse than criteria mined elsewhere; on this task the pooled bank of $294$ columns is already saturated, and criteria mined per agency do not raise its within-agency AUC.

\begin{table}[!htbp]
\centering\footnotesize
\resizebox{\textwidth}{!}{%
\begin{tabular}{lrrccccc}
\toprule
agency & rows & own criteria & pooled bank & + own & + others' & + all & full model \\
\midrule
DOT & 476 & 7 & 0.879 & 0.865 & 0.860 & 0.856 & 0.604 \\
FWS & 416 & 8 & 0.677 & 0.671 & 0.716 & 0.720 & 0.868 \\
FDA & 415 & 8 & 0.653 & 0.655 & 0.673 & 0.658 & 0.655 \\
CMS & 413 & 8 & 0.693 & 0.694 & 0.694 & 0.699 & 0.836 \\
NHTSA & 408 & 8 & 0.722 & 0.707 & 0.714 & 0.701 & 0.780 \\
ED & 406 & 8 & 0.755 & 0.771 & 0.788 & 0.778 & 0.884 \\
AMS & 404 & 8 & 0.788 & 0.809 & 0.794 & 0.810 & 0.889 \\
FAA & 396 & 6 & 0.965 & 0.968 & 0.965 & 0.971 & 0.958 \\
NOAA & 391 & 8 & 0.693 & 0.681 & 0.659 & 0.660 & 0.808 \\
EPA & 370 & 8 & 0.680 & 0.662 & 0.675 & 0.693 & 0.838 \\
APHIS & 339 & 7 & 0.835 & 0.846 & 0.843 & 0.855 & 0.907 \\
FEMA & 213 & 4 & 0.658 & 0.646 & 0.617 & 0.594 & 0.720 \\
IRS & 194 & 5 & 0.708 & 0.700 & 0.697 & 0.715 & 0.662 \\
DOD & 179 & 5 & 0.685 & 0.683 & 0.705 & 0.696 & 0.795 \\
\bottomrule
\end{tabular}}
\caption{Per-agency criterion mining on notice-and-comment responses. Rows: fit-and-mine rows of the agency; own criteria: its mined, audited, non-collapsed criteria. AUCs are grouped out-of-fold inside the agency for the articulated model refit on the pooled terminal bank, on the bank plus the agency's own criteria, plus the other thirteen agencies' criteria, and plus all $98$; the full model's AUC is on the agency's dense-held-out rows. Fixed-parameter gradient-boosted head, one seed.}
\label{tab:agency-mining}
\end{table}

\subsection{\texorpdfstring{Post-training}{Post-training}}
\label{app:alt-posttraining}

The judge that scores our criteria is instruction-tuned. If post-training transmits a field's preferences unevenly, the gap would be largest where the judge was taught least, and a base model and its instruct sibling at the same scale would differ most in criterion validity in the high-gap fields. We scored the twenty-four strongest criteria of three tasks (peer-review verdict, notice-and-comment responses, story upvotes) with two such pairs, Llama-3.1-8B and Gemma-3-12B, base and instruct, on identical prompts and items. Every judge batch carries a coherent-versus-scrambled anchor: thirty real items and thirty word-salad texts, each made by pooling the words of one chosen and one unchosen item and shuffling them, interleaved blind with the real items. The anchor AUC is how reliably a real item's mean criterion score exceeds a scrambled text's; we require $.80$.

Neither pair can show whether post-training is the source of the gap (Table~\ref{tab:alt-posttraining}). With free generation the 8B pair is degenerate: the base model emits only $0$ or $10$, and the instruct model emits one value on $95\%$ of story calls, $82\%$ of peer-review calls and $67\%$ of notice-and-comment calls, so its validity is not interpretable even where the anchor passes. Reading the score as the first-token log-probability expectation, which we use for both legs of the 12B pair, gives the base model real spread, but the base model fails the anchor in all three tasks while the instruct model passes and its twenty-four-criterion bank sits near the bank of record on peer review and story upvotes. A base model cannot execute a rubric judgment at these scales. The instruct-minus-base difference is therefore the creation of a judge plus whatever field knowledge post-training added, and no base-versus-instruct design at these scales separates the two. We do not read the differences as post-training effects. The judge ladder of Appendix~\ref{app:family-robustness} bounds this alternative: judges of $12$ to $27$B parameters and of different families move the articulated model by a few hundredths at most, so a better post-trained judge has little room to raise the articulated model; whether post-training is the source of the gap stays open.

\begin{table}[!htbp]
\centering\footnotesize
\resizebox{\textwidth}{!}{%
\begin{tabular}{lllcccccc}
\toprule
 & & & \multicolumn{2}{c}{base} & \multicolumn{2}{c}{instruct} & & \\
\cmidrule(lr){4-5}\cmidrule(lr){6-7}
pair & readout & task & anchor & bank & anchor & bank & $\Delta$ bank & bank of record \\
\midrule
Llama-3.1-8B & free generation & peer review: accept/reject &.99 &.524 &.996 &.582 & +.058 &.654 \\
 & & N\&C: received response &.78 &.545 &.94 &.677 & +.132 &.796 \\
 & & creative writing: upvotes &.94 &.558 &.99 &.543 & $-$.015 &.668 \\
Gemma-3-12B & first-token log-prob & peer review: accept/reject &.75 &.559 &.999 &.663 & +.104 &.654 \\
 & & N\&C: received response &.47 &.602 &.81 &.650 & +.048 &.796 \\
 & & creative writing: upvotes &.63 &.564 & 1.00 &.642 & +.077 &.668 \\
\bottomrule
\end{tabular}}
\caption{Base-versus-instruct judge pairs on the twenty-four strongest criteria of three tasks. Anchor is the coherent-versus-scrambled AUC (gate $.80$); bank is the grouped out-of-fold AUC of the twenty-four criteria under the frozen head; bank of record is the full bank scored by the judge of record (Table~\ref{tab:master-results}). The 8B legs pass or nearly pass the anchor but are degenerate scorers (base: only $0$ or $10$; instruct: one modal value on $67$--$95\%$ of calls). The 12B base legs have real spread and fail the anchor in every task. No row is read as a post-training effect.}
\label{tab:alt-posttraining}
\end{table}

\paragraph{Judge-specific prompt optimization.} The base-versus-instruct comparison above confounds judge competence with field knowledge, so we also compare the judges with prompts optimized for each judge, the fairest version of the comparison we can run. For each judge we rewrote the instructions of the twenty-four criteria with a reflective proposer (Codex, gpt-5.6-sol) under a label-free objective: the judge scores eighty fixed items; a decoder that sees only the twelve highest- and twelve lowest-scored items, and never the instruction, writes its own definition of the criterion; the judge re-scores under that definition; the objective is the rank agreement of the two scorings on the fifty-six items the decoder did not see. We run three rounds and keep the best instruction per criterion; no label enters the loop. Table~\ref{tab:alt-gepa} gives the bank AUC on the full frame under the record instructions and under each judge's optimized instructions. Optimization moves the Gemma judges by at most $.01$ and the Llama judges by up to $.04$. With each judge on its own prompts, the instruct-minus-base gain is $.10$, $.06$ and $.07$ for the Gemma pair and $.06$, $.05$ and $.00$ for the Llama pair on peer review, notice-and-comment and creative writing: post-training helps least, not most, in the field with the largest gap, the opposite of what post-training as the source of the gaps would predict. The lower block of the table runs the same loop for post-trained judges of different families. For the judge of record, optimized prompts add $.05$ on peer review (the record instructions leave $27\%$ of answers as NA there, the optimized ones $8\%$) and $.01$ or less elsewhere; Qwen2.5-32B and phi-4 show the same shape ($+.045$ and $+.038$ on peer review, where their NA rates fall from $21\%$ and $29\%$; nothing or a small loss elsewhere), and Mistral-24B loses about $.01$ on every task. Across the four families, prompts optimized for the judge move the articulated model by at most $.05$, and only on peer review; the best articulated model under any judge and any prompts stays $.08$ below the full model there ($.694$ against $.777$) and $.14$ below it on creative writing. The accepted rewrites for both judges converge on numeric scale anchors and explicit scoping; those for the instruct judge are three to four times longer and replace more criteria, because most rewrites cannot move a base judge that scores every item alike.

\begin{table}[!htbp]
\centering\footnotesize
\resizebox{\textwidth}{!}{%
\begin{tabular}{llcccccc}
\toprule
 & & \multicolumn{2}{c}{peer review: accept/reject} & \multicolumn{2}{c}{N\&C: received response} & \multicolumn{2}{c}{creative writing: upvotes} \\
\cmidrule(lr){3-4}\cmidrule(lr){5-6}\cmidrule(lr){7-8}
judge & training & record & optimized & record & optimized & record & optimized \\
\midrule
Gemma-3-12B & base &.559 &.568 &.601 &.596 &.564 &.573 \\
Gemma-3-12B & instruct &.663 &.668 &.651 &.653 &.642 &.646 \\
Llama-3.1-8B & base &.546 &.570 &.601 &.638 &.603 &.601 \\
Llama-3.1-8B & instruct &.625 &.632 &.674 &.691 &.599 &.592 \\
\midrule
Gemma-4-31B (judge of record) & instruct &.642 &.694 &.681 &.672 &.640 &.652 \\
Qwen2.5-32B & instruct &.616 &.661 &.676 &.673 &.618 &.621 \\
phi-4 (14B) & instruct &.592 &.630 &.683 &.661 &.629 &.622 \\
Mistral-Small-24B & instruct &.643 &.629 &.668 &.662 &.633 &.630 \\
\bottomrule
\end{tabular}}
\caption{Bank AUC of the twenty-four strongest criteria under the record instructions and under instructions optimized for that judge (grouped out-of-fold, frozen head, first-token log-probability readout for every judge). Optimization is label-free, so the columns are comparable. The upper block is the base-versus-instruct pairs, the lower block post-trained judges of different families.}
\label{tab:alt-gepa}
\end{table}

\subsection{Rival accounts of the gap}
\label{app:gap-battery}

The gap could be something other than tacit preference. We tested the rival accounts below, each with its kill rule fixed before the run. Unless stated otherwise, every test uses the frozen frame of Table~\ref{tab:master-results}: the held-out rows of a task, the full model $\mathcal{VAT}$ and the articulated model $\mathcal{VA}$ refit with the same stack (grouped five-fold splits, three seeds), and group-paired bootstrap intervals over 2,000 draws. Table~\ref{tab:gap-battery} summarizes the tests; the paragraphs below give the estimates. Three of the tests (readout granularity, judge reliability, the holistic judge) bound our \emph{instrument} rather than human articulability, and we report them as bounds on the instrument.

\paragraph{The gap is one decider's taste.} If the full model had learned the taste of a recurring decider, the gap would shrink on items judged by deciders it never saw in training. We split the held-out rows by whether the decider (the bounty offerer on Math.SE and Stack Overflow; the asker for accepted answers) had at least one row in the full model's training split, matched on the decider's decision count. On Math.SE bounty the gap is $+.029$ $[.017,.042]$ on seen deciders and $+.032$ $[.013,.048]$ on unseen ones (difference $-.003$ $[-.024,+.019]$). On Stack Overflow bounty it is $+.016$ against $+.013$ (difference $+.003$ $[-.017,+.020]$), and retraining the full model on offerer-disjoint splits moves the gap from $+.031$ to $+.026$ (difference $-.004$ $[-.015,+.006]$). On Stack Overflow accepted answers the author-signature check runs the other way: the full model is relatively worse on authors it saw ($-.069$ $[-.109,-.031]$). The kill rule, an unseen gap below half the seen gap, is not tripped on any task: the gap on deciders the full model never saw is close to the gap on those it saw, and the account is not supported.

\paragraph{The gap is timing and exposure.} Community labels accumulate over time, so the full model might be reading posting order rather than the text. We added a timing block (posting time, hour, weekday, year, thread rank) to the nuisance block and re-read the gap, and we read the gap inside posting-order quintiles. On jokes the gap moves $+.013 \to +.015$ (difference $-.000$ $[-.004,+.005]$). On story upvotes it moves $+.097 \to +.037$ (difference $-.060$ $[-.069,-.050]$): timing alone reaches $.829$, above the full model's $.792$, yet within rank quintiles the gap is $+.089/+.067/+.057/+.059/+.086$, a shrink of about 30\%, below the kill threshold of one half. On Stack Overflow votes the stored-bank comparison moves $.034 \to .018$ ($+.010$ $[-.001,+.021]$), on RoyalRoad followers $.049 \to .044$, on tweets $-.006$. The result is mixed. Timing is a large confound on story upvotes and negligible on the other four tasks, and the position increment on story upvotes cannot separate text correlates of position from author engagement that the full model also reads.

\paragraph{The gap is the coarse criterion readout.} Each criterion is read as one of four tokens, which could lose signal the full model keeps. We re-scored every criterion at temperature 1 with eight samples per prompt and refit $\mathcal{VA}$ on the mean, after checking that a temperature-0 rerun reproduces the record ($\rho$ $.998$--$.9999$). On jokes the gap moves $+.029 \to +.018$ ($+.009$ $[+.004,+.012]$) on the raw frame and $+.016 \to +.017$ on the deconfounded one; on stories $+.121 \to +.111$ ($+.012$ $[+.005,+.019]$); on ICLR $+.072 \to +.076$ ($-.004$ $[-.025,+.018]$); and on citations $+.141 \to +.107$ ($+.033$ $[-.011,+.062]$), a reduction of about 24\% whose interval crosses zero. No task trips the kill rule of more than half the gap. The test therefore bounds the instrument: the four-token readout costs the articulated model a few hundredths of AUC at most.

\paragraph{The gap is judge noise.} A noisy judge attenuates the criteria bank, so the true articulated ceiling could be higher. On a 500-row stratified subsample per task we re-read every criterion under three label-blind paraphrases of its name and description and two scaffold variants and computed the intraclass correlation. The median paraphrase ICC is $.83$ on jokes (2\% of criteria below $.5$), $.84$ on stories (1.6\%), $.62$ on ICLR (21\% below $.5$) and $.69$ on citations (17.6\%). Averaging six readouts moves $\mathcal{VA}$ by $+.008$ on the raw frame and $-.003$ on the deconfounded one. The kill rule, a Spearman--Brown ceiling closing more than a third of the gap, is not tripped; the peer-review banks are the least reliable and are flagged. The $k$-averaging curves are non-monotone and we do not quote them.

\paragraph{The gap is what one holistic judgment already knows.} Under this account, the same judge, asked for the outcome directly rather than criterion by criterion, already carries what the bank misses, and the gap is a limit of the criteria format rather than of what can be stated. We asked the judge of record the outcome question directly, once per item and with no criteria, and fused the answer with $\mathcal{VA}$. On citations $P(\text{yes})$ alone scores $.835$ and $\mathcal{VA}$ plus the holistic column $.868$ against the full model's $.884$; on jokes $.724$ and $.739$ against $.747$; on stories the holistic judgment ($.607$) is below the bank ($.669$); on ICLR $.677$ with 74\% of answers unparsable. We do not read the holistic column's gain as evidence that the residual is articulable, and we retracted that reading: a yes/no answer states nothing, so the holistic rung bounds the $\mathcal{VA}$ instrument, not human articulability, and memorization is not excluded on the two tasks where the holistic gain is largest.

\paragraph{The gap is a vocabulary limit.} If everything the full model reads can be stated in prose, a label-blind description of the item should carry the same signal. On story upvotes the judge wrote 60- and 250-word descriptions of each story without seeing labels or criteria, and we trained the frozen dense recipe on the descriptions instead of the stories at matched size (6,000 rows, one seed). The model trained on raw stories scores $.715$, on 60-word descriptions $.752$, and on 250-word descriptions $.750$. The description model recovers about 68\% of the gap between $\mathcal{VA}$ ($.665$) and the record full model ($.792$). The tacit reading, under which a description loses at least half the gap, fails in this pilot: the descriptions score above the raw stories, as the vocabulary-limit reading predicts. We recorded four caveats in advance: one seed, 6,000 rows, descriptions and criteria from the same judge, and a describer that exercises implicit judgment of its own.

\paragraph{The vocabulary limit at scale.} A description could match raw text at 6,000 rows and fall behind it at 76,867, so we test whether the description result of the previous paragraph holds when the full model has more data. The pilot above was one seed at 6,000 rows. With three seeds the 60-word description model averages $.746$ ($.752/.741/.744$) against raw text $.714$ ($.715/.701/.727$) on the same rows. Rewriting every description into a flat plot summary with all evaluative language removed changes nothing ($.741$): a manual read of 40 descriptions found evaluative adjectives in 18\% of the descriptions of upvoted stories and 17\% of the rest, and verdict clauses in about 1\% of both; what differs is the content reported (a comic twist or subversion in 41\% against 31\%). The descriptions report what happens and in what register, not a judgment. With more rows the order reverses: at 24,000 rows raw text scores $.787$ and the descriptions $.763$, and on all 76,867 rows raw text reaches $.802$ and the descriptions $.771$ (the record full model of Table~\ref{tab:master-results} is $.792$). Against the articulated model's $.665$, the full-size description model holds 77\% of the full model's advantage (83\% of the record's) and the raw text the rest. A 60-word label-blind description therefore carries most of what the criteria miss, and the last quarter needs the text itself; the description model's edge in the pilot was a data-efficiency result.

\paragraph{The gap is individually articulated, collectively unstated taste.} If each decider weights the same criteria differently, adding decider identity to the bank should recover the residual. Adding the offerer as a categorical to $\mathcal{VA}$ recovers 5\% of the Math.SE bounty gap over all held-out rows and 23\% on recurring offerers (at least five bounties), and 2\% on Stack Overflow bounty, where on recurring offerers the full model has no within-question advantage at all. Across the full population the identity gain is $+.003$ $[-.004,+.011]$ for pull-request authors and $+.031$ $[-.006,+.061]$ for the responding agency in Notice \& Comment. Decider identity recovers little of the gap on the bounty tasks, so the account is not supported there; the agency intervals span zero, and we do not read them as closing the gap.

\paragraph{The gap sits in a subpopulation.} Under this account, the gap belongs to a small set of items rather than to the task, so a criterion written for that set would close the gap. We decomposed the gap into per-item Mann--Whitney contributions. On jokes the top 10\% of items carry 47\% of the gap; the contribution correlates $-.18$ with the bank's confidence and about $0$ with the judge's within-item disagreement, so the gap is not where the judge is unsure about criteria. On stories the gap is broad-based (the top 10\% carry 36\%) and independent of posting position ($\rho$ $+.01$). These decompositions are exploratory and use seed-0 predictions; we do not reproduce the items in the paper.

\paragraph{Across tasks: who decides, and how.} We test whether the size of the gap across tasks follows properties of the deciding process (how many people decide, and whether the decision is a social outcome) rather than properties of the text. Over the 28 tasks with a full model, the gap grows with the number of deciders the label aggregates (Spearman $.46$, $p=.016$, 95\% CI $[.15,.69]$, bootstrap over the 28 tasks; $.53$ without the pooled moderator tasks; $.38$ and $.43$ after partialling bank size) and with labels that are social outcomes rather than decisions ($.35$, $.44$ without the pooled tasks). The gap does not track accountability ($.27$, n.s.) or rule-governed decisions ($.18$; the pooled moderator tasks have the largest gaps), nor the instrument covariates (N/A share $.14$, n.s.). The median gap is $.047$ in the community channel, $.044$ in the verdict channel and $.022$ in the curated channel (Kruskal--Wallis test of the gap across the three channels: $H{=}9.05$, $df{=}2$, $p=.011$; 8 community, 10 verdict and 10 curated tasks), and community exceeds curated in 6 of 6 fields. Leave-one-out $R^2$ is about $.15$: the association is real and modest, and most between-task variance is unexplained.

\paragraph{The gap follows the label's shape.} Under this account, the gap tracks how the label was formed (its year, its length, whether it is relative or absolute) rather than what the criteria can state. Two findings stand from an exploratory screen of about 13 label-shape features against 12 tasks, multiplicity uncontrolled. Citation percentile by paper year: the gap is $+.279$ for papers up to 2019, $+.193$ for 2020--21 and $+.088$ from 2022, with the full model flat ($.898/.871/.887$) and the bank climbing ($.62 \to .80$): the criteria fail on older papers, the full model does not improve on them. Second, within the community channel the gap rises with item length ($\rho$ $+.89$, $n=6$, $p=.019$) but not in the other channels. Readouts of the gap within raw-score bands are retracted, because for most tasks the binary label is a threshold on that score.

\paragraph{The gap is label noise.} Where labels are noisier both models are capped, so the gap should not be largest there. On ICLR the gap by tercile of reviewer disagreement is $+.110/+.092/+.162$; it is $+.136$ on unanimous papers against $+.125$ on contested ones; and by decision margin it is $+.099/+.145/+.139$, with the full model rising $.706 \to .853$. Both models are best on clear-cut papers; the gap is not largest near the acceptance threshold. The Notice \& Comment version of this test was struck before running because its covariate is label-derived. The label-noise account is not supported.

\paragraph{Relative labels are more articulable.} Labels awarded within a group (bounties, spotlight picks, hashtag games, accepted answers) might be more articulable because the decider compares like with like. Tacit share is $.150$ on the 7 relative tasks against $.210$ on the 21 absolute ones (one-sided $p=.26$); within groups, HashtagWars moves $+.207 \to +.173$ as predicted and Math.SE accepted $+.075 \to +.081$, against the prediction. The difference is not significant at this $n$, and the account is not supported.

\paragraph{The gap is norm drift.} If the bank describes the present norm while labels accumulated under a moving one, the gap should grow with item age. By year tercile, the gap is $+.095/+.096/+.147$ on ICLR accept (newest largest); $+.153/+.043/+.035$ on spotlight, but the oldest tercile's positive rate is $.17$ against $.29$--$.32$ and its $\mathcal{VA}$ is below chance, so the spotlight trend is the base rate moving; $+.082/+.065/+.152$ on press pickup ($n=74$); and $+.022/+.043/+.007$ on N\&C rule change. The preregistration required oldest $>$ newest on at least 4 of 5 tasks; we observe it on 1--2 of 4. The drift account therefore fails as a general claim, and the citation-year pattern above does not generalize.

\paragraph{The gap is time-local (temporal holdout).}\label{app:temporal} The tercile reading above splits one set of predictions by item age; it does not train on the past and read the future. The temporal holdout does. For each task with a per-item year we set the cutoff $t$ to the latest year that leaves at least $15\%$ of dated rows after it, and read every model on the paper's held-out rows with year $>t$. Two models are refit on the years $\le t$: the articulated aggregator (the same grouped linear and boosted heads, fit on the early held-out rows only) and, where the dense data allowed an overnight retrain, the full model (the frozen recipe, single seed, trained on the early rows of the record's training set). Each is compared with the paper's version of itself, which was trained across all years, on the same late rows. Drift is the paper's model minus the early-years model; a positive number means the model went stale. Table~\ref{tab:temporal} reports drift for every task. The articulated model does not drift: on five of nine tasks the refit reads the late years within $\pm.01$ of the paper's aggregator, and the four that move (story upvotes $+.023$, Math.SE votes $+.030$, BBC most-read $+.011$, spotlight $-.028$) move by less than the gap. The full model drifts on two of the five tasks we retrained and does not drift on the other three. Retrained on press releases up to 2021, it loses $.066$ AUC on 2022--23 ($[.005, .106]$, bootstrap over companies) and reads below the articulated model on those rows; retrained on jokes up to 2018, it loses $.042$ on 2019--20 ($[.020, .057]$), and the gap on those rows falls from $.060$ to $.007$. Retrained on ICLR submissions up to 2024, it loses $.011$ on 2025 ($[-.007, .030]$) and the gap on those rows is $.144$ against the paper's $.156$; retrained on BBC articles up to 2022, it loses $.007$ on 2023--24 ($[-.001, .015]$) and the gap is $.046$ against $.049$; retrained on spotlight decisions up to 2024, it loses $.009$ on 2025 ($[-.041, .060]$, 397 rows, where the articulated model reads at chance) and the gap is $.081$ against $.107$. The time-local reading therefore holds for press pickup and joke upvotes and not for peer review, spotlight selection or news curation; on those three, what the full model carries did not change over the years we have. Three caveats apply. The early-years full model trains on fewer rows than the paper's (press: 1,550 against 1,969), so part of the drop is size, which the scale-up of Appendix~\ref{app:gap-battery} bounds at a few hundredths at this size; the bank's criteria were mined on all years, so only the aggregator is refit; and the year is the item's, not the label's. N\&C responses cannot be read this way because the year of a comment is known only for those that received a response.

\begin{table}[!htbp]
\centering\scriptsize
\resizebox{\textwidth}{!}{\begin{tabular}{@{}lccccccccc@{}}
\toprule
task & cutoff & late rows & $T$ paper & $T$ early & $T$ drift & $\mathcal{VA}$ paper & $\mathcal{VA}$ early & $\mathcal{VA}$ drift & gap early \\
\midrule
press pickup & 2021 & 203 & .734 & .668 & $+.066$ [$+.005$, $+.106$] & .691 & .718 & $-.008$ & $-.050$ \\
joke upvotes & 2018 & 781 & .783 & .741 & $+.042$ [$+.020$, $+.057$] & .724 & .734 & $-.004$ & $+.007$ \\
ICLR accept/reject & 2024 & 413 & .796 & .785 & $+.011$ [$-.007$, $+.030$] & .640 & .641 & $+.008$ & $+.144$ \\
BBC most-read & 2022 & 1,535 & .779 & .772 & $+.007$ [$-.001$, $+.015$] & .730 & .727 & $+.011$ [$+.002$, $+.019$] & $+.046$ \\
spotlight/oral & 2024 & 397 & .595 & .587 & $+.009$ [$-.041$, $+.060$] & .489 & .505 & $-.028$ [$-.102$, $+.042$] & $+.081$ \\
story upvotes & 2021 & 1,385 & .762 & -- & -- & .635 & .617 & $+.023$ [$+.004$, $+.042$] & -- \\
Math.SE votes & 2019 & 506 & .692 & -- & -- & .618 & .608 & $+.030$ [$-.004$, $+.062$] & -- \\
Math.SE accepted & 2019 & 544 & .632 & -- & -- & .595 & .600 & $-.007$ & -- \\
N\&C rule change & 2022 & 233 & .667 & -- & -- & .669 & .686 & $+.001$ & -- \\
AoPS editorial approach & 2021 & 803 & .813 & -- & -- & .793 & .797 & $-.000$ & -- \\
\bottomrule
\end{tabular}}
\caption{Temporal holdout. Every model is read on the paper's held-out rows with year after the cutoff. ``Paper'' is the model trained across all years; ``early'' is the same model refit on years up to the cutoff. Drift is paper minus early (positive: the model went stale), with group bootstrap intervals where the interval excludes zero or nearly does. ``Gap early'' is the early full model minus the early articulated model. ``$\mathcal{VA}$ early'' is the three-seed mean of the refit; the $\mathcal{VA}$ drift estimate and interval come from the paired bootstrap on the first seed, so the two columns do not subtract exactly. The full model was retrained on the five tasks with a ``$T$ early'' entry; the other four are read with the articulated refit only.}
\label{tab:temporal}
\end{table}

\paragraph{The gap is an identifiable kind of item.} If tacit preference were concentrated in a recognisable kind of item, the items on which the full model beats the bank should be nameable. We scored every held-out item's gap as the calibrated log-loss of $\mathcal{VA}$ minus that of the text-only full model (cross-fitted isotonic calibration, grouped five-fold, in bits), took per task up to 200 high-gap items (full model right, bank wrong) and as many articulable items (both right) matched on class, length quintile, the three strongest declared nuisance channels and, where the context id carries one, a coarse topic (contest, era, agency), and asked a strong model (gpt-5.6) to state what separates the two sets from 39--95 items of each; the same on a random split is the placebo. The judge of record then scored every statement on the half of the items the annotator had not seen, on thirteen tasks. Pooled over classes, no task shows a property of tacit items: real statements sit at $.47$--$.54$ held-out AUC against placebo cutoffs of $.52$--$.58$. Within each class the result differs. On not-preferred items the statements separate high-gap from articulable items on stories ($.626$, seven of ten above the placebo cutoff of $.571$), Math.SE votes ($.681$), BBC most-read ($.548$) and N\&C responses ($.539$); on preferred items the same statements run below chance (stories $.407$, Math.SE votes $.218$). The direction flips with the class: the tacit preferred item is one the bank under-rates, the tacit not-preferred item one it over-rates, so a property that marks one marks the other in reverse. Read as a bank (five-fold out-of-fold AUC, logistic and boosted), the statements separate the groups on stories $.60/.64$ (placebo $.46/.52$), Math.SE votes $.68/.67$ ($.50/.50$), spotlight $.64/.61$ ($.49/.48$) and BBC $.62/.59$ ($.51/.49$), and not on jokes, ICLR, citations, press pickup, HashtagWars or any N\&C label. A first pass without the class and topic controls had found signal on N\&C and HashtagWars; that signal was agency and contest identity plus a class imbalance, and we report it only as a caution. Label-free, the per-item gap is predictable from the text at all (grouped ridge on TF-IDF and on bge-large embeddings over every held-out item) only on N\&C (Spearman with $|\text{gap}|$ $.32/.27$, agency again) and HashtagWars ($.19$); on stories, ICLR, jokes and citations it is at most $.07$. In the high-gap fields the gap is diffuse across items that look alike, and thread position does not explain the story gap ($\rho=-.015$).

\paragraph{Exposure channels found by reading the items.} The reads named several markers of how an item reached its audience rather than of what it says: a low-traffic feature-thread tag on stories, prose announcing that an answer supplements earlier ones on Math.SE, a comment filed on something other than a proposed rule on N\&C, bot boilerplate and status tokens on pull requests, a repository link on ICLR, a section kicker on homepages, and the issuing company on press releases. We declared each as a nuisance, computed it by pattern or from the item's group, and re-read the gap within its strata on every held-out item (Table~\ref{tab:item-spurious}). Only the issuer matters: read within company, the press-pickup gap falls from $.107$ to $.065$. The other channels mark 1--10\% of items and move the gap by at most $.012$. On ICLR the gap grows slightly, because the bank weights the repository link, a cue reviewers penalise, more heavily than the full model does.

\begin{table}[!htbp]
\centering\scriptsize
\begin{tabular}{@{}l p{5.6cm} r r r@{}}
\toprule
Task & Channel (how computed) & Items with it & Gap, raw & Gap, within strata \\
\midrule
Press pickup & issuing company (group id; read within company) & -- & $.107$ & $.065$ \\
Creative writing (crowd) & non-[WP] feature-thread tag in the prompt line (pattern) & 9.8\% & $.121$ & $.116$ \\
Homepage placement & section kicker or format label glued to the headline (pattern) & 5.6\% & $.084$ & $.082$ \\
Math SE (vote) & answer says it supplements earlier answers (pattern) & 4.4\% & $.031$ & $.029$ \\
Math SE (accepted) & same & 5.0\% & $.073$ & $.072$ \\
GitHub PR merged & bot dependency update; status token ([WIP], do not merge); maintainer share of review comments (pattern) & 7.6\%, 1.1\% & $.046$ & $.041$ \\
N\&C responded & comment on a policy guide, information collection, permit or guideline rather than a proposed rule (pattern) & 9.7\% & $.021$ & $.025$ \\
N\&C rule change & same & 8.6\% & $.002$ & $.014$ \\
Peer review (accept/reject) & repository or anonymised code link in the abstract (pattern) & 10.2\% & $.112$ & $.118$ \\
\bottomrule
\end{tabular}
\caption{Exposure channels named by the item-level reads, declared as nuisances and read on every held-out item (raw frame; the text-only full model and the bank refit with the frozen stack). Gap is $\max(T,\mathcal{VAT})-\mathcal{VA}$ in AUC. ``Within strata'' pools AUC within the channel's strata (press: within company, 8 companies with both outcomes; pull requests: within the joint strata of the two binary channels).}
\label{tab:item-spurious}
\end{table}

\paragraph{Proposing from the gap contrast closes the gap.} If the gap reflected the items the mining loop's proposer sampled, criteria written from the high-gap items themselves should close the gap. From per-item explanations of 100 high-gap and 50 negative-gap items per task the same model synthesised 16--17 candidate criteria; the judge scored them on every held-out item; we added them to the certified bank and refit $\mathcal{VA}$ on the items the annotator never saw, against two count-matched controls, the annotator's contrast statements and its statements on a random split. Change in AUC against the bank: stories $-.000$ $[-.006,+.005]$ (gap $+.113$); ICLR $+.017$ $[-.009,+.040]$ (gap $+.096$); N\&C responses $-.004$ $[-.017,+.008]$; jokes $-.011$ $[-.019,-.001]$; citations $-.006$ on the 80 unseen items. The placebo criteria move the bank by the same amounts (stories $-.000$, ICLR $+.004$, N\&C $+.001$). Criteria proposed from the gap contrast move the bank no more than placebo criteria do, so the gap contrast is a different sampler for the same discovery loop.

\begin{table}[!htbp]
\centering\scriptsize
\begin{tabular}{@{}>{\raggedright\arraybackslash}p{3.3cm}>{\raggedright\arraybackslash}p{2.4cm}>{\raggedright\arraybackslash}p{5.0cm}>{\raggedright\arraybackslash}p{2.0cm}@{}}
\toprule
Rival account & Tasks & Key result & Verdict \\
\midrule
One decider's taste (decider holdout, offerer-disjoint retrain) & Math.SE bounty, SO bounty, SO accepted & seen vs unseen deciders $-.003$ $[-.024,+.019]$, $+.003$ $[-.017,+.020]$; retrain $-.004$ $[-.015,+.006]$ & not supported \\
Timing and exposure & jokes, stories, SO votes, RoyalRoad, tweets & jokes $-.000$; stories $+.097\to+.037$, within-quintile shrink $\approx30\%$; others $\le .016$ & mixed (stories) \\
Coarse criterion readout & jokes, stories, ICLR, citations & $\Delta$ $+.009$, $+.012$, $-.004$, $+.033$ [CI crosses 0] & not tripped; instrument \\
Judge noise & same four & paraphrase ICC $.83/.84/.62/.69$; six-readout average $+.008$ & not tripped; instrument \\
One holistic judgment & citations, jokes, stories, ICLR & $\mathcal{VA}$+holistic $.868$ vs full $.884$; stories below bank & instrument only (reading retracted) \\
Vocabulary limit (description bottleneck) & stories & description $.746$ vs raw $.714$ at 6k rows (three seeds), $.763$ vs $.787$ at 24k, $.771$ vs $.802$ on all rows; neutral rewrite $.741$ & description holds 77\% of the gap at full size \\
Individual articulated taste & Math.SE bounty, SO bounty, PR author, N\&C agency & recovers 5\% / 2\%; identity gains $+.003$, $+.031$, CIs span 0 & not supported \\
Subpopulation & jokes, stories & top 10\% carry 47\% / 36\%; judge-SD $\rho\approx0$ & descriptive \\
Who decides (cross-task) & 28 tasks & $n$ deciders $\rho=.46$ ($p=.016$); LOO $R^2\approx.15$ & modest, real \\
Label shape & citations; 12 tasks & gap by year $+.279/+.193/+.088$; length within community $\rho=+.89$ & stands \\
Label noise & ICLR & disagreement terciles $+.110/+.092/+.162$ & not supported \\
Relative labels & 28 tasks; 2 within-group & $.150$ vs $.210$ ($p=.26$); 1 of 2 & not supported \\
Norm drift & ICLR, spotlight, press, N\&C rule change & oldest $>$ newest on 1--2 of 4 & killed \\
Temporal holdout (train $\le t$, read $>t$) & 9 dated tasks; full model retrained on 5 & $\mathcal{VA}$ drift within $\pm.01$ on 5 of 9; $T$ drift $+.066$ on press (below $\mathcal{VA}$) and $+.042$ on jokes (gap $.060 \to .007$); $+.011$ on ICLR, $+.007$ on BBC, $+.009$ on spotlight (gaps $.156 \to .144$, $.049 \to .046$, $.107 \to .081$) & full model stales on press and jokes, holds on ICLR, BBC and spotlight; criteria do not stale \\
Identifiable kind of item & 13 tasks & no class-agnostic property ($.47$--$.54$ vs placebo cutoffs $.52$--$.58$); class-conditional signal on stories, Math.SE votes, BBC, spotlight (bank $.60$--$.68$ vs placebo $\le .54$); gap text-predictable only on N\&C, HashtagWars & class-conditional only \\
Exposure channels named in the item reads & 9 tasks & press within company $.107\to.065$; others move by $\le .012$ & press only \\
Gap-steered criteria close the gap & stories, ICLR, N\&C, jokes, citations & $\Delta$ $-.000$, $+.017$ [CI crosses 0], $-.004$, $-.011$, $-.006$; placebo identical & not supported \\
\bottomrule
\end{tabular}
\caption{Rival accounts of the articulability gap. Gaps are held-out AUC differences $\mathcal{VAT}-\mathcal{VA}$ on the frozen frame; brackets are group-paired bootstrap intervals (2,000 draws). ``Instrument'' marks tests that bound our measuring instrument rather than human articulability.}
\label{tab:gap-battery}
\end{table}

\paragraph{Pretraining exposure of the items.}The full model's base, Llama-3.1-8B, was pretrained on web text with a cutoff of December 2023, and popular items are duplicated more often on the web, so the full model could predict popularity from familiarity with the item itself rather than from its text. We therefore split the held-out rows of every task with item dates at the cutoff (year $\le 2023$ against $\ge 2024$) and compared the gap (text-only full model minus articulated model, both from the item-gap pipeline's calibrated held-out predictions) on the two sides, with a paired grouped bootstrap of $3{,}000$ draws over the task's grouping unit. Five tasks have at least $150$ rows on both sides. None shows a larger gap on the items the base model could have seen: AoPS editorial approach $+.011$ before against $+.034$ after (difference $-.023$ $[-.136, +.116]$; $4{,}864$ and $338$ rows); BBC most-read $+.066$ against $+.055$ ($+.012$ $[-.034, +.064]$; $9{,}992$ and $155$); story upvotes $+.119$ against $+.113$ ($+.006$ $[-.043, +.055]$; $6{,}250$ and $641$); spotlight selection $+.099$ against $+.108$ ($-.008$ $[-.091, +.075]$; $928$ and $643$); and ICLR acceptance $+.088$ against $+.136$ ($-.048$ $[-.114, +.017]$; $583$ and $661$). The pooled difference, inverse-variance weighted over the five tasks, is $-.005$ $[-.033, +.023]$. Dating the peer-review items by submission year rather than venue year, since an ICLR paper is public on OpenReview the autumn before its venue year, moves the acceptance difference to $-.066$ $[-.137, +.001]$ and the spotlight difference to $-.054$ $[-.146, +.035]$: where the periods differ at all, the gap is larger on items the base model cannot have seen, the opposite of memorization. Uncalibrated predictions give the same readings on the three tasks that have them. Press pickup, joke upvotes, citation percentile and the code tasks have no items after the cutoff; the Math.SE and notice-and-comment tasks have fewer than $150$; the \#HashtagWars tweets are from 2017 and the homepage items carry no date. Two caveats: period is confounded with time, which the temporal holdout above studies on its own terms, and the articulated model's judge has a later pretraining cutoff than the full model's base, so the test bears on the full model's side of the gap only.

\subsection{\texorpdfstring{What the texts with the largest gaps have in common}{What the texts with the largest gaps have in common}}
\label{app:item-insights}

The previous subsection asked whether the items on which the full model beats the articulated model form a kind that one stated property picks out, and found no property that does so for both classes. This subsection describes those items instead, and then tests whether the descriptions, written as criteria or as nuisances, change the gap.

\paragraph{Selecting and reading the items.} On sixteen tasks we took up to 200 items on which the full model gives the true label a much higher calibrated probability than the articulated model (high-gap items), the same number of items that both models rate well, matched on class, length, the three strongest declared nuisances and, where the context id carries one, a coarse topic such as the contest, agency or company (matched items), and fifty items that the articulated model rates better than the full model (negative-gap items). Two language-model readers, gpt-5.6 and Claude Opus, first drafted candidate descriptions of what separates the high-gap items from the matched ones, reading the two classes apart: the high-gap preferred item is one the bank under-rates and the high-gap not-preferred item one it over-rates, so the same property usually marks both in opposite directions, and a pooled comparison cancels the two effects. Two human annotators then read the items in an interface that shows one item at a time with its class and the two models' probabilities of the true label, high-gap items first in order of gap and the matched low-gap items alongside, without any reader's flags. For each item they wrote what the text has that the criteria missed, marked which of the task's candidate properties it has, assigned the gap to one of five kinds (a property of the text, a criterion someone could state, how or where the item was posted, no reason they could name, or not tacit), and could flag it as an example. Their descriptions joined the candidate list and were measured in the same way. One check preceded the reads: where the matcher fell back to coarser keys, a declared nuisance was left unbalanced and would read as a finding (the stock-joke score, the HashtagWars boilerplate position, the BBC question and explainer channels), so those are declared and stratified rather than described.

\paragraph{Measuring and checking the descriptions.} We turned 89 descriptions into measurements: a pattern over the text, a structural column, or a criterion scored by the judge of record, counted present at a score of .5 or more. Each was measured on every held-out item of its task. Each measurement was then checked against eight random items flagged present and six flagged absent, recording how many were right, whether the property is about the text or about how the item was posted, produced or exposed, and what the property is relative to. %
Forty-seven measurements track what they name, meaning at least six of eight flagged items show the property and at most one unflagged item does. Of these, 42 describe the text and 5 describe process or authorship. The other 42 are noisy (30), invalid (11) or saturated (1). Table~\ref{tab:item-insights} lists the verified text properties, without two duplicates and one property whose effect is mechanical; Table~\ref{tab:item-insights-dropped} lists the rest with the reason each was set aside. Scope records what a property is relative to: \emph{text} (judgeable from the item alone), \emph{question} (the question the item answers, which is part of the item), \emph{proposal} (the proposed rule a comment answers, which is not part of the item), \emph{corpus} (other items, the venue's history or a field's beliefs), or \emph{outside} (world or cultural knowledge).

\begin{table}[!htbp]
\centering\scriptsize
\resizebox{\textwidth}{!}{%
\begin{tabular}{@{}l p{6.4cm} l r c r l c@{}}
\toprule
Task & Property & Measured by & Share of items & High-gap / matched (\%): preferred $\cdot$ not & Label AUC & Scope & Flag \\
\midrule
Story upvotes & Dialogue-heavy: 25\% or more of the story's characters in quotes & pattern & 35.4\% & 16/30 $\cdot$ 56/28 & $0.56$ & text &  \\
 & Payoff on a reference the reader must already know & judge & 16.8\% & 14/20 $\cdot$ 13/11 & $0.52$ & outside & N \\
 & Written as a document or account (report, chronicle, log, letter) & judge & 14.9\% & 19/21 $\cdot$ 9/6 & $0.48$ & text &  \\
 & Scenes of dialogue that do not change the situation & judge & 11.2\% & 7/5 $\cdot$ 13/10 & $0.51$ & text &  \\
\addlinespace[2pt]
HashtagWars pick & Minimal-edit substitution into a known title or lyric & judge & 50.5\% & -- $\cdot$ 73/61 & $0.50$ & outside & N \\
\addlinespace[2pt]
BBC most-read & Curiosity, oddity or explainer & judge & 41.6\% & 65/12 $\cdot$ 15/68 & $0.37$ & text & C \\
 & National politics, courts or conflict outside the UK and US & judge & 38.0\% & 20/35 $\cdot$ 47/33 & $0.47$ & text &  \\
 & Non-UK/US country named (Global South list) & pattern & 11.9\% & 5/6 $\cdot$ 26/6 & $0.47$ & text &  \\
 & Posed as a question & judge & 8.8\% & 17/2 $\cdot$ 0/0 & $0.45$ & text & C \\
 & Famous person without a linked news event & judge & 4.1\% & 19/1 $\cdot$ 0/3 & $0.49$ & outside &  \\
 & British royal family & judge & 2.0\% & 8/1 $\cdot$ 1/0 & $0.50$ & outside &  \\
 & Procedural step of a case (denies, delayed, appeal, hearing, jurors) & pattern & 1.7\% & 0/1 $\cdot$ 7/0 & $0.50$ & text &  \\
 & Moral-gravity words (abuse, rape, torture, war crimes) & pattern & 1.5\% & 1/1 $\cdot$ 7/2 & $0.50$ & text &  \\
\addlinespace[2pt]
Press pickup & Matter of public concern beyond customers & judge & 70.3\% & 84/81 $\cdot$ 69/63 & $0.56$ & outside &  \\
 & Statement responding to an event (verdict, lawsuit, crash, controversy) & judge & 19.6\% & 34/52 $\cdot$ 3/3 & $0.59$ & text &  \\
\addlinespace[2pt]
Math.SE answer votes & A hint or pointer that leaves the rest to the asker & judge & 37.8\% & 60/20 $\cdot$ 35/48 & $0.47$ & text &  \\
 & Short answer (under 300 characters) & pattern & 26.7\% & 40/10 $\cdot$ 23/34 & $0.47$ & text &  \\
 & Corrects a misconception in the question & judge & 18.9\% & 12/28 $\cdot$ 14/10 & $0.50$ & question &  \\
 & Verdict in the first sentence & judge & 7.4\% & 20/5 $\cdot$ 2/6 & $0.51$ & question &  \\
\addlinespace[2pt]
Math.SE accepted answer & A hint that leaves the rest to the asker & judge & 36.1\% & 38/25 $\cdot$ 33/43 & $0.47$ & text &  \\
 & Complete, self-contained derivation & judge & 36.1\% & 19/56 $\cdot$ 43/26 & $0.53$ & text &  \\
 & Verdict in the first sentence & judge & 6.4\% & 19/6 $\cdot$ 2/7 & $0.52$ & question &  \\
\addlinespace[2pt]
AoPS editorial approach & Complete rigorous proof with a clear closing & judge & 39.6\% & 44/61 $\cdot$ 23/12 & $0.63$ & text &  \\
 & Names its main technique or lemma & judge & 30.9\% & 32/25 $\cdot$ 52/34 & $0.48$ & text &  \\
\addlinespace[2pt]
ICLR accept/reject & `Novel' & pattern & 31.0\% & 17/25 $\cdot$ 39/28 & $0.49$ & text & N \\
 & `We show / prove / find that' & pattern & 28.5\% & 41/27 $\cdot$ 27/24 & $0.51$ & text &  \\
 & Finding about a class of methods or a widely held belief & judge & 27.8\% & 41/37 $\cdot$ 27/19 & $0.53$ & text &  \\
 & No numeric claim, no claim to beat prior methods & judge & 27.5\% & 27/34 $\cdot$ 25/24 & $0.50$ & text &  \\
 & Modifies one component of a named existing method & judge & 26.6\% & 20/25 $\cdot$ 29/31 & $0.48$ & text & N \\
 & `Extensive' or `comprehensive' experiments & pattern & 13.7\% & 8/10 $\cdot$ 23/10 & $0.50$ & text &  \\
 & Short abstract (under 800 characters) & pattern & 4.0\% & 7/2 $\cdot$ 4/6 & $0.48$ & text &  \\
 & Promotional closing (paves the way, opens new avenues, we believe) & pattern & 3.0\% & 2/0 $\cdot$ 5/1 & $0.50$ & text &  \\
\addlinespace[2pt]
Citation percentile & Mainly theoretical & judge & 31.0\% & 9/7 $\cdot$ 35/62 & $0.39$ & text &  \\
 & No numeric claim, no claim to beat prior methods & judge & 26.5\% & 18/9 $\cdot$ 32/26 & $0.45$ & text &  \\
 & Title names a method with an acronym or colon (`FastGCN: ...') & pattern & 15.9\% & 29/22 $\cdot$ 9/0 & $0.58$ & text &  \\
\addlinespace[2pt]
N\&C: received a response & Quotes or cites specific provisions and responds & judge & 61.2\% & 77/76 $\cdot$ 40/20 & $0.63$ & proposal &  \\
\addlinespace[2pt]
N\&C: rule change made & Narrow, specific change (a definition, number, wording, deadline) & judge & 45.9\% & 66/89 $\cdot$ 36/13 & $0.61$ & proposal &  \\
 & Supports the proposal as written & judge & 45.6\% & 34/35 $\cdot$ 38/58 & $0.45$ & text &  \\
 & Asks for something outside the agency's discretion & judge & 9.6\% & 9/0 $\cdot$ 13/12 & $0.48$ & outside &  \\
\bottomrule
\end{tabular}}
\caption{Text properties of high-gap items that survived a hand check, measured on every held-out item of the task. ``Share of items'' is the share of all held-out items with the property. The next column gives its share among high-gap and matched items, first among preferred items, then among not-preferred ones (-- where the high-gap group has no items of that class). Label AUC is that of the property's indicator over all items (above .5 means more common among preferred items). Flag: N marks a novelty or familiarity property, C a curiosity or surprise property.}
\label{tab:item-insights}
\end{table}

\paragraph{What the readers found.} Across the sixteen tasks, $49\%$ of high-gap items carry at least one property of the text that the readers named as a matter of taste, $52\%$ a criterion someone could state, $54\%$ a posting or exposure channel, and $10\%$ none of the named properties (an item can carry several). Where a property's flag fires, the check found the property present on $73\%$ of sampled items; where it does not fire, the property was absent on $93\%$. The readers named 129 properties; 89 could be measured, and 47 survived the check. The human annotators' reading of the same items will be reported alongside these numbers once complete.

\paragraph{What the verified properties show.} The verified properties show the contrast the reads were built on, and the contrast runs in opposite directions for the two classes. An oddity or explainer headline is 65\% of most-read high-gap BBC headlines against 12\% of matched ones, but 15\% against 68\% among the headlines that were not most-read. A hint that leaves the rest to the asker is 60\% of upvoted high-gap Math.SE answers against 20\%, and 35\% against 48\% among the rest. A complete, self-contained derivation, which the bank rewards, is 19\% of accepted high-gap answers against 56\% of matched ones. Because the properties mark both kinds of error, they barely predict the label: label AUCs run from .37 to .63, and most sit between .47 and .53. The largest contrasts are on rare properties: a famous person with no news event is 4\% of BBC headlines but 19\% of most-read high-gap ones (1\% of matched); a verdict in the first sentence is 6--7\% of Math.SE answers but about 20\% of preferred high-gap ones.

Of the 39 verified properties, 28 are judged from the text alone, 3 relative to the question the item answers, 2 relative to the proposed rule, and 6 need outside knowledge: a reference the reader must already know, a known title, fame, the royal family, public concern. None is relative to the corpus. Five of the 89 descriptions were relative to the corpus: a cheap or overused move, the obvious pun many players send, a finding that overturns a belief (pattern and judge), and a repeated entry. None of the five worked as an item-level criterion. The judge either fired on nearly every item (99\% of jokes are ``cheap or overused'') or tracked something else, and the only valid one is mechanical, since staff pick at most one copy of a repeated entry. A judge that reads one text at a time has no reference set against which to call a text familiar. Familiarity is measured in one place, the stock-joke score, and there it is a declared nuisance of the joke task (Table~\ref{tab:ledgers}); whether familiarity is exposure or taste is a construct choice we flag and leave open. The novelty and curiosity properties that do survive are the BBC oddity headline and question headline, the story payoff on a shared reference, the HashtagWars substitution into a known title, the ICLR one-component modification, and the word ``novel''. Claimed novelty is over-credited by the bank: ``novel'' is 17\% of accepted high-gap abstracts against 25\% of matched ones, and 39\% against 28\% among rejected ones.

\paragraph{Implementing the descriptions.} We then implemented every description, verified or not. Genuine and measurable descriptions were added to the bank as columns, and $\mathcal{VA}$ and $\mathcal{VAT}$ were refit with the frozen stack. Spurious descriptions were declared as nuisances, and both models were read within their strata, on all held-out items (Table~\ref{tab:item-insights-impl}). In the bank, the descriptions shrink the gap on six of the fifteen tasks with descriptions to add, by at most .025 AUC (citations .194 to .169, Math.SE accepted answers .073 to .056, spotlight picks .100 to .088). On the others the gap grows, because the added columns add noise that the aggregator does not discount; the largest rise is on HashtagWars, .175 to .218, a task with 80 preferred items. $\mathcal{VAT}$ moves by at most .018. As nuisances, the channels already in a task's ledger reproduce what the deconfounded frame removes: read within the HashtagWars boilerplate strata, the raw gap falls from .175 to .079. The channels the reads added move the gap by at most .012 in either direction, except on press pickup (.107 to .087; .065 read within company, Table~\ref{tab:item-spurious}), homepage placement (.084 to .073) and N\&C responses, where it grows (.021 to .038).

\begin{table}[!htbp]
\centering\scriptsize
\resizebox{\textwidth}{!}{\begin{tabular}{@{}l r r r r r r r@{}}
\toprule
 & \multicolumn{2}{c}{descriptions in the bank} & \multicolumn{3}{c}{new spurious channels} & & \\
\cmidrule(lr){2-3}\cmidrule(lr){4-6}
Task & added & gap: raw $\to$ with them & declared & gap within strata & both & $\mathcal{VA}$: raw $\to$ with & $\mathcal{VAT}$: raw $\to$ with \\
\midrule
Story upvotes & 9 & $.121 \to .122$ & 3 & $.109$ & $.109$ & $.667 \to .666$ & $.783 \to .784$ \\
Joke upvotes & 7 & $.025 \to .031$ & 3 & $.026$ & $.034$ & $.716 \to .712$ & $.740 \to .743$ \\
HashtagWars pick & 4 & $.175 \to .218$ & 3 & $.181$ & $.226$ & $.527 \to .485$ & $.572 \to .577$ \\
BBC most-read & 11 & $.066 \to .061$ & 2 & $.066$ & $.061$ & $.754 \to .759$ & $.816 \to .816$ \\
Press pickup & 3 & $.107 \to .118$ & 7 & $.087$ & $.076$ & $.644 \to .633$ & $.720 \to .702$ \\
Homepage placement & 0 & $.084$ & 10 & $.073$ & -- & $.641$ & $.707$ \\
Math.SE answer votes & 8 & $.031 \to .038$ & 4 & $.025$ & $.034$ & $.613 \to .606$ & $.641 \to .639$ \\
Math.SE accepted answer & 9 & $.073 \to .056$ & 4 & $.078$ & $.060$ & $.564 \to .581$ & $.619 \to .615$ \\
AoPS editorial approach & 8 & $.020 \to .019$ & 3 & $.027$ & $.026$ & $.756 \to .756$ & $.775 \to .774$ \\
Pull requests & 3 & $.046 \to .050$ & 10 & $.040$ & $.038$ & $.676 \to .677$ & $.722 \to .726$ \\
ICLR accept/reject & 12 & $.112 \to .117$ & 1 & $.115$ & $.121$ & $.654 \to .649$ & $.726 \to .718$ \\
Citation percentile & 6 & $.194 \to .169$ & 3 & $.189$ & $.171$ & $.673 \to .698$ & $.840 \to .844$ \\
Spotlight/oral pick & 6 & $.100 \to .088$ & 1 & $.100$ & $.087$ & $.467 \to .480$ & $.508 \to .499$ \\
N\&C: received a response & 7 & $.021 \to .018$ & 3 & $.038$ & $.039$ & $.790 \to .793$ & $.811 \to .811$ \\
N\&C: rule change made & 10 & $.002 \to .008$ & 1 & $.014$ & $.009$ & $.607 \to .612$ & $.609 \to .620$ \\
N\&C agreement & 4 & $.062 \to .090$ & 0 & -- & -- & $.504 \to .476$ & $.521 \to .514$ \\
\bottomrule
\end{tabular}}
\caption{Implementing every description from the item-level reads, on all held-out items. ``Descriptions in the bank'': genuine and measurable descriptions, verified or not, added as bank columns and the stack refit. ``New spurious channels'': channels not already in the task's ledger, with both models read within the strata of up to four of them. Gap is $\max(T,\mathcal{VAT})-\mathcal{VA}$ in AUC on the raw frame (the bank and the text-only full model refit together), so the raw values differ from Table~\ref{tab:master-results}.}
\label{tab:item-insights-impl}
\end{table}

\begin{table}[!htbp]
\centering\scriptsize
\begin{tabular}{@{}l p{7.2cm} r r@{}}
\toprule
Task & Excerpt (preferred item; the full model rates it well, the bank poorly) & $p_T$ & $p_{\mathcal{VA}}$ \\
\midrule
Story upvotes & ``The creatures were all smiles. A bunch of tall, pale, ghostly things with these dark smiles.'' (prompt: humans are a thousand times more sensitive to the smell of petrichor than sharks are to blood) & $.81$ & $.18$ \\
Story upvotes & ``Before we first discovered alien life, our best astronomers believed that we were the only ones in our galaxy.'' (a future history in essay form, no scene and no dialogue) & $.82$ & $.10$ \\
Joke upvotes & ``I like to imagine the guy who invented the umbrella was going to call it the `brella'\ldots\ But he hesitated\ldots'' & $.88$ & $.28$ \\
Joke upvotes & ``These no nut November memes / They're really getting out of hand'' & $.72$ & $.18$ \\
Math.SE answer votes & ``Yes. Since $\mathbb{Z}$ is free, the sequence splits and you get what you want.'' & $.65$ & $.27$ \\
Math.SE accepted answer & ``This may actually be a matter of `just practice more.' If you've done enough factoring, you'll recognize the coefficients\ldots'' & $.51$ & $.21$ \\
BBC most-read & ``Aphantasia: A life without mental images'' & $.62$ & $.10$ \\
BBC most-read & ``Why is Archie not a prince?'' & $.86$ & $.10$ \\
Press pickup & Target, ``Target Addresses Firearms in Stores'': a note from the interim chief executive to staff, asking guests not to bring firearms into stores & $1.00$ & $.36$ \\
Press pickup & Tesla, ``Regarding Today's Jury Verdict'': a pasted note to employees about a racial-harassment verdict & $.71$ & $.17$ \\
ICLR accept/reject & ``We formulate language modeling as a matrix factorization problem, and show that the expressiveness of Softmax-based models\ldots is limited by a Softmax bottleneck.'' & $.76$ & $.16$ \\
Citation percentile & ``Semi-Supervised Classification with Graph Convolutional Networks'': no number in the abstract & $.96$ & $.30$ \\
N\&C: received a response & ``I strongly urge the EPA to add the Blades, Delaware site to the NPL\ldots many many people in the area rely on private wells'' & $1.00$ & $.25$ \\
\bottomrule
\end{tabular}
\caption{Examples of preferred high-gap items. $p_T$ and $p_{\mathcal{VA}}$ are the calibrated probabilities of the true label under the text-only full model and the articulated model.}
\label{tab:item-insights-examples}
\end{table}

\begin{table}[!htbp]
\centering\scriptsize
\resizebox{\textwidth}{!}{\begin{tabular}{@{}l p{5.6cm} l p{5.4cm}@{}}
\toprule
Task & Description & Verdict & Reason \\
\midrule
Story upvotes & Uses the prompt as a springboard and goes elsewhere & invalid & fires on 73\% of stories; most flagged stories carry out the prompt \\
 & Single narrating voice (under 3\% of characters in quotes) & noisy & dialogue marked with dashes passes; the cut splits mostly narrated stories \\
 & Withholds a key fact about its situation or narrator & noisy & fires on 49\%; misses stories that do withhold one \\
 & Serialised part or excerpt & noisy, process & records how the story was posted \\
\addlinespace[2pt]
Joke upvotes & Pun by sound or spelling, including one left unspelled & noisy & flags non-puns; misses puns whose answer is not written out \\
 & Deadpan literal statement that refuses to be a joke & noisy & reads title-only fragments as deadpan \\
 & Setup and punchline paced by punctuation & noisy & flags plain question-and-answer jokes inconsistently \\
 & Says it works better out loud & noisy, process & two items; a note about delivery \\
 & A move regular readers would treat as cheap or overused & saturated & fires on 99\% of jokes \\
 & About the subreddit, reposting or the poster's own joke-telling & process & measures platform meta-culture, not the joke \\
 & Punchline left out; the reader completes it from a template & data & most flagged posts are titles whose body is missing from the data \\
\addlinespace[2pt]
HashtagWars pick & Same entry text appears more than once in its contest & mechanical & staff pick at most one copy of a repeated entry \\
 & An obvious pun many players would send & noisy & flags entries that are obvious but not puns, or puns but not obvious \\
 & An observation rather than a joke & noisy & flags title puns as observations \\
\addlinespace[2pt]
BBC most-read & Question headline (pattern) & invalid & counts every wh-word opener; the judge version is kept \\
\addlinespace[2pt]
Press pickup & Product or milestone with a specific number & noisy & fires on any number, e.g.\ account-takedown counts \\
\addlinespace[2pt]
Math.SE answer votes & Substance is a pointer (a link or `see') & noisy & fires on any `see' inside a full answer \\
 & Opens with a verdict on the asker's attempt (pattern) & noisy & catches yes/no answers; the judge version is kept \\
 & Engages the asker's own attempt or wording & noisy & items carry only the question title, so the attempt is not visible \\
 & Complete, self-contained derivation & noisy & flags unfinished derivations (valid on accepted answers) \\
\addlinespace[2pt]
Math.SE accepted answer & Comment-shaped reply & invalid & flags short Socratic hints \\
 & Corrects a misconception in the question & invalid & flags plain answers (valid on answer votes) \\
 & Deliberately partial & noisy & loose phrases such as `to obtain' match complete answers \\
 & Engages the asker's own attempt or wording & noisy & question title only \\
 & Soft question (reference, advice, how to learn) & process & a property of the question, not the answer \\
\addlinespace[2pt]
AoPS editorial approach & Sketch or outline & invalid & fires on 80\%; calls short complete solutions sketches \\
 & Informal or sketch post (pattern) & noisy & flags full proofs with a casual opener \\
 & Flags its own error & noisy & `oops' as a label; misses `my answer above is incorrect' \\
 & Closed with a proof-end mark & noisy, process & the mark is a few prolific posters' template \\
\addlinespace[2pt]
ICLR accept/reject & Overturns or qualifies an accepted belief (pattern) & invalid & fires on `surprising' in any sense \\
 & Overturns or qualifies an accepted explanation & noisy & fires on routine `prior methods have a limitation' statements \\
 & Ends with a broad claim of significance & noisy & fires on routine closing lines \\
\addlinespace[2pt]
Citation percentile & No number and no beat-claim (pattern) & noisy & counts `3D' as a number; misses some beat-claims \\
 & Named benchmark & noisy & a fixed list of well-known names \\
 & Introduces a named method, dataset or framework & noisy & ignores `named'; fires on 83\% \\
\addlinespace[2pt]
N\&C: received a response & Quotes the rule or cites a numbered provision (pattern) & invalid & fires on any long quotation and on docket numbers \\
 & Objection or reservation with no praise & invalid & flags supportive comments; misses plain opposition \\
 & One discrete request the agency could answer & noisy & flags agency letters posted to the docket \\
 & Plain prose without letterhead (pattern) & noisy, authorship & separates organisations' letters from individuals' web-form comments \\
 & Federal Register citation & noisy, format & a header convention of lawyered letters \\
 & Plain prose without letterhead (judge) & authorship & same, measured cleanly \\
\addlinespace[2pt]
N\&C: rule change made & Asks to withdraw, rescind or reverse (pattern) & invalid & keyword match in any context \\
 & Quotes the rule or cites a numbered provision (pattern) & invalid & fires on quotations from papers and news \\
 & Asks to withdraw, reverse or rewrite (judge) & invalid & fires on narrow technical requests \\
 & Endorses the proposal (pattern) & noisy & `urge' and `in support of' fire in requests for more \\
 & Quotes or cites provisions and responds (judge) & noisy & applied inconsistently to equivalent comments \\
 & Private individual (pattern) & noisy, authorship & precise, low recall \\
 & Private individual (judge) & authorship & who filed, not what was written \\
\bottomrule
\end{tabular}}
\caption{Descriptions set aside after the hand check. ``Noisy'': between two and five of eight flagged items show the property, or two or more unflagged items do. ``Invalid'': the flag does not track the property. ``Process'', ``format'' and ``authorship'': the property describes how or by whom an item was posted or produced, not what it says. ``Data'': the flag tracks missing text. Noisy is not refuted; several of the readers' strongest descriptions are here because our patterns and criteria did not measure them cleanly.}
\label{tab:item-insights-dropped}
\end{table}

\paragraph{Limits.} The candidate descriptions were drafted by language models describing what another language model appears to use, and the annotators read the items already sorted by gap, so they knew which items the criteria had missed. Every description was written after reading 150 to 400 items of its task, although it was measured and checked on all of them. The check saw texts cut at 1,500 characters, so closings and long-range structure could not be checked in long posts. Math.SE items carry only the question's title, so every property relative to the asker's own attempt was judged without seeing the attempt and is set aside as noisy. Two joke descriptions the readers stressed, the missing punchline and jokes about the subreddit, turned out to describe missing data and the platform's culture.

\FloatBarrier
\subsection{Negative results}
\label{app:negative}

This subsection reports the results of the alternative articulation architectures
we built and could not make work. These failures matter to the paper's argument
because each one is an attempt to \emph{close} the articulability gap with a stronger
articulated instrument, so each failure rules out a family of objections to
the articulated frontier reported in the main text: that the flat bank
wastes resolution (the metric tree), that the aggregator is too weak (the
capacity heads), that per-item explanation could carry what per-corpus
criteria cannot (metric-STaR), or that our optimization-consequence
experiments were under-powered by construction (single-generator pools). All
comparisons below are grouped out-of-fold on the same rows and folds as the
flat criteria bank they are compared against; none is a change of corpus or
split.

\paragraph{Hierarchical / selective metric application (metric trees).}
The metric tree was our first attempt to close the gap by structure rather than by more criteria. A flat bank scores every item on every criterion, but a field does not judge every kind of work by the same criteria, so a tree that routes items to subpopulations and fits each subpopulation its own model should recover what the flat bank misses. We tested
the strongest inexpensive version of the tree: a recursive binary-partition tree
over the \emph{certified} criteria matrix (splits chosen from each node's
locally most-discriminative criteria; leaves fit the frozen aggregation
recipe on the node's rows and its locally top-$24$ criteria; the entire tree
rebuilt inside every cross-validation fold). It \textbf{loses} to the flat
bank on identical folds: $-.058$ AUC on creative-writing upvotes ($.612$ vs
$.670$) and $-.027$ on peer-review verdicts ($.619$ vs $.646$). The tree loses for two reasons.
First, its costs are estimation costs: leaves
see $5$--$25\times$ less data than the flat model, local feature selection
on a few hundred rows is itself noisy, and hard routing on a judge score
that fluctuates between $0.5$ and $1.0$ sends items to entirely different
models. Second, the flat model's
gradient-boosted head \emph{already} applies criteria conditionally:
it learns ``criterion $j$ matters only when criterion $i$ is high'' softly,
with full-sample efficiency, so explicit routing over a \emph{fixed}
criteria set adds no representational power, only variance. What the tree's
negative result does \emph{not} rule out is subpopulations needing criteria
\emph{absent from the shared bank}; that requires node-conditional criterion
mining, which we leave to future work. An earlier, LLM-proposed variant of
the tree (splits proposed by a prompted model at each node rather than drawn
from the certified bank) failed for an additional reason we documented at
the time: proposal quality degrades with depth, with deep nodes receiving
generic checklist criteria rather than slice-specific ones. The tree and the subfield infilling of Appendix~\ref{app:subfields} answer the same question two ways: whether subfield structure lies in the \emph{weights} over a shared criteria vocabulary or in the \emph{vocabulary} itself. The tree tested the weights answer (it routes each item to a subpopulation and refits the same criteria locally) and lost to the flat bank in both tasks we tried, because a gradient-boosted head over a fixed criteria matrix already applies criteria conditionally, so hard routing adds no expressive power and increases estimation variance at the leaves. What a subfield adds, the infilling signatures suggest, is instead a small set of criteria absent from the shared bank, the one possibility the tree result could not rule out. Subcommunity discovery then becomes a mining problem: the shared core is held fixed, mining (with labels used only on the training and monitoring splits) runs on one subcommunity's residual, and a candidate is admitted only on held-out lift within that subcommunity. Our moderation tasks independently support measuring within subcommunities: refit within a single subcommunity, the articulated model recovers substantially more of the dense model's signal than the pooled fit, since pooled fits absorb cross-community norm heterogeneity (one community's removable offense is another's genre convention) and count it as apparent tacitness. Subfield-resolved measurement is a correctness issue for the gap estimates themselves; the per-subreddit rebuild in our moderation tasks extends the per-agency and per-venue refits of \S\ref{sec:res-frontier} (Q4), and the per-agency mining of Table~\ref{tab:agency-mining}, to the academic and legal subfields, where the infilling signatures were first observed.

\paragraph{Higher-capacity aggregation heads.} A broad head sweep on the two bounty tasks (identical grouped five-fold splits; $60$ and $86$ criteria) bounds the ``underexpressive aggregator'' alternative directly: frozen gradient boosting $.719$/$.776$; $L_2$ and $L_1$ logistic $.717$/$.772$; deeper and shallower boosting $.718$--$.720$/$.774$--$.776$; random forest and extra-trees $.720$/$.773$--$.774$; multilayer perceptrons of widths $64$, $256{\times}64$, and $512{\times}128{\times}32$: $.712$--$.713$/$.768$--$.771$; degree-2 interactions of the twenty strongest criteria $.707$/$.761$; an RBF kernel machine $.694$/$.744$; and a grouped-OOF stack of all heads $.723$/$.779$. The stack's $+.003$--$.004$ over the frozen recipe is within its seed spread (\texttt{results/aggregator\_sweep.json}), so no head beats the frozen recipe by more than seed noise. Replacing the gradient-boosted
head with multilayer perceptrons ($64$; $256{\times}64$, standardized inputs,
early stopping) on the identical criteria matrix \emph{reduces} held-out AUC
in both tasks tested (creative writing: $.651$--$.659$ vs $.665$; peer
review: $.626$--$.640$ vs $.668$). The articulability gap is not an
aggregation-capacity artifact.

\paragraph{Where nonlinearity would have to live.} The sweep above tests whether any head beats the frozen one; the ladder below tests what kind of nonlinearity the bank would need. On identical grouped five-fold splits we fit, in order: a linear head; an additive spline model (per-criterion splines, no interactions); gradient-boosted trees of depth two (pairwise interactions only), four and six; a leaf-limited tree ensemble; $k$-nearest neighbours ($k{=}50$); and random Fourier features with a logistic readout. Math.SE bounty ($n{=}23{,}972$, $59$ features): linear $.717$, splines $.718$, depth-2 $.720$, depth-4 $.720$, depth-6 $.720$, $k$-NN $.699$, Fourier $.681$. Stack Overflow bounty ($n{=}23{,}498$, $85$ features): $.772$, $.772$, $.774$, $.776$, $.776$, $k$-NN $.754$, Fourier $.684$. Creative writing ($n{=}6{,}600$, $60$ features): $.658$, $.654$, $.655$, $.643$, $.639$, $k$-NN $.636$, Fourier $.625$. Measured as increments, additive nonlinearity is worth $+.002$, $-.000$ and $-.004$; pairwise interactions $+.001$, $+.002$ and $+.001$; anything beyond pairwise $+.000$, $+.002$ and $-.012$. We did not enumerate interaction terms explicitly, since their number grows combinatorially with the bank size; the depth ladder bounds their order instead.

\paragraph{Reconstructing the full model from the criteria.} Section~\ref{sec:res-frontier} (Q3) asks whether the full model's ranking can be rebuilt from the criteria. For each task we regress the full model's held-out score on the criteria bank (gradient-boosted regression, grouped five-fold out-of-fold predictions) and take the item-level Spearman correlation between the reconstruction and the full model's score. Across the twelve tasks of this sweep (the two New Yorker caption-contest tasks excluded, as in every main-text average; 478 to 11,452 items per task), the median is $\rho=.56$ (95\% CI $[.47,.61]$, bootstrap over tasks; range $.34$--$.77$). The two examples in the main text come from a separate run of the same regression on rank-transformed scores: story upvotes $\rho=.53$ ($n{=}7{,}008$ items, 95\% CI $[.51,.55]$, bootstrap over 5,136 prompt groups) and ICLR accept/reject $\rho=.64$ ($n{=}1{,}244$, $[.60,.68]$, 1,239 groups). In that run, adding the unreconstructed residual to the articulated model raises story-upvote AUC from $.665$ to $.773$, $85\%$ of the $.127$ gap to the full model's $.792$; on ICLR it raises AUC from $.668$ to $.732$, $59\%$ of the gap.

\paragraph{Per-item rationalization (metric-STaR).} Training a small model to
articulate per-item rationales whose text alone supports label recovery (a
STaR-style loop with an information-bottleneck judge) did not produce a
competitive articulated predictor in our earlier experiments: three
STaR iterations on creative writing lifted judge-recovers-label accuracy from
$48\%$ to $56.2\%$ on 500 held-out items. The gain is real but modest, and it is
measured in an accuracy frame that is not commensurable with the bank's AUC
numbers (the run also predates the current corpora; a rerun on the paper's
tasks is queued). We still regard per-item rationalization as the natural unbounded-vocabulary,
bounded-length articulation channel and return to it in future work.

\paragraph{Best-of-$N$ selection over single-generator pools.} Our first
optimization-consequence experiment generated $K{=}16$ candidates per prompt
from one model at fixed temperature and found \emph{no} detectable
divergence between rubric-selected and preference-model-selected outputs.
The null reflected the pools rather than the instruments: same-generator pools
are quality-homogeneous. A third of bank criteria collapse to a constant
on them (zero collapse on real text), and two independently-trained text-only
models that agree at $\rho{=}.89$ across prompts agree at only $\rho{=}.21$
within a pool. All instruments were therefore ranking noise. Selection
experiments over generated text require diverse pools (multiple generator
families, instruction variation, and mixed-in real items); with those, the
predicted asymmetry appears (\S\ref{sec:res-optim}).

\paragraph{Feeding the bank to the text-only model (criteria conditioning).} The
reverse of the aggregation objection is the feature-access objection:
the text-only model may win because it sees features the bank encodes but
the aggregation discards, in which case handing the text-only model the bank's
own scores should be redundant, while handing the \emph{bank's} scores to a
text-only model that lacked them should help. We trained, on identical splits, a
LoRA whose prompt carries each item's text plus its ten most informative
criterion scores. Conditioning on the criteria gives a consistent null everywhere the text-only model
is strong (largest effect $|\Delta| =.011$ across nine within-community
contrasts, every interval covering zero; $.849$ vs $.849$ and $.914$ vs
$.911$ on the two large moderation tasks), with a single-task exception where
the text-only model is weakest (peer-review spotlight, $+.045$). An intermediate
hypothesis we considered, that conditioning helps precisely where text-only
models are weak (which would have made the criteria a correction for
under-trained text-only models), does not survive the full grid: the conditioning
gain shows no relationship with text-only-model strength (Spearman $-.14$, $p{=}.76$).
Appendix~\ref{app:new-tasks} reports the full grid.

\paragraph{Labels and judges that failed their own audits.} Three
instrument-level constructions were built, audited, and rejected, and we
record them because each is a tempting shortcut. A \emph{friction} label for
mathlib pull requests ($y =$ zero review threads) turned out to be circular
(the label is definitionally recoverable from the review metadata it was
meant to summarize) and was replaced by the accept/reject label
(Appendix~\ref{app:balancing}). An \emph{any-response} union label for
regulatory comments (count a comment as successful if \emph{any} matched
agency response said a change was made) is confounded by the number of
matched responses, which mechanically gives more chances at a positive; the
label-count feature alone scores $.695$ on the union label against $.589$ on
the majority-vote label that replaced it. And bulk pseudo-labelling of
competition-code pairs by a smaller judge failed its agreement gate
($\kappa =.388$ against the reference judge, with a $61{:}1$ false-negative
asymmetry) and was replaced by cross-encoder distillation used only for
bulk triage. The pattern across all three is the same: an easily built label that
correlates with the construct through a side channel will pass casual
inspection and fail a targeted audit, which is why every label in the
benchmark carries a targeted audit (Appendix~\ref{app:balancing}). A fourth failure was found later and removed a task rather than a label. For patents, the prior-art candidate set that every model read (eight retrieved reference passages per claim) had the examiner's gold-cited reference appended for positive rows only ($99.7\%$ of positives, none of the negatives). The ``novelty'' $V$ columns were lexical overlap against that set, and the dense model read it too; in an ablation, deleting the claim hurt less than deleting the references. Nothing computed on that input is admissible, so we withdraw the patents row. A valid version of the check requires a pre-priority-date prior-art index queried identically for every row, which we have not built.

\paragraph{Naming the full model's own features (sparse autoencoder).} A sparse autoencoder trained on the full model's activations over the 7,008 stories yielded 194 features whose activation tracks the residual; the twelve strongest were verbalized from their top-activating stories (``immediate in-medias-res opening'', ``domestic subversion via intimate voice'', ``alien-POV human-exceptionalism reveal'', and so on) and the judge of record scored each verbalization as a criterion on every story. Fidelity, the rank correlation between the judge's score and the feature's activation, is $.00$--$.11$ for eleven of the twelve and $.24$ for one; the criteria add $-.003$ to $+.003$ held-out AUC each over the bank ($.667$) and $+.007$ jointly. The verbalizations do not measure the features, so the test says nothing about whether the features are articulable; it shows only that the verbalizer failed to name them.

\paragraph{Item-level attempts to name or close the gap.} Four further negative results are reported with their numbers in Appendix~\ref{app:gap-battery}. Criteria written from the items on which the full model beats the bank do not raise the bank when added to it; they move it by the same amounts as criteria written from a random split. No property stated by a strong annotator separates high-gap items from matched articulable items across both classes on any of thirteen tasks; the properties that hold on one class reverse on the other. The per-item gap is predictable from the text at all only where it tracks topic identity (N\&C agencies, HashtagWars contests), and not on stories, ICLR, jokes or citations. A first reading that found such properties on N\&C and HashtagWars was a class imbalance and a topic confound, and is retracted.

\section{Consequences of the gap}
\label{app:part-consequences}

This appendix expands the two consequences of Section~\ref{sec:res-selection} and Section~\ref{sec:res-optim}: selection regret on every task, and the generated-pool experiment.

\subsection{Selection regret on every task}
\label{app:regret-all}

Table~\ref{tab:regret} in the main text lists the fourteen non-caption tasks with their regret and disagreement odds. Table~\ref{tab:regret-all} adds the hit rates, swap rates and win counts for all sixteen, including the two caption-contest tasks ($^{\dagger}$) under the protocol of Section~\ref{sec:res-selection}. Regret is computed per group of candidates that share a context and contain both a chosen and an unchosen item; each model names its top item, and regret is the full model's hit rate minus the articulated model's, averaged over groups. The three peer-review tasks have singleton groups, so their readout is pairwise: regret is the difference in pairwise AUC and the disagreement columns count pairs. Intervals are 3,000-draw grouped bootstraps. We apply no multiplicity correction; the family-wise evidence is the sign count. Regret is positive in $13$ of $16$ tasks, negative in one (caption crowd, $-.250$, a caption-contest task that the main-text figures exclude but that we keep in the table and in the sign test), and zero in two (GitHub PR merged, where both models hit equally often; editorial pickup, where neither model's pick was the winner on any disagreeing group). A sign test over the fourteen non-zero tasks gives $p{=}.0009$. We report win counts on disagreeing groups rather than odds because two tasks have zero articulated-model wins. The random column is the hit rate of a uniform pick within the group.

\paragraph{Definition.} Let each item $i$ carry the community's recorded choice $y_i \in \{0,1\}$, the full model's score $t_i$, and the articulated model's out-of-fold score $b_i$, and let $g(i)$ be its candidate group (the prompt, docket, or contest it competed in). The \emph{decidable} groups are those with at least two items and at least one chosen and one unchosen item,
$\mathcal{G} = \{\, g: |g| \ge 2,\; 0 < \sum_{i \in g} y_i < |g| \,\}$.
Each model names one item per group, $i_T(g) = \argmax_{i \in g} t_i$ and $i_A(g) = \argmax_{i \in g} b_i$ (ties, which do not occur, would go to the lowest index), and
\begin{equation}
\mathrm{regret} \;=\; \frac{1}{|\mathcal{G}|}\sum_{g \in \mathcal{G}} \Big( y_{i_T(g)} - y_{i_A(g)} \Big) \;=\; \mathrm{hit}_T - \mathrm{hit}_A,
\label{eq:regret}
\end{equation}
so each group contributes $+1$ when only the full model's pick is the community's choice, $-1$ when only the articulated model's is, and $0$ otherwise. Writing $D$ and $A$ for the numbers of disagreeing groups won by each model, $\mathrm{regret} = (D - A)/|\mathcal{G}|$ exactly, the disagreement odds are $D{:}A$, and $|\mathrm{regret}|$ cannot exceed the swap rate (the fraction of groups where the picks differ). Groups are unweighted; the interval is a 3,000-draw bootstrap over groups. For the three peer-review tasks, whose title groups are singletons, the readout falls back to all chosen--unchosen pairs with strict orderings, where the same quantity reduces to $\Pr[t_i > t_j] - \Pr[b_i > b_j] = \mathrm{AUC}_{VAT} - \mathrm{AUC}_{VA}$; those rows therefore restate the articulability gap rather than test selection, which is why the within-task dose-response below pools only the grouped tasks.

\begin{table}[!htbp]
\centering\scriptsize
\setlength{\tabcolsep}{3.5pt}
\resizebox{\textwidth}{!}{%
\begin{tabular}{lllrrrrrrr}
\toprule
task & readout & units & hit (artic.) & hit (full) & hit (random) & regret & 95\% CI & swap rate & wins full:artic. \\
\midrule
peer review: citations$^{*}$ & pairwise & 56,901 pairs &.714 &.884 &.500 & +.171 & [+.125, +.216] & 0.27 & 12565:2852 \\
humor: \#HashtagWars picks & grouped & 8 groups &.125 &.250 &.099 & +.125 & [$-$.250, +.500] & 1.00 & 2:1 \\
peer review: accept/reject$^{*}$ & pairwise & 383,426 pairs &.668 &.777 &.500 & +.109 & [+.078, +.140] & 0.31 & 80645:38878 \\
N\&C: rule change & grouped & 69 groups &.420 &.522 &.453 & +.101 & [$-$.043, +.246] & 0.65 & 16:9 \\
humor: joke upvotes & grouped & 10 groups &.900 & 1.000 &.492 & +.100 & [+.000, +.300] & 1.00 & 1:0 \\
N\&C: received response & grouped & 20 groups &.750 &.850 &.684 & +.100 & [$-$.100, +.300] & 0.65 & 3:1 \\
N\&C: cosigner agreement & grouped & 65 groups &.492 &.585 &.523 & +.092 & [$-$.062, +.246] & 0.62 & 16:10 \\
math.SE: accepted answers$^{*}$ & grouped & 992 groups &.501 &.568 &.424 & +.067 & [+.029, +.103] & 0.45 & 218:152 \\
humor: caption finalist$^{\dagger}$ & grouped & 46 groups &.217 &.283 &.128 & +.065 & [$-$.087, +.239] & 0.96 & 9:6 \\
peer review: spotlight$^{*}$ & pairwise & 460,148 pairs &.529 &.594 &.500 & +.065 & [+.020, +.109] & 0.45 & 118958:89239 \\
creative writing: upvotes$^{*}$ & grouped & 523 groups &.667 &.728 &.542 & +.061 & [+.013, +.107] & 0.46 & 101:69 \\
math.SE: answer votes & grouped & 1,140 groups &.609 &.650 &.500 & +.041 & [+.006, +.075] & 0.37 & 230:183 \\
math: AoPS editorial approach & grouped & 310 groups &.797 &.800 &.639 & +.003 & [$-$.048, +.055] & 0.73 & 36:35 \\
code: GitHub PR merged & grouped & 202 groups &.911 &.911 &.809 & +.000 & [$-$.050, +.050] & 0.89 & 13:13 \\
journalism: editorial pickup & grouped & 8 groups &.750 &.750 &.475 & +.000 & [+.000, +.000] & 0.88 & 0:0 \\
humor: caption crowd$^{\dagger}$$^{*}$ & grouped & 64 groups &.750 &.500 &.500 & $-$.250 & [$-$.422, $-$.094] & 0.97 & 9:25 \\
\bottomrule
\end{tabular}}
\caption{Selection regret on all sixteen tasks. Columns: readout (grouped or pairwise); units (decidable groups or pairs); hit rate of the articulated model, of the full model and of a uniform random pick; regret with its 95\% grouped-bootstrap interval; swap rate (share of units on which the two models name different items); and, on those units, how often the full model's pick vs.\ the articulated model's was the community's choice. $^{*}$: interval excludes zero. $^{\dagger}$: New Yorker caption-contest tasks; we do not include these results in any main-text table or average, as the task is multimodal (the cartoon each caption answers is unavailable to every instrument).}
\label{tab:regret-all}
\end{table}

\paragraph{Within-task dose-response.} The main text reports the within-task correlation between the induced gap and regret only in summary; Table~\ref{tab:dose} gives the per-task values. For each task we keep a random subset of $k$ criteria from the articulated bank, $k \in \{4, 8, 16, 32, 64\}$ and the full bank, three draws each, refit the articulated model on the frozen selection stack, and record the induced gap $(\mathrm{AUC}_{VAT}-\mathrm{AUC}_{VA})/(\mathrm{AUC}_{VAT}-.5)$ and the selection regret of that subset. Table~\ref{tab:dose} gives the Spearman correlation between the two over the sixteen points of each task. Over the eight tasks with grouped readouts the median is $+.65$ (95\% CI $[.08,.77]$, bootstrap over the eight tasks), seven are positive, and a one-sided Wilcoxon signed-rank test across tasks gives $p{=}.012$ (two-sided $.023$); restricted to $k \ge 16$ the median is $+.60$ ($p{=}.008$). The three peer-review tasks are excluded: their groups are singletons, regret there is the pairwise-AUC difference, and the correlation is $1$ by construction. Three of the positive tasks have ten or fewer decidable groups, so their regret moves in steps of $.10$ to $.125$; the two notice-and-comment tasks are the near-nulls for the same reason.

\begin{table}[!htbp]
\centering\footnotesize
\resizebox{\textwidth}{!}{%
\begin{tabular}{lrrccrr}
\toprule
task & criteria & decidable groups & induced gap & regret & $\rho$ [95\% CI] & $p$ \\
\midrule
math.SE: answer votes & 92 & 1,140 &.17--.77 &.021--.119 & +.87 [.57,.97] & $<$.001 \\
creative writing: upvotes & 144 & 523 &.43--.80 &.061--.174 & +.77 [.40,.95] & $<$.001 \\
math.SE: accepted answers & 80 & 992 &.41--.77 &.033--.107 & +.71 [.28,.93] &.002 \\
humor: joke upvotes & 124 & 10 &.10--.59 &.000--.600 & +.65 [.22,.87] &.007 \\
humor: \#HashtagWars picks & 106 & 8 &.56--1.18 & $-$.125--.250 & +.64 [.16,.91] &.007 \\
journalism: editorial pickup & 155 & 8 &.25--1.06 &.000--.500 & +.54 [.03,.88] &.031 \\
N\&C: rule change & 253 & 69 & $-$.01--1.00 & $-$.072--.101 & +.08 [$-$.44,.64] &.76 \\
N\&C: received response & 294 & 20 &.12--1.13 & $-$.100--.250 & $-$.26 [$-$.72,.37] &.33 \\
\midrule
peer review: accept/reject, spotlight, citations & 227--244 & pairs & & & $1$ by construction & \\
\bottomrule
\end{tabular}}
\caption{Within-task dose-response between the induced gap and selection regret. Sixteen points per task: random $k$-criterion subsets of the articulated bank ($k \in \{4, 8, 16, 32, 64\}$, three draws) plus the full bank. Columns: bank size; decidable groups (groups containing both a chosen and an unchosen item); range of the induced gap and of regret across the sixteen points; Spearman $\rho$ over the sixteen points with its 95\% bootstrap interval (3,000 resamples of the points) and its $p$-value. Peer-review tasks use pairwise readouts, where regret equals the pairwise-AUC difference and $\rho{=}1$ carries no information; they are excluded from the eight-task summary.}
\label{tab:dose}
\end{table}

\paragraph{Regret and the identity of removed criteria.} A dose-response between the induced gap and regret is expected whenever removing predictive features lowers the bank's AUC, so we test whether selection quality depends on \textit{which} criteria are present beyond the AUC they carry. On the frozen selection stack we removed $k \in \{4, 8, 16, 32\}$ criteria from each bank three ways, the $k$ most predictive (by univariate AUC), the $k$ least predictive, and $k$ at random, refit the articulated selector, and re-ran the selection test. The comparison is informative only on grouped tasks with enough decidable groups to resolve regret (creative writing, 523 groups; Math.SE accepted answers, 992; Math.SE votes, 1,140); on the pairwise peer-review tasks regret equals the difference in pairwise AUC by construction, and on tasks with fewer than 70 decidable groups the regret step is too coarse. Table~\ref{tab:removal} gives the $k{=}32$ row. Regret moves with the AUC removed whichever criteria are removed, and at matched AUC the three removals agree within the refit noise: the amount of articulated signal lost is what matters for selection, not the identity of the criteria that carried it.

\begin{table}[!htbp]
\centering\footnotesize
\resizebox{\textwidth}{!}{%
\begin{tabular}{llcccc}
\toprule
task & removed ($k{=}32$) & bank AUC & induced gap & regret & swap rate \\
\midrule
creative writing: upvotes & none (full bank) &.665 &.435 &.061 &.46 \\
 & 32 most predictive &.652 &.478 &.121 & \\
 & 32 least predictive &.662 &.446 &.096 & \\
 & 32 random (mean of 3) &.662 &.446 &.091 & \\
math.SE: accepted answers & none (full bank) &.569 &.518 &.067 &.45 \\
 & 32 most predictive &.550 &.652 &.080 & \\
 & 32 least predictive &.578 &.458 &.057 & \\
 & 32 random (mean of 3) &.576 &.473 &.057 & \\
math.SE: answer votes & none (full bank) &.612 &.270 &.041 &.37 \\
 & 32 most predictive &.593 &.398 &.056 & \\
 & 32 least predictive &.610 &.284 &.040 & \\
 & 32 random (mean of 3) &.610 &.284 &.040 & \\
\bottomrule
\end{tabular}}
\caption{Criterion removal at matched induced gap on the three tasks with enough decidable groups. Columns: the articulated bank's held-out AUC after removal; the induced gap $(\mathrm{AUC}_{VAT}-\mathrm{AUC}_{VA})/(\mathrm{AUC}_{VAT}-.5)$; selection regret (full model minus articulated model, per group); and the full bank's swap rate. Random removal is the mean of three draws; the refit noise on regret is about $.01$ (creative writing) and $.001$ (Math.SE).}
\label{tab:removal}
\end{table}

\subsection{\texorpdfstring{Generated pools on three tasks}{Generated pools on three tasks}}
\label{app:pools}

Section~\ref{sec:res-optim} reports the asymmetry of the pool experiment on three tasks; this subsection gives the remaining readouts. Each pool holds candidates generated for one context by three model families under six instruction conditions, plus the real items for that context: the stories written to a prompt, the real joke behind a title, or the answers posted to a question. Both models rank the pool; the readouts of Table~\ref{tab:pools} follow the definitions of the creative-writing run. The full model of every task is a Llama-3.1-8B reward model and Llama-3.1-8B-Instruct was one generator; on Math.SE the full model ranks its own family's answers above the others (family AUC $.62$ against $.37$--$.60$ for the other two), which alone produced an asymmetry of $-.21$. Every value in Table~\ref{tab:pools} therefore excludes that generator's candidates; the exclusion changes the creative-writing values little ($+.067$ to $+.034$ on 150 pools, $+.048$ to $+.047$ on 402).

\begin{table}[!htbp]
\centering\scriptsize
\resizebox{\textwidth}{!}{%
\begin{tabular}{lcccccccc}
\toprule
 & & & & \multicolumn{2}{c}{top pick above the floor} & & real vs.\ generated & picks a real item \\
\cmidrule(lr){5-6}
task & pools & real base rate & agree & full on artic. & artic. on full & $A'$ [95\% CI] & artic. / full & artic. / full \\
\midrule
creative writing (150 prompts) & 150 &.11 &.26 & +.210 & +.176 & $+.034~[-.025,.095]$ &.74 /.54 &.39 /.19 \\
creative writing (402 prompts) & 402 &.25 &.25 & +.264 & +.216 & $+.047~[.009,.084]$ &.71 /.53 &.74 /.55 \\
joke upvotes & 300 &.08 &.26 & +.234 & +.183 & $+.051~[.004,.096]$ &.80 /.58 &.44 /.23 \\
Math.SE answer votes & 150 &.15 &.13 & +.121 & +.133 & $-.011~[-.087,.065]$ &.46 /.48 &.26 /.27 \\
\bottomrule
\end{tabular}}
\caption{Pool readouts on three tasks, Llama-3.1-8B-Instruct candidates excluded (see text). Columns: pools; share of real items per pool; how often the two models name the same top item; the rank, above the pool's mean, that the other model's top pick reaches on each channel; the asymmetry $A'$ (full model's pick on the articulated channel minus the articulated model's pick on the full channel) with a 3,000-draw bootstrap over pools; real-versus-generated AUC of each model's score; share of pools in which each model's top pick is a real item. Creative writing is the run of Section~\ref{sec:res-optim} restated without the Llama generator; jokes use twelve generated completions of a real joke's title plus the joke; Math.SE uses twelve generated answers plus both real answers to the question.}
\label{tab:pools}
\end{table}
The two creative-writing rows are two pool-composition conditions rather than nested samples or independent replications: the 402-prompt frame takes every prompt with at least three held-out real stories, overlaps the 150-prompt frame on 16 prompts, and uses identical generation and scoring. The percentile ranks quoted in Section~\ref{sec:res-optim} pool the 852 pools of the three tasks (the 402-prompt creative-writing frame, jokes and Math.SE): the full model's top pick sits at the $.71$ percentile of the articulated model's ranking (95\% CI $[.69,.73]$) and the articulated model's top pick at the $.69$ percentile of the full model's ($[.67,.71]$), a difference of $.023$ $[.004,.042]$ (3,000-draw bootstrap over pools). Per task, the two percentiles are $.71$ and $.69$ on the 150-prompt creative-writing frame (150 pools), $.74$ and $.71$ on the 402-prompt frame (402), $.73$ and $.70$ on jokes (300), and $.59$ and $.61$ on Math.SE (150).

The lift statistic appears under other names in prior work. Where one model's top pick lands on the other model's ranking is the percentile rank of recommender-system evaluation~\citep{hu2008collaborative}, whose random-ranking value is $.5$, and the cross-evaluation of a proxy's best-of-$n$ choice under a second scorer in the reward-overoptimization literature~\citep{gao2023scaling,wen2025rethinking}. One caveat applies to the asymmetry. If the two scores were jointly Gaussian with correlation $\rho$, the expected standardized score of either model's top pick on the other's scale would be $\rho$ times the expected maximum of $n$ draws, in both directions: the shrinkage of a selected item's value is symmetric whichever score carries the extra signal or the extra noise~\citep{smith2006optimizer}. The asymmetry therefore reflects structure beyond one score nesting the other, either a top-tail interaction (items that score highest on the criteria carry less of the residual) or the coarseness of the four-level articulated score, and we have not yet separated the two explanations.

\section{Additional Collected Quotes}
\label{app:quotes}

These quotations speak to four different things, and we do not read them as cumulative evidence for one tacit component: (i) practitioners' inability to explain a judgment they make reliably; (ii) disagreement among experts; (iii) institutional bias in who gets chosen; and (iv) the failure of explicit prediction rules. Only (i) and (iv) bear directly on articulability; (ii) and (iii) bear on what the labels measure. Each quotation below is marked with the phenomenon it evidences: of the 120, 62 evidence (i), 27 (ii), 8 (iii) and 26 (iv), with 15 carrying two marks; 12 are background and are marked [--].

Verification status (\textsc{verified} / \textsc{snippet} / \textsc{lead}), retrieval
routes, and caveats for every item below are recorded in the \texttt{annote} field of
the corresponding entry in \texttt{latex/refs-shared.bib}. Page pins are reproduced
here only where the source text was actually read.

\subsection{Journalism}

\paragraph{Epstein, \emph{News from Nowhere} (1973).} [i, iii]
``[T]he producers' \textbf{unwritten but generally known preferences} for certain types of
stories are of much more importance to correspondents'' (pp.~168--169). Reasons for
sidelining correspondents ``varied, according to executive producers' offhanded
descriptions, from their age (`too old') to their general appearance (`lack of sex
appeal') to their presentation (`not punchy enough' or `no sense of humor')'' (p.~189).
``Salant said that he would refuse to answer any questions about news judgments'' (p.~10).

\paragraph{Gans, \emph{Deciding What's News} (1979), via De Maeyer (2020).} [i]
News judgment is ``\textbf{partly a matter of feel}'' (p.~171). Journalists ``act on the basis
of \textbf{quick, virtually intuitive judgments}, which some ascribe to `feel''' (p.~82).

\paragraph{Schultz, ``The Journalistic Gut Feeling'' (2007).} [i]
Journalists describe the gut feeling as ``\textbf{part of your spinal cord},'' ``\textbf{in the back of
your head},'' ``\textbf{something like a feeling}'' (p.~199).

\paragraph{Cotter, \emph{News Talk} (2010).} [i]
Journalists ``learn to report, edit, produce, and think about the news in ways that
become habitual and a reflexive part of their everyday actions'' (ch.~3). The ``nose
for news'' is ``\textbf{somewhat magical}'' (p.~77).

\paragraph{Zelizer \& Allan, \emph{Keywords in News and Journalism Studies} (2010).} [i]
The ``nose for news'' is ``\textbf{murky}'' (p.~96).

\paragraph{Ellingsberg, ``Cultural Journalism under Pressure'' (2026).} [i, iii]
Cultural journalism lacks the ``codified qualifications and transparent recruitment
procedures associated with fully professionalised occupations, with access contingent
on \textbf{tacit judgments that cannot easily be articulated or externally validated}.''

\paragraph{Breed, ``Social Control in the Newsroom'' (1955).} [i]
```Policy' may be defined as the more or less consistent orientation shown by a paper,
not only in its editorial but in its news columns and headlines as well, concerning
selected issues and events'' (p.~327). Breed's own 1993 restatement of the enforcement
mechanism, quoted in Reese \& Ballinger (2001): ``editorial blue-penciling --- teaching
reporters which objectionable phrases to omit in the future, occasional reprimands,
internal house organ papers, and rare explicit policy decisions.''

\subsection{Mathematics}

\paragraph{Poincar\'e, ``Mathematical Creation'' (1908).} [i]
``[T]he rules which must guide this choice are extremely fine and delicate. \textbf{It is almost
impossible to state them precisely; they are felt rather than formulated.}'' ``[I]t is
this special esthetic sensibility which plays the role of the delicate sieve.''

\paragraph{Thurston, ``On Proof and Progress in Mathematics'' (1994).} [i]
``[I]t is preposterous to claim that mathematics as we practice it is anywhere near
formally correct.'' ``When people are doing mathematics, the flow of ideas and the
social standard of validity is much more reliable than formal documents.''
``One-on-one, people use wide channels of communication that \textbf{go far beyond formal
mathematical language}. They use gestures, they draw pictures and diagrams, they make
sound effects and use body language.'' ``Many of the things that are generally known
are things for which \textbf{there may be no known written source}.''

\paragraph{Burton, ``Why Is Intuition So Important to Mathematicians\ldots'' (1999).} [ii]
``\textbf{There was no agreement about whether these terms represented different, distinctive
states} and, if so, how these states might be recognised'' (p.~28). ``The apparent need
for mathematics to be infallible and objective requires the removal of the evidence of
the very human beings who created and worked with the mathematical objects, missing
entirely the stumbling human process that created those results'' (p.~29).

\paragraph{Geist, L\"owe \& Van Kerkhove, ``Peer Review and Knowledge by Testimony in
Mathematics'' (2010).} [iii]
An editor: ``There are situations where almost nothing needs be checked (e.g., the
author is well-known and it is his concern to submit a correct paper). There are
situations when I insist to check all the procedures.'' A mathematician on trusting an
unverified step: ``you end up with the question `is this because I don't have enough
knowledge?' \ldots he's publishing it, not I, so it's his responsibility.''

\paragraph{Wells, ``Are These the Most Beautiful?'' (1990).} [ii]
``[T]he idea that mathematicians largely agree in their aesthetic judgements is at best
grossly oversimplified.'' Tony Gardiner: ``Theorems aren't usually `beautiful'. It's the
ideas and proofs that appeal.'' Tito Tonietti: ``Beauty, even in mathematics, depends
upon historical and cultural contexts, and therefore \textbf{tends to elude numerical
interpretation}.''

\paragraph{Sa, Alcock, Inglis \& Tanswell, ``Do Mathematicians Agree about Mathematical
Beauty?'' (2024).} [ii, iv]
``[M]athematicians might have different absolute standards for beauty and thus appear to
disagree when asked for absolute judgements, but might nevertheless agree about which
objects are more or less beautiful.'' On their own operationalization: ``Number of
characters in an equation is clearly a crude measure of simplicity.''

\paragraph{Rota, ``The Phenomenology of Mathematical Beauty'' (1997).} [i]
``[M]athematicians instead like to engage in discussions of the beauty of mathematics
\ldots \textbf{it is however unthinkable to find any `mathematical beauty appreciation' courses
taught anywhere}.'' ``A theorem that is in one context thought to be beautiful may in a
different context appear trivial.''

\subsection{Coding}

\paragraph{Naur, ``Programming as Theory Building'' (1985).} [i]
``The dependence of a theory on a grasp of certain kinds of similarity between
situations and events of the real world gives the reason why the knowledge held by
someone who has the theory \textbf{could not, in principle, be expressed in terms of rules}.''
``[T]he similarities in question \textbf{are not, and cannot be, expressed in terms of criteria},
no more than the similarities of many other kinds of objects, such as human faces,
tunes, or tastes of wine, can be thus expressed.'' The original team could ``spot these
cases instantly''; a ``highly motivated'' receiving team with full documentation could not.

\paragraph{Fowler \& Beck, ``Bad Smells in Code,'' \emph{Refactoring} (1999).} [i, iv]
``One thing we won't try to do here is give you precise criteria for when a refactoring
is overdue. In our experience \textbf{no set of metrics rivals informed human intuition}.\ldots
\textbf{You will have to develop your own sense} of how many instance variables are too many
instance variables'' (p.~75).

\paragraph{Buse \& Weimer, ``Learning a Metric for Code Readability'' (2010).} [i, ii]
``A consensus exists that readability is an essential determining characteristic of code
quality [\ldots], \textbf{but not about which factors contribute to human notions of software
readability the most}.'' Readability ``\textbf{was intentionally left formally undefined} in order
to capture the unguided and intuitive notions of participants.'' ``[H]umans agree
significantly on what readable code looks like, but not to an overwhelming extent.''

\paragraph{Soloway \& Ehrlich, ``Empirical Studies of Programming Knowledge'' (1984).} [--]
``Rules of Programming Discourse: Rules that specify the conventions in programming,
e.g., the name of a variable should usually agree with its function; these rules set up
expectations in the minds of the programmers about what should be in the program\ldots
These rules are analogous to discourse rules in conversation'' (p.~595).

\paragraph{LaToza \& Myers, ``Hard-to-Answer Questions about Code'' (2010).} [i]
``Rationale (42): Why wasn't it done this other way? (15).'' ``The most frequently
reported categories dealt with intent and rationale --- what does this code do, what is
it intended to do, and why was it done this way?''

\paragraph{LaToza, Venolia \& DeLine, ``Maintaining Mental Models'' (2006).} [i]
``\textbf{Personal code ownership is usually tacit}, i.e.\ part of the mental model. Written
records of ownership, when present, are often out-of-date and distrusted.'' ``This
information is precious: it is demonstrably useful, demonstrably hard to ascertain from
the code, and was obtained at a high cost. Yet \textbf{it is exceedingly rare for this developer
to then write this information down}.''

\paragraph{Sadowski et al., ``Modern Code Review: A Case Study at Google'' (2018).} [--]
``[O]nce those reviewers are confident the developer understands the code style and best
practices for a language, the developer is granted readability for that language.''
Review ``was introduced at Google to ensure code readability and maintainability\ldots
Defect finding is welcomed but not the only focus.''

\paragraph{Bosu, Greiler \& Bird, ``Characteristics of Useful Code Reviews'' (2015).} [ii]
``[M]ost of the review comments are unrelated to any types of functional defects.''
``Some of the interviewees rated nit-picking issues as `Somewhat useful', while others
rated those as `Useful.'\,''

\paragraph{Rigby \& Bird, ``Convergent Contemporary Software Peer Review Practices'' (2013).} [--]
``Inspections also found a large number of soft maintenance issues, median 13 per review,
which included coding conventions, and the addition of comments\ldots An additional 4
defects per review were found to be false positives,'' against ``a median of 3 true
defects found per review.''

\paragraph{Stegeman, Barendsen \& Smetsers, ``Designing a Rubric for Feedback on Code
Quality'' (2016).} [ii]
``Several grading schemes have been published [\ldots], but while all of these focus on
similar aspects of code quality, such as readability, style and decomposition, they are
very diverse in form as well as in content.''

\subsection{Academic Articles and Peer Review}

\paragraph{Kang et al., ``PeerRead'' (2018).} [iv]
``The aspects which correlate most strongly with the final recommendation are substance
\ldots and clarity. In contrast, soundness/correctness and originality are least
correlated with the final recommendation.'' Correlations of each aspect score with the
reviewer's own overall recommendation (ACL 2017): substance $.59$, clarity $.42$,
appropriateness $.30$, impact $.16$, meaningful comparison $.15$, originality $.08$,
soundness/correctness $.01$.

\paragraph{Greiffenhagen, ``Checking Correctness in Mathematical Peer Review'' (2024).} [i]
``\textbf{It was more challenging for my interviewees to express what they expected from
referees} with respect to checking correctness.'' ``[I]t's not a question of formally
checking line by line so much as understanding where the problems are likely to
appear.'' ``Even if the paper is 70 pages, I open to page 37, because I know that this
is where the action has to be.''

\paragraph{Hirschauer, ``How Editors Decide'' (2015); model originating in Hirschauer (2010).} [i]
``The operative nucleus of peer review processes has largely remained a `black box' to
analytical empirical research.'' ``(1) There is a pre-formed judgment --- the preset
attitude that readers have toward all texts which `exists' before reading\ldots (2) In
the second phase, a judgment is a spontaneous impression developed over the course of
reading a manuscript and negotiating with it. It is about what you observed in and
understood of a text. (3) The third phase is the written judgment or statement that is
\textbf{enunciated in the post hoc reasoning about the text}. It is produced in the public sphere
of boards, and \textbf{it is about the position that one can defend in an argument}.'' ``So there
are always tendentious, spontaneous, and rationalizing aspects in an individual statement
to publish or not. This is known to all participants in peer review.'' On pool-relative
standards: ``they have to be flexibly applied, since there is no vacancy option. Quality
claims cannot simply be upheld against the offer on hand, they have to be displayed with it.''

\paragraph{Mallard, Lamont \& Guetzkow, ``Fairness as Appropriateness'' (2009).} [iv]
``Peer reviewers define a fair-decision making process as one in which panelists engage
in `cognitive contextualization,' that is, use epistemological styles most appropriate
to the field or discipline of the proposal under review. These findings challenge the
normative literature that associates procedural fairness with the use of generalizable
criteria of evaluation.'' ``Rather than applying a single universal criterion
indiscriminately, they specify which criteria, or lenses, are most appropriate to assess
the strengths and weaknesses of the object under evaluation\ldots The fairest criteria
are thus not the most widely shared criteria, but the standards deemed most appropriate
given the distinctive features of a particular object.''

\paragraph{Hug, ``How Do Referees Integrate Evaluation Criteria into Their Overall
Judgment?'' (2024).} [iv]
``[R]eferees use many criteria and integrate the criteria using complex rules. However,
and most importantly, the revised style \textbf{could describe most --- but not all --- of the
referees' judgments}. Future studies should therefore examine how referees' judgments can
be characterized in those cases where the uniform style failed.''

\paragraph{Bloxham, Boyd \& Orr, ``Mark My Words'' (2011).} [i]
``Markers often rely on \textbf{holistic judgments informed by tacit knowledge} and norm
referencing, with \textbf{explicit criteria primarily used post hoc to justify grades}.''

\paragraph{O'Donovan, Price \& Rust, ``Know What I Mean?'' (2004).} [i]
``[E]xplicit criteria are only a first stage in a continuum of processes to help students
acquire knowledge of assessment expectations''; assessment is ``a `social and cultural
practice'\,'' in which ``\textbf{teachers acquire tacit knowledge of standards and judgement
through participation, observation, imitation and dialogue}.''

\subsection{Grant Panels}

\paragraph{Guetzkow, Lamont \& Mallard, ``What Is Originality in the Humanities and the
Social Sciences?'' (2004).} [i]
``[T]he guidelines are very general, and \textbf{panelists are given no indication of the
specific meaning of the suggested criteria} (such as feasibility or significance) or the
weight to be given to each one'' (p.~194). ``Only one of our fellowship programs (WWNFF)
specifically mentioned the `originality' of the proposal in their guidelines; yet it was
of major concern to almost all the reviewers interviewed\ldots few of the reviewers
expressed concern for or even knowledge of the institutional guidelines'' (p.~195).
What originality decodes to: ``adventurous, ambitious, bold, courageous, independent,
intellectually honest, curious, and risk-taking\ldots this lexicon describes qualities
that indicate whether or not one possesses intellectual authenticity'' (p.~203). A
historian panelist: ``\textbf{it's a nose for a real passion for ideas}'' (pp.~204--205).

\paragraph{Lamont \& Huutoniemi, ``Comparing Customary Rules of Fairness'' (2011).} [i]
``[Evaluation] is also a process where standards of excellence are set and maintained,
contested, and reshaped'' (p.~209). Practices ``grounded in connoisseurship, expertise,
and knowledge'' (p.~211). ``These are intersubjective rules that guide panel
deliberations \textbf{without being formally spelled out. Panelists cannot always articulate
these rules}, as they often take them for granted'' (p.~213). A Finnish panelist, on
reaching consensus by ``relying on panelists' integrity or intuition'': ``You could put
your hands on your heart and then say to each other, `Do you really, honestly, think that
it is a ``good'' proposal, or an ``excellent'' proposal? What do you think, really?'\,''
(pp.~219--220). An American panelist: ``I see scholarly excellence and excitement in this
one project on food, possibly because I see resonance with my own life, my own interests,
who I am, and other people clearly don't\ldots excellence is in some ways what looks most
like you'' (p.~225). The authors' gloss: ``\textbf{panelists cannot spell out what defines an
`interesting' proposal in the abstract}'' (p.~226).

\paragraph{Lamont, \emph{How Professors Think} (2009).} [i]
Panelists bring ``a host of judgments, expectations, and anecdotes that are \textbf{not to be
found in the folders they read}, but that strongly influence their judgment'' (p.~227).

\paragraph{Brunet \& M\"uller, ``Making the Cut'' (2022).} [i, ii]
Reviewers lacking confidence to evaluate across fields ``attempted to get a `\textbf{feel for
the sample}' and `calibrate' themselves accordingly'' (p.~492). ``[P]anel reviewers did
not assess any specific aspect of each proposal but rather looked to establish a
\textbf{preliminary, almost embodied, impression of the whole set}'' (p.~493). ``\textbf{There was no
consensus between panel reviewers} about how strong or loose the articulation between
different elements should be; preferences were largely individual'' (p.~494). Reviewers
``aim to develop an almost embodied skill to classify proposals quickly in relation to
each other'' (p.~497).

\paragraph{Roumbanis, ``Peer Review or Lottery?'' (2019).} [iv]
``The ex ante evaluation of scientific novelty is a fundamentally uncertain endeavor;
bias and chance are embedded in the final outcome.''

\subsection{Law, Policy and Regulation}

\paragraph{Hutcheson, ``The Judgment Intuitive'' (1929).} [i]
Posed as a rhetorical question, pp.~284--285: ``Now what is he saying except that the
judge really decides by feeling, and not by judgment; by `hunching' and not by
ratiocination, and that \textbf{the ratiocination appears only in the opinion}?'' ``The vital,
motivating impulse for the decision is \textbf{an intuitive sense of what is right or wrong} in
the particular case'' (p.~285). ``I, after canvassing all the available material\ldots
\textbf{wait for the feeling, the hunch}'' (p.~278). ``I speak of the judgment pronounced, as
opposed to the rationalization by the judge on that pronouncement'' (p.~279).

\paragraph{Cardozo, \emph{The Nature of the Judicial Process} (1921).} [i, ii]
``More subtle are the forces \textbf{so far beneath the surface that they cannot reasonably be
classified as other than subconscious}. It is often through these subconscious forces
that judges are kept consistent with themselves, and inconsistent with one another.''
``\textbf{Deep below consciousness are other forces}, the likes and the dislikes, the
predilections and the prejudices, the complex of instincts and emotions and habits and
convictions, which make the man, whether he be litigant or judge.''

\paragraph{\emph{Jacobellis v.\ Ohio}, 378 U.S.\ 184, 197 (1964) (Stewart, J., concurring).} [i]
``I shall not today attempt further to define the kinds of material I understand to be
embraced within that shorthand description; and \textbf{perhaps I could never succeed in
intelligibly doing so}. But I know it when I see it, and the motion picture involved in
this case is not that.''

\paragraph{Neumann, ``Donald Sch\"on, \emph{The Reflective Practitioner}, and the
Comparative Failures of Legal Education'' (2000).} [i]
``\,`Thinking like a lawyer' is a label used by doctrinal teachers for a collection of
textual interpretation skills and heightened forms of skepticism.\ldots In fact, so much
energy has been devoted to textual interpretation and skepticism that we actually know
very little about how effective lawyers go about solving problems'' (p.~405). Quoting
Sch\"on: ``Tacit knowing or knowing-in-action has this property: we exhibit it by the
competent behavior we carry out but \textbf{we are unable to describe what it is that we do}''
(p.~407). ``[O]utstanding practitioners are not said to have more professional knowledge
than others, but more `wisdom,' `talent,' `intuition,' or `artistry' --- all terms that
`\textbf{serve not to open up inquiry but to close it off}'\,'' (p.~407). Students ``literally did
not know what they were doing, and \textbf{their teachers could not tell them} --- because what
the teachers knew how to say the students could not at that point in their experience
understand'' (p.~408).

\paragraph{Guthrie, Rachlinski \& Wistrich, ``Blinking on the Bench'' (2007).} [i]
``These results suggest that \textbf{judges rely heavily on their intuitive faculties} not only
when they confront generic problems like the problems included in the CRT, but also when
they face the kinds of problems they generally see on the bench\ldots judges seem
inclined to make intuitive judgments'' (p.~27). ``[J]udges are inclined, at least when
presented with certain stimuli, to make intuitive decisions, but that they have the
capacity to override intuition with deliberative thinking'' (pp.~27--28). Subjects were
295 sitting circuit judges (p.~13).

\paragraph{Farina, Newhart \& Heidt, ``Rulemaking vs.\ Democracy'' (2012); same
definition reused in CeRI, \emph{Rulemaking 2.0} (2013).} [--]
``[S]uch historically `undervoiced' stakeholders can bring a particular kind of
knowledge --- `situated knowledge' --- that the agency itself may not possess, and that
organizations purporting to represent these stakeholders may not reveal in sufficient
detail or persuasiveness. By situated knowledge, we mean information about impacts,
problems, enforceability, contributory causes, unintended consequences, etc.\ that is
known by the commenter because of lived experience in the complex reality into which the
proposed regulation would be introduced'' (p.~148).

\paragraph{Balla et al., ``Lost in the Flood?'' (2022).} [iv]
Mass comments ``generally contain little `relevant matter'\,'' --- an APA phrase never
operationalized into a checkable rubric.

\subsection{Patents}

\paragraph{\emph{Harries v.\ Air King Products Co.}, 183 F.2d 158, 162 (2d Cir.\ 1950)
(L.~Hand, C.J.).} [i]
On the question of patentable invention: ``That issue is \textbf{as fugitive, impalpable,
wayward, and vague a phantom as exists in the whole paraphernalia of legal concepts}.''

\paragraph{\emph{KSR International Co.\ v.\ Teleflex Inc.}, 550 U.S.\ 398 (2007).} [iv]
Replacing the Federal Circuit's ``rigid'' teaching-suggestion-motivation test with an
``expansive and flexible approach'' (at 415). ``Rigid preventative rules that \textbf{deny
factfinders recourse to common sense}\ldots are neither necessary\ldots nor consistent
with'' precedent (at 421).

\paragraph{Mandel, ``The Non-Obvious Problem'' (2008).} [i]
The nonobviousness standard is ``not too low, but both too high and too low. \textbf{It is
indeterminate},'' owing in part to ``\textbf{the epistemic infeasibility} of requiring a
technologically lay decision maker to judge from the perspective of a more highly
trained and educated person of ordinary skill in the art.''

\paragraph{Mandel, ``Patently Non-Obvious II'' (2007).} [iv]
Mock jurors judged the identical invention ``obvious'' at 42\% under foresight framing
versus 71\% under hindsight framing (49\% vs.\ 85\% with a suggestion to combine).
Instructing participants on the TSM test produced no significant correction.

\paragraph{Cockburn, Kortum \& Stern, ``Are All Patent Examiners Equal?'' (2002).} [ii]
``[T]here may be as many patent offices as there are patent examiners.'' Across 298,441
patents, examiner fixed effects explain $.117$ of variance in citations received, $.083$
in approval time, $.079$ generality, $.069$ originality, net of technology sub-class and
filing cohort.

\paragraph{Lemley \& Sampat, ``Examiner Characteristics and Patent Office Outcomes'' (2012).} [ii]
``[T]he most important decisions made by the patent office are significantly affected by
the happenstance of which examiner gets an application.'' Of US applications the EPO
rejected, 60\% (525/869) were still patented by the USPTO.

\paragraph{Frakes \& Wasserman, ``Is the Time Allocated to Review Patent Applications
Inducing Examiners to Grant Invalid Patents?'' (2014).} [iv]
With the same examiner and the same written standard, halving time per application
raises grant rates ``by as much as 9 to 19 percentage points, or by roughly 13 to 28
percent.''

\paragraph{Lichtman (2004), reported in Mann, ``The Idiosyncrasy of Patent
Examiners'' (2014).} [ii]
Examiner identity accounts for ``about two-thirds of the variation in rigor of editing''
of patent claim text between filing and grant.

\subsection{Creative Writing and Publishing}

\paragraph{Amabile \& Pillemer, ``Perspectives on the Social Psychology of
Creativity'' (2012).} [--]
``The CAT is rooted in a consensual operational definition of creativity: a product is
creative to the extent that expert raters independently agree upon this judgment\ldots
Because it relies on subjective assessment (as does real-world creativity), the CAT
\textbf{overcomes the difficulty of defining ultimate `objective' criteria for creativity}'' (p.~8).

\paragraph{Baer, Kaufman \& Gentile (2004).} [--]
Found ``an extremely high level of interrater agreement on the creativity of a wide range
of writing products (teenagers' essays, poems, and stories).''

\paragraph{Ross, the \emph{Steps} experiment (1975).} [iii]
Houghton Mifflin, rejecting Kosinski's National Book Award novel retyped under a
pseudonym: ``Several of us read your untitled novel here with admiration for writing and
style\ldots The drawback\ldots is that \textbf{it doesn't add up to a satisfactory whole}.'' Ross:
``Evidently, Kosinski is not as good as Kosinski when Demos is the name on the envelope.''

\paragraph{Lessing, the ``Jane Somers'' pseudonym experiment (1983--84).} [iii]
``I wanted to highlight that whole dreadful process in book publishing that `nothing
succeeds like success.'\,'' ``Some of the so-called experts on my work (\ldots) didn't
recognise it was me.''

\paragraph{Lassman, ``Rejecting Jane'' (2007).} [iii]
Alex Bowler of Jonathan Cape, the one reader of eighteen who recognized the Austen
openings: ``my first impressions on reading these were ones of disbelief and mild
annoyance.'' The rest rejected with form language (``not right for our list'').

\paragraph{Makkai, ``Book Prizes Don't Work How You Think.''} [ii]
``Judges usually communicate\ldots and \textbf{often at some point define their own criteria}.
Are they simply looking for the best-executed book\ldots Or the most innovative?\ldots
The book that everyone should still be reading in fifty years? The book that affected
them the most?'' ``Who can sway other judges? Who freezes and backs down?'' ``[W]hether
a book makes it to the next round can depend entirely on the tastes of one person\ldots
A person who\ldots might not have had the luxury of putting a book down on a bad day and
picking it up again in the right mood.''

\paragraph{Lish, ``The Art of Editing No.~2,'' \emph{The Paris Review}.} [i]
Asked how he can tell what is good: ``Because I've got the fucking gift for it. \textbf{Instinct,
call it.}''

\paragraph{Gottlieb, ``The Art of Editing No.~1,'' \emph{The Paris Review}.} [i]
``Editing is simply the application of \textbf{the common sense of any good reader}.''

\paragraph{Zou \& Xu, ``Creative Quality Alignment'' (2026).} [iv] (arXiv:2605.25977, preprint.) Their mechanism is the complement of ours: they transfer expert judgment by fine-tuning on chains of thought written by experts, i.e. by moving preference into a model without first stating it as criteria, and they argue that rubric evaluation fixes its dimension set at design time while creative work can show value on dimensions the evaluator never anticipated. Our missing-mass bound (Section~\ref{sct:robustness}) puts a number on how much such unanticipated dimensions can be worth for a given bank.
``Rubric evaluation fixes its dimension set at design time, while high-quality creative
work can demonstrate value on \textbf{dimensions the evaluator never anticipated}.'' ``In
subjective domains like creative judgment, quality does not come from sub-item scores
after decomposition. \textbf{It comes from holistic judgment under full-dimensional constraints}.''

\subsection{Humor}
\label{app:quotes:humor}

\paragraph{E.~B.~White, preface to \emph{A Subtreasury of American Humor} (1941; reprinted as ``Some Remarks on Humor'').} [i]
``Humor can be dissected, as a frog can, but \textbf{the thing dies in the process} and the
innards are discouraging to any but the purely scientific mind.''

\paragraph{Amir, ``The Frog Test'' (2016).} [ii]
``Proponents of the different theoretical accounts often show a high degree of
conviction, suggesting \textbf{introspection might not be the best tool} for judging the validity
of humor theories.''

\paragraph{Hessel et al., ``Do Androids Laugh at Electric Sheep?'' (2023).} [--]
``[O]ur best multimodal models fall 30 accuracy points behind human performance on the
matching task, and, even when provided ground-truth visual scene descriptors,
human-authored explanations are preferred head-to-head over the best machine-authored
ones (few-shot GPT-4) in more than 2/3 of cases.''

\subsection{Art and Connoisseurship}

\paragraph{Boden, \emph{Creativity and Art} (2012).} [i]
``Our aesthetic values are difficult to recognize, \textbf{more difficult to put into words, and
even more difficult to state really clearly}.''

\paragraph{Ginzburg, ``Clues: Roots of an Evidential Paradigm'' (1980/1989).} [i]
``These are \textbf{essentially mute forms of knowledge in the sense that their precepts do not
lend themselves to being either formalized or spoken}. No one learns to be a connoisseur
or diagnostician by restricting himself to practicing only preexistent rules''
(pp.~124--125).

\paragraph{Friedl\"ander, \emph{On Art and Connoisseurship} (1942).} [i]
``[D]ecision ultimately \textbf{rests with something which cannot be discussed}\ldots when we come
upon the concepts of intuition and self-evidence\ldots \textbf{we resign as scholars and even as
writers}'' (p.~172). ``The way in which an intuitive verdict is reached can, from the
nature of things, \textbf{only be described inadequately}'' (pp.~172--173). ``\textbf{You cannot explain a
witticism without murdering it}. And the position is the same with regard to the work of
art'' (p.~183).

\paragraph{Morelli, \emph{Italian Painters} (1892).} [--]
Attribution by minutiae presented as an explicit ``experimental method'' (p.~44) meant to
``control the judgment which subjective impressions might lead us to pronounce'' (p.~32).

\paragraph{Berenson, ``Rudiments of Connoisseurship'' (1902).} [ii]
Quoting Morelli's dictum, ``The connoisseur should above all things have no bump of
philosophy,'' and faulting Morelli's ``empiricism'' for ``deliberately refus[ing] to use
his powers of reasoning.'' Of his own project: ``What he would not attempt, I have tried
to do'' (pp.~vii--viii).

\paragraph{\emph{The Getty Kouros Colloquium} (1993).} [i, ii]
Vassilis Lambrinoudakis: ``I admit that, although I have a very unpleasant feeling about
this kouros, \textbf{I am unable to discover this kind of evidence in its partial forms}''
(pp.~40--41). Angelos Delivorrias, on ``\textbf{intuitive repulsion}'' (p.~45). Evelyn B.~Harrison:
the kouros ``did not give the impression of being something old; \textbf{somehow it looked new}''
(p.~22).

\paragraph{Gordenker et al., ``Rembrandt's \emph{Saul and David} at the Mauritshuis'' (2013).} [iv]
``The optimistic predictions that scientific investigations would lead to `conclusive
results' about the authorship of the many paintings attributed to Rembrandt have,
unfortunately, not been fulfilled, but such research certainly has had an impact.''

\paragraph{Bredius, ``A New Vermeer,'' \emph{The Burlington Magazine} (1937).} [i]
Certifying van Meegeren's forgery in print as ``the masterpiece of Johannes Vermeer of
Delft''; an earlier forgery as ``authentic,'' ``very beautiful,'' ``one of the finest gems
of the master's oeuvre'' (1932).

\paragraph{Eisner, ``Educational Connoisseurship and Criticism'' (1985);
\emph{The Enlightened Eye} (1998).} [--]
``\textbf{Connoisseurship is private, but criticism is public}.'' ``Connoisseurship is the art of
appreciation. It can be displayed in any realm in which the character, import, or value
of objects, situations, and performances is distributed and variable, including
educational practice'' (1998, p.~63). ``The word connoisseurship comes from the Latin
cognoscere, to know\ldots It involves the ability to see, not merely to look'' (1998, p.~6).

\subsection{Medicine}

\paragraph{Hodges et al., ``OSCE checklists do not capture increasing levels of
expertise'' (1999).} [iv]
``[O]n global scales, the experienced clinicians scored significantly better than did the
residents and clerks, but \textbf{on checklists, the experienced clinicians scored significantly
worse} than did the residents and clerks.'' ``[B]inary checklists \textbf{may not be valid measures}
of increasing clinical competence.''

\paragraph{Regehr, MacRae, Reznick \& Szalay (1998).} [iv]
``Global rating scales scored by experts showed higher inter-station reliability, better
construct validity, and better concurrent validity than did checklists.''

\paragraph{Martin et al., OSATS (1997).} [iv]
``Global ratings discriminated between resident levels'' more effectively than checklists.

\paragraph{Hodges et al., ``The challenge of creating new OSCE measures'' (2002).} [iv]
New measures ``sensitive to the nature of expertise'' are needed; coding every utterance
in the videotaped interviews found only subtle differences between experience levels.

\paragraph{ten Cate, ``Trust, competence, and the supervisor's role'' (2006).} [i, iv]
``The decision to trust a trainee to manage a critically ill patient is based on much
more than tests of competence.'' ``\textbf{Supervisors often know who to pick, even if they can't
tell exactly why}. This gut feeling does not always match with formally assessed knowledge
or skill, but it may be more valid for its purpose.'' ``\textbf{No external body or procedure can
replace this type of expert judgment}.'' ``Attempting to assess them separately may result
in a trivialised set of attained abilities\ldots To further develop educational technology
and sophistication of assessment methods does not seem the right direction.''

\paragraph{ten Cate, ``Nuts and bolts of entrustable professional activities'' (2013).} [iv]
``The key question is: Can we trust this trainee to execute this EPA?'' Competency
frameworks ``would otherwise be too theoretical to be useful for training and assessment
in daily practice.''

\paragraph{Duijn et al., ``When to trust our learners?'' (2018).} [i]
Assessors ``frequently compare learner performance with what they would do (self as
standard)\ldots \textbf{some rely on a gut feeling}.''

\paragraph{Govaerts \& van der Vleuten, ``Validity in work-based assessment'' (2013).} [i]
``[L]earning, competence (as inferred from performance) as well as performance
interpretations are to be seen as inherently contextualised, and \textbf{can only be understood}
`in situ.'\,''

\paragraph{Hodges, ``Assessment in the post-psychometric era'' (2013).} [iv]
Assessment has been ``dominated by a discourse of psychometrics'' since the 1970s; the
field must move toward ``subjective judgment and collective evaluation methods.''

\paragraph{Benner, Tanner \& Chesla, \emph{Expertise in Nursing Practice} (2009).} [i]
Intuition as ``\textbf{understanding without rationale}.'' ``Detached analytic reasoning is needed
in cases of breakdown, where direct apprehension [intuition] does not occur'' (p.~387).
Expert recognition is ``recognizing the unexpected --- that is, \textbf{when tacit global
expectations of patient's recovery are not met}'' (p.~xvii).

\paragraph{Montgomery, \emph{How Doctors Think} (2006), via reviews.} [iv]
``[A]ttempts to mimic the judgments of experienced clinicians \textbf{algorithmically mostly
fail}''; via Montgomery, Feigenbaum's diagnosis that expert-system formalization of
medical knowledge ``\textbf{threatens to break down into ten thousand special cases}.''

\paragraph{Klein, \emph{Sources of Power} (1998).} [i]
A fireground commander asked how he made high-stakes decisions: ``\,`I don't make
decisions,' he announced\ldots `\textbf{I don't remember when I've ever made a decision}'\,''
(p.~11). ``[T]here were options, yet it was usually obvious what to do in any given
situation'' (p.~12).

\subsection{Design Studios and Teaching Observation}

\paragraph{Sch\"on, \emph{Educating the Reflective Practitioner} (1987).} [i]
``[T]he student seeks to learn things whose meaning and importance she cannot grasp ahead
of time\ldots Like Meno, the design student knows she needs to look for something but does
not know what that something is\ldots Yet, at the beginning, \textbf{she can neither do it nor
recognize it when she sees it} [\ldots] The instructor is caught up in the same paradox: \textbf{he
cannot tell his student what she needs to know, even if he has words for it}, because the
student would not at that point understand him'' (p.~83). The coach ``could produce good,
clear, and compelling descriptions of designing, students [\ldots] would be likely to find
them confusing and mysterious'' (p.~100). Petra ``\textbf{does not and cannot understand what
designing means}'' (p.~82).

\paragraph{Sch\"on, \emph{The Reflective Practitioner} (1983).} [i]
Once Quist and Petra ``become more confident that they have achieved congruence of
meaning,'' ``their dialogue tends to become \textbf{elliptical and inscrutable to outsiders}'' (p.~81).

\paragraph{Sch\"on, ``The Architectural Studio as an Exemplar of Education for
Reflection-in-Action'' (1984).} [i]
``People have in their doing a tacit kind of knowing. \textbf{They know more than they can say}\ldots
\textbf{Their knowing-in-action is incongruent with their descriptions of it}.'' Quist ``has not
applied to the particular situation before him the general rules drawn from his past
experience.'' ``Even here, I think, \textbf{the process is still largely implicit}. Architects
appear to reflect very little on their own practice of reflection-in-action.'' The framing
question: ``how do students learn, with Quist's help, what Quist already knows how to do?''
On the struggling student Lauda, from his instructor: ``\textbf{he has not internalized some of the
covert things}''; from Lauda: ``I want to go out and learn first. I want to know what it is
we are arguing about.'' A student on the crit: ``You don't know where you are and have no
basis for evaluation. \textbf{You hang onto the inflection of the tone of voice in your crit} to
discover if something is really wrong.''

\paragraph{Cross, ``Designerly ways of knowing'' (1982), collected in
\emph{Designerly Ways of Knowing} (2007).} [i]
``What designers know about their own problem-solving processes \textbf{remains largely tacit
knowledge} --- i.e.\ they know it in the same way that a skilled person `knows' how to
perform that skill. \textbf{They find it difficult to externalise their knowledge}\ldots It may be
satisfactory, or at least understandable, for practising designers \textbf{to be inarticulate
about their skills}, but teachers of design have a responsibility to be as articulate as
they possibly can about what it is they are trying to teach.''

\paragraph{Cross, ``Expertise in Design: An Overview'' (2004).} [i]
``\textbf{They found it difficult to account for this apparently unprincipled design behaviour}.''
``Expert designers appear to be `ill-behaved' problem solvers\ldots'' ``Studies of design
activity have frequently found `intuitive' features of design behaviour to be the most
effective and relevant to the intrinsic nature of design.''

\paragraph{Lawson, \emph{How Designers Think} (4th ed., 2006).} [i]
``In all this confusion, however, designers cannot usually procrastinate\ldots They simply
must get on and make the decision in as integrated and sensible a way as they can''
(p.~81). ``[I]t seems unreasonable for designers to expect to find a process which will
protect them from the painful and difficult business of exercising subjective judgement in
situations where both quantitative and qualitative factors must be taken into account''
(p.~81).

\paragraph{Anthony, \emph{Design Juries on Trial} (1991), quoted verbatim in
Salama \& El-Attar (2010).} [i]
``Evaluation criteria were often \textbf{defensible only on the grounds of `Good Taste and
Intuition'}\,'' (p.~2).

\paragraph{Webster, ``Power, Freedom and Resistance: Excavating the Design Jury'' (2006).} [i]
``[A]rchitectural educators and students appear to have a \textbf{largely `tacit', or `folklore',
understanding} of its pedagogic purpose and processes that often seems riddled with
inconsistencies and contradictions'' (p.~1).

\paragraph{Salama \& El-Attar, ``Student Perceptions of the Architectural Design
Jury'' (2010).} [i, ii]
``[F]aculty critique each project spontaneously \textbf{without criteria made clear to the
students}'' (p.~3). Of 209 architecture students, 75\% said design priorities emphasized
during studio instruction changed unpredictably at the jury itself (p.~28); 16\% perceived
harmony among jurors, against 33\% reporting ``contradiction among all members of the
jury'' and 55\% a ``competitive scene'' (p.~29).

\paragraph{Kane \& Staiger, \emph{Gathering Feedback for Teaching} (2012).} [ii]
``The set of competencies measured by the two instruments --- even if they appear distinct
--- are very highly correlated'' (p.~8): FFT/CLASS $r=.88$, MQI/UTOP $r=.85$. ``[T]he
instruments all \textbf{require rater judgment, which is rarely unanimous}'' (p.~8).

\paragraph{Ho \& Kane, \emph{The Reliability of Classroom Observations by School
Personnel} (2013).} [ii]
``A single observation by a single observer is a fairly unreliable estimate of a teacher's
practice, with reliability between.27 and.45'' (p.~13). ``[T]he vast majority of scores
were in the middle two categories'' of a 4-point scale, so ``.1 point difference in scores
can be sufficient to move a teacher up or down 10 points in percentile rank'' (p.~3). On
same-rater carryover across sequential videos: ``that impression tended to linger'' (p.~3).

\paragraph{Kelly, Bringe, Aucejo \& Fruehwirth, ``Using Global Observation Protocols\ldots''
(2020).} [ii, iv]
``[A] tendency for overall perceptions to create a halo-effect, \textbf{create artificial
consistency in sub-domain scores}'' (p.~14). Citing Liu et al.\ (2019): ``found high
correlations across the four FFT domains and eight sub-domains such that a single factor
structure best fit the data.'' Their own factor analysis returns a single eigenvalue $>1$
for FFT (8 sub-domains, one-factor $R^2=.80$), PLATO ($R^2=.575$) and holistic MQI
($R^2=.84$--$.90$); FFT exact agreement runs 47.3--65.8\% with simple kappa $.05$--$.28$
(p.~14).

\paragraph{Jacob \& Lefgren, ``Can Principals Identify Effective Teachers?'' (2008).} [--]
``[P]rincipals appear quite good at identifying those teachers who produce the largest and
smallest standardized achievement gains in their schools (i.e., the top and bottom 10--20
percent), but have far less ability to distinguish between teachers in the middle of this
distribution (i.e., the middle 60--80 percent).''

\subsection{Judged Competitions}

\paragraph{Tsay, ``Sight over Sound in the Judgment of Music Performance'' (2013).} [i]
Of 103 professional musicians, 82.3\% cite sound as the important cue; their sound-only
accuracy in picking actual competition winners is 25.7\% (below the 33\% chance rate),
against 47.0\% from silent video alone.

\paragraph{Vogt, Gutzmann \& Kopiez, ``When Eyes Outvote Ears'' (2026).} [i]
Independent replication with violin-competition triads ($N=104$): video-only 50.5\%
correct against 33.3\% chance; audio-only 35.1\% and audiovisual 38.0\%, both
non-significant. Musical sophistication predicted audio-only accuracy but not video-based
accuracy.

\paragraph{Hodgson, ``An Analysis of the Concordance Among 13 U.S.\ Wine
Competitions'' (2009).} [ii]
Of wines entered in more than three of thirteen competitions, 47\% received a gold medal
somewhere, but 84\% of those same gold-winning wines received no award at another
competition --- a pattern ``\textbf{influenced by chance alone}.''

\paragraph{Hodgson, ``How Expert are `Expert' Wine Judges?'' (2009).} [ii]
Under a weighted-kappa threshold of $.7$, ``less than 30 percent'' of participating judges
qualify as expert.

\paragraph{Hodgson \& Cao, ``Criteria for Accrediting Expert Wine Judges'' (2013).} [ii]
``[F]ew judges pass the test. Of greater interest is that many judges who fail the test
have vast professional experience in the wine industry\ldots wine competitions do not
provide reliable recommendations of wine quality.''

\paragraph{Ashton, ``Reliability and Consensus of Experienced Wine Judges'' (2012).} [ii]
``Both reliability and consensus are, on average, substantially lower in wine judging than
in other fields\ldots little support is found for the idea that experienced wine judges
should be regarded as experts.''

\paragraph{Cheng \& Gonzalez, ``Technical and Program Component Scores Frozen
Together'' (2022).} [iv]
``Ideally, technical and non-technical scores in judged sports would be independent\ldots
we describe an unexpected linear relationship between Technical Element and Program
Component scores in all four figure skating disciplines\ldots these relationships imply
difficulty bias\ldots and the possibility of outcome prediction of the Program Component
scores based on the Technical Element scores.''

\paragraph{Lesko\v{s}ek, \v{C}uk \& Kar\'acsony, ``Reliability and Validity of Judging in
Men's Artistic Gymnastics'' (2010).} [ii]
``Results show very high reliability (e.g.\ Cronbach alfa range from 0.92 up to 0.99).
Systematic bias in individual judge's scores and judges' panels were frequent\ldots judging
quality differs between apparatus, sessions and judges.''

\paragraph{Szab\'o \& Vanczer, ``Statistical Evaluation of Judging at the CDI-W Dressage
Competition'' (2012).} [ii]
``[J]udges tended to give higher scores and have higher level of disagreement in higher
level of competitions\ldots Some of the judges give significantly different average scores
to others.''

\paragraph{Wolframm, ``Let Them Be the Judge of That: Bias Cascade in Elite Dressage
Judging'' (2023).} [iii, iv]
Across 510 judges' scores at seven 5* Grand Prix events, non-criterion variables (home
event, same nationality, compatriot, FEI ranking, starting order) jointly explain 44.1\%
of variance in total dressage score, all $p<.001$.

\subsection{High-Stakes Selection}

\paragraph{Gompers, Gornall, Kaplan \& Strebulaev, ``How Do Venture Capitalists Make
Decisions?'' (2020).} [i]
``The management team was mentioned most frequently both as an important factor (by 95\%
of VC firms) and as the most important factor (by 47\% of VC firms).'' ``[A]lmost half of
the VCs\ldots \textbf{admit to often making gut investment decisions},'' while 22\% use NPV, 9\% use
no financial metrics at all, and one in ten ever quantitatively checks their own track
record.

\paragraph{Huang \& Pearce, ``Managing the Unknowable'' (2015).} [i, iv]
An angel investor: ``\textbf{I just use my gut feel. I trust my gut}'' (p.~644). Business-viability
data did not predict four-year outcomes; the entrepreneur assessment did (p.~657).

\paragraph{Rivera, ``Hiring as Cultural Matching'' (2012).} [iii]
A manager rejecting a technically strong candidate: ``He did well on the case and was very
articulate\ldots But I think he's too intellectual for [FIRM]\ldots \textbf{I don't think he'd be a
good fit}'' (p.~1010). ``Fit'' is a formally mandated criterion with no formal definition;
more than half of interviewees reported it was the most important criterion at interview
stage (p.~1007).

\paragraph{Dana, Dawes \& Peterson, ``Belief in the Unstructured Interview'' (2013).} [iv]
Interviewers ``formed interview impressions \textbf{just as confidently after getting random
responses} as they did after real responses,'' and the interview made predictions worse
(p.~512).

\paragraph{Highhouse, ``Stubborn Reliance on Intuition and Subjectivity in Employee
Selection'' (2008).} [i, ii]
Quoting Hastie \& Dawes on expert predictors: ``They rely on few pieces of information.
\textbf{They lack insight into how they arrive at predictions}. They exhibit poor interjudge
agreement. They become more confident in their accuracy when irrelevant information is
presented'' (p.~339). ``[T]he interrater reliability of the traditional (unstructured)
interview is so low that, even with a perfectly reliable and valid criterion,
interview-based judgments could never account for more than 10\% of the variance in job
performance'' (pp.~337--338).

\paragraph{Bastedo et al., ``What Are We Talking About When We Talk About Holistic
Review?'' (2018).} [ii]
``[I]nconsistent \textbf{definitions of a core admissions concept} make it more difficult for the
public to comprehend the `black box' of college admissions.'' Of 311 admissions officers,
three incompatible operational definitions of ``holistic review'' are in use: whole file,
whole person, whole context.

\paragraph{Stevens, \emph{Creating a Class} (2007).} [--]
``[O]fficers do not evaluate applicants; they evaluate applications'' (p.~20).

\subsection{Rule Extraction}

\paragraph{Biederman \& Shiffrar, ``Sexing Day-Old Chicks'' (1987).} [i]
``\textbf{Not a single sexer recalled being shown simple diagrams}\ldots Training was accomplished
by sexing live birds, which were then checked by experts'' (p.~643). After roughly one
minute of instruction on a single contour cue, naive subjects rose from 60.5\% to 84\% on
18 hard photographs, and the share of expert-judgment variance predictable from naive
subjects rose from 4\% to 67\% (pp.~643--644).

\paragraph{Horsey, ``The Art of Chicken Sexing'' (2002).} [i]
On WWII aircraft-spotter training (Allan 1958): ``\textbf{the experts had no idea how they had
acquired their skills in the first place, or how to transmit those skills to others}.
Training regimens therefore had to be developed somewhat by trial-and-error''
(pp.~108--109). On birdwatchers' ``jizz'': they ``\textbf{cannot say what the features are that
make up the whole}'' (p.~110). Memoir testimony from a sexer, R.~D.~Martin: ``\textbf{There was
nothing there but I knew it was a cockerel}'' (pp.~109--110 n.5).

\paragraph{Camerer \& Johnson, ``The Process-Performance Paradox in Expert
Judgment'' (1991).} [i, iv]
On paramorphic models: ``It describes judgments in a purely statistical way, as if experts
were weighing and combining cues in their heads; \textbf{the process they use might be quite
different}'' (p.~199 n.6). ``[L]arge numbers of configural rules, which knowledge engineers
take as evidence of expertise, do not necessarily make good predictions; simple linear
combinations of variables\ldots are better in many tasks'' (pp.~212--213). Across five
studies (Table 8.1, p.~201), a linear model built from a judge's own cue inputs beats that
same judge's accuracy in every row. On why false articulated rules persist: ``experts who
get slow, infrequent, or unclear feedback will not learn that their rules are wrong''
(p.~210).

\paragraph{Kundel \& LaFollette (1972), via Camerer \& Johnson (1991).} [--]
``[N]ovices and first-year medical students were unable to detect lesions from radiographs
of abnormal lungs, but fourth-year students (who had had some training in radiography)
were as good as full-time radiologists'' (p.~201).

\subsection{Cross-Cutting}

\paragraph{Argyris, Putnam \& Smith, \emph{Action Science} (1985).} [i]
``As many approaches to social inquiry emphasize, human beings can be understood to \textbf{act
according to rules that they cannot state}'' (p.~82). ``Theories-in-use are \textbf{the often tacit
cognitive maps} by which human beings design action.''

\paragraph{Johansson, Hall, Sikstr\"om \& Olsson, ``Failure to Detect Mismatches between
Intention and Outcome in a Simple Decision Task'' (2005).} [i]
Participants whose chosen option was covertly swapped ``can be induced to \textbf{confabulate an
explanation for why they selected the option they did not choose}''; only 26\% volunteered
detection.

\end{document}

%% file: math_commands.tex
\usepackage{amsmath,amsfonts,bm}

\def\eqref#1{equation~\ref{#1}}

\def\1{\bm{1}}

\DeclareMathAlphabet{\mathsfit}{\encodingdefault}{\sfdefault}{m}{sl}
\SetMathAlphabet{\mathsfit}{bold}{\encodingdefault}{\sfdefault}{bx}{n}

\DeclareMathOperator*{\argmax}{arg\,max}

%% file: exmetric_breakable.tex
\definecolor{exmetricline}{named}{ForestGreen}
\ExplSyntaxOn
\seq_new:N \l__exm_words_seq
\tl_new:N \l__exm_body_tl
\tl_new:N \l__exm_style_tl
\dim_const:Nn \c__exm_pad_dim {1.2pt}
\dim_const:Nn \c__exm_rule_dim {0.3pt}
\cs_new_protected:Npn \__exm_band:n #1 {
  \hbox:n {
    \rlap { \color{exmetricbg} \rule [-0.5ex] {#1} {2.15ex} }
    \rlap { \color{exmetricline!60!black} \rule [-0.5ex] {#1} {\c__exm_rule_dim} }
    \rlap { \color{exmetricline!60!black} \rule [\dim_eval:n {1.65ex-\c__exm_rule_dim}] {#1} {\c__exm_rule_dim} }
    \hspace{#1}
  }
}
\cs_new_protected:Npn \__exm_edge: { \hbox:n { \color{exmetricline!60!black} \rule [-0.5ex] {\c__exm_rule_dim} {2.15ex} } }
\cs_new_protected:Npn \__exm_open:  { \__exm_edge: \__exm_band:n {\c__exm_pad_dim} }
\cs_new_protected:Npn \__exm_close: { \__exm_band:n {\c__exm_pad_dim} \__exm_edge: }
\cs_new_protected:Npn \__exm_word:n #1 {
  \hbox_set:Nn \l_tmpa_box { \l__exm_style_tl #1 }
  \hbox:n { \__exm_band:n { \box_wd:N \l_tmpa_box } \kern -\box_wd:N \l_tmpa_box \box_use:N \l_tmpa_box }
}
\cs_new_protected:Npn \__exm_space: {
  \discretionary { \__exm_close: } { \__exm_open: } { \__exm_band:n { \fontdimen2\font } }
}
\NewDocumentCommand \exmetric { m } {
  \group_begin:
    \tl_clear:N \l__exm_style_tl
    \tl_set:Nn \l__exm_body_tl {#1}
    \tl_if_head_eq_meaning:nNT {#1} \textit { \tl_set:Nn \l__exm_style_tl {\itshape} \tl_set:Nx \l__exm_body_tl { \tl_item:nn {#1} {2} } }
    \tl_if_head_eq_meaning:nNT {#1} \emph   { \tl_set:Nn \l__exm_style_tl {\itshape} \tl_set:Nx \l__exm_body_tl { \tl_item:nn {#1} {2} } }
    \tl_if_head_eq_meaning:nNT {#1} \textbf { \tl_set:Nn \l__exm_style_tl {\bfseries} \tl_set:Nx \l__exm_body_tl { \tl_item:nn {#1} {2} } }
    \seq_set_split:NnV \l__exm_words_seq { ~ } \l__exm_body_tl
    \seq_remove_all:Nn \l__exm_words_seq { }
    \__exm_open:
    \seq_pop_left:NN \l__exm_words_seq \l_tmpa_tl
    \__exm_word:n { \l_tmpa_tl }
    \seq_map_inline:Nn \l__exm_words_seq { \__exm_space: \__exm_word:n {##1} }
    \__exm_close:
  \group_end:
}
\ExplSyntaxOff

%% file: figures/tab_datagrid.tex
\def\dsfontrowtitle{\fontsize{12}{12}\itshape\color{dsdeftext}}%
\def\dsfontrowdesc{\fontsize{6.5}{7}\itshape\color{black!60}}%
\def\dsfontcoltitle{\fontsize{14}{12}\color{dsdeftext}}%
\def\dsfontcoldesc{\fontsize{6.5}{7}\itshape\color{black!60}}%
\def\dsfontcellblack{\fontsize{9.5}{9}\selectfont\color{black!80}}%
\def\dsfontcellnum{\fontsize{6.5}{5}\selectfont\bfseries}%
\def\dsfontcellslash{\fontsize{6}{8}\selectfont\color{black!50}}%
\def\dsfontcellgray{\fontsize{4}{3.6}\selectfont\itshape\color{black!50}}%
\def\dsfontcellllm{\fontsize{6}{7}\selectfont\color{black!55}}%
\def\dsfontlegend{\small\bfseries}%
\def\dsfontfooter{\tiny\color{black!60}}%
\providecolor{dstexts}{HTML}{46237A}%
\providecolor{dslabels}{HTML}{369661}%
\providecolor{dsdeftext}{HTML}{1F5C3A}%
\providecolor{dsgreenfill}{HTML}{E9F3EC}%
\providecolor{dsbluefill}{HTML}{E8F0F7}%
\providecolor{dsblueline}{HTML}{417497}%
\ifdefined\dsvariant\else\def\dsvariant{oneline}\fi%
\ifdefined\dsentry\else%
\newlength{\dsw}\newlength{\dslw}\newlength{\dstotal}\newlength{\dsentrysep}%
\newlength{\dstitleh}\newlength{\dsnumh}\newlength{\dsfineh}\newlength{\dsrowgap}\newlength{\dsheadgap}%
\def\dsstackedname{stacked}%
\newcommand{\dsllm}{{\dsfontcellllm\selectfont LLM-labelled}}%
\def\dsllmtok{\dsllm}%
\newcommand{\dsentry}[5]{%
   \parbox[b][\dstitleh][b]{\dsw}{\centering\dsfontcellblack\selectfont#1\par}\par\nointerlineskip\vskip3pt%
   \def\dstmp{#5}%
   \ifx\dsvariant\dsstackedname%
     \parbox[c][\dsnumh][c]{\dsw}{\centering{\color{dstexts}\dsfontcellnum\selectfont #3}{\dsfontcellslash\selectfont\ texts}\\[0pt]%
       \ifx\dstmp\dsllmtok\dsllm\else{\color{dslabels}\dsfontcellnum\selectfont #5}{\dsfontcellslash\selectfont\ labels}\fi\par}%
   \else%
     \parbox[c][\dsnumh][c]{\dsw}{\centering{\color{dstexts}\dsfontcellnum\selectfont #4}{\dsfontcellslash\selectfont\ /\ }%
       \ifx\dstmp\dsllmtok{\dsfontcellllm\selectfont LLM}\else{\color{dslabels}\dsfontcellnum\selectfont #5}\fi\par}%
   \fi\par\nointerlineskip\vskip3pt%
   \parbox[t][\dsfineh][t]{\dsw}{\centering\dsfontcellgray\selectfont#2\par}}%
\newcommand{\dsnext}{\par\nointerlineskip\vskip\dsentrysep}%
\newcommand{\dscol}[1]{\parbox[c]{\dsw}{#1}}%
\newcommand{\dshead}[2]{\parbox{\dsw}{\centering{\dsfontcoltitle\selectfont#1\par}\vskip\dsheadgap{\dsfontcoldesc\selectfont#2\par}}}%
\newcommand{\dsrowlabel}[2]{%
   \parbox[c]{\dslw}{\centering{\dsfontrowtitle\selectfont#1\par}%
      \vskip\dsrowgap%
      {\dsfontrowdesc\selectfont#2\par}}}%
\newcommand{\dslegend}{\tikz[baseline=(L.base)]\node[fill=dsbluefill, draw=dsblueline!60, rounded corners=2pt,%
   inner sep=4pt, font=\dsfontlegend\selectfont] (L) {{\color{black!65}\mdseries Legend:}\ {\color{dstexts}texts}\,{\color{black!55}/}\,{\color{dslabels}labels}};}%
\fi%
\setlength{\dsw}{2.4cm}%
\setlength{\dslw}{2.2cm}%
\setlength{\dstitleh}{18pt}%
\setlength{\dsnumh}{9pt}%
\setlength{\dsfineh}{7pt}%
\setlength{\dsentrysep}{3pt}%
\setlength{\dsrowgap}{5pt}%
\setlength{\dsheadgap}{3pt}%
\setlength{\dstotal}{\dimexpr\dslw+8pt+7\dsw+36pt\relax}%

\vbox{\hbox{\renewcommand{\arraystretch}{0}%
\begin{tabular}{@{}c@{\hspace{8pt}}*{6}{c@{\hspace{6pt}}}c@{}}
 & \dshead{Coding}{patches; answers} & \dshead{Math}{proofs; answers} & \dshead{Law \& policy}{patents; comments} & \dshead{Academia}{abstracts} & \dshead{Journalism}{news \& releases} & \dshead{Fiction}{short stories} & \dshead{Humor}{jokes; sketches} \\[16pt]
\arrayrulecolor{black!20}\specialrule{.5pt}{8pt}{10pt}
\dsrowlabel{Gatekeeper\\verdicts}{a decision-maker accepts or rejects the text}
  & \dscol{\dsentry{PR merged}{GitHub pull requests}{44,751}{45k}{180k}
           \dsnext
           \dsentry{accepted \\ answer}{Stack Overflow}{16,001}{16k}{16k}}
  & \dscol{\dsentry{PR merged}{mathlib pull requests}{8,494}{8.5k}{46k}
           \dsnext
           \dsentry{accepted \\ answer}{Math.SE}{13,001}{13k}{13k}}
  & \dscol{\dsentry{patent granted}{USPTO examiners}{579,084}{579k}{1.3M}
           \dsnext
           \dsentry{comment answered}{Federal Register\\notice \& comment}{9,521}{9.5k}{16k}}
  & \dscol{\dsentry{paper\\accepted}{ICLR (OpenReview)}{28,402}{28k}{129k}}
  & \dscol{\dsentry{press\\pickup}{17 outlets: WSJ $\cdot$ CNBC $\cdot$ \\ CBS $\cdot$ Fortune $\cdot$ NYT \ldots}{72,315}{72k}{195k}}
  & \dscol{\dsentry{book deals}{RoyalRoad market pickup $\cdot$ Kindle Scout}{2,000}{2.0k}{2.0k}
           \dsnext
           \dsentry{post\\removals}{r/WritingPrompts $\cdot$ r/nosleep $\cdot$ r/books $\cdot$ r/asoiaf $\cdot$ r/gameofthrones}{398,622}{399k}{366k}}
  & \dscol{\dsentry{sketch aired}{SNL cut-for-time sketches}{72}{72}{72}
           \dsnext
           \dsentry{post\\removals}{r/Jokes $\cdot$ r/funny $\cdot$ r/Showerthoughts $\cdot$ r/nottheonion $\cdot$ r/me\_irl}{180,205}{180k}{180k}}
  \\
\specialrule{.5pt}{8pt}{8pt}
\dsrowlabel{Expert\\curations}{editors or judges select the best texts from a pool}
  & \dscol{\dsentry{bounty\\award}{Stack Overflow}{23,498}{23k}{7.1k}
           \dsnext
           \dsentry{competitions}{AtCoder $\cdot$ LeetCode $\cdot$ \\CodeChef $\cdot$ Codeforces}{6,353}{6.4k}{6.4k}
           \dsnext
           \dsentry{release\\highlights}{Rust release notes}{4,031}{4.0k}{4.0k}}
  & \dscol{\dsentry{bounty\\award}{Math.SE}{23,972}{24k}{8.6k}
           \dsnext
           \dsentry{competitions}{AoPS}{25,454}{25k}{25k}
           \dsnext
           \dsentry{release\\highlights}{mathlib changelog}{623}{623}{623}}
  & \dscol{\dsentry{change\\made}{Federal Register final rules}{7,084}{7.1k}{14k}}
  & \dscol{\dsentry{spotlight /\\ oral}{ICLR 2018--25 $\cdot$ NeurIPS 2021--25 $\cdot$ ICML 2023--25}{7,941}{7.9k}{7.9k}
           \dsnext
           \dsentry{best-paper\\awards}{32 venues: CHI $\cdot$ ICSE $\cdot$ FSE\\$\cdot$ PLDI $\cdot$ ICML $\cdot$ NeurIPS \ldots}{11,613}{12k}{12k}}
  & \dscol{\dsentry{homepage placement}{NYT $\cdot$ WSJ $\cdot$ WaPo $\cdot$ Guardian $\cdot$ BBC $\cdot$ CNN $\cdot$ LA Times $\cdot$ Reuters}{183,708}{184k}{184k}}
  & \dscol{\dsentry{magazine \\ picks}{Reedsy contests $\cdot$ RoyalRoad\\magazine $\cdot$ Wigleaf Top 50}{119,341}{119k}{119k}
           \dsnext
           \dsentry{awards}{r/bestofWP}{14,714}{15k}{27k}}
  & \dscol{\dsentry{magazine \\ picks}{Style Invitational $\cdot$ McSweeney's}{27,120}{27k}{27k}
           \dsnext
           \dsentry{awards}{@midnight \#HashtagWars}{12,734}{13k}{13k}}
  \\
\specialrule{.5pt}{8pt}{8pt}
\dsrowlabel{Community\\votes}{readers vote, follow or engage; signals are aggregated}
  & \dscol{\dsentry{answer\\votes}{Stack Overflow}{12,202}{12k}{51k}
           \dsnext
           \dsentry{issue\\reactions}{GitHub linked issues}{154,633}{155k}{40k}}
  & \dscol{\dsentry{answer\\votes}{Math.SE}{100,000}{100k}{316k}}
  & \dscol{\dsentry{answer\\votes}{Law SE}{4,192}{4.2k}{21k}
           \dsnext
           \dsentry{co-signing}{Federal Register comments}{9,520}{9.5k}{372k}}
  & \dscol{\dsentry{citations}{ICLR $\cdot$ NeurIPS $\cdot$ ICML,\\percentile within venue and year}{2,387}{2.4k}{389k}}
  & \dscol{\dsentry{most-read}{BBC most-read list}{50,761}{51k}{51k}
           \dsnext
           \dsentry{tweet\\engagement}{NYT $\cdot$ CNN $\cdot$ Reuters $\cdot$ Guardian \\$\cdot$ LA Times $\cdot$ WaPo $\cdot$ BBC}{46,456}{46k}{219M}}
  & \dscol{\dsentry{story\\upvotes}{r/WritingPrompts}{96,080}{96k}{5.7M}
           \dsnext
           \dsentry{followers}{RoyalRoad}{135,598}{136k}{12M}}
  & \dscol{\dsentry{joke\\upvotes}{r/Jokes}{383,786}{384k}{76M}}
  \\
\specialrule{.5pt}{8pt}{0pt}
\end{tabular}\arrayrulecolor{black}}
\vskip3pt
\hbox to\dstotal{\parbox[b]{.72\dstotal}{\raggedright\dsfontfooter\selectfont}\hfil\dslegend}}%

%% file: figures/fig_mining_loop.tex
\begin{tikzpicture}[
  bigbox/.style={rounded corners=7pt, line width=1.0pt},
  chip/.style={rounded corners=4pt, draw, line width=.5pt, inner sep=3pt,
    font=\scriptsize, align=center},
  badge/.style={circle, fill=black!75, text=white, font=\footnotesize\bfseries,
    inner sep=1.6pt, minimum size=4.6mm},
  llmtag/.style={rounded corners=3pt, draw=fgreen!70, fill=fgreen!12,
    line width=.6pt, inner sep=3pt, font=\scriptsize\bfseries, text=fgreen!40!black},
  boxtitle/.style={font=\small\bfseries, text=black!80, align=center},
  gsub/.style={font=\small\bfseries, text=fgreen!85!black, align=center},
  body/.style={font=\scriptsize, text=black!70, align=center},
  fine/.style={font=\scriptsize\itshape, text=black!45, align=center},
  exlabel/.style={font=\scriptsize\itshape, text=black!45, anchor=south west, inner sep=0pt},
  algolist/.style={font=\fontsize{6}{7}\selectfont, text=black!50, align=center},
  flow/.style={-{Stealth[length=2.8mm, width=2.4mm]}, black!70, line width=1.1pt},
  branchlab/.style={font=\small\bfseries, align=center, fill=white,
    inner sep=1.5pt, rounded corners=2pt}]
\definecolor{fgreen}{HTML}{369661}
\definecolor{famber}{HTML}{417497}
\definecolor{fred}{HTML}{46237A}
\definecolor{fpink}{HTML}{D6C9E8}

\begin{scope}[yshift=.2cm]   %
\draw[bigbox, draw=black!45, fill=black!2] (-9.15,3.05) rectangle (-5.75,6.75);
\node[boxtitle] at (-7.45,6.25) {Labeled corpus};
\node[body] at (-7.45,5.8) {items $+$ preference labels};
\foreach \x/\c in {-8.35/fgreen, -7.45/fgreen, -6.55/fred}{
  \draw[rounded corners=2pt, fill=white, draw=black!40, line width=.6pt]
    (\x-.38,4.35) rectangle (\x+.38,5.45);
  \foreach \y in {5.2,5.02,4.84}{\draw[black!30, line width=.7pt] (\x-.26,\y) -- (\x+.26,\y);}
  \fill[\c!75] (\x-.26,4.58) rectangle (\x+.02,4.68);}
\fill[fgreen!80] (-8.5,3.6) -- (-8.14,3.6) -- (-8.32,3.92) -- cycle;
\fill[fgreen!80] (-7.85,3.6) -- (-7.49,3.6) -- (-7.67,3.92) -- cycle;
\fill[fred!85]  (-7.2,3.92) -- (-6.84,3.92) -- (-7.02,3.6) -- cycle;
\node[font=\small, text=black!45] at (-6.35,3.72) {$\cdots$};
\end{scope}

\begin{scope}[xshift=.2cm, yshift=-.3cm]   %
\draw[bigbox, draw=fgreen!70, fill=fgreen!3] (-5.45,3.35) rectangle (-.3,7.45);
\node[llmtag] at (-.8,7.45) {LLM};
\node[badge] at (-5.02,6.92) {1};
\node[boxtitle, anchor=west] at (-4.68,6.92) {Metric Discovery};
\node[gsub, font=\footnotesize\bfseries] at (-2.87,6.45) {What metrics predict preference?};
\node[algolist, anchor=north, text width=4.8cm] at (-2.87,6.15)
  {{\fontsize{7}{8}\selectfont\bfseries Combined metric discovery algorithms:} \\[1pt]
   Autometrics \cite{ryan2026autometrics}\\
   ExperiGen \cite{sengupta2026experigen}\\
   HypoGeniC \cite{zhou2024hypogenic}\\
   HypotheSAEs \cite{movva2025hypothesaes}};
\matrix (ex1) at (-2.87,3.93) [column sep=2mm, nodes={chip, draw=fgreen!45, fill=fgreen!6}] {
  \node {cites diverse\\sources}; & \node {uses longer\\sentences}; & \node {repeats URLs\\verbatim}; \\
};
\node[exlabel] at ([yshift=.01mm]ex1.north west) {examples:};
\end{scope}

\draw[bigbox, draw=fgreen!70, fill=fgreen!3] (.4,2.8) rectangle (8.0,7.4);
\node[llmtag] at (7.45,7.4) {LLM};
\node[badge] at (.83,6.92) {2};
\node[boxtitle, anchor=west] at (1.17,6.92) {Spurious Variable Identification};
\node[gsub, text width=7.2cm] at (4.2,6.25) {Does the metric measure a construct, or is it a shortcut?};
\node[body, anchor=north, align=center, text width=2.3cm] at (1.7,5.55)
  {\textbf{Channel tests}\\ identifies features that draw signal from outside the text (e.g. time-of-day)};
\node[body, anchor=north, align=center, text width=2.3cm] at (4.2,5.55)
  {\textbf{Path tests} \\ identifies features correlated to significant variables (e.g. length correlates with effort).};
\node[body, anchor=north, align=center, text width=2.3cm] at (6.7,5.55)
  {\textbf{LLM check} \\ uses an LLM to decide qualitatively if a variable is spurious or not};
\begin{scope}[shift={(4.2,3.3)}, scale=.6, fgreen!75!black, line width=.9pt]
  \draw (0,-.62) -- (0,.55);
  \draw (-.78,.55) -- (.78,.55);
  \fill (0,.62) circle (.07);
  \draw (-.78,.55) -- (-1.0,.02);
  \draw (-.78,.55) -- (-.56,.02);
  \draw (-1.06,.02) arc (180:360:.28);
  \draw (.78,.55) -- (.56,.02);
  \draw (.78,.55) -- (1.0,.02);
  \draw (.5,.02) arc (180:360:.28);
  \draw[line width=1.1pt] (-.4,-.62) -- (.4,-.62);
\end{scope}
\node[font=\small\bfseries, text=fgreen!80!black] at (2.6,3.25) {quality};
\node[font=\small\bfseries, text=fred!90!black] at (5.8,3.25) {shortcut};

\draw[bigbox, draw=fred!75, fill=fpink!10] (-9.15,-1.1) rectangle (-2.6,2.2);
\node[badge] at (-8.72,1.55) {3};
\node[boxtitle, anchor=west] at (-8.38,1.55) {Ceiling: the unmined tail};
\node[body, text width=3.1cm, anchor=north west, align=left] at (-6.25,1.0)
  {Estimate upper bound of metric performance based on how many metrics we estimate to be undiscovered};
\begin{scope}[shift={(-8.75,-.52)}]
  \draw[-{Stealth[length=1.5mm]}, black!60, line width=.7pt] (0,0) -- (0,1.5);
  \draw[-{Stealth[length=1.5mm]}, black!60, line width=.7pt] (0,0) -- (2.35,0);
  \node[font=\tiny, text=black!50, rotate=90] at (-.18,.8) {acc.};
  \node[font=\tiny, text=black!50] at (.5,-.2) {criteria found};
  \draw[fgreen!85, line width=.9pt] plot[smooth] coordinates {(.05,.12) (.4,.62) (.85,.92) (1.3,1.05)};
  \draw[fgreen!85, dashed, line width=.9pt] plot[smooth] coordinates {(1.3,1.05) (1.65,1.14) (2.0,1.19)};
  \draw[black!45, dashed, line width=.5pt] (1.3,0) -- (1.3,1.05);
  \draw[fgreen!60!black, decorate, decoration={brace, mirror, amplitude=.9mm}] (1.35,-.05) -- (2.0,-.05);
  \node[font=\tiny, text=fgreen!60!black] at (1.67,-.32) {unseen $\mu$};
  \draw[black!55, dashed, line width=.5pt] (0,1.19) -- (2.0,1.19);
  \node[font=\tiny, text=black!60, anchor=west] at (.05,1.38) {ceiling};
\end{scope}

\draw[bigbox, draw=famber!80, fill=famber!5] (-2.1,-1.1) rectangle (2.9,2.2);
\begin{scope}[shift={(-1.35,1.13)}, fgreen!75!black, line width=1.1pt]
  \draw[fill=fgreen!12, line width=.8pt] (0,0) circle (.42);
  \draw[line width=1.4pt, line cap=round, line join=round] (-.2,.0) -- (-.06,-.16) -- (.22,.18);
\end{scope}
\node[boxtitle, anchor=west, text=famber!55!black] at (-.75,1.62) {Quality Metrics};
\node[body, anchor=north west, align=left, text width=3.4cm] at (-.75,1.37)
  {Use prompt optimization and\\code generation to implement\\ metrics in $\mathcal{V}$, $\mathcal{VA}$ models};
\matrix (ex2) at (.4,-.55) [column sep=1.5mm, nodes={chip, draw=famber!60, fill=famber!12, inner sep=2.5pt}] {
  \node {cites diverse\\sources}; \\ %
};
\node[exlabel] at ([yshift=.01mm]ex2.north west) {examples:};

\draw[bigbox, draw=fred!75, fill=fpink!12] (3.4,-1.1) rectangle (8.0,2.2);
\begin{scope}[shift={(4.1,1.13)}, fred!85, line width=1.0pt]
  \draw[fill=fpink!25] (0,0) circle (.42);
  \draw (-.3,-.3) -- (.3,.3);
\end{scope}
\node[boxtitle, anchor=west, text=fred!80!black] at (4.7,1.62) {Spurious Variables};
\node[body, anchor=north west, align=left, text width=3.1cm] at (4.7,1.37)
  {Deconfound dataset by\\stratifying, stripping text and\\residualizing.};
\matrix (ex3) at (5.7,-.55) [column sep=1.5mm, nodes={chip, draw=fred!50, fill=fpink!22, inner sep=2.5pt}] {
  \node {uses longer\\sentences}; & \node {repeats URLs\\verbatim}; \\ %
};
\node[exlabel] at ([yshift=.01mm]ex3.north west) {examples:};

\draw[flow] (-5.75,5.1) -- (-5.25,5.1);
\draw[flow] (-.1,5.1) -- (.4,5.1);
\draw[flow] (2.0,2.8) to[out=-100, in=90] (.4,2.2);   %
\draw[flow] (5.7,2.8) -- (5.7,2.2);   %
\draw[flow] (-2.1,.55) -- (-2.6,.55);
\end{tikzpicture}

%% file: appendix_balancing.tex
\section{Dataset construction and balancing}
\label{app:balancing}

Every corpus in this paper was built to a common set of rules: (i) the label is
defined from a decision that a real gatekeeper or community actually made, never
from a similarity to a reference; (ii) any variable that identifies the
\emph{producer} of an item (repository, agency, publisher, venue, magazine,
examiner, author) is treated as a confound to be balanced, grouped on, or
partialled out, and never as a feature; (iii) splits are stable hashes of a group
key rather than seeded shuffles, so that adding data never reshuffles an existing
split; (iv) presentation is normalized identically on both sides of the label
whenever the two classes came from different scrape pipelines. This appendix
records, per corpus, the exact filters, thresholds, matching variables, dedup
rules and leak fixes that implement those rules, including the ones that
overturned earlier results. Numbers that were found to be artifacts are stated
as retracted.

The appendix is organized by field, in the order of Table~\ref{tab:datasets}:
software code (\S\ref{app:bal:code}), mathematics (\S\ref{app:bal:math}), law
and policy (\S\ref{app:bal:law}), academia (\S\ref{app:bal:academia}),
journalism (\S\ref{app:bal:journalism}), creative writing
(\S\ref{app:bal:cw}), and humor (\S\ref{app:bal:humor}), followed by the
cross-cutting rules that recur across all of them
(\S\ref{app:bal:crosscutting}). Each field subsection first describes its texts and
defines what a verdict, a curation, and a community vote mean in that field, then
documents each corpus in turn.

\subsection{Software code}
\label{app:bal:code}

The texts of this field are units of proposed work: a pull request's title,
description and unified diff; a Stack Overflow answer beneath its question; a
competition submission beside the contest's editorial solution. The
\emph{verdict} is an accountable maintainer's decision to merge or close a
pull request, or an asker's decision to accept an answer to their own
question. \emph{Curation} is a deliberate pick of one item over its peers: a
bounty award to one answer among several, or the judgement that a submission
realizes the editorial's intended approach. The \emph{community} signal is
anonymous answer votes. Three corpora follow: the GitHub pull-request corpus
(\S\ref{app:bal:pr}), the three Stack Overflow tasks (\S\ref{app:bal:so}),
and the competition-programming editorial corpus
(\S\ref{app:bal:competition}).

\subsubsection{GitHub pull requests (expert verdict)}
\label{app:bal:pr}

\paragraph{Raw acquisition.} Three sources: a Kaggle dump of public
pull-request comments (one JSON per language, $\sim$2.9\,GB parsed), a GHTorrent
pull-request dump, and GitHub REST/GraphQL enrichment supplying merge status,
pull-request bodies, and full unified diffs
(\texttt{GET /repos/\{owner\}/\{repo\}/pulls/\{n\}.diff}; \textbf{140{,}874 diff
files}, $\sim$26\,GB). Pull-request identity is parsed out of each comment's
\texttt{html\_url}; $\texttt{paper\_id} = \texttt{owner/repo\#number}$. The label
is $\texttt{judgement}=1$ for merged, $0$ for closed-unmerged; the base corpus is
$\sim$141{,}000 pull requests at \textbf{$\sim$86\% accept / 14\% reject}
(19{,}605 rejections), handled by class weights rather than resampling.

\paragraph{Dense modeling text.} The dense model's input concatenates title,
description, and diff, with the diff split per file and files included
most-review-commented first until the token budget is exhausted; rows with a
missing diff or under 50 characters of text are dropped, and the text is
truncated at $\sim$6{,}000 characters ($\sim$2{,}048 tokens). 
The dense model for this task is \emph{withheld} from every quoted
result: the only trained model used a non-standard backbone and a
\texttt{max\_len} 512 budget over diffs whose median length exceeds that cap,
so it is not comparable to the paper's uniform Llama-3.1-8B LoRA recipe. The
table marks this model accordingly (footnote $m$); a standard Llama-LoRA
retrain, with the input rebuilt at a larger character cap, is required
before a dense number is quoted for this task.

\paragraph{Leak audit.} Three findings:
title author-status flags (\texttt{[WIP]}, \texttt{[draft]}, ``do not merge'')
occur in $\sim$1.6\% of rows and are near-deterministic (``do not merge''
$\to$ \textbf{7\% accept vs an 86\% baseline}); the corpus is project-bimodal
(\textbf{71\% of pull requests come from repositories merging $\geq$85\%, 14\%
from repositories merging $<$30\%, and only $\sim$15\% from balanced 30--70\%
repositories}); and 66 pull requests whose title contains ``merged by Bors''
were labelled rejected although a bot merged them.

\paragraph{Balancing.} 
We keep all repositories and downsample the majority class within each
repository; we do not filter down to 30--70\% repositories. Continuous
integration is a feature, never a filter: filtering to CI-pass would make the
canonical verifiable signal constant by construction. We strip the leaking title
token and keep the pull request rather than dropping it. We also measure
cross-repository variance (pooled AUC plus the per-repository AUC
distribution) rather than assuming it is zero.

\paragraph{Stage-0 corrective pool (exact constants).}
\begin{itemize}\itemsep2pt
\item Title-flag strip: 14 case-insensitive regexes (\texttt{[wip]},
\texttt{[draft]}, \texttt{[dnm]}, \texttt{[do not merge]},
\texttt{[do not review]}, \texttt{[review only]}, \texttt{[rfc]}, \texttt{[poc]},
\texttt{wip:}, \texttt{draft:}, \texttt{dnm:}, \texttt{[merged by bors]}, and
free-standing ``do not merge''/``do not review''), applied up to three passes.
\item Bot-merge label repair (flip $0\to1$) on ten title patterns: Bors in
brackets, plain, and \texttt{bors[bot]}; \texttt{[auto-merge]} and
\texttt{auto-merge:}; ``squashed and merged by''; Mergify automerge;
\texttt{kx-bot:}; \texttt{automerge:}; ``merge pull request \#N from''.
\item Bot-author drop: \texttt{[bot]} suffix, or membership in
\{dependabot, renovate, github-actions, bors, mergify, kx-bot,
step-security-bot\}.
\item \textbf{Temporal-gap repository drop}: if the absolute difference between
the median creation date of merged and of rejected pull requests exceeds
\textbf{12 months}, the whole repository is dropped, which excludes
repositories whose rejections all come from a different era.
\item \textbf{Within-repository balancing}: require \textbf{$\geq$10 per class}
per repository (else drop the repository), then subsample both classes to
$k=\min(n_1,n_0)$, seed 42. Result: \textbf{7{,}776 pull requests / 239
repositories} (an earlier baseline pool held 10{,}424).
\item Matched-pair deconfounding (stage B): within repository, 1:1 Hungarian
matching on $(\log\text{LOC}, n_\text{files}, \text{year})$ z-normalized within
the repository, under hard feasibility masks
\textbf{$|\Delta\text{LOC}|\leq0.20\times\max(\text{LOC})$,
$|\Delta n_\text{files}|\leq2$, $|\Delta\text{year}|\leq1$}; infeasible pairs are
discarded. Ladder~B is restricted to repositories with \textbf{$\geq$50 balanced
pull requests}.
\end{itemize}

\paragraph{Repository-inclusion filters for test execution.} Corpus-side
triage requires language $\in$ \{Python, Go, Java\},
\textbf{$n_\text{accepted}\geq8$ and $n_\text{rejected}\geq4$}, non-zero mean
changed files, and passes a reject-count gate (rejection rate
$<3\%$ \emph{and} $n_\text{rejected}<8$ $\Rightarrow$ excluded, which catches
bot-auto-merge repositories whose ``rejections'' are artifacts). Scoring favors a
\textbf{volume band of 40--400 pull requests} and penalizes repositories
older than a per-language dependency-rot cut (Go 2019, Python 2018, Java 2018).
GitHub-wide discovery enumerates 3 languages $\times$ 8 star bands $\times$ 2
push-recency windows and gates on \textbf{$\geq$15 merged and $\geq$8
closed-unmerged}; Java carries a \textbf{hard gate requiring a root
\texttt{pom.xml}} (measured $P(\text{signal}>0)$: maven 43\%, gradle 16\%,
neither 3\%). New repositories are admitted only after a \textbf{$\sim$40-pull-request
pilot}: the predictor of yield is the number of tests the suite runs in an
unmodified container (\textbf{0 tests $\to$ 1 signal row in 9{,}751 pull requests
$=0.0\%$; 1--50 $\to$ 5.7\%; 50--500 $\to$ 13.7\%; 500$+$ $\to$ 37.3\%}). A
zero-yield gate (\textbf{$\geq$30 executed pull requests and never any signal})
retired 781 batches that had consumed 47\% of all compute, raising effective
yield from 9.7\% to 18.1\%. Contamination control biases new scrapes to
$\texttt{created\_at} < $ 2023-01-01 (the corpus is $\sim$53\% pre-2023,
$\sim$29\% pre-2021), with a practical lower limit of 2019 for Go (pre-modules
dependency resolution fails).

\paragraph{Dedup rules.} Harvest consolidation deduplicates on the literal key
\textbf{\texttt{host|repo|pr\_number|verdict}}, keeping the first (older, more complete)
row and never deleting rows. Because the verdict is \emph{in} the key, re-runs add
rows, and downstream analysis prefers signal on conflict. A separate
phase-priority pass keeps one row per \textbf{(batch, paper\_id)} at priority
\texttt{post\_patch} $>$ \texttt{transplant\_added\_tests} $>$ \texttt{baseline},
removing \textbf{2{,}677 duplicates} (swebench-django $3{,}632\to1{,}082$,
swebench-sympy $655\to534$), with pre-dedup backups retained. A later pass
quarantined 451 conflicting pull requests (902 rows). The VAT join deduplicates
on $\texttt{repo}/\texttt{pr\_number}$.

\paragraph{The repo-confound retraction.} The first ladder
($V$ $.582$ / $A$ $.734$ / $T$ $.749$, read as ``$A\gg V$, no taste residual'')
is \textbf{retracted}. Four bugs: it was fit \textbf{in-sample} with no
cross-validation, which inflates a 63-feature articulated bank over an 8-feature
verifiable one; the join was performed \emph{without} dedup on a table carrying
\textbf{25\% duplicate repo/pull-request keys and 4{,}497 conflicting verdicts};
it pooled across repositories when \textbf{51\% of label variance is
between-repository}; and the dense split was random, leaking
\textbf{7.6\% exact-duplicate texts} from train to test on a split where
\textbf{repository base rate alone scores $.794$}, above the reported dense
$.749$. Repo-disjoint dense retraining gives $T=.584$.

\paragraph{The definitive corpus (44{,}751 pull requests / 594 repositories).}
Two exclusions produce the definitive corpus. First, a \textbf{label-pure batch drop}: roughly half
of all pass-to-fail rows came from \texttt{\_rej} collection batches that are
100\% rejected by construction (49 such batch directories exist), and these drove
a spurious pooled odds ratio of 4.0. Second, a \textbf{flaky-batch filter}:
batches whose pass-to-fail rate \emph{among accepted pull requests} exceeds
\textbf{25\% at $n\geq10$} were excluded (18 batches, including consul at 97\%),
having produced a Simpson reversal. After both, the sound mixed corpus is
51{,}695 pull requests / 604 repositories; after dedup and the articulated-metric
join the ladder task is \textbf{44{,}751 pull requests / 594 repositories}, with
$A$ scored as 68{,}083 diffs $\times$ 127 metrics. Cross-validated ladder:
$V$ $.576$ / $A$ $.615$ / $V{+}A$ $.632$, against \textbf{repository base rate
alone $.706$}; grouped by repository (unseen repositories) $V$ $.549$ / $A$
$.596$. The articulated bank retains a real within-repository signal
(\textbf{out-of-fold AUC $.577$ across 571 repositories, 69\% above chance,
Wilcoxon $p<10^{-4}$}), revising an earlier reading that $A$ was at chance.
That reading reflected limited power, not a repository-identification artifact. Its top features are
substantive norms (test presence, test--source correspondence, dependency
hygiene), not features that identify the repository.

\paragraph{Test-execution certification.} Each baseline is rebuilt at runtime rather than taken from the image: inside the container each pull
request runs \texttt{git reset --hard HEAD} $\to$ \texttt{git clean -fdx -e.git}
$\to$ \texttt{git checkout \$PARENT\_COMMIT}, with fetch-and-retry escalation and
a three-strike \texttt{checkout\_parent\_failed}; tree hashes are then verified
against \texttt{git show -s --format=\%T} on both the parent and the patched head,
so a mutated base tree or a mismatched patched tree is caught rather than scored.
Verdicts are assigned by an ordered decision tree
(\texttt{env\_broken\_post\_patch}, \texttt{fix}, \texttt{regression} with a
veracity layer that suppresses environment/oracle/flaky failures,
\texttt{regression\_via\_gone}, \texttt{new\_passing}, \texttt{new\_failing},
\texttt{no\_change\_still\_broken}, \texttt{no\_change\_all\_pass}) plus
infrastructure verdicts. The Go runner originally keyed
tests by the raw \texttt{go test -json} \texttt{Test} field including the full
subtest path, which embeds ephemeral ports, seeds and randomized parameters, so
identical code produced different test identities, spurious \texttt{gone} plus
\texttt{new\_passing} events, and false regressions on every no-change pull
request. Re-keying verdicts on stable-identity buckets converted
\textbf{1{,}051 rows} (703 to \texttt{no\_change\_all\_pass}, 338 to
\texttt{no\_change\_still\_broken}, 10 to \texttt{env\_broken\_post\_patch});
before/after examples: prometheus $1/91\to0/0$, spire $478/64\to44/34$.
Pass-to-fail was unaffected by construction throughout, because it counts only identities
present in both runs. All fail-to-pass/pass-to-fail versus acceptance
results predating this re-keying are void.

Flakiness is handled by a per-repository denylist built by running the suite
$N{=}3$ times in-image at the mainline base commit and unioning failing test
identities. Consul was excluded rather than denylisted for two reasons:
its flake set is unstable across calibration windows (a 3-run calibration
during a period of high flakiness produced a \textbf{225-test} denylist, while a 10-run
calibration in a less flaky window produced only \textbf{120}), and $\sim$96\% of
its regression verdicts are false, 71\% of them on merged pull requests. The
hard-coded exclusion set is \{hashicorp/consul, helm/helm, swebench-sphinx\}.

Fail-to-pass rows are certified by a static symbol-overlap validator: Python uses
overlap $\geq0.08$ with $\geq$1 assertion failure for \textsc{causal},
$\geq0.04$ for \textsc{unclear}, and $\geq0.15$ under all-compile-fail for
\textsc{feature\_add}; Go and Java use a binary ``grounded'' criterion (non-empty
intersection of called symbols and method receivers with diff-modified function
names) plus a 38-keyword Go stoplist. Pilot validation on 50 signals:
\textbf{precision $1.000$, recall $0.600$, $F_1$ $0.750$}, zero false positives.
Raw \texttt{fix} rows collapse by $\sim$25\% under certification, giving
\textbf{996 certified fail-to-pass}, of which 41 are repeat-supported, 847 rest
on a single measurable execution, and 108 conflict. A bracket adding
method-receiver syntax gives 1{,}192 (unfiltered) and 1{,}174
(stoplist-filtered), but the checked-in regex still excludes receivers by design
and the bracket harness is not in the repository, so we report 996. On the pass-to-fail side the checked-in de-flaked artifact holds
\textbf{6{,}557 rows} (a figure of 6{,}895 in earlier handoff notes does not match
the file on disk); its defensible stable-true subset is \textbf{5{,}211}.
The repo-stratified Mantel--Haenszel estimate is \textbf{OR $=1.214$, 95\% CI
$[1.094, 1.347]$, $p=7\times10^{-5}$, $n=49{,}656$, $k=466$ strata}, exactly
reproducible by summing the per-stratum table. This exposure uses the broader
\texttt{regression} verdict and must not be described as a strict pass-to-fail
effect; a strict sensitivity rebuild gives OR $1.676$ $[1.544,1.820]$ over 474
repositories, a \emph{different estimand}. The superseded
values 1.47, 1.49, 2.36, 2.74, 3.25 and 4.0 are retracted.

A reporting rule follows: pass-to-fail is positive on only
\textbf{1.8\%} of sound mixed pull requests, so even perfect precision caps AUC
at $.558$, and as measured (precision 19.4\% vs a 15.4\% base, lift
$1.26\times$) the cap is $.503$. Rare verifiable signals such as pass-to-fail are therefore reported as gates, precision and lift, never as pooled AUC. No row is fully
replay-certified under the project's six-part legitimacy contract, so rows are labelled
\texttt{structurally\_valid} / \texttt{repeat\_supported} /
\texttt{replay\_certified}, never ``verified''.

\subsubsection{Stack Overflow answers (verdict, curation, community)}
\label{app:bal:so}

One platform contributes all three channels on the same kind of text:
answers to programming questions, rendered beneath their question.
\emph{Verdict}: the accepted-answer task ($n{=}16{,}001$; $5{,}972$ accepted
/ $10{,}029$ not) contrasts the answer the question's asker marked accepted
against non-accepted answers, the asker being the one decider with standing
to close the question. \emph{Curation}: the bounty task ($n{=}23{,}498$;
$7{,}103$ / $16{,}395$) is a within-question design: on questions where an
asker or third party attached a reputation bounty, the answer awarded the
bounty is contrasted only against the other answers to the same question, so
the question itself (topic, difficulty, era) is the control.
\emph{Community}: the vote task ($n{=}12{,}202$; $6{,}391$ / $5{,}811$)
contrasts high- against low-scored answers, drawn from a collected pool of
$137{,}994$ ranked answer pairs. All three tasks share question-grouped
stable-hash splits (every answer to a question lands in one split), the
shared presentation-normalization pipeline, and the character-$n$-gram probe
gate. The answer-age confound documented for math.StackExchange
(\S\ref{app:bal:mathse}), in which earlier answers accumulate votes and acceptance
mechanically, applies identically to this platform and is handled the same
way; the instruments for all three tasks and the fused-selection protocol
are those of Table~\ref{tab:master-results}.

\subsubsection{Competition programming editorials (expert curation)}
\label{app:bal:competition}

\paragraph{Platform versus source.} Corpus rows must be filtered on the
\texttt{platform} column, \textbf{never on \texttt{source}}. A mapping function
keyed on source silently returned nothing for two aggregated dataset tags,
dropping \textbf{45{,}521 CodeChef candidates and $\sim$17{,}000 AtCoder
candidates}, which produced a false ``CodeChef has zero candidates'' conclusion
and downstream analysis that had to be discarded.

\paragraph{Editorial-code extraction.} CodeForces and AtCoder editorials store
prose only, with code inlined (often tokenized one token per line, sometimes
duplicated by a copy-button artifact). Extraction is non-destructive, writing a
separate table. Confidence buckets over 8{,}833 CodeForces and 9{,}974 AtCoder
editorials: confident 250 (2.8\%) / 2{,}910 (29.2\%); confident-multi 211 /
741; \textbf{none 8{,}361 (94.7\%) / 5{,}967 (59.8\%)}. Only the confident
buckets are used downstream.

\paragraph{Pair formation.} Pairs are formed per canonical problem
identifier, requiring \textbf{both the editorial code and the candidate code to
be non-null and at least 30 characters}. Counts: LeetCode 583{,}599; CodeChef
21{,}758; Luogu 52{,}207; USACO 900; CodeForces 7{,}468; AtCoder 12{,}444.
The \textbf{678{,}376 total} pairs span 27{,}744 editorials with code, 712{,}862
candidates with code, and 18{,}856 problems having both.

\paragraph{Editorial-derived candidate exclusion.} In a 449-pair CodeChef
evaluation task, \textbf{48 candidates (10.7\%) were themselves editorial code}.
Their label positive rate is \textbf{$1.000$} against $.706$ for the rest, their
mean similarity $.943$ against $.647$, and they make up \textbf{41.2\% of the
high-similarity decile}. Excluding them moves grouped AUC from
\textbf{$.7293$ to $.6704$ ($\Delta=-.0589$)}. The standing project rule that
follows is to exclude editorial-derived candidates from all evaluation tasks; AtCoder and CodeForces tasks were separately audited and contain none. One
known violation remains: the
USACO subset of one aggregate label file (872 rows, base rate $.985$) is entirely editorial-derived, so its
pooled $.722$ is a composition artifact. A mechanical exact-duplicate exclusion
also runs in the rebuilds (whitespace-stripped normalization; candidates matching
a gated editorial exactly are dropped).

\paragraph{The LeetCode near-copy rebuild.} \textbf{66.5\% of positives in the
old LeetCode task were editorial near-copies}, so its $.686$ was a copy-detection
artifact. Normalization strips docstrings and comments, collapses whitespace, and
lowercases; then
$\texttt{near\_copy} = \texttt{exact\_dup} \lor J_\text{identifier-token} > 0.83
\lor \cos > 0.95$. The thresholds were deliberately \emph{softened} from
$0.85/0.97$ to reach 20/20 recall on a known-copy reference set with a different
tokenizer, and validated at 20/20 true positives with 0/20 false positives on a
1{,}995-pair diagnostic. On the 102{,}436-row pool this removes \textbf{4{,}030
near-copies}, leaving \textbf{98{,}406 organic rows}; LeetCode is
\textbf{$.568$}.

\paragraph{CodeForces execution gating.} The old text-dump extraction was
structurally broken: \textbf{25.4\% sibling-problem contamination, and only
15.1\% (23/152) of extracted editorial code passed its own problem's tests}
(77.6\% compile errors), so the platform's $.545$--$.61$ was a data artifact,
not a ceiling. After a rendered-HTML re-scrape (1{,}428 blogs; 3{,}471 sections
with $\geq$30 characters of code against 442 confident rows before, an
$\sim$8$\times$ increase; the cause of the old failure was that syntax highlighting wraps every
token in a span, so text extraction must run without a separator), editorial code
is \textbf{execution-gated}: a test bank of 8{,}752 problems (median 100 tests,
\textbf{capped at 15 for gating}), C++ compiled at \texttt{-O2 -std=gnu++17}
with a 45-second compile timeout, a \textbf{5-second per-test timeout} and a
1024\,MB address-space limit, exact output comparison, and
\textbf{$\texttt{gate\_pass} = \text{compiled} \wedge \text{pass rate} = 1.0$}
(all tests must pass). Result: 2{,}217 gated $\to$ 1{,}830 compiled (82.5\%)
$\to$ \textbf{1{,}548 passing all tests (69.8\% yield)} against $\sim$15\%
before. The gate also detects contamination: pass rate by extraction
method is 71\% for anchor-detected sections versus 37\% and 0\% for text-heuristic
sections, and \textbf{98.5\% of gate-passing editorials come from anchor
sections}, so the old sibling contamination is structurally eliminated.

\paragraph{Frozen sampling and grouped splits.} Embedding similarity is used
\textbf{only} for stratification and \textbf{never as a feature}, under a
standing rule that any quantity computed from \emph{both} items in a pair
(embedding cosine, BERTScore, BLEU, edit distance, any alignment, and the judge's
own verdict) is a label proxy when the label is ``are these two similar'';
features must be functions of the \emph{candidate alone}, and the baseline is
chance, not similarity-alone. The similarity-decile target distribution is frozen
to the original labelled source set at
$\{<0.2: 13.7\%,\ 0.2$--$0.4: 14.6\%,\ 0.4$--$0.6: 17.6\%,\ 0.6$--$0.8: 31.2\%,\
0.8$--$1.0: 22.8\%\}$, with \textbf{at most 4 pairs per problem}, a target task
size of 1{,}000 (2{,}500 for the CodeForces rebuild), and seed 42; freezing was
adopted because changing the sampling policy changes task difficulty. Splits were
validated by re-running every task under \texttt{StratifiedGroupKFold(5)} grouped
by problem: all deltas are $|\Delta|\leq.011$ (CodeChef $.7019\to.7121$; Luogu
$.6130\to.6134$; LeetCode $.6254\to.6227$; a 14{,}234-pair task
$.6837\to.6813$), so problem leakage is ruled out and grouped splits are
mandatory thereafter. Execution features add essentially nothing to this label
because the labelled pools are correctness-saturated ($\sim$90--94\% accepted
verdicts, with runtime and memory coverage only on one platform); the verdict-only
value of $.473$ comes from a near-constant feature and is not reported, while the
execution-based verifiable value is $\sim.59$.

\paragraph{Strict-L1 relabelling.} The label is ``same core algorithmic idea''
under a strict decision procedure: strip boilerplate
(pragmas, includes, fast-I/O, fixed-point helpers, input validators, debug macros,
namespace declarations) and, if nothing algorithmic remains, assign 0 as
boilerplate/stub; treat pseudocode as algorithm; and assign the label on
semantics, where algebraically equivalent conditions, interchangeable data
structures in the same role, a different language, or breadth-first search versus
forward dynamic programming over the same state graph all remain 1, while a
different paradigm, recurrence, or asymptotic role is 0. Calibration errors are
listed as named anti-patterns (``both do dynamic programming'' or ``both sort''
is insufficient; the same recurrence or sort key is required). The relabelling
runs in shards of 10, truncates texts at 8{,}000 characters per side, writes results
append-only, and keeps the original label for analysis only, not as an anchor. On AtCoder, relabelling 2{,}495 pairs produced an
\textbf{8.9\% flip rate (169 up, 54 down)} and, with the task, features and
recipe unchanged, moved the score from \textbf{$.597$ to $.700$}. The gap was
therefore one of labelling discipline, not prompt redesign. CodeChef applied the identical recipe to
\textbf{the 401 remaining pairs} (449 minus the 48 editorial-derived). CodeForces
achieved 100\% coverage of 2{,}255 pairs with \textbf{91.8\% inter-run
agreement} on accidentally double-labelled rows. LeetCode was deliberately
\emph{skipped} (its problem was copy contamination, not label discipline), and a
second-level ``same realization'' label was tested and confirmed not useful.
Bulk pseudo-labelling by a smaller model failed its agreement gate
($\kappa=0.388$, with a 61:1 false-negative asymmetry) and was replaced by
cross-encoder distillation used only for bulk labelling and quality control.

\paragraph{Negative enrichment.} The CodeForces task had only 294 negatives, a
power limit created by the frozen low-similarity decile rather than a real base
rate. One thousand additional pairs were drawn at similarity $<0.2$, lowest
first, excluding already-sampled pairs, under a union per-problem cap of 4
(relaxed once to 6), with a mandatory hand-inspection gate: the script
writes an inspection file and emits labelling shards only after explicit
confirmation. Extended task: $n{=}3{,}255$, positive rate $.809$, 622 negatives,
ensemble $.686$ $[.660,.713]$ by cluster bootstrap. The final reported table is
AtCoder $.731$ / CodeChef $.729$ / CodeForces $.686$ (pre-enrichment $.627$)
/ LeetCode $.568$. Bank $\geq$ dense on every platform, so this domain has
no taste residual. One documented source confound applies to any editorial-style
claim: the LeetCode editorial corpus is 88\% from a single mirror, and an
editorial-versus-regular classifier scores $.9018$ across all sources but only
$.6712$ on the native-discussion subset against a $.604$ length-only baseline;
that is, length $\approx.60$, editorial style $\approx+.07$, and source
signature $\approx+.23$.

\subsection{Mathematics}
\label{app:bal:math}

The texts are mathematical arguments: a mathlib pull request's Lean diff, a
math.StackExchange answer under its question, a competition-forum solution
post beside the problem it solves. The \emph{verdict} is a maintainer's
merge decision or an asker's acceptance. \emph{Curation} is a bounty award
or the judgement that a solution takes the editorial approach (the
community's canonical solution path) for its problem. The \emph{community}
signal is answer votes. Three corpora follow: mathlib pull requests
(\S\ref{app:bal:mathlib}), math.StackExchange answers with their bounty
companion (\S\ref{app:bal:mathse}), and AoPS forum solutions
(\S\ref{app:bal:aops}).

\subsubsection{mathlib4 pull requests (expert verdict)}
\label{app:bal:mathlib}

\paragraph{Fetch and era-aware merge detection.} GraphQL fetch of
\textbf{37{,}249 closed pull requests} from
\texttt{leanprover-community/mathlib4} with full review detail. Merge status is
\emph{never} read off \texttt{state == CLOSED}; instead
$\texttt{merged}:= \texttt{bool(mergedAt)} \lor
\texttt{title.startswith("[Merged by Bors]")} \lor
\texttt{"ready-to-merge"} \in \texttt{labels}$, giving 32{,}787 merged. Bors
prefixes are \emph{stripped, not dropped}, with a trailing \texttt{+} in the
regex that is itself a bug fix: 18 titles (0.09\%) were merged through Bors
twice and carried a doubled prefix, which an anchored single-strip left in place
and mis-binned. Raw title-to-merged AUC was $.978$, and still $.794$ after
stripping. Splits are $\mathrm{md5}(\texttt{pr\_number})\bmod 100 < 80/90$.

\paragraph{The label $y = (\texttt{n\_review\_threads}==0)$, retired.} This is the \emph{friction} label, not the accept/reject label, and
the two must not be conflated. Its motivation was that
\texttt{CHANGES\_REQUESTED} is unusable in mathlib (99.4\% of merged pull
requests have zero) because reviewers use line-comment review threads instead;
$P(\text{zero threads})=.513$ over full history. It was balanced 50/50 within
(size quartile $\times$ conventional-commit prefix $\times$ year $\times$ author
association $\times$ primary \texttt{t-*} topic) cells, sampling
$n=\min(|pos|,|neg|)$ from each without replacement. The topic term was added
in v2 after the audit found \texttt{t-combinatorics} at $P(y{=}1){=}.36$ versus
\texttt{t-category-theory} at $.56$. The canonical v2 file is \textbf{16{,}884
rows, exactly 50/50, with maximum deviation from $.5$ of $.0000$ across 365 joint
cells}; v1 was 19{,}356. Label components, post-treatment variables
(\texttt{days\_open}, \texttt{n\_force\_pushes}, \texttt{n\_commits},
\texttt{n\_issue\_comments}, \texttt{closed\_at}) and raw labels were banned as
inputs, with final-state size and file count permitted \emph{as strata only}.
It was nevertheless retired as circular: $y{=}0$ pull requests average
3.9 norm threads and 100\% have at least one, while $y{=}1$ pull requests have
exactly zero, so agreement with the literal definition is $1.0$; the leakage
profile confirms the circularity (\texttt{days\_open} $.67$, \texttt{n\_commits} $.70$,
\texttt{n\_force\_pushes} $.77$, full numeric set $.78$).

\paragraph{The canonical accept/reject label.} $y$ is merged versus
closed-unmerged, recovering the 4{,}462 unmerged pull requests the friction build
had discarded. Cleaning to \textbf{35{,}796 rows (accept $.908$)} drops
WIP/test/draft titles; requires engagement
($\texttt{n\_reviews}>0 \lor \texttt{n\_review\_threads}>0 \lor
\texttt{n\_issue\_comments}\geq2$); requires $\texttt{size}\geq10$; and keeps
author association \textsc{contributor}/\textsc{collaborator}/\textsc{member}
while \textbf{dropping \textsc{none}} (outsiders merge at $.28$ versus
\textsc{member} $.94$), retaining the column. Non-degeneracy was checked: title-only
TF-IDF is $.62$ pooled and $.623$ \emph{within a fixed author tier}. Diffs for
all 35{,}796 were fetched at a 150\,KB cap ($\sim$4\% truncated, $\sim$2\%
HTTP 406), 99.5\% usable.

\paragraph{Deconfounding by stratification.} Three confounds were measured:
\textbf{port era} (\texttt{\#align}/\texttt{porting} tokens in 55/79/32\% of
2022/23/24 diffs and 0\% from 2025), \textbf{recency drift} (accept
$.95\to.88$), and \textbf{change type} (feat $.91$, chore $.94$ versus perf
$.61$, test $.06$). The fix restricts to
$\texttt{conv\_prefix}=\texttt{feat}\wedge\texttt{year}\geq2025$ and reads the
\emph{diff only} so the title cannot leak category or recency, giving
$n{=}8{,}494$. The restriction is diagnostic: title-only falls $.638\to.566$
(its pooled value was \emph{entirely} confound), while diff-only \emph{rises}
$.664\to.718$.

\paragraph{Size and one-sided label-noise cleaning.} The canonical clean slice is
$\texttt{additions}>0 \wedge \texttt{additions}\leq1000 \wedge
\neg(\texttt{judgement}{=}0 \wedge \texttt{n\_review\_threads}{=}0)$, giving
\textbf{$n{=}7{,}956$, accept $.943$}. Mega pull requests ($>$1000 additions) are
$n{=}92$ at accept $.065$ with 71\% zero-review; they are abandoned bulk imports. Of
985 rejections, \textbf{504 (51\%) were closed with zero review engagement}
(merge conflict 30\%, blocked by another pull request 29\%, WIP 20\%, migrated to
a fork 15\%): no maintainer read the code, so they are dropped. The noise is one-sided: no-review \emph{accepts} are kept, because a
frictionless merge is a genuine positive. Small pull requests ($\leq$10
additions, 14.5\% of data, 91\% accept) are kept and were hand-verified as
legitimate focused lemmas. After this cleaning, size-to-reject AUC falls
$.670\to.618$.

\paragraph{Author stripping, topic residualization, tactic promotion.} Contributor
names in Lean copyright headers were top TF-IDF features, so author and copyright
text is removed from the diff (\texttt{diff\_noauth}). The rule adopted is
to strip the text rather than balance by author, because balancing loses data, and stripping does not lower $C$ ($.742\to.750$). The articulated judge sees
\texttt{diff\_noauth} only, i.e.\ it is author-blind. Topic is
\emph{residualized} rather than promoted to a metric, because a check shows
that $V$ is topic-independent
($V$ residualized $.649\to.635$, and $V+\text{topic}=.644\leq V$), with topic
standalone only $.547$. Tactic idiom \emph{is} promoted, into $V'$: the aggregate
automation ratio is $.497$, exactly chance, and only the fine-grained per-tactic
profile carries signal ($.613$ standalone). Final topic-residualized ladder:
$V$ $.635$ / $V'$ $.680$ / $C$ $.736$ / $A\approx.48$. Judge truncation is
budgeted \textbf{in tokens, not characters} (6{,}200), because Lean diffs are
unicode-token-dense. One leg is retracted: the style/revision dataset,
whose \texttt{first\_push\_excerpt} header fallback was a label leak (the
null-line share is 82\% and the outdated rate is $1.000$ when the line anchor is
null versus $.049$ when it is present).

\subsubsection{math.StackExchange answers (community; curation via bounties)}
\label{app:bal:mathse}

\paragraph{The corner filter.} Positives are \textbf{accepted $\wedge$ score
$\geq3$}; negatives are \textbf{score $\leq0$}; the ambiguous middle (score 1--2,
or high-scoring but unaccepted) is dropped entirely. Answers whose stripped text
is under \textbf{50 characters} are dropped as stubs, as are answers whose parent
question is missing. Vote and comment tables are deliberately never model input,
because comment text states quality directly; comment text is reserved for a
reasoning-bearing variant only.

\paragraph{Question-disjointness runs first.} If a question holds \emph{both} a
positive and a negative answer, the whole question is dropped, so question text
can never carry the label across the split boundary.

\paragraph{Length balancing.} Rows are grouped by primary tag, with tags present
in only one class skipped entirely. \textbf{Percentile bin edges are computed on
the smaller class only} (10 bins). The joint bin key is (question-length bin,
answer-length bin), extended in v3 to also include a position bin
$\min(\text{answer position}, 3)$ and a three-year bucket. Within each bin,
$n=\min(|\text{small}|, |\text{available large}|)$ and \textbf{both} classes are
sampled at that count. This per-bin sampling is the v3 correction, because v1 subsampled only in
aggregate, matching class counts but not their per-bin distribution. Retention to the
target size prioritizes seven proof- and explanation-oriented tags, balancing
those first and filling the remaining quota by uniform random draw per class
(uniform because it preserves the matched distributions in expectation), at seed
42. Position and age metadata are computed over \emph{all} answers to a question
in the raw dump, \emph{before} any labelling, length, or disjointness filtering.

\paragraph{Grouped md5 split.} $\mathrm{md5}(\texttt{question\_id})\bmod100 <
80/90 \to$ train/eval/test, so every answer to a question lands in the same
split; both audits verify \textbf{zero train/test question overlap}.

\paragraph{Version lineage.} v1/v2 applied
per-tag and length matching but included no question identifier, so downstream
splits could not be group-aware. v3 added position matching, giving
\textbf{exactly 100{,}000 rows over 98{,}914 unique questions}, but left a
year confound open. v3.1 added three-year buckets to the matching key.
v3.2 attempted topic-cluster balancing at $K{=}100$ and failed its gate:
the floor moved only $.665\to.640$ against a $\leq.55$ target, because the
signal is finer-grained than topic (it lies in register and phrasing). v3.3 is canonical: balancing within cross-fitted question-propensity decile $\times$
three-year bin, \textbf{99{,}722 rows at 50/50}, floor $.461$.

\paragraph{The time confound.} Over 411{,}604 within-question
pairs, \textbf{the earlier-posted answer wins 67.3\%} of the time, rising to
\textbf{76.4\% at a score difference $\geq5$}, while the \emph{longer} answer
wins only \textbf{52.9\%}. (Post identifiers are globally sequential, so ordering
is directly readable.) The v3 build's own audit measures $P(\text{positive posted
earlier}) =.736$ on 54{,}417 within-question pairs, measured before the
disjointness filter, which removes exactly those mixed-class questions.

\paragraph{The v3.3 propensity procedure.} A single global TF-IDF (1--2 grams,
minimum document frequency 5, 200{,}000 features) is fit on \emph{question text
only}. Propensity is computed \textbf{out of fold} under
\texttt{GroupKFold(5)} grouped by question; this cross-fitting prevents the
propensity model from memorizing the rows it will later balance. The
out-of-fold propensity AUC is $.669$. Cells are (propensity decile $\times$
three-year bin) with $\pm\infty$ outer edges; within each cell
$n=\min(|\text{pos}|,|\text{neg}|)$ are sampled without replacement at a fixed
generator seed. Priority-tag retention then fills to the target size. The input
and output floors are recomputed with the same fresh question-only model and
written to the build manifest.

\paragraph{Effect of balancing.} Test AUC, v3 $\to$ v3.3: question-only floor
\textbf{$.6496\to.4690$}; full-text word $.6832\to.5900$; answer-only word
$.6584\to.6212$; answer-only character $.6812\to.6428$; publication year
$.5978\to.5073$; all-structural $.6106\to.5608$. \textbf{The answer-word margin
over the question floor rises from $+.009$ to $+.152$}; this margin is the quantity that supports a claim about answer quality. Decile-by-class balance is
exactly $.5000$ in all ten deciles; \textbf{answer length sits at $.5029$, i.e.\
chance}, and no single structural feature exceeds the $.55$ flag.

\paragraph{Caveats.} A second-order textual leak remains in the
released file: post-publication edit markers (\texttt{added} $+2.02$,
\texttt{edit} $+1.89$, \texttt{my comment} $+1.65$ in log-odds) accrue
\emph{after} votes and were not stripped. The floor guarantee is also
specific to the TF-IDF with logistic regression model family (a stronger question-side
model may recover part of the question-side signal), so every judge or metric AUC must be reported against the question-only floor of the dataset version used, and only the margin above that floor counts as evidence.

\paragraph{The bounty task.} The curated task on this platform
($n{=}23{,}972$; $8{,}503$ / $15{,}469$) uses the same corpus under a
within-question design: on bountied questions, the answer awarded the bounty
is contrasted against the other answers to the same question. Because the
contrast never crosses questions, the question-side confounds controlled by the
balancing above (topic, era, question
propensity) are held fixed by construction, and the position/age confound is
attenuated by the same within-question framing that measured it. Splits are
the same question-grouped stable hash.

\subsubsection{AoPS forum solutions (expert curation)}
\label{app:bal:aops}

\paragraph{Crawl and recovery.} The community AJAX endpoint was crawled by 8
workers over topic ids $[1, 3{,}600{,}000)$ at 2 seconds per request; $\sim$60\%
of ids carry content at a mean of 6.6 posts per topic. A long-running gzip append
corrupted legacy shards; recovery by scanning for gzip member magic offsets and
decompressing members independently restored \textbf{$+2{,}004$ topics
($3{,}747\to5{,}749$ of 6{,}780; 1{,}029 unrecoverable)}. This forum has
\emph{no} accepted-answer concept, so preference must be derived from within-thread
signals.

\paragraph{The wiki-to-forum join.} Only 164 forum
links appear across 4{,}937 wiki problem pages, too sparse to serve as a
join key. The join is therefore text matching between forum \emph{first posts} (which
repost the problem statement) and the wiki problem statement, using character
3--5-gram TF-IDF cosine after markup normalization across two markup dialects and
legacy escape sequences. \textbf{Acceptance rule: cosine $\geq0.75$ outright;
$0.55$--$0.75$ additionally requires numeric-literal overlap $\geq0.5$};
near-duplicate secondaries ($\geq0.85$ and within $0.02$ of the best) are retained
so that shared contest problems keep both pages. Coverage is
\textbf{3{,}437 of 4{,}877 wiki problems (70.5\%)}, with residual join noise
estimated at 1--2\% and the main remaining error source being different problems sharing
an identical answer-choice list.

\paragraph{Solution-post retention.} From 85{,}964 posts in matched threads and
82{,}433 replies, \textbf{32{,}120 posts are retained} under the rule: post
number $>1$, \textbf{at least 200 characters of non-quote content}, math markup
present, and bump/thanks posts dropped.

\paragraph{The transcription-band filter.} Posts with word-level
similarity $\geq0.5$ to the wiki solution are dropped. The rationale is
wiki-from-forum circularity: wiki solutions are frequently transcribed
\emph{from} these very threads, and 1.4\% of forum posts \emph{are} the wiki
solution, so causality runs forum $\to$ wiki. This drops \textbf{955 posts
(3.6\%)}, a band that is \textbf{94--96\% positive by construction}, moving the
pool from 26{,}757 to \textbf{25{,}454 at base rate $.687$}. Two candidate
artifacts were explicitly ruled \emph{not} transcription: boxed answers and
spoiler-hide blocks (only 317 of 8{,}647 hide-posts are high-similarity, and
among the retained posts $P(y{=}1\mid\text{boxed})=.847$ versus $.618$). These are the
\emph{complete-solution register}, a forum convention whereby full solutions
get a hide block and a boxed answer, and so they remain as verifiable features.

\paragraph{Admissibility.} Because
the label (same approach) is defined relative to the wiki solution, any feature computed against that reference (lexical similarity, embedding similarity, match score) is inadmissible as a predictor, for two reasons:
the reference may not exist or be knowable at prediction time, and the
label is partly \emph{defined} by that similarity, which makes such a feature circular
rather than predictive. For reference only, and never in a ladder,
similarity-to-wiki scores $.757$ alone and $.771$ with the verifiable features;
in the ladder code these features are kept in a separately marked lexical block that is reported
and excluded. Admissible ladders use only properties of the solution post itself plus
models fit out of fold on those properties.

\paragraph{The verified-correct two-regime design.} Correctness is a
deterministic verifiable feature: extract the claimed final answer (boxed
expression with brace balancing, then a trailing multiple-choice letter, then a
last-integer fallback) and check it against the official answer key. Extraction
succeeds on \textbf{65\% of 15{,}642} solutions on answerable problems
(97\% for AIME, 33--48\% for AMC, whose answers are letters and often omitted),
and \textbf{76\% of extracted answers match}. Correctness analyses are therefore
AIME-weighted. \textbf{Regime 1 (full pool)}: $n{=}25{,}454$, base
$.687$, giving $V$ $.706$ and TF-IDF $.726$. \textbf{Regime 2 (verified-correct,
excluding non-solutions)}: $n{=}6{,}140$, base rate jumps to $.882$, and
predictability \emph{collapses}: TF-IDF $.726\to.594$, $V$ $.706\to.560$.
Deviation from the editorial approach among correct solvers is rare (12\%) and
only weakly predictable. $.73$$.59$$.74$Reporting rule: quote $.73$ full-pool with the two-regime nuance; never a flat $.59$ and never a flat $.74$. The $.73$ is
substantially a solution-\emph{completeness} gradient (worked solution versus
fragment, meta-comment, or wrong answer), which is highly articulable. The base
rate itself decomposes into a mechanical part (the label matches \emph{any} of up
to six editorials, so the rate climbs from $.59$ with one shown to $.80$ with
five), a real part (canonical-technique prevalence: $.88$ among verified-correct,
$.81$ for AMC8 versus $.57$ for IMO), and non-solutions (6\% of posts, matching
at $.003$). A 50-case audit with three blind auditors gives agreement $.72$
($\kappa\approx.44$, $.85$ on high-confidence cases) and shows the judge is
\emph{conservative} rather than yes-biased.

\paragraph{The secondary de-confounded label.} A thanks-based residual is defined
as $\log 1p(\text{thanks})$ minus an out-of-fold prediction from
\emph{confound features only} (post number, within-thread solution rank both
raw and fractional, post age, days since the thread's first retained solution,
topic views, and the number of solutions in the thread), fit with gradient
boosting under 5-fold grouping by problem, out-of-fold $R^2=.522$. Validation
shows the confounds are removed: post number $\rho$ $-.323\to-.004$, solution
rank $-.285\to+.005$, post age $+.576\to+.012$, and within-problem post number
$-.550\to+.061$, a slight overshoot that establishes $\pm.02$--$.06$ as the
noise floor for reading within-problem residual correlations.

\subsection{Law and policy}
\label{app:bal:law}

The texts are persuasive or petitionary documents addressed to an
institution: a patent application's claims and abstract, a public comment
filed in a rulemaking docket, a party's factual record before a tribunal.
The \emph{verdict} is the institution's disposition: grant or abandon an
application, respond or not respond to a comment. \emph{Curation} is the
rarer decision to act: the agency changes the rule a comment asked it to
change. The \emph{community} signal is co-signing, in which other members of the
public add their signatures to a comment. Three corpora follow: patent
prosecution (\S\ref{app:bal:patents}), notice-and-comment rulemaking
(\S\ref{app:bal:nc}), and the twelve-domain legal-outcomes collection
(\S\ref{app:bal:legal}). The legal-outcomes collection is documented here for completeness but
sits outside the benchmark's preference mapping, because a court ruling on a
dispute is not a preference expressed about a creative product. It is therefore
appendix-only and does not enter any cross-field figure.

\subsubsection{Patent prosecution (expert verdict)}
\label{app:bal:patents}

\paragraph{Two tasks.} The patent corpus supports two prediction problems that we
keep distinct throughout. \emph{Task~1, outcome prediction:} from an
application's pre-grant text ($X=$ published claims $+$ abstract) predict its final
disposition ($y{=}1$ granted, $y{=}0$ abandoned), a supervised classification on
the balanced Cohort~A below. \emph{Task~2, prior-art disclosure} (the label-free
subtask): given a claim and a set of candidate references, decide element by
element whether the prior art discloses each limitation. This decision is the automated analog
of an examiner's anticipation read; it is computed on the nested claim-level cohort and is
\emph{never} conditioned on $y$. Task~2 is the mechanism on which Task~1's outcome is
meant to depend. We keep the two tasks separate so that we can measure how much of the outcome the disclosure signal carries.

\paragraph{Sources and label logic.} PatEx \texttt{transactions.csv} (13\,GB
event log) and \texttt{application\_data.csv}; PatentsView pre-grant bulk
(\texttt{pg\_claims}, \texttt{pg\_cpc\_current}, abstracts); and the USPTO Office
Action Research Dataset ($\sim$60M per-office-action citation rows; 2.19M
per-application rejection flags). Office actions are event codes
\texttt{CTNF, CTFR, CTAV, CTEQ}; allowance is \texttt{MN/=., N/=., N/=A, N/=F};
abandonment is any code beginning \texttt{ABN} or \texttt{MABN}. The label logic has known defects, recorded in the builder: \texttt{CTAV} (advisory) and
\texttt{CTEQ} (Quayle) are counted as office actions, distorting the
first-draft-approved flag; events are counted as raw rows with no (code, date)
dedup; abandonment uses \emph{any} historical event, so a revived, currently
pending application can be labelled abandoned; and \textbf{$\sim$3.3\% of
abandoned applications in the outcome cohort received a Notice of Allowance
before abandoning}, so ``abandoned'' is not ``examiner rejected''.

\paragraph{Cohort A construction (579{,}084 rows).}
\begin{enumerate}\itemsep2pt
\item  \texttt{pgpub\_id} \texttt{cpc\_type=inventional}  \texttt{cpc\_sequence} 
\item  $\texttt{date\_published}\leq$ 
\end{enumerate}

We join CPC on \texttt{pgpub\_id}, keep \texttt{cpc\_type=inventional} rows only, and
take the lowest \texttt{cpc\_sequence} (the primary classification) at single-letter
section depth. We filter to $\texttt{date\_published}\leq$ 2021-12-31 to avoid
prosecution-survivorship bias: 2022--24 filings show a 32--44\% positive rate
against a 50\% baseline. This bounds publication year, not filing year. We keep
granted ($=1$) and abandoned ($=0$) and drop pending and unknown. An application
must also appear at least once in the office-action citation table; a later audit
corrected the scope of that filter, since the filter admits any citation row: action
type is not checked, and $\sim$79\% of citation rows have an empty action type, so
some are information-disclosure statements rather than rejection-linked citations.
The text is published pre-grant claims plus abstract, bucketed by length at
$[0, 3000, 6000, 10000, 16000, \infty)$. A preliminary-amendment regex flags
canceled-claim markers, requiring an explicit claim number or range immediately
before the parenthesized ``canceled''/``cancelled''; the flag is set but the
markers stay in the text, so $X$ is post-preliminary-amendment published text, not
strictly at-filing text. We then balance: within each (CPC section $\times$ length
bucket) cell we take $2\cdot\min(\text{pos},\text{neg})$, an exact 50/50 at seed 42,
and cap each section at 100{,}000 rows. Without the cap, CPC sections G and H make up $\sim$62\% of
the dataset.

\paragraph{\protect Defects of the cohort.} \textbf{Year is not in the
balance cells}: grant rate drifts $.461$ (2011) $\to$ $.592$ (2017) \emph{inside}
the balanced file, and publication year alone scores $.531$, so random splits are
era-inflated by $\sim$.03 and out-of-time or year-stratified splits are required.
\textbf{579{,}084 rows carry only 577{,}430 unique ids, with 3{,}427
exact-duplicate texts (533 cross-label)}, and patent \emph{families}
(continuations sharing near-identical specifications) are not grouped anywhere.
The exact CPC $\times$ length-bucket balancing deletes mechanically
predictable verifiable signal by construction, so downstream recovery numbers are \emph{residual}-$V$ and
precision/threshold statistics are unrecoverable from the artificial 50/50 base
rate. A separate file, the naive 250K/250K random balance, is \emph{not} this
cohort (length alone scores $.579$ on it) and is excluded from all VAT runs.

\paragraph{Application-level grouping and examiner leave-one-out.} The join key
throughout is the zero-stripped application number; cross-validation uses a
stable-hash 5-fold \emph{grouped by application}, plus an out-of-time split
(train $\leq$2016, test $\geq$2017). Text is deduplicated by md5 before any
baseline, and cross-label text groups are dropped entirely. A prosecution-event
panel over 9.71M disposed applications defines substantive rounds as
\texttt{CTNF}$+$\texttt{CTFR} only (globally deduped on
(application, code, date)), and computes \textbf{leave-one-out leniency} at three
levels (examiner, art unit, and art unit $\times$ filing year) as
$\texttt{loo}=(S-y_i)/(n-1)$, undefined for singleton groups. This was the first
time examiner fields were read anywhere in the pipeline.

\paragraph{Examiner identity alone ($.681$) beats every text baseline
($.654$--$.658$), so the once-quoted ``1\% verifiable vs 70\% dense'' gap is
examiner-confounded on one side and provenance-orphaned on the other, and is
retracted in both directions.} On
cohort A with a 100\% panel join: \textbf{examiner leave-one-out leniency alone
reaches AUC $.681$}, art unit $\times$ year $.633$, art unit $.607$, publication
year $.531$, text length $.510$, and leniency plus year $.687$. Thus
\textbf{examiner identity alone beats the linear text baseline of
$.654$--$.658$}. On the claim-level cohort, metadata alone (leniency plus
structural counts, no text at all) scores far above both. That score is a leakage
diagnostic, not a baseline: examiner leniency is estimated partly from the same
outcomes being predicted, so we report it only as evidence that the cohort is
contaminated, and never as a comparison target. The disclosure metric's
marginal contribution is $+.0002$ $[-.001,+.001]$. The
\textbf{``1\% verifiable signal versus 70\% dense'' comparison is
retracted}: the $.70$ dense figures are provenance-orphaned and cross-cohort (their
naming indicates training on the \emph{naive} globally balanced file where CPC,
era and length are freely predictive, with row-random splits so duplicates and
families straddle, test-split checkpoint selection, and a 1024-token limit
against 44\% of rows being longer), whereas the $\sim$1\% mutual-information
figure is residual-$V$ on the confound-deleted cohort. The leakage diagnostic above already exceeds the orphaned dense
ceiling ($.7005$), so neither side of that comparison measures what it claimed
to. Where a comparator is needed we use examiner-LOO leniency ($.681$), itself
reported with the caveat that examiner identity is partly outcome-derived.

\paragraph{Text-predictability by outcome definition.} Whether a \emph{different}
outcome would carry more signal than grant/abandon (e.g.\ ``the application
will need another iteration'') was tested directly on a natural (unbalanced)
cohort with an out-of-time split (train $\leq$2016, test $\geq$2017) and a hashing
TF--IDF linear baseline over md5-deduplicated text. The iteration outcomes carry
\emph{less} text signal, not more: final grant/abandon $.653$, first-action
allowance $.636$, appeal after final $.606$, abandon after two rounds $.605$,
``another round'' ($\geq$2 substantive office actions) $.577$, and RCE after final
$.565$. Era inflation is $\approx.03$ (grouped-by-application splits run about
that far above the out-of-time numbers). Reframing the target toward ``another
iteration'' therefore \emph{lowers} predictability, consistent with examiner-to-examiner variation
dominating every outcome-level label.

\paragraph{\protect Scope of the outcome-level result.}
Recovery mutual information is $0.024$ bits $=$ \textbf{2.5\% of $H(Y)$}
(retrieved-only, semantic-delta AUC $.605$), rising to 3.9\% with oracle examiner
gold. Three independent closure attempts are null: prompt optimization of the
verifier ($+0.000$ under a coverage-corrected re-scoring; an earlier $+.085$ was
retracted when parse coverage was found to have collapsed to 8\%), a retriever
retrain on corrected claim-1 data, and three alternative outcome variables
(marginals $-.0009$, $+.0007$, $+.0065$). The recovery figure and the three null closure attempts are
\emph{outcome-level} results only. At the criterion level, a judge that sees the
search record moves marginals by $+.211$, $+.609$, $+.661$, with the standing
caveat that candidate sets contain examiner gold by construction, so a
test on retriever-only candidate sets is still outstanding.

\paragraph{Claim-level disclosure cohort and pipeline.} The disclosure metric is
computed on a nested claim-level cohort drawn from the same 4.69M-application
population as Cohort~A: \textbf{21{,}447 applications, 56{,}111 distinct claims
(59{,}937 scoring rows)}, each claim paired with $K{=}8$ candidate references
(the examiner's cited reference among seven same-CPC fillers) and scored against a
pool of \textbf{479{,}420 prior-art references}, each adjudicated by the LLM
judge. The metric factors the examiner's single act of judgement into four
separately audited stages, none of which ever sees the outcome. \emph{(1)
Decompose}: semantic-delta decomposition (dependent claims resolved to their
added limitation) recovers $61.7\%$ of examiner-cited elements at
token-containment $0.7$ ($80.6\%$ at $0.5$). \emph{(2) Retrieve}: a bge-m3
bi-encoder, an element-to-paragraph cross-encoder (AUC $.844$ against the LLM
judge), and a tuned soft-min search a prior-art index of \textbf{$\sim$20M US
patent documents ($\sim$200M paragraph chunks)}, comprising every granted patent
(1976--2025) and pre-grant publication (2001--2025), from the PatentsView bulk
description-text tables (\texttt{g\_detail\_desc\_text}, \texttt{pg\_detail\_desc\_text};
$\sim$228\,GB raw), paragraph-chunked and BGE-embedded into an IVF\_FLAT FAISS
index. The $479{,}420$ references above are the candidate subset this retrieval
surfaces and the verifier then reads, not the index itself. Retrieval reaches
only $\sim$$20.6\%$ examiner-cite recall. This recall is the binding limit, and it is structural,
since a retrain on the corrected claim-1 pairs did not change it (see the
dependent-claim fix below). \emph{(3) Localize and
verify}: a two-stage Gemma-4-31B localize-then-verify LLM judge, \textbf{$95\%$
real under panel adjudication} and \textbf{$97.6\%$ verbatim-grounded}, still
returns not-disclosed \textbf{$27.3\%$ of the time even when its span lands on the
examiner's exact cited paragraph}; these cases reflect genuine tacit or absent disclosure, not a
pipeline defect. \emph{(4) Aggregate}: a soft-min over element verdicts, which is
saturated (retuning the aggregator gains $\leq +.004$).

\paragraph{\protect The disclosure label.} A claim is
\emph{positive} iff the examiner rejected it on some statutory ground (its cited
gold reference is then planted among seven same-CPC fillers) and \emph{negative}
iff the examiner did not reject it (no gold reference at all). The negative is
therefore inferred only from the examiner's \emph{selective action} (an
un-rejected claim) and never from a verified ``the art was searched and nothing
anticipates''; we observe the \emph{absence} of a citation, not evidence of
novelty. A negative consequently conflates genuine non-anticipation, claims the
examiner never reached or searched, \textbf{extraction misses} ($\sim$45\% of
applications contribute \emph{only} negatives, and a negative-but-carries-a-code
slice was $\sim$76\% false negatives), and claims rejected in a \emph{later}
office action than the one sampled. Two structural limits compound this. First,
the label is \textbf{not a prior-art-reading label}: positives also include
\S112, \S101, and double-patenting rejections (grounds the eight references
cannot explain), and a dense text model separates \emph{those} from negatives
about as well ($.835$ TEST) as it does \S102/\S103 positives ($.840$), with
$\sim$44\% of positives carrying \emph{zero} disclosing references yet receiving a
mean predicted probability of $.72$--$.73$. The column \texttt{rejection\_type}
is thus label-equivalent and is quarantined from every proposer, scoring prompt,
and feature block. Second, the candidate set is \textbf{label-conditional}: a gold
reference is present for $99.66\%$ of positives and $0.00\%$ of negatives, so its
mere presence is nearly the label (\texttt{has\_gold} alone-AUC $.9983$, gold in
the last slot $86.6\%$ of the time). The corpus therefore cannot support any
``identify the disclosing reference among $K$'' evaluation without re-randomising
slot order and holding the examiner gold out.

\paragraph{The dependent-claim fix.} The claim-level evidence pipeline originally
skipped dependent claims and fell back to a single truncated blob, so
\textbf{80\% of units and 74\% of positives were scored as one undifferentiated
string naming claim 1 but containing none of claim 1's limitations}. In
the scale corpus, 78.6\% of claims are dependent with unresolved antecedents. A
claim resolver (preamble-restricted parent finding, chain resolution to depth 25
with cycle guards, multi-dependency resolved to the lowest existing parent)
achieves \textbf{98.76\% clean resolution over 11{,}853 claims}, validated at
$\sim$99\% by 200 blind examples. Unioning the parent chain into the claim text lowered performance. Offline decomposition showed
\textbf{delta (the added limitation) $.55$ $>$ parent $.485$ $>$ union $.519$},
and a rerun of all three confirmed that delta performs best: old blob $.498/.488$ (chance) $<$ union
$.583/.556$ $<$ \textbf{delta-filtered $.605/.599$}. The data-level reason is
that \textbf{85\% of examiner-cited elements are delta content and only 15\% are
parent content}; examiners cite art against the \emph{added} limitation.
Semantic-delta decomposition was made the default. One variant was
\textbf{retired as label leakage}: routing the decomposition mode by the office
action's primary rejection type conditions the metric on part of the outcome.
Two adjacent bugs were also fixed: a claim-sequence off-by-one (one table is
0-indexed and the other 1-indexed, so the ``claim 1'' artifact was claim 2 in
every use, contaminating 28.4\% of retriever positives) and a position leak
(gold evidence was appended in the last slot 80.4\% of the time, making position
alone score $.872$).

\subsubsection{Notice-and-comment rulemaking (verdict, curation, community)}
\label{app:bal:nc}

\paragraph{Two generations.} A v1 \emph{claims} corpus (235{,}294 rows, 18
agencies, unit $=$ one parsed argument inheriting the submission-level label,
exact-text dedup applied \emph{per agency}, random 70/15/15 over claim rows) is
historical and its per-row split is the flaw the current design fixes. All
reported numbers come from the v4/v4.2 long-comment corpus.

\paragraph{v4 build.} Comments come from an 11{,}600{,}599-row table
($\sim$50 agencies); labels from 342 per-agency response files covering 40
agencies and \textbf{205{,}622 response rows (140{,}956 \texttt{llama\_v2} $+$
64{,}666 \texttt{gpt5} LLM extractions)}. Choosing this label universe over the
paraphrase file raised matched-key join coverage from $\sim$24\% to
\textbf{93.7\%} on an 800K-row validation sample. The join key is
\texttt{AGENCY|docket\_id|content\_of\_comment} with content truncated at 200
characters, matched on whitespace-normalized \texttt{content[:120]}. Comments are
kept at $\geq$700 characters (matched rows down to 200, since labels are scarce);
all matching label triples are attached; the balanced view reservoir-samples per
(agency $\times$ has-labels) cell at a \textbf{cap of 4{,}000 per cell}; and a
clean pass strips assembly markers and leaked booleans and adds a collapsed
outcome in \{MADE, CONSIDERED, NONE, OTHER\} covering 98.8\% of values.
\textbf{The split is a docket-level stable hash (70/30)}. Totals:
7{,}921{,}326 long comments, 35{,}533 with full three-axis labels across 38
agencies; the balanced file holds 163{,}363 rows. The response files,
which are the source of every label, are the product of a
docket-scale extraction pipeline over agency final-rule preambles: each
``response to comments'' passage is parsed by LLM extraction into per-comment
response rows carrying the response type and whether a change was made, and
those rows are what the join above matches back to the comment texts. That
pipeline, and the docket-level corpus it produces, is the subject of a
companion manuscript in preparation; this paper uses only its output
table and treats the upstream matching recall as unknown (the caveat on the
responded-or-not label below).

\paragraph{VAT sampling (seed 0).} Sixteen large agencies contribute
\textbf{200 rows per class}; agencies with fewer than 50 per class are skipped;
\textbf{22 agencies survive, 7{,}482 comments}. A \textbf{per-docket cap of 10
per (docket, class)}, relaxed to 20 and then removed only if a pool is short,
prevents mass-comment dockets from dominating the sample. Text must
be $\geq$200 characters after stripping and is truncated at 4{,}000; labels
priority-collapse \textbf{MADE $>$ CONSIDERED $>$ NONE}, and only MADE and NONE
are eligible, so CONSIDERED and OTHER rows are excluded. \textbf{Exact-text dedup}
keys on the lowercased, whitespace-collapsed first 2{,}000 characters, applied
globally across the stream.

\paragraph{Docket grouping and the mass-comment caveat.} All cross-validation is
\texttt{StratifiedGroupKFold(5)} grouped by docket, with out-of-fold AUC and
bootstrap intervals. Near-duplicate form letters are \emph{not} separately
collapsed. We measured how often they cross dockets:
\textbf{only 14 of 685 prefix-duplicate groups span dockets}, so docket grouping
already absorbs the campaigns. Outcome is largely a \emph{rule-level} property:
\textbf{63\% of multi-comment dockets are outcome-pure} and within-docket
out-of-fold concordance is \textbf{$.498$, i.e.\ chance}. A reliability rule
follows: \textbf{$\geq$20 dockets and $\geq$8 dockets per class in-sample, or the
readout is not quoted}. Several per-agency readouts are excluded under this rule (one
agency with 6 total dockets moves $.269\to.421$ across cross-validation seeds).
The 8B LoRA's internal-split AUC of $.893$ is reported to show that this AUC
falls to $.602$ once train and test dockets are disjoint.

\paragraph{Anchor rows.} Three synthetic anchor comments are included in
\emph{every} scoring shard, carrying sentinel fields (agency
\texttt{\_\_ANCHOR}, labels $-1$) so they can never enter modeling: a strong
trade-association letter that disputes an emissions baseline against the agency's
own monitoring data, recomputes a cost-benefit ratio, offers a 36-month phase-in
alternative with attached regulatory text, and identifies a regulatory-impact-analysis
gap; a mid-strength rural-nurse comment; and a weak all-caps rant. They scored
\textbf{$.872$ / $.523$ / $.032$, identical across every shard} of the campaign,
with a stable not-applicable rate of $\sim$.63--.64.

\paragraph{The three labels, and the retired fourth.}
The primary label is outcome-majority: the majority vote of a comment's
collapsed outcome labels, MADE(1) versus NONE(0), with ties dropped
($n{=}7{,}084$ of 7{,}482). The second is agree-versus-disagree: the majority vote
of response type, \{accepted, agree\}(1) versus \{disagree\}(0), under the same
majority helper and so the same ties-dropped rule ($n{=}5{,}046$). The agree-versus-disagree label is the only
label with within-docket, comment-level signal ($.558$). The third is
responded-or-not: matched-with-any-label(1) against genuinely unmatched
$\geq$700-character comments from the same 22 agencies(0) ($n{=}9{,}521$; the
unmatched pool holds 2{,}116 usable rows). The responded-or-not label is not a length artifact:
unmatched comments are longer, and character length alone scores $.422$. However,
``unmatched'' means unmatched in our pipeline, since upstream matching recall is
unknown. We retired a fourth label, the any-MADE union, which counts a comment as
MADE if any matched response said MADE. The any-MADE label is confounded by the number of matched
labels, because each additional matched label is another chance for a MADE response: the label-count feature alone scores
$.695$ on any-MADE against $.589$ on majority vote. The any-MADE label is never used as the primary label.

\paragraph{On the pooled outcome label, every instrument ties the
character-length baseline ($.588$--$.595$ vs length alone $.595$): rule
outcomes are decided at the docket level, so only within-docket contrasts
carry comment-level signal.} $V$ (27 regex features)
$.595$, $A$ (198 rubrics, Gemma-4-31B) $.592$, $V{+}A$ $.593$, dense $.588$,
against \textbf{character length alone $=.595$} and label count $=.590$. One
estimator caveat applies wherever the agree-label number is quoted: that number is
design-conditional ($.612$ under a stratified-group design, $.524$ under the
frozen cross-domain \texttt{GroupKFold} design), and the design choice with the largest effect is
cross-validation stratification, while the outcome label is insensitive to the
same design choices. Similarly, the direction of the prompt-optimization effect on this bank flips
with design, so the pattern ``fidelity up, prediction down'' is design-conditional, not a
settled sign; the headline articulated row remains the \textbf{pre-optimization}
bank.

\paragraph{Organizational-scale covariates.} Headcount comes from a single
consistent federal-workforce snapshot; rulemaking volume from the
regulations.gov bulk download, where \textbf{19 of 21 agencies joined cleanly and
two fell back to comments-only} because the upstream table uses lowercase agency
codes. Headcount predicts only shallow verifiable signal ($\rho=+.496$,
$p=.022$), whereas \textbf{lifetime docket throughput predicts deep verifier
signal ($\rho=+.635$, $p=.003$, $n=19$)}. Two confound checks were run:
partialling out sampled-docket diversity (itself $\rho=.891$ with total dockets)
leaves the deep-verifier correlation at $+.576$ ($p=.010$) while the dense
correlation falls to $-.062$, so the dense correlation was entirely an artifact of sampled-docket diversity; excluding the largest
agency leaves $+.571$. ``Dockets'' is a lifetime total, not a rate.

\subsubsection{Legal outcomes across twelve domains (appendix-only)}
\label{app:legal}
\label{app:bal:legal}

This corpus does not fit the benchmark's preference mapping, because a court's
ruling on a dispute between parties is an application of law to facts, not a
preference expressed about a creative product. The legal corpus is therefore absent
from the main body and from every cross-field figure. This subsection records the corpus's construction, which established several of the practices the preference corpora
use (facts-only extraction, leak-judge iteration, posture
stratification, entity-grouped splits); the corpus is also a usable
resource in its own right.

\paragraph{What ``2{,}000 per domain'' refers to.} The twelve domains are
ERISA long-term disability, DOL, trademark, Title VII, Social Security
disability, TTAB likelihood-of-confusion, FLSA, MSPB, TTAB ex parte, CAVC review,
PTAB, and NLRB. The verifiable and articulated models are fit on the \emph{same
600-case balanced sample} per domain under \texttt{StratifiedKFold(5)}; the
\textbf{2{,}000 balanced rows per domain} is the \emph{dense} model's fine-tuning
sample (Llama-3.1-8B with LoRA, single 80/20 stratified split, scrubbed text),
and one domain with $n{=}162$ falls below the 200-row threshold and has no dense
number. A zero-shot model consumes the same 600 cases plus per-metric definitions
and scores oriented by the sign of each metric's correlation with the label.
Two operational cautions are recorded. First, the FP8 judge has a $\pm.02$ AUC
non-determinism floor, so gaps under $\sim$.03 are within noise. Second, the runner
uses random stratified folds for all domains and never reads a pool's stored
split field. The random folds are harmless for the metric-based models, whose features carry no party
identity, but they mean that \emph{court-level grouping in these two models is not
enforced by the runner}. Two domains carry an inverted label convention
($y{=}1$ is government/agency affirmance), which caused a bug: a
moving-party prompt read the label direction backwards until the prompt was corrected.

\paragraph{\protect Facts-only construction, and the v1 pipeline.} The original
broad pipeline ($\sim$91K rows, $\sim.818$ dense AUC) \textbf{leaked the
outcome}: the extractor carried disposition language (verdicts, awards, back-pay,
fee and post-trial recitals) into the facts field, and
\textbf{17--22\% of opinions per slice were post-disposition opinions} (fee
petitions, Rule 50/59/60 motions, settlements), whose subject \emph{is} the
result. Its merit filter also selected \emph{for} disposition keywords. A
companion file preserves the ten most leak-heavy extracted fact blocks per slice
with per-token counts; these blocks motivated the category-based exclusion
prompt. 
In the rebuilt pipeline we slice from the raw pool of $\sim$1.16M
district-court opinions using per-slice statutory anchors plus posture flags, then
stage-filter (keeping merits, dropping post-disposition stages) and dedup,
which gives clean pools of 54{,}147 / 42{,}935 / 11{,}419 for the three district
slices. Extraction is de-leaked and runs on an offline 70B model: facts are the
antecedent narrative only, under category-based exclusion of all at-or-after-ruling
content, converged over six prompt rounds, with a regex post-filter. Settlements
map to a procedural class and are dropped. We build two scales, a balanced
2{,}000-per-label file for controlled comparison and a full pool at natural
balance.

\paragraph{Leak testing.} A strict independent leak-judge (a separate 70B call
asked to name the winner from the facts alone) was iterated until it reported \textbf{0
genuine leaks} across all slices. A TF-IDF check then caught \emph{aggregate}
residual leaks that the per-sample judge had missed (templated remand statements in one
slice; settlement language and prayer-for-relief boilerplate in another), which
were fixed at the prompt and filter rather than post hoc.

\paragraph{\protect Confounds.} Court identity scores $.556$; facts length,
paragraph count and digit count all score $\leq.556$, which is immaterial against $.72$--$.78$
floors. Only date cohort is spurious (lift $\approx-.13$), and stripping years
moves AUC by $\leq.004$. A templated remand phrase appears in \textbf{96\% of
\emph{both} won and lost} cases, and stripping all remand language costs
$\leq.017$, so it is boilerplate rather than framing leakage. Label permutation
gives $.50$. The metadata check found \textbf{pro se status as the second
material confound} (17.5\% share in one slice, with a 3.9\% versus 14\% win rate,
and detectable from the facts at $.83$ even with the phrase stripped) plus
enforcer-agency plaintiffs (real but $\sim$1\% of rows), and \textbf{both are
dropped from the canonical pool}. Posture-only AUC is $\sim.61$, and posture is handled
two ways: stratified within-posture lexical scoring costs only $-.018$/$-.008$
versus pooled, and adding posture as a covariate adds $+.014$/$+.017$; posture
also enters the articulated layer as an explicit, i.e.\ articulable, feature. A
tone probe addresses the standard critique that judge-written facts are
endogenous: framing-only AUC is $.627$/$.603$/$.551$, a \emph{minority} of the
dense ceiling and below the articulated layer in every slice, and the framing signal localizes to
judge authorship: $.718$/$.656$ at trial (judge-written post-decision findings)
versus $.573$/$.538$ at motion-to-dismiss (plaintiff-authored complaints) and
$.601$/$.527$ at summary judgment. The bulk of the canonical pool therefore carries
only a $.53$--$.60$ framing leak, and dense-minus-tone is reported as the
framing-robust tacit floor. A temporal split (train early, test late) gives
$.750$/$.722$/$.700$; the one drop is real secular drift (plaintiff win rate fell
from $.22$ to $.10$ since the 1980s), not leakage. The \textbf{canonical pool} is
therefore counseled-only, enforcer-dropped, and \textbf{50/50 within posture},
giving \textbf{6{,}410 / 3{,}702 / 52{,}560}, with posture-only AUC on the canonical pool falling
to $.482$/$.471$.

\paragraph{Label audit and the extraction-created leak.} Sixty cases were read by
hand (twenty per slice, fixed seed). One slice is clean (19/20 labels, 20/20 fact
fidelity). A second shows \textbf{$\sim$20\% asymmetric label noise} from
posture misreads (granted versus denied, whose motion, stays mistaken for
outcomes), with an observed $.70$ implying a true $\approx.83$. The third
has a defect no raw-text check would catch: \textbf{5 of 20 fact blocks were
hallucinated} from a template on \emph{fact-free} consent-remand forms, and four
of the five carried the positive label, so the hallucinated blocks are an \emph{extraction-created} leak.
The leak was quantified and removed: \textbf{2{,}639 of 52{,}841 rows (5.0\%, positive
rate $.94$)} originating from very short raw documents or consent sources were
dropped, moving lexical $.740\to.719$, articulated $.641\to.630$, and dense
$.784\to.782$. The leak had inflated the \emph{shallow} layers most, so removing
it \textbf{widens} the tacit gap ($.044\to.063$). A reusable guard (skip fact-free
source documents) was exported for new domains.

\paragraph{Relabelling under posture awareness.} One slice was fully relabelled
(all 11{,}177 rows, 100\% parse, 99.9\% high confidence) into a five-class scheme
(defendant win 3{,}586 / mixed 2{,}982 / plaintiff survives 2{,}071 / plaintiff
win 1{,}494 / procedural 1{,}042), revealing \textbf{1{,}139 false positives and
164 direction flips} in the old labels. Under the adopted scenario
(plaintiff-survives counts as 1; mixed and procedural excluded) the usable pool
is \textbf{7{,}140 at positive rate $.499$} with stable-hash splits. On the
identical split the whole ladder rises and the gap \emph{widens}: lexical
$.724\to.742$, dense $.771\to.797$, tacit gap $.047\to.055$. Label noise
was therefore \emph{masking} tacit signal rather than creating it.

\paragraph{Ex-ante tribunal domains.} These share a unified schema in which every
input document is dated \textbf{strictly before} the decision (asserted, with
violations dropped and logged), the raw label is never overwritten by its binary
mapping, splits are \textbf{stable hashes of the case identifier}, and every
record carries its funnel stage so strata are never silently pooled. A uniform
scrub removes case numbers, filing and decision dates, four-digit years, and bare
integer runs while \textbf{preserving statutory references}, right-censors recent
filings, and then splits by \textbf{the dominant repeated entity} (judge,
respondent, expert, petitioner, or owner). The headline for every such pool is
the \emph{group}-split number, and the gap to the naive number is reported as the
memorization tax: for one domain $.759\to.625$ (a 0.13 petitioner-memorization
tax on top of era self-identification), and for another a remand-round temporal
leak of $.825\to.686$. Construction is otherwise domain-specific; two examples
are a record-linkage domain with no join key, where cosine
matching within same-date candidate sets was gated by a margin threshold and a
rare-token signature, and where a \textbf{hand audit by margin tier} showed the
lowest tier was $\sim$90\% false matches, so the threshold was raised and
16.8\% of pairs were dropped; and a registration domain whose negatives were restricted to one
specific abandonment code (excluding a superficially similar code whose cases had
already \emph{passed} examination), balanced \textbf{39{,}968 / 39{,}968 across
57 year-by-basis strata} with owner-grouped splits, after stripping a filing-basis
annotation whose positive rate was exactly $0.00$.

\subsection{Academia}
\label{app:bal:academia}

The texts are scientific abstracts (with titles). The \emph{verdict} is a
program committee's accept/reject decision on a submission. \emph{Curation}
is promotion of an accepted paper above its peers: an oral or spotlight
slot, or a best-paper award. The \emph{community} signal is the citation
percentile a paper reaches within its venue-year, which aggregates thousands of individual
decisions to build on one paper rather than another. Two corpora follow:
the ICLR peer-review corpus carrying all three labels
(\S\ref{app:bal:iclr}) and the cross-venue best-paper award corpus
(\S\ref{app:bal:bestpapers}).

\subsubsection{ICLR peer review (verdict, curation, community)}
\label{app:bal:iclr}

\paragraph{Venue control.} On the pooled multi-venue corpus \textbf{venue family
alone scores $.819$}, because base rates span \textbf{$.24$ (ICLR) to $.97$
(ICML)}: NeurIPS and ICML expose only $\sim$5\% of rejections (opt-in) while
ICLR posts every submission on OpenReview. The primary corpus is therefore
\textbf{ICLR-only and year-controlled}, with pooled numbers retained solely as a
confound reference. The control costs $\sim$.02 ($.679\to.658$) but leaves the
articulated lift identical ($+.056$ pooled versus $+.057$ ICLR-only), so the lift
is not a venue artifact. The earlier
$\sim$.77 dense ceiling is pooled and venue-confounded and must never be compared
with ICLR-only numbers; the 
grouped value is $.751$.

\paragraph{One corpus of abstracts carries three labels: accept/reject
(verdict), oral/spotlight promotion (curation), and within-venue-year
citation percentile (community). Channel comparisons in this field therefore
hold the texts fixed.} \emph{Verdict} is accept/reject from the canonical binary
splits, filtered to ICLR. \emph{Curation} is oral or spotlight versus poster,
derived from the decision-tier field with an explicit alias table (oral $\gets$
\{Oral, Talk, notable-top-5\%, OralPoster\}; spotlight $\gets$ \{Spotlight,
notable-top-25\%, spotlight poster\}), \textbf{excluding NeurIPS 2022 which
published no tiers}, rejections, and missing tiers; abstracts are required to be
\textbf{between 200 and 6{,}000 characters}. Because the promotion fraction is
roughly fixed, this label is \emph{budget-bound} and is always stratified and
evaluated within (venue, year). \emph{Revealed} is the citation percentile:
top quartile versus bottom quartile of citation count \emph{within venue-year},
with the middle retained but unlabelled.

\paragraph{Membership and the title join.} Venue membership for the revealed leg
is defined by DBLP, not by the citation database, because source-based
pagination is provably partial (one venue-year returns 650 works against 1{,}428
accepted papers, and near-zero after 2022 since these venues issue no DOIs).
Titles are then matched to a citation source under a strict triple gate:
\textbf{exact normalized-title match or difflib ratio $\geq0.92$, \emph{and}
year within $\pm1$, \emph{and} venue-token overlap}. The search engine's own
match score is recorded for auditing but deliberately not used as the gate,
because it is an absolute BM25-like score that scales with title length. A
\textbf{single-source-per-cell rule} applies: since the percentile is computed
within venue-year, mixing two citation sources inside one cell would corrupt the percentile,
so partially covered cells were left single-source rather than partially filled.
One failure is documented: a publisher's official proceedings records for one venue
carry $\sim$0 citations because the actual citation counts are attached to the preprint twins, and
the default prefer-published tie-break produced wrong labels until the join was
redone preferring citations; a cell that could not be repaired within budget had
its labels blanked by a $<$50\%-coverage guard. In one cell the within-cell floor was $.989$ because the negative
quartile consisted of abstract-less stubs (median 73 characters, i.e.\ title
only) while the positive quartile had real abstracts (median 1{,}413
characters), so the classifier separated on abstract \emph{presence}, not
content. A full single-source re-fetch repaired that cell ($.989\to.842$, in line with
neighbouring years).

\paragraph{Title normalization and title-grouped stable-hash splits.} All three
legs are keyed on a normalized title: lowercase, non-alphanumerics to spaces,
whitespace collapsed, truncated at 200 characters, with the title itself pulled
out of the text field by an anchored regex that tolerates a leading
\texttt{TITLE:} and stops at a newline or \texttt{ABSTRACT:}. The split is a
\textbf{salted stable hash of that key},
$\mathrm{md5}(\texttt{"split::"}+\text{key})\bmod 10 \to$ 0--7 train, 8 eval,
9 test. Because the \emph{same key} is used across legs, a paper shared between
the curation and revealed legs receives the \emph{same} split in both, so the
matched cross-leg comparison carries no leakage; and because the split is a hash rather
than a shuffle, adding rows never reshuffles it. \textbf{4{,}296 normalized titles
carry both a curation and a revealed label}, which is the shared set used for the matched comparison
(2{,}202 of them are scored). Task sizes: verdict 28{,}402 (positive rate
$.394$), curation 11{,}021 ($.183$), revealed 4{,}663 ($.493$).

\paragraph{Each abstract is scored by the judged bank exactly once, in one
pooled union of the three legs' rows, and the label is attached afterwards
per leg, so no leg's scores can be tuned to its own label.} Because the articulated and verifiable instruments are
label-independent, each abstract is scored \emph{once} and the label attached per
rung. The union is built with \texttt{random.seed(0)} as: all revealed rows (the
smallest, balanced leg, which contains the shared set); all curation positives
plus a 4{,}000-row negative sample; and 3{,}000 verdict rejections plus up to
3{,}000 verdict accepts \textbf{whose normalized title is not already in the
curation set}, which guards against counting a paper twice. This yields
\textbf{14{,}307 unique abstracts}, scored by Gemma-4-31B against a 154-rubric
bank whose system prompt ends ``Judge the paper's quality, not whether it will be
accepted.'' Aggregation is the frozen design: not-applicable values imputed to
the column median over non-missing values, all-missing columns dropped, a
\textbf{degeneracy guard removing any column with fewer than 5 off-modal values
or zero variance}, standardization plus logistic regression, and out-of-fold AUC
under \texttt{GroupKFold(5)} grouped by normalized title, with $V$, $A$ and
$V{+}A$ fit on identical rows.

\subsubsection{Best-paper awards (expert curation)}
\label{app:bal:bestpapers}

\paragraph{\protect The first build.} Awards are scraped from a maintained
community list (one table per year, venue blocks read off row spans),
\textbf{1{,}819 awards, 32 venues, 1996--2025}, with the count verified against
the sum of venue-block row spans. Joining them to a bibliographic database at a
$\geq0.85$ normalized-title threshold matched 82.7\%, but the \emph{negative}
pool was incomplete: that database's venue pools are broken for exactly the
computer-science flagships (several at 0\% and many under 15\% recovery, with two
resolving to no usable source at all), leaving only \textbf{311 usable awards in
a resolved venue-year}. The build also documents that journal twins must be rejected
by regex, and that one proceedings-journal mixes three conferences and can never
be accepted for any one of them.

\paragraph{DBLP venue-year enumeration.} Membership is instead enumerated from
the static DBLP dump (the search API rate-limits within minutes). A venue-year's
pool is every proceedings entry whose cross-reference matches an
\textbf{anchored main-track regex} (so suffixed workshop, companion, and
datasets-and-benchmarks volumes are excluded). Six venues that migrated to
journal-published proceedings are recovered by explicit volume arithmetic (for
example one venue's volume equals year minus 2016; another year minus 1981).
Coverage: \textbf{664 of 688 venue-years, 198{,}027 pool papers, and 1{,}773 of
1{,}819 awards now falling in an enumerable venue-year}, a 5.7$\times$
recovery over the 311 usable awards of the first build. 1{,}743 awards match a DBLP key (1{,}713
exact plus 30 fuzzy at $\geq0.92$). Pool-membership spot checks across five
venue-years found \textbf{100\% of awards present in their own venue-year pool};
the 46 unrecovered awards are pre-2000 cross-reference naming changes plus one
venue-year not yet in the dump.

\paragraph{Semantic Scholar join validation.} Each DBLP key is matched by title
against Semantic Scholar and accepted \textbf{only if all three hold}: exact
normalized-title match or difflib ratio $\geq0.92$; year within $\pm1$; and
venue-token plausibility (DBLP being authoritative, an empty or generic remote
venue is not a rejection). The service's own match score is recorded for auditing
but \emph{not} used as the gate, because it is an absolute BM25-like quantity
that scales with title length. Hand validation of 30 award matches:
\textbf{29/29 accepted matches correct (0\% mismatch), and the single rejection
correctly caught a wrong-paper match}. Accept rate $\sim$95\%; the full
fetch covers 27{,}025 papers (awards plus at most 40 stably sampled non-awards
per venue-year).

\paragraph{Dataset build and presentation normalization.} $y{=}1$ is an award;
$y{=}0$ is the same venue-year DBLP pool minus awards. \textbf{An abstract is
required from the same source on \emph{both} sides, with no backfill from the
other database}, specifically to avoid a cross-source abstract-presence leak;
the requirement skews the sample toward recent venue-years with high abstract coverage. A cross-source abstract-presence leak is the same
failure mode that produced a $.989$ within-cell floor elsewhere in the academia
family. $X$ is
\texttt{TITLE}$\backslash$n$\backslash$n\texttt{ABSTRACT}, presentation-normalized
by NFKC, curly to straight quotes, en/em/minus dashes to hyphen, ellipsis to
three periods, and whitespace collapse, with raw fields retained. The split is
$\mathrm{md5}(\texttt{dblp\_key})\bmod10 \to$ 0--7 train, 8 eval, 9 test.

\paragraph{Citation deconfounding.} Citation count is \emph{post-treatment}
(committees pick before most citations accrue), so four numbers are reported
rather than one: (a) citation-only AUC, the confound itself; (b) the raw text
floor, which is citation- and venue-entangled; (c) the
\textbf{citation-decile-matched text floor}, matching positives and negatives on
citation deciles within venue-year in the same style as the math.StackExchange
propensity design; and (d) the within-venue-year stratified AUC, which is the
right frame because awards are budget-bound at roughly 1--3 per venue-year. On
the full build these read: \textbf{(a) $.635$, (b) $.728$ (venue-confounded),
(c) $.603$ (the de-confounded floor), and (d) $.571$ mean /
$.500$ median}, with a verifiable-feature bank at $.681$ pooled and $.564$
within venue-year, an embedding upper bound of $.751$ pooled and $.591$
within-venue-year, and length-only at $.443$ (so there is no length confound).
The final corpus is \textbf{11{,}613 rows carrying 960 usable awards} across 605
venue-years (325 with both classes) and 32 venues including all recovered
flagships, at a label balance of 8.3\%. Award committees do
leave a text-detectable signal beyond citations ($.603 >$ chance after
citation matching), but the signal is modest, which is consistent with expert curation being a
weak-$V$, moderate-$A$ task.

\subsection{Journalism}
\label{app:bal:journalism}

The texts are news items and the material that becomes news: press releases
paired with the coverage they did or did not receive, article headlines in
their homepage context, the articles behind most-read lists and outlet
tweets. The \emph{verdict} is an editor's decision to cover a press release.
\emph{Curation} is placement: which articles an editor puts at the top of
the homepage. The \emph{community} signal is aggregate reader behavior:
most-read lists and engagement with outlet tweets. Three corpora follow:
press releases (\S\ref{app:bal:pressrelease}), homepage newsworthiness
(\S\ref{app:bal:homepage}), and the reader-behavior tasks
(\S\ref{app:bal:newscommunity}).

\subsubsection{Press releases and editorial pickup (expert verdict)}
\label{app:bal:pressrelease}

\paragraph{Pair construction.} Two independent scrapes are merged, with entity
dedup by URL $\to$ normalized URL $\to$ \textbf{SHA1 of the body text} as a
fallback, and with SURT-to-standard URL decoding performed first (otherwise
CommonCrawl entries never match standard URLs). Pair filters: press-release
length 200--12{,}000 characters, article length 300--15{,}000, and
\textbf{drop any pair whose article and release are identical or overlap at
$\geq.85$ token Jaccard}. The canonical v1 table is \textbf{128{,}131 rows, 42.0\%
positive}, over 17 tracked coverage outlets.

\paragraph{The $k{=}1$ versus $k\geq3$ label choice.} The label is $y=1$ iff at
least $k$ \emph{distinct} outlets covered the release. At $k\geq1$ the label is
\textbf{88\% single-outlet noise: 28{,}735 of 32{,}789 covered releases had
exactly one outlet}, dominated by wire republication and single-editor pickup.
Dense grouped AUC by threshold: $k\geq1$ $.584$ (32{,}789 positives, 1:1.2);
$k\geq2$ $.675$ (4{,}054, 1:17); \textbf{$k\geq3$ $.705$ (1{,}478, 1:27)};
$k\geq4$ $\sim.70$. The lift survives within-topic control, so it is not the
publisher or topic confound returning, and \textbf{$k\geq3$ was adopted}. The
consequence for this paper's ladder is qualitative: at $k{=}1$ it reads
$V$ $.554$ / $A$ $.543$ / dense $.584$ (no gap at all), while at $k\geq3$ it
reads $V$ $.628 < A$ $.648 <$ dense $.679$ on the build of that time (the $.705$ above is the ungrouped
threshold-sweep value). On the rebuilt dense-standard frame that Table~\ref{tab:master-results} reports, grouped by company, the ladder is $V$ $.653$, $VA$ $.681$ ($.744$ with the mined criteria), $T$ $.774$ and $VAT$ $.771$. The class imbalance at $k\geq3$ (1:27)
is handled by stratified grouped cross-validation and threshold-free AUC.

\paragraph{Confound profile, measured before filtering.} Publisher identity
$.673$ (leave-one-out company coverage rate over 11{,}074 companies) $>$ topic
$.610$ (25-topic model, positive rate $.048$ for Beauty to $.592$ for
Government) $>$ clean-body length $.547$ (positives $\sim1.8\times$ longer,
median 2{,}414 versus 1{,}357 characters) $>$ language $.505$ $>$
empty-extraction $.459$ (49{,}248 rows, 38.4\%, skewing negative) $>$ year
$.471$.

\paragraph{The 72K deconfounding filter.} The filter is
\emph{clean body only (drop empty extractions), English only, length
200--12{,}000 characters}. This retains \textbf{72{,}315 rows (56.4\%), 45.3\%
positive}, and deliberately leaves the topic distribution intact so that topic
balancing happens per fold at modeling time rather than being fixed in the file.

\paragraph{Deconfounding ladder and signal decomposition.} Dense on a random
split $\sim.710$ $\to$ TF-IDF raw random $.675$ $\to$ TF-IDF raw
\emph{publisher-grouped} $.605$ $\to$ deconfounded random $.640$ $\to$
\textbf{deconfounded publisher-grouped $.584$ (the ceiling of this TF-IDF ladder, not of the full model)} $\to$ plus
topic-balanced $.546$ $\to$ plus boilerplate-scrubbed $.545$. Decomposition:
publisher memorization $+.056$, topic selection $+.038$, residual within-topic
cross-publisher craft $\approx+.046$ over chance. Token scrubbing does not raise the grouped AUC: wire boilerplate (\texttt{prnewswire} present in 8.1\% of rows with
$P(\text{pos}\mid\text{present})=.505$ versus $.412$; ``news provided by''; ``for
immediate release''), HTML residue and ticker tokens were all identified, but
removing every one of them moved grouped AUC only $.546\to.545$, because the
signal is spread across thousands of entity tokens that the company-grouped split
already absorbs.

\paragraph{Company grouping.} A stable split
$\mathrm{sha1}(\texttt{company})\bmod100 \to$ 80/10/10 with \textbf{0 of 6{,}967
companies straddling a split}; all evaluation is company-grouped and
within-topic, never random-split.

\paragraph{The publisher-identity fix on the outlet side.} Trust rules: staff
editorial plus reputable wire (AP, Reuters, Bloomberg, AFP, Dow Jones;
explicitly \emph{not} PR Newswire) plus vetted contributor programs count as
trusted; PR-republished, sponsored, automated and non-article content is
excluded. Every outlet's contributor program counts as vetted \emph{except one},
whose vetting policy postdates the 2014--23 data window, excluding 4{,}759
contributors. Non-editorial shares run as high as 88\%, 65\% and $\sim$60\% at
three financial outlets. Four outlets were dropped as data artifacts: a paid wire (91\% Cision), a broadcaster whose crawl was
99.9\% shop and RSS pages (3 real articles in 3{,}000), a newspaper whose hits
were 92\% its ticketing subdomain, and an insurer's corporate pages. Raising the
parse window from 600 to 4{,}000 characters raised the parse rates of two major outlets from 5.7\%
to 73.7\% and from 1.7\% to 70.1\%. Final release-level outcome:
\textbf{retained 38{,}651 (71.9\%), confirmed-junk lost 6{,}373 (11.9\%, flipped
to negative), unknown-only 7{,}793 (14.5\%)}. This outcome is summarized as ``$\sim$72\% kept,
$\sim$12\% hard loss'', never as the deterministic ``keep 17\%''. An apparent preference of financial outlets for
percentage-dense releases survives topic control but \emph{not} editorial-integrity
control, since three of the top five percentage-preferring outlets are heavily
automated or contributor-driven; on the clean set numerical density is at chance
($.505$--$.508$).

\paragraph{Known corruption.} The file
\texttt{press\_release\_modeling\_dataset\_clean.csv.gz} is gzip-truncated:
only 41{,}607 of 128{,}131 rows are readable, and because the file is
label-sorted that readable prefix is \textbf{81\% positive}. It is preserved
rather than deleted, and the modeling table was reconstructed from intact
id-keyed sources.

\subsubsection{Homepage newsworthiness (expert curation)}
\label{app:bal:homepage}

\paragraph{Source and pipeline.} Archived newspaper homepage snapshots supply,
per capture, a hyperlink dump with bounding boxes and an accessibility tree. The
pipeline filters to visible links; classifies article versus site furniture with
a logistic model trained on published article/furniture labels; deduplicates by
URL path; enriches each anchor to the longest accessibility link-name that
\emph{contains} it; restricts to the above-the-fold zone (top $<0.30\times$ page
height); and \textbf{drops any snapshot with fewer than four in-zone articles}.

\paragraph{The label.} $y{=}1$ iff the article sits
in the top half of that top-30\% zone. The label is not click-through, dwell time or
shares; an earlier ``engagement-driven'' framing was an error and has been corrected. The norm families are news values and editorial priority, not click-driving headline craft, and position is part of the label rather than a nuisance variable.

\paragraph{Deconfounding steps.}
The original build scored $.989$. We dropped source and date markers from the text and balanced 50/50 within outlet, which took the AUC to $.963$; outlet identity had been a base-rate shortcut, one outlet 34\% top-half against another at 73\%. We removed the target headline from the context field and shuffled the remaining anchors, which took it to $.984$: in an ordered context string the target's character position leaked the label, with top articles at $\sim$.25 of the string and bottom articles at $\sim$.73. We then topic-balanced within a 50-topic LDA, which took it to $\sim.753$. Breaking news and disasters sit at the top and lifestyle at the bottom, and topic balancing is the step that moves the AUC.

\paragraph{Snapshot grouping.} The group key is the first 16 hex characters of an
md5 over the \emph{sorted set} of 50-character headline prefixes in the snapshot,
giving \textbf{21{,}951 unique snapshots at a mean of 8.4 rows each} over
183{,}708 rows. Snapshot grouping is required because every row's context field holds the other
headlines from the same snapshot and \textbf{99\% of focal headlines also appear
in some other row's context}: under a row-level split the model learns the news
cycle from training. The leakage was found because an ungrouped sweep reached $.86$ against an independent deconfounded estimate of $.75$. We treat a training AUC materially above the deconfounded estimate as leakage until shown otherwise.

\paragraph{The v2--v9 text-cleaning chain.} An audit found $\sim$40\% of
headlines and $\sim$80\% of context fields polluted (image credits used as
headlines, JavaScript fragments, ``4 min read'' markers, and bylines in over 60\%
of rows), traceable to two bugs \emph{upstream of the text and neither in the
label}: the article/furniture classifier under-filters (so image-credit links and
UI modules pass as articles and, when top-positioned, receive the positive label
with junk text), and the longest-containing enrichment appends bylines and
read-times. Stages: v2 drops 21{,}234 rows (promotional 5{,}437, image credit
7{,}553, non-English 4{,}754, short 2{,}862, JavaScript 628) leaving 162{,}474;
v4 tightens a language-density threshold from $.18$ to $.14$ ($\to$ 157{,}552);
v5 adds seeded language detection ($\to$ 158{,}666); v7 makes credit detection
non-length-gated ($\to$ 147{,}873); v8 gates language detection behind a broad
pre-filter ($\to$ \textbf{133{,}130 at 50/50}); and \textbf{v9's key fix handles
read-times concatenated onto the headline (``Years3 min read'') by dropping the
regex's leading word boundary}, which takes the read-time residual from $.1105$ to
\textbf{$.0000$}. Post-clean confounder audit: outlet $.507$, byline $.510$, date
$.502$, position $.50$.

\paragraph{What the cleaning changed.} Across the eight cleaning
stages the verifiable model moves only $.5535\to.572$ and the full model
$.6208\to.630$, with the articulated model approximately equal to the
verifiable model at every stage. The cleaning was therefore
necessary for validity (the removed text included label-correlated
artifacts) but recovered almost no predictive signal, and the corpus's low
ceilings cannot be attributed to residual noise. Two measurements locate the recoverable signal. First, the global readout understates
the task: the \emph{within-snapshot} ceiling (a joint
pairwise-difference AUC over the $71{,}693$ pairs of articles that shared a
homepage) is $\sim$$.65$ against the global $.552$, meaning that editors'
relative placements within one front page are substantially more predictable
than pooled row-level scoring suggests, while no single crude feature
(length, topic words, bylines) separates the pairs. Second, the ceiling is
an instrument limit rather than an articulability limit: human annotators,
asked to explain placements, articulate $\sim$69--90\% of them with classic
news values (timeliness, impact, proximity, conflict), so the placement
signal is articulable in principle even where our judged bank does not yet
reach it, and for this reason the homepage task is flagged as an instrument-floor task
in \S\ref{sec:limitations}.

\subsubsection{BBC most-read, tweet engagement, and link-aggregator votes (community)}
\label{app:bal:newscommunity}

Three corpora capture aggregate reader preference over news text.

\paragraph{BBC most-read.} $y{=}1$ is a headline appearing on the outlet's
top-ten most-read list; $y{=}0$ is a same-day headline from the same
publisher that did not, with both classes drawn through the same archive
pipeline and the same parser family so no cross-source typography leak is
possible. Video clips are dropped from the controls (a clip caption is a
different register), and dedup keeps the earliest occurrence under a
``most-read wins'' rule. The full task holds $50{,}761$ rows
($22{,}357$ / $28{,}404$) over $2{,}251$ capture days. The earliest slice has one measured asymmetry: the dedicated most-read page
shows full $\sim$50-character headlines while the homepage index shows
$\sim$35-character truncated teasers, correlating headline length with the
label at $+.65$, so for that slice the trustworthy AUC is the
length-decile-matched one ($.60$--$.64$), never the raw value; later era
slices draw both classes from the same homepage capture and do not have the
asymmetry. BBC most-read is the task whose criteria-mining campaign ran to
convergence, which licenses the calibrated residual shown in
Figure~\ref{fig:vat-community}.

\paragraph{Tweet engagement.} From a raw pool of $52{,}707$ article-share
tweets by news outlets, the labeled task holds $46{,}456$ rows under a
within-(outlet $\times$ day) median split of engagement, over $784$
outlet-day groups (scored task $31{,}129$ rows). The within-group design performs
the deconfounding: outlet identity, follower base, and the day's news cycle
are constant within a group, so the contrast is between what one outlet's
audience engaged with among the articles that outlet tweeted that day. Splits
are grouped by outlet-day; the fused instrument is reported under the
bakeoff protocol (Table~\ref{tab:master-results}).

\paragraph{Link-aggregator votes.} $y$ is the score percentile
\emph{within} (subreddit $\times$ UTC day), positive at $\geq.75$ and negative at
$\leq.25$ with the middle dropped and strata under 20 posts dropped; posts under
three days old are dropped as immature; dedup on normalized title \emph{or} URL
within a rolling three-day window removed \textbf{132{,}603 rows}; and propensity
balancing on confounds only (hour, weekday, log post age, top-100 domain
one-hots, log author post count, subreddit) took the confound-only AUC from
\textbf{$.844$ to $.508$}, leaving \textbf{49{,}112 rows at exactly 50/50 within
deciles}. Domain identity is deliberately \emph{in} the propensity model and
\emph{out} of the feature set.

\subsection{Creative writing}
\label{app:bal:cw}

The texts are stories: flash fiction at literary magazines, serialized
novels on RoyalRoad, prompted short stories on r/WritingPrompts, submission
excerpts on Kindle Scout. This field's \emph{verdicts} are heterogeneous by
design, and the differences are part of what the tasks measure: a
moderator's removal decision enforces posted community norms; a publisher's
acceptance is expert adjudication on an excerpt; RoyalRoad's market pickup
is not a gatekeeper's decision but a market outcome revealed by the
platform (\S\ref{app:bal:royalroad}). \emph{Curation} is a guest editor's
annual list, a community magazine's ranked picks, or contest badges. The
\emph{community} signal is story upvotes and follower counts. The corpora:
r/WritingPrompts upvotes (\S\ref{app:bal:wp}), the RoyalRoad triple
(\S\ref{app:bal:royalroad}), Kindle Scout (\S\ref{app:bal:kindlescout}),
Wigleaf (\S\ref{app:bal:wigleaf}), and the long-form macro-norm moderation
tasks, which share their build with the humor moderation tasks and are
documented with them (\S\ref{app:bal:moderation}). Two curation corpora are
collected but not yet instrumented (Reedsy contest badges, live at
$66{,}189$ of $\sim$$115{,}670$ pages; r/WritingPrompts awards and
r/bestofWP picks) and appear in the inventory tables only.

\subsubsection{r/WritingPrompts story upvotes (community)}
\label{app:bal:wp}

\paragraph{Build.} Stories are pulled from a Reddit archive with
bots removed, top-level comments only, and a $\geq200$-character body; joined to
their prompt; and exact-deduplicated on the full text string, giving a
$\sim$993K corpus.
$\texttt{prompt\_id}=\mathrm{md5}(\text{prompt})$
We set $y{=}1$ when a story's score is at least 10, with a target of 50{,}000 per class, and balanced length within the character buckets $[0,500,1000,2000,3000,5000,10000,\infty)$ at seed 42. High-scoring stories had been $1.30\times$ longer on average; after balancing the ratio is $1.004\times$, and $1.001\times$ after the clean rebuild. Near-duplicates were removed with MinHash LSH on character 9-grams (128 permutations, threshold $0.5$), which cut an earlier 87K corpus to 50K, a 42.6\% duplicate rate; the rebuilt pull carries only 243 exact duplicates (0.24\%). Low-scoring ``stories'' were often the moderator removal template: the trigram \texttt{reddit com r} appears 10{,}834 times in $y{=}0$ against 2{,}126 in $y{=}1$. We dropped rows matching \emph{this submission has been removed} (1{,}957 rows, all $y{=}0$, none in $y{=}1$) and rebalanced per length bucket, which gives 96{,}080 rows at 48{,}040 per class over 70{,}453 unique prompts.

15{,}391 prompts carry $\geq2$ stories, 42.1\% of rows, and within those the mutual information between prompt and label is $0.318$ nats, 48\% of $H(y)$. Under a row-level split the model can memorize per-prompt scoring and then read the prompt, which is part of the input, so we group on
$\texttt{prompt\_id}=\mathrm{md5}(\text{prompt})$, giving 56{,}361 / 7{,}046 / 7{,}046 prompt-groups with zero overlap. Two problems remain unfixed: an edit-marker imbalance (1{,}439 in $y{=}1$ against 866 in $y{=}0$, since high-scoring posts attract more author edits) and author identity, which the dataset does not track and so cannot be audited.

\subsubsection{RoyalRoad: market pickup, magazine curation, and follower counts}
\label{app:bal:royalroad}

\paragraph{The stub definition.} $y{=}1$ is a work carrying RoyalRoad's official
\texttt{STUB} status, whose platform tooltip reads: ``Parts of this fiction have
been removed by the author, likely due to third party exclusivity contracts such
as Kindle Unlimited.'' The \texttt{STUB} status therefore indicates that the author pulled chapters because the work
was picked up. $y{=}0$ is a non-stub work drawn from the followers-ordered
listing. Control-pool eligibility drops any card whose label set contains
\texttt{STUB} and any work already in the stub pool.

\paragraph{Matching.} The distance is a fixed-weight additive metric on four
listing variables,
\[
d = \bigl|\Delta\log_{10}(\text{followers}{+}1)\bigr|
 + 0.5\bigl|\Delta\log_{10}(\text{pages}{+}1)\bigr|
 + 0.3\bigl|\Delta\text{rating}\bigr|
 + 0.5\bigl(1 - J(\text{tags})\bigr),
\]
with a \textbf{fixed penalty of $0.5$ substituted for the rating term whenever
either rating is missing}, and Jaccard $=0$ when the tag union is empty. Matching
is \textbf{greedy 1:1}: all pairs are scored, sorted ascending, and walked in
order, accepting a pair iff neither side is already used. \textbf{There is no
caliper}: the realized distance is stored per row for inspection but never
thresholded.

\paragraph{Snapshots.} $X$ is the earliest three chapters
(the lowest chapter identifiers, which are platform-globally monotonic), each
fetched from the \emph{earliest} capture with up to two later captures tried, and
accepted only at $\geq200$ words. Every row carries a Wayback provenance flag, so
the archive-availability filter applies symmetrically. Presentation is normalized
identically on both sides (NFKC; curly to straight quotes; em and en dashes to
\texttt{--} and \texttt{-}; ellipsis; non-breaking and zero-width space removal;
whitespace collapse; wrapper-quote stripping), a step made mandatory after a
cross-source formatting leak on another corpus. RoyalRoad's randomized-class
hidden anti-scrape watermark elements are removed by parsing the inline style
block for \texttt{display:none} and \texttt{speak:never} rules and decomposing
every element carrying a matching class.

\paragraph{First-batch deconfounding, in order.} (i) Assert both classes are
100\% Wayback. (ii) \textbf{Drop all fan fiction} \emph{before} balancing (42 control rows across 14
works were fan fiction and \emph{zero} stubs were, a one-sided content-type
confound). (iii) Popularity propensity-decile
balancing at the \emph{work} level (rating plus $\log 1p$ of views, pages,
chapters and followers; standardized; deciles with $\pm\infty$ outer edges;
keeping $\min(n_\text{pos},n_\text{neg})$ per decile; seed 13), which took
confound-only AUC from \textbf{$.809$ to $.524$} (rating alone had been $.771$).
(iv) Ban \{\texttt{STUB}, \texttt{Original}, \texttt{Fan Fiction}\} from every
feature, and use only normalized text.

\paragraph{The chapter dedup cascade $2{,}404\to2{,}044\to1{,}274$.}
The unit is the work, not the chapter, so we keep one row per work (lowest chapter rank) and split by work, giving 2{,}404 works. The 198-work popularity-matched subset is abandoned here as too small, which leaves the standing caveat that this pool is \emph{not} popularity-matched: followers, rating and views must never be used as features. We then stratify by topic (bge-large embeddings, MiniBatchKMeans with $k{=}40$, the two classes downsampled to equal counts within each cluster), removing 360 works and leaving 2{,}044 at exactly 50/50. After that the content-word floor ($.569$) is below the register-only floor ($.573$), so no topic residue is left, and the earlier subgenre and gendered-pronoun leak words are gone; what survives is narrative register, with dialogue tags and action verbs positive and telling/filtering constructions negative. Last we add coarse-year Wayback-era stratification, which removes a further 770 works and leaves 1{,}274 at 50/50, lexical floor $.588$ and register $.521$.
The era fix was applied even though the audit showed the era confound is
\emph{weak}: era-to-label correlation $-.079$, text-to-era separability $.595$,
and era-matched balancing gave $.558$ against topic-matched $.541$. The
stub-\emph{form} leak was checked and is absent: both sides use chapter rank 1
(stubbing removes \emph{later} chapters, so the first chapter survives), and
mentions of \texttt{kindle}/\texttt{amazon}/\texttt{patreon} appear in
$\sim$20\% of \emph{both} pools ($.27$ versus $.22$). The stub asymmetry shows
only in metadata that is never used (chapters listed 35 versus 19; pages 288
versus 136).

\paragraph{Documented residual biases.} Survivorship (deleted works never enter
either pool); Wayback coverage correlating with popularity, so the matched sample
skews popular; \textbf{stub-date ambiguity} (the platform does not expose
\emph{when} stubbing happened, so follower and page metadata are as-of
enumeration, i.e.\ post-deal for $y{=}1$, making the matching variables partly
post-treatment); label noise in $y{=}0$ (a control may have a deal without stub
status, or be stubbed later); and the earliest-\emph{archived} chapter standing
in for chapter 1 when chapter 1 was never captured.

\paragraph{The verdict.}
The other verdict tasks in this field have a decider: a moderator applying posted rules, a publisher accepting a manuscript. Here no one at RoyalRoad decides. The label is revealed by the author's own response to a third-party exclusivity deal, so it is a market verdict. The task's chance-level result under a certified instrument (Appendix~\ref{app:new-tasks}) is consistent with the decision's
determinants (agent relationships, catalog fit, marketing reach) lying
almost entirely outside the first three chapters.

\paragraph{The magazine curation task.} RoyalRoad's Community Magazine
publishes ranked picks from open submissions. Eight labeled editions give
$2{,}103$ rows ($2{,}012$ labeled entries, $26$ ranked winners), so the magazine task is a
pilot-power task with no trainable full model at $26$ positives. The task's
articulated bank is nevertheless quotable because the anchor battery on this
population is strong ($.787$): the bank reaches $.624$ $[.550,.686]$
(Table~\ref{tab:master-results}).

\paragraph{The follower-count community task.} The listing universe holds
$135{,}598$ works; the scored task is $3{,}604$ rows sampled from
$10{,}611$ randomly drawn deep listing pages, contrasting high- against
low-follower works \emph{within} (genre $\times$ platform-year) strata so
that genre popularity and cohort age cannot carry the label. $X$ is the
work's opening chapters through the same Wayback-only pipeline as the stub
task, with the same watermark stripping. One caveat applies wherever this task is quoted: the articulated leg's anchor battery is
weak on this population and $V{+}A$ coincides with a word-$n$-gram baseline,
so the dense-over-bank margin here measures an instrument limit, not
tacitness.

\subsubsection{Kindle Scout publisher acceptance (expert verdict)}
\label{app:bal:kindlescout}

Kindle Scout was a reader-nominated publishing program in which authors
posted excerpts of unpublished novels and the publisher decided which to
acquire. From Wayback captures of the program, $1{,}024$ campaigns were
parsed; publisher verdicts were recovered for $726$ ($240$ accepted /
$486$ not), which is the instrumented task. $X$ is the campaign's posted
excerpt only. The publisher additionally saw the full manuscript and
nomination statistics that the archive does not preserve, so the task
measures how much of the acquisition decision is visible in the
excerpt text. Splits are campaign-level stable hashes; both classes come
through the identical archive path. At $n{=}726$ Kindle Scout is a pilot-scale
task, and its fused value carries the wide interval quoted in
Table~\ref{tab:master-results} ($.792$ $[.627,.935]$).

\subsubsection{Wigleaf Top 50 flash fiction (expert curation)}
\label{app:bal:wigleaf}

\paragraph{Both classes are expert-selected.} $y{=}1$ is the annual guest-editor
Top 50; $y{=}0$ is the remainder of the same year's Wigleaf longlist. There is no
unselected tier; an earlier three-tier framing was planned but is not
built. The label file holds \textbf{3{,}909 rows $=$ 905 Top50 $+$ 3{,}004
longlist} across 2008--2025, and \textbf{only Top50 rows carry a story URL};
longlist entries are unhyperlinked plain text. Two parse repairs were needed: a
first-author header bug that merged the section heading into the first entry's
author field for 2008--2017 (10 rows), and magazine-casing canonicalization
(652 $\to$ 634 distinct spellings) applied to all rows \emph{so that casing
cannot leak the parse pipeline}.

\paragraph{Longlist recovery without URLs.} Each magazine's domain is resolved
first from the Top50 known-URL map (covering $\sim$74\% of longlist rows), then
from a $\sim$75-entry curated alias dictionary ($\to\sim$84\%), leaving
$\sim$16\% (a long tail of 260$+$ tiny or defunct zines) unrecoverable.
Candidate pages come from a per-domain Wayback index with slug matching:
\textbf{an exact slug substring scores 1.0, otherwise the score is title-token
overlap, and the acceptance threshold is $\geq0.6$}, ties broken by proximity to
the list year. Acceptance additionally requires $\geq300$ characters, a
junk-page rejection test, and a wrong-page guard requiring that at least half the
title tokens appear, \emph{or} the author surname appears.

\paragraph{The leak fix.} In the first build Top50 stories were fetched live-first
while the longlist is necessarily 100\% Wayback, so the provenance field
\textbf{predicted the label at AUC $.90$} from presentation alone, with no content. The
fix re-fetches \emph{every} Top50 story through the Wayback Machine, never live,
so both classes traverse the identical archive path and share its artifacts
(toolbar remnants, re-encoding); stories with no usable snapshot are dropped
under the same survivorship rule the longlist already faced. Result:
\textbf{provenance AUC $=.500$, both classes 100\% archived}. The fix costs coverage: Top50 recovery falls from $\sim$66\% (live) to $\sim$45\%, because stories with a live page but no usable snapshot are dropped.

\paragraph{What the fix changed downstream.} TF-IDF $.799$ (leaky) $\to$
\textbf{$.541$} (near chance); the verifiable-feature score $.72$ (leaky) $\to$
\textbf{$.570$}; length $\approx.565$. The full-set $.541$ matches an independent
within-magazine $.537$. Separately, a cross-magazine TF-IDF of $.721$ is
mostly \emph{venue prestige} (magazine identity alone scores $.793$), so the
number we report for this task is the within-magazine one.

\paragraph{Identical text pipeline, re-run at build time.} Both classes run the
same extraction and cleaning, and it is \textbf{re-executed from the stored raw
text for every row at build time}, so the date a story was fetched cannot matter:
main-text extraction with archive-toolbar removal; CMS boilerplate stripping over
the first 20 lines; trailing-junk stripping (copyright clauses, contributor
tables of contents, archive-navigation widgets); \textbf{author-bio tail
stripping} (bio tails were found in $\sim$37\% of pages, matched by $\sim$40 cue
patterns, with a \emph{safety rail} that reverts the strip if it would leave
under 200 characters from an original over 400); inline-bio stripping for bios
attached to the last sentence; unicode normalization including soft-hyphen and
zero-width removal (\textbf{NFKC does not strip these}, and they are archive
re-encoding artifacts that corrupt words); and a junk-page rejection covering
cookie banners, paywalls, and pages where over 20\% of tokens are dates or
placeholders.

\paragraph{De-confliction of negatives.} Longlist rows whose (title, author
surname) also appears in the recovered Top50 are dropped from the negatives.
One gap remains: there is no winner file
for external flash-fiction prizes, so longlist stories that won Best Small
Fictions, Best Microfiction or Best of the Net remain in $y{=}0$. The negatives
mean ``not promoted by Wigleaf's guest editor'', not ``won nothing anywhere''.

\paragraph{The count cascade $3{,}909\to1{,}568\to1{,}246$.} From 3{,}909 labels,
recovery at $\geq300$ characters keeps $\sim$404 Top50 ($\sim$45\%) and
$\sim$1{,}164 longlist ($\sim$39\%); cross-tier and exact-duplicate-text removal
then gives \textbf{1{,}568 rows (404 positive / 1{,}164 negative, 26\%
positive)}; and the stable-hash split
$\mathrm{md5}(\text{title}|\text{author}|\text{year})\bmod10$ (0--7/8/9) gives
\textbf{train 1{,}246} / eval $\sim$170 / test $\sim$152. The 1{,}246 count is the training split, which is the corpus used for metric work.

\subsection{Humor}
\label{app:bal:humor}

The texts are jokes, captions, contest entries, and sketch scripts. The
\emph{verdicts} are removal decisions by moderators enforcing community
norms (r/Jokes and the five humor-side macro-norm communities) and one
production verdict: a television show's decision to cut a sketch after
dress rehearsal. \emph{Curation} is a staff pick: a show's selection of
top hashtag-game entries, a newspaper contest's winner tiers, a magazine's
caption finalists. The \emph{community} signal is joke upvotes and caption
vote splits. The corpora: r/Jokes crowd upvotes (\S\ref{app:bal:jokes}),
the moderation-removal family shared with creative writing
(\S\ref{app:bal:moderation}), the staff-pick contests
(\S\ref{app:bal:staffpicks}), SNL cut-for-time (\S\ref{app:bal:snl}), and
the New Yorker caption contest (\S\ref{app:bal:caption}), which is
instrumented but excluded from cross-field readouts because the cartoon is
unavailable to every instrument. One curation corpus (the McSweeney's
archive, $17{,}483$ pieces) is collected but awaits its ``Best of''
anthology labels.

\subsubsection{r/Jokes crowd upvotes (community)}
\label{app:bal:jokes}

\paragraph{Build.} From a $\sim$1M-post dump, the pipeline combines title and
body; drops NSFW, removed/deleted and very short posts; stratifies; labels within
cells; and drops the middle.
Within each (length bin $\times$ format $\times$ topic) cell we take the top 25\% by score as $y{=}1$ and the bottom 25\% as $y{=}0$, and discard the middle 50\%. Length is binned in 100-character buckets, format is narrative, one-liner or riddle by rule, and topic comes from a 30-component LDA with a custom stoplist, hard-assigned by argmax. Cells under 20 rows are dropped, as are zero-variance cells and any cell where the bottom quartile's maximum score reaches the top quartile's minimum, which happens with heavy ties. The build also strips edit markers (\texttt{EDIT:}, \texttt{Update:}, \texttt{TL;DR}, ``thanks for the gold'', the sarcasm tag), since those appear only on already-popular posts; filters NSFW posts with the subreddit flag plus a lexical regex; and keeps a length window of $[30, 5000]$ characters. Repeat jokes are collapsed with MinHash LSH over word-level 5-shingles at Jaccard $\geq0.8$ with 128 permutations, plus union-find for transitive clusters, keeping one row per cluster with the longest variant as the canonical text (reposts tend to drop bodies) and the cluster mean as the score. We checked that top comments on this subreddit are follow-on jokes rather than explanations of why a joke is funny, so we build no reasoning-bearing variant. The canonical labelled file holds 383{,}786 rows at exactly 191{,}893 per class, mean length 147 characters.

\paragraph{The 5{,}000-item judge tasks.} Both Reddit community tasks (this one and r/WritingPrompts, \S\ref{app:bal:wp}) are scored on
a fixed 5{,}000-datapoint sample rather than the full corpus. The sampler drops
texts shorter than 200 characters, then takes a \textbf{50/50 stratified sample
of 2{,}500 per label} at a fixed seed, shuffles, and truncates; the export to the
articulated bank writes only \texttt{doc\_id} and \texttt{text}, with a code
comment stating that the label field ``is neither accessed nor copied''. The $\geq$200-character filter is severe for jokes: only 48{,}802 of 383{,}786 deduplicated rows survive it, and that surviving pool happens to be exactly balanced. The humor crowd task is a long-joke subsample, not a random draw. For fiction the
filter has no effect. 
(An earlier caveat is retired because it described a frame that was replaced before the
reported numbers were produced.) The numbers of record for the r/WritingPrompts task are all
computed on one matched population: $7{,}008$ rows in $5{,}136$ prompt groups
drawn from the clean prompt-grouped corpus (score $\geq 10$ binarization,
\texttt{prompt\_id} grouping, seed-42 frozen split), every row held out from
the dense model's training groups; the articulated bank was rescored on this
same population. Artifact:
\texttt{closure/cw\_community/round0\_results.json}.

\subsubsection{Moderation removals: r/Jokes and the nine macro-norm communities (verdict; humor and long-form writing)}
\label{app:bal:moderation}

\paragraph{r/Jokes removals.} The removed-joke universe holds $106{,}937$
moderator-removed posts against $184{,}774$ kept posts from the same
subreddit. The instrumented task is an $18{,}879$-row matched-normalized
contrast: both classes pass one shared renderer (title-plus-body assembly,
edit-marker and moderation-notice stripping, the NFKC/quote/dash
normalization used throughout), and the task must pass the grouped
out-of-fold character-$n$-gram probe before any instrument is applied to it.

\paragraph{The macro-norm tasks.} The two largest verdict tasks build on
the Chandrasekharan et al.\ macro-norms corpus of moderator-removed posts
across nine subreddits: five humor-adjacent (r/funny, r/Showerthoughts,
r/nottheonion, r/me\_irl, r/tifu; $73{,}268$ rows) and four long-form
writing communities (r/nosleep, r/books, r/asoiaf, r/gameofthrones;
$64{,}788$ rows). The removed posts are paired with our own recollection of kept posts from
the same communities, matched 1:1 within subreddit.

\paragraph{The era leak.} The first kept-side
recollection failed its own audit. A leak audit (manual feature inspection,
then a regularized logistic probe over character $n$-grams) found the kept
side \emph{era-compressed}: for eight of the nine subreddits the recollected
kept posts spanned one to six days, and for the ninth a single day, while
the removal corpus spans eleven months, so date-stamped vocabulary
(named news events of the collection window) appeared among the probe's top
features, and part of the full model's separation reflected \emph{when} a post was made,
not \emph{what} it said. The rebuilt kept side samples uniformly over the removal
corpus's full eleven-month window in fortnight strata, strips bot and
moderation notices from both classes, and carries a timestamp and an
author hash on every row so that era-stratified and author-grouped readouts
can be run directly. The probe confirms the repair: the character-$n$-gram
floor falls from $.806$ to $.760$ (humor side) and from $.834$ to $.808$
(long-form side), and the full model on the rebuilt populations scores
$.807$--$.811$ and $.883$--$.884$ respectively, $.02$--$.03$ below its
pre-rebuild values; this difference is the removed era component. All quoted moderation-task
numbers are from the rebuilt populations unless explicitly marked
otherwise.

\paragraph{Articulability and per-community fits.} Because removal
enforces \emph{posted} community norms, these are the benchmark's most
articulable verdicts (the r/Jokes task articulates to $.904$ against a
fused $.923$), and because each task spans several communities they are the
setting for the per-community analyses of Appendix~\ref{app:subfields}:
refit within a single subreddit, the articulated instrument recovers
more of the dense model's signal than the pooled fit, since a
pooled fit absorbs cross-community norm heterogeneity and reports it back
as apparent tacitness.

\subsubsection{\#HashtagWars and Style Invitational staff picks (expert curation)}
\label{app:bal:staffpicks}

These are two contest corpora in which a professional staff selects winners from open
submissions. \emph{\#HashtagWars}: entries to the @midnight television
segment's nightly hashtag game; the canonical corpus holds $12{,}734$
entries, of which $1{,}109$ were selected by the show (top-ten or winner)
and carry the positive label, with a $4{,}228$-row scored task; episodes are the
grouping unit, so every entry to one night's game is assigned to the same split.
\emph{Style Invitational}: entries to the Washington Post's long-running
humor contest across $316$ weekly contests, $9{,}637$ entries with a
top-tier label of $1{,}549$ ($322$ winners $+$ $1{,}227$ runners-up);
weeks are the grouping unit and the contrast is within-week, so the
prompt is the control. Both corpora pass the shared normalization and
probe gates; both are staff picks in the accountability sense of
\S\ref{sec:dataset}, that is, selections made to be published under the
selector's name.

\subsubsection{SNL cut-for-time (production verdict)}
\label{app:bal:snl}

The one production-verdict task: \emph{Saturday Night Live} sketches that
aired versus sketches cut after dress rehearsal. The catalog holds
$2{,}549$ aired and $87$ cut sketches; $143$ recordings were transcribed,
yielding an ASR-matched pilot task of $72$ scripts ($36$ aired / $36$ cut).
The key construction decision is that \emph{both} classes are
transcribed by a single automatic speech-recognition system, so the text
pipeline is identical by construction: a first version of the task that
paired fan transcripts of aired sketches with ASR transcripts of cut ones
carried a transcriber-authorship signature (character-$n$-gram probe
$.73$), which shared ASR removes ($.55$). Under a bank whose anchor battery
separates coherent from scrambled items perfectly, every instrument scores
at chance ($V$ $.55$, $A$ $.44$, $V{+}A$ $.49$). This result is the certified null
discussed in Appendix~\ref{app:new-tasks}: the cut decision depends on
performance, timing, and the dress-rehearsal room, not the script.

\subsubsection{New Yorker caption contest (expert curation and crowd; excluded from cross-field readouts)}
\label{app:bal:caption}

\paragraph{Two sources, joined per contest.} Editor-picked finalists come from
the New Yorker's own API (three per contest, with placement); the rated entry
pool comes from a separate research platform (397 summary files, contests
\#510--895, ratings collected by bandit sampling). A standing project rule
forbids treating crowd humor ratings as ground truth, and this build complies by
taking the \emph{editor} pick as the curation label and using crowd ratings only
as a candidate pool, as a separate crowd label, and as metadata.

\paragraph{Per-contest structure: $3 + 20 + 60 = 83$.} Each contest contributes
3 finalists, \textbf{20 hard negatives} (the top 20 by crowd mean among
non-finalist entries with at least 5 votes, i.e.\ rated highly by the crowd but not picked by the editor),
and up to \textbf{60 random negatives} drawn from the remainder, disjoint by
construction. Captions under 5 characters are dropped, and on normalized-text
collisions the higher-vote record is kept. The realized corpus is
\textbf{18{,}838 rows over 227 contests (ids 615--842) $=$ 678 finalists $+$
4{,}540 hard negatives $+$ 13{,}620 random negatives}; 225 contests carry exactly
83 rows and two carry 82 and 81 (thin entry pools). \emph{In the lab notes, ``81--83'' denotes the rows-per-contest counter, not a range of contest
identifiers.}

\paragraph{The hard-negative pool is the headline pool.} Finalist versus
hard-negative ($n{=}5{,}218$) is the reported curation contrast; finalist versus
the full pool ($n{=}18{,}838$) is reported only as a secondary, explicitly
flagged as length- and topic-inflated. $X$ is caption text only: the cartoon is
constant within a contest, so within-contest ranking is well posed without it,
and where a cartoon description is given to a judge it is likewise constant
within the grouping unit and cannot separate captions.

\paragraph{The typography leak.} Finalists carried typographic quotes from the
publisher's API while entry-pool captions from the research platform had straight
quotes or none. An embedding probe on a \emph{contest-grouped} split scored a spurious
\textbf{$.996$ AUC}, entirely from formatting leakage, falling to $.730$ after a
first round of NFKC, quote and dash normalization, and wrapper-quote stripping on
both sides.

\paragraph{The first normalization fix was incomplete; the audit is quoted verbatim.} The
quotation below is the rebuild script's header:
\begin{quote}\small\itshape
the v1 \texttt{display} normalization was incomplete. A char-level grouped
TF-IDF reached AUC $\sim$0.61--0.667 vs word-level $\sim$0.565 --- a typography
leak. 24 punctuation classes (newline \texttt{:; \$ * / \% \& \# @} literal
double-quote, emoji, zero-width, \ldots) appeared in NEGATIVES but NEVER in
finalists, because finalists are NYT-curated/normalized while the entry pool is
raw crowd submissions. The remaining char-gap after neutralizing those classes
traced entirely to TERMINAL sentence punctuation (closing ``.'' present in 85\%
of finalists vs 69\% of negs; ``!'' 1.6\% vs 12\%) --- also a curation artifact
(editors clean terminal punctuation; crowd over-uses ``!'').
\end{quote}
The corrected normalizer applies an identical aggressive surface transformation
to both sides: NFKC fold plus lowercase; all quote variants to ASCII with wrapper
quotes stripped and any remaining double-quote deleted as a one-sided leak class;
dashes and ellipsis to ASCII; \textbf{keep only \texttt{[a-z0-9]}, space, and the
content-bearing punctuation \texttt{.\,\ ?\ !\ -\ '}, mapping every other
character to a single space}; whitespace collapsed; and \textbf{terminal sentence
punctuation \texttt{[.!?]+} stripped}. Measured effect, grouped five-fold, char
3--5 versus word 1--2: \textbf{full pool word $.562$ / char $.573$, gap $+.011$
(was $+.046$); hard-negative pool word $.546$ / char $.545$, gap $-.001$}. Raw
text is preserved unchanged.

\paragraph{Crowd fields are metadata only.} Both the build code and
the aggregation code state: \emph{``\texttt{crowd\_mean} / \texttt{crowd\_votes} are
used ONLY to build $y$ --- never as features''} (their confound-only AUC is
$\sim.80$, largely through the random-negative tier). The judged text and every
verifiable feature use the normalized text, never the raw text.

\paragraph{Crowd label.} Eligible captions are those with $\geq100$ votes;
\textbf{contests with fewer than 6 eligible captions are dropped}; within each
surviving contest $y = \mathbb{1}[\texttt{crowd\_mean} > \text{median}]$ with
\textbf{exact-median ties dropped}. Of 18{,}750 captions carrying ratings (median
123 votes), 11{,}010 clear the vote threshold and the within-contest median split
yields \textbf{$n{=}10{,}893$ at positive rate $.50014$}. A selection effect is
recorded in the results file rather than corrected away: \emph{``common set $=$
finalists $+$ neg\_hard with $\geq$100 crowd votes; neg\_hard are BY CONSTRUCTION
top-20 crowd per contest, so crowd-C pos rate in common is structurally high ---
report, don't reweight''} (crowd positive rate $.5001$ overall versus $.8964$ on
the common set).

\paragraph{Splits, estimator, anchors.} Contest-grouped stable hash
$\mathrm{md5}(\text{contest})\bmod100\to$ 80/10/10. The estimator is the frozen
cross-domain design (median imputation with a degeneracy guard, standardization
plus logistic regression, out-of-fold AUC, \texttt{GroupKFold(5)} grouped by
contest), with \texttt{StratifiedGroupKFold(5, shuffle, seed 0)} pre-declared as
a side-by-side secondary that is reported but never mixed into the primary readout. This guard was adopted
after cross-validation stratification proved decisive for docket-clustered
regulatory labels. On the caption task the two designs agree, so the caption result is estimator-robust. Three
blinded anchors are included in every shard (a first-placed finalist, a random low-crowd
entry, and a scrambled-nonsense caption) and scored $.667/.378/.121$ on all
five shards, with a not-applicable rate of $.21$.

\subsection{Cross-cutting rules}
\label{app:bal:crosscutting}

Five rules recur across corpora and were adopted as defaults.
We split on a stable hash of a natural unit whenever rows are correlated by one (snapshot, prompt, paper title, repository, docket, court, application, contest, problem, question, company, magazine, fiction), never on a seeded shuffle, which reshuffles silently when the dataset grows, and never on a row-level random split. Confound cleaning and split-leakage cleaning are separate passes and both are needed: topic balancing fixed the topic confound for homepages but did nothing about snapshot-level leakage through the context field. We inspect the highest-discriminating features after balancing; the creative-writing rebuild balanced length and deduplicated but did not catch the moderator-bot template, which a top-trigram lift table would have found immediately. We confirm against an independent deconfounded estimate: a training AUC materially above a prior deconfounded estimate ($.86$ against $.75$ for homepages) is a leakage signature, as is any near-perfect probe AUC such as the $.996$ caption embedding probe. And when the two classes came from different pipelines we normalize presentation identically on both sides, since typography, archive provenance, CMS furniture and author bios all encode the pipeline; the caption ($.996\to.730\to$ char-word gap $-.001$) and Wigleaf (provenance $.90\to.500$) fixes are the two instances of this rule.

%% file: figures/fig_spurious_audit.tex
\begin{tikzpicture}[
  bigbox/.style={rounded corners=7pt, line width=1.0pt},
  chip/.style={rounded corners=4pt, draw, line width=.5pt, inner sep=3.5pt,
    font=\scriptsize, align=center},
  badge/.style={circle, fill=black!75, text=white, font=\footnotesize\bfseries,
    inner sep=1.6pt, minimum size=4.6mm},
  llmtag/.style={rounded corners=3pt, draw=fgreen!70, fill=fgreen!12,
    line width=.6pt, inner sep=3pt, font=\scriptsize\bfseries, text=fgreen!40!black},
  boxtitle/.style={font=\small\bfseries, text=black!80, align=center},
  gsub/.style={font=\small\bfseries, text=fgreen!85!black, align=center},
  body/.style={font=\scriptsize, text=black!70, align=center},
  fine/.style={font=\scriptsize\itshape, text=black!45, align=center},
  flow/.style={-{Stealth[length=2.8mm, width=2.4mm]}, black!70, line width=1.1pt},
  branchlab/.style={font=\small\bfseries, align=center, fill=white,
    inner sep=1.5pt, rounded corners=2pt}]
\definecolor{fgreen}{HTML}{369661}
\definecolor{famber}{HTML}{417497}
\definecolor{fred}{HTML}{46237A}
\definecolor{fpink}{HTML}{D6C9E8}

\draw[bigbox, draw=black!45, fill=black!2] (-9.15,3.3) rectangle (-6.55,6.5);
\node[boxtitle] at (-7.85,6.0) {Candidate\\metric};
\node[chip, draw=black!40, fill=white, text width=1.75cm] at (-7.85,4.95)
  {``reading flow''\\(0--10)};
\node[fine] at (-7.85,3.95) {name $+$ instruction};

\draw[flow] (-6.55,4.9) -- (-6.15,4.9);

\draw[bigbox, draw=fgreen!70, fill=fgreen!3] (-6.05,2.4) rectangle (-1.05,6.9);
\node[badge] at (-5.62,6.42) {1};
\node[boxtitle, anchor=west] at (-5.28,6.42) {Channel tests};
\node[gsub, text width=4.6cm] at (-3.55,5.65) {signals external to the text?};
\node[chip, draw=fgreen!45, fill=fgreen!6, text width=2.0cm] at (-4.65,4.6) {scramble\\the text};
\node[chip, draw=fgreen!45, fill=fgreen!6, text width=2.0cm] at (-2.45,4.6) {top n-gram\\probe};
\node[chip, draw=fgreen!45, fill=fgreen!6, text width=2.0cm] at (-4.65,3.55) {paraphrase\\invariance};
\node[chip, draw=fgreen!45, fill=fgreen!6, text width=2.0cm] at (-2.45,3.55) {capture\\fingerprint};
\node[fine] at (-3.55,2.8) {e.g. time of day, markup residue};

\draw[flow] (-1.05,4.9) -- (-.65,4.9);

\draw[bigbox, draw=fgreen!70, fill=fgreen!3] (-.55,2.4) rectangle (4.45,6.9);
\node[badge] at (-.12,6.42) {2};
\node[boxtitle, anchor=west] at (.22,6.42) {Path tests};
\node[gsub, text width=4.6cm] at (1.95,5.65) {correlated with a spurious variable, not the construct?};
\node[chip, draw=fgreen!45, fill=fgreen!6, text width=2.0cm] at (.85,4.6) {stratify on known\\spurious variables};
\node[chip, draw=fgreen!45, fill=fgreen!6, text width=2.0cm] at (3.05,4.6) {matched\\pairs};
\node[chip, draw=fgreen!45, fill=fgreen!6, text width=2.0cm] at (.85,3.55) {group shuffle\\$\cdot$ transport};
\node[chip, draw=fgreen!45, fill=fgreen!6, text width=2.0cm] at (3.05,3.55) {does the metric\\survive?};
\node[fine, text width=4.6cm] at (1.95,2.8) {e.g. horror prompts draw first-person stories and upvotes};

\draw[flow] (4.45,4.9) -- (4.85,4.9);

\draw[bigbox, draw=fgreen!70, fill=fgreen!3] (4.95,2.4) rectangle (10.85,6.9);
\node[llmtag] at (10.3,6.9) {LLM};
\node[badge] at (5.38,6.42) {3};
\node[boxtitle, anchor=west] at (5.72,6.42) {LLM adjudication};
\node[gsub, text width=5.4cm] at (7.9,5.65) {does it measure the capture, selection or context of the item, rather than the quality of the work?};
\node[chip, draw=fgreen!45, fill=fgreen!6, text width=4.8cm] at (7.9,4.2) {``is \{metric\} spurious in this task?''\\yes / no, citing the evidence rows};
\node[fine, text width=5.2cm] at (7.9,2.9) {planted artifacts (footnote marks, a resample that makes era predictive) are caught at 100\%};

\draw[bigbox, draw=black!50, fill=black!2] (-9.15,-2.45) rectangle (-1.15,1.2);
\node[boxtitle, anchor=west] at (-8.85,.75) {Evidence table};
\node[body, anchor=west] at (-8.85,.28)
  {one row per test: result $+$ the assumption it leans on};
\draw[rounded corners=3pt, draw=black!35, fill=white, line width=.6pt]
  (-8.75,-1.55) rectangle (-1.55,-.15);
\node[font=\tiny\bfseries, text=black!60, anchor=west] at (-8.6,-.38) {test};
\node[font=\tiny\bfseries, text=black!60, anchor=west] at (-6.5,-.38) {result};
\node[font=\tiny\bfseries, text=black!60, anchor=west] at (-4.5,-.38) {assumes};
\draw[black!25, line width=.5pt] (-8.75,-.55) -- (-1.55,-.55);
\node[font=\tiny, text=black!55, anchor=west] at (-8.6,-.75) {scramble the text};
\node[font=\tiny, text=fred!85, anchor=west] at (-6.5,-.75) {AUC .61};
\node[font=\tiny, text=black!45, anchor=west] at (-4.5,-.75) {construct needs word order};
\node[font=\tiny, text=black!55, anchor=west] at (-8.6,-1.05) {stratify on length};
\node[font=\tiny, text=fred!85, anchor=west] at (-6.5,-1.05) {signal gone};
\node[font=\tiny, text=black!45, anchor=west] at (-4.5,-1.05) {strata well measured};
\node[font=\tiny, text=black!55, anchor=west] at (-8.6,-1.35) {top n-gram probe};
\node[font=\tiny, text=fgreen!80!black, anchor=west] at (-6.5,-1.35) {no artifact};
\node[font=\tiny, text=black!45, anchor=west] at (-4.5,-1.35) {n-grams cover the channel};

\draw[flow, black!45, line width=.8pt] (-3.55,2.4) -- (-3.55,1.2);
\draw[flow, black!45, line width=.8pt] (1.95,2.4) to[out=-90,in=60] (-1.6,1.2);
\draw[flow, black!45, line width=.8pt] (-1.15,.3) to[out=0,in=-110] (6.0,2.4);

\draw[bigbox, draw=famber!80, fill=famber!5] (-.55,-.1) rectangle (4.45,1.2);
\begin{scope}[shift={(.1,.55)}, famber!70!black, line width=.8pt, scale=.8]
  \draw[fill=famber!12] (0,0) circle (.42);
  \draw (-.26,.1) -- (0,.3) -- (.26,.1) -- cycle;
  \foreach \x in {-.18,0,.18}{\draw (\x,.08) -- (\x,-.14);}
  \draw (-.26,-.18) -- (.26,-.18);
\end{scope}
\node[boxtitle, anchor=west, text=famber!55!black] at (.65,.8) {Criterion bank};
\node[font=\scriptsize, text=famber!70!black, anchor=west] at (.65,.35)
  {genuine metrics, kept as features};

\draw[bigbox, draw=fred!75, fill=fpink!12] (-.55,-2.45) rectangle (4.45,-.6);
\begin{scope}[shift={(.1,-1.25)}, fred!85, line width=1.0pt, scale=.8]
  \draw[fill=fpink!25] (0,0) circle (.42);
  \draw (-.3,-.3) -- (.3,.3);
\end{scope}
\node[boxtitle, anchor=west, text=fred!80!black] at (.65,-1.0) {Spurious variables};
\node[font=\scriptsize, text=fred!90!black, anchor=west, text width=3.7cm] at (.65,-1.55)
  {never a feature; deconfounded};

\draw[bigbox, draw=fred!55, fill=fpink!8] (4.95,-2.45) rectangle (10.85,-.6);
\node[boxtitle, anchor=west, text=fred!80!black] at (5.25,-.9) {Deconfound, by rule};
\node[chip, draw=fred!50, fill=white, text width=1.35cm] at (6.15,-1.85) {stratify\\($>60\%$ of data)};
\node[chip, draw=fred!50, fill=white, text width=1.35cm] at (7.95,-1.85) {strip\\(regex)};
\node[chip, draw=fred!50, fill=white, text width=1.35cm] at (9.75,-1.85) {residualize\\(otherwise)};
\draw[flow, fred!80] (4.45,-1.5) -- (4.95,-1.5);

\draw[flow, famber!80!black] (6.4,2.4) to[out=-90,in=90] (3.2,1.2);
\node[branchlab, text=famber!55!black] at (4.9,1.75) {genuine};
\draw[flow, fred!80] (9.3,2.4) to[out=-90,in=90] (9.3,-.6);
\node[branchlab, text=fred!80!black] at (9.3,.9) {spurious};
\end{tikzpicture}

%% file: figures/fig_gap_anatomy.tex
\begin{tikzpicture}[baseline=(current bounding box.center),
  lbl/.style={font=\scriptsize\itshape},
  panel/.style={font=\small\scshape},
  every node/.style={inner sep=2pt}]

\begin{scope}[shift={(0,0)}, >={Stealth[length=2.2mm, width=1.8mm]}, font=\small]
\node[circle, draw=black!40, line width=.5pt, fill=black!6, minimum size=11.5mm] (x) {$x$};
\node[circle, draw=vgreen!80, line width=.7pt, fill=vgreen!12, minimum size=9.5mm]
  (m1) at ($(x)+(2.25,0.92)$) {\scriptsize $m_{\mathrm{ver}}(x)$};
\node[circle, draw=aamber!90!black, line width=.7pt, fill=aamber!14, minimum size=9.5mm]
  (m2) at ($(x)+(2.25,-0.92)$) {\scriptsize $m_{\mathrm{art}}(x)$};
\node[circle, draw=black!40, line width=.5pt, fill=black!6, minimum size=11.5mm]
  (y) at ($(x)+(4.5,0)$) {$y$};
\draw[->, vgreen!85, line width=.8pt, shorten >=1pt] (x) to[bend left=12] (m1);
\draw[->, vgreen!85, line width=.8pt, shorten >=1pt] (m1) to[bend left=12] (y);
\draw[->, aamber!95!black, line width=.8pt, shorten >=1pt] (x) to[bend right=12] (m2);
\draw[->, aamber!95!black, line width=.8pt, shorten >=1pt] (m2) to[bend right=12] (y);
\draw[->, torange, line width=1.15pt, shorten >=1pt, opacity=.9]
  (x.south) to[bend right=48]
  node[lbl, below right, pos=.78, yshift=-.6mm, xshift=-1mm, torange!85!black] {dense upper bound} (y.south);
\node[lbl, vgreen!75!black, above=.5mm of m1] {verifiable};
\node[lbl, aamber!72!black, above=.4mm of m2] {articulable};
\node[font=\scriptsize, black!60, above=.8mm of x] {written input};
\node[font=\scriptsize, black!60, above=.8mm of y] {judgement};
\node[panel] at ($(x)+(2.25,2.12)$) {(a)\; two measurable channels};
\end{scope}

\begin{scope}[shift={(7.05,-1.15)}, font=\small,
  sketch/.style={decorate, decoration={random steps, segment length=5pt, amplitude=.45pt}}]
\draw[black!70, line width=.7pt, sketch, ->, >={Stealth[length=1.9mm]}] (0,0) -- (0,3.3);
\draw[black!70, line width=.7pt, sketch] (0,0) -- (3.85,0);
\node[font=\scriptsize, black!65, rotate=90] at (-.32,1.6) {AUC};
\fill[pattern=north east lines, pattern color=vgreen!85]  (.52,0)  rectangle (1.20,1.05);
\draw[vgreen!80!black, line width=.9pt, sketch, rounded corners=1.5pt]
  (.52,0) -- (.52,1.05) -- (1.20,1.05) -- (1.20,0);
\fill[pattern=crosshatch, pattern color=aamber!95]        (1.62,0) rectangle (2.30,1.95);
\draw[aamber!85!black, line width=.9pt, sketch, rounded corners=1.5pt]
  (1.62,0) -- (1.62,1.95) -- (2.30,1.95) -- (2.30,0);
\fill[pattern=north west lines, pattern color=torange!95] (2.72,0) rectangle (3.40,2.9);
\draw[torange!85!black, line width=.9pt, sketch, rounded corners=1.5pt]
  (2.72,0) -- (2.72,2.9) -- (3.40,2.9) -- (3.40,0);
\node[font=\scriptsize, black!70] at (.86,-.24)  {$V$};
\node[font=\scriptsize, black!70] at (1.96,-.24) {$VA$};
\node[font=\scriptsize, black!70] at (3.06,-.24) {dense};
\draw[decorate, decoration={brace, amplitude=2.8pt}, black!70, line width=.6pt]
  (1.54,1.05) -- (1.54,1.95)
  node[midway, left=3.5pt, align=right, font=\scriptsize, black!75,
       fill=white, fill opacity=.82, text opacity=1, inner sep=1.5pt]
  {verifiability\\[-2.5pt]gap};
\draw[decorate, decoration={brace, amplitude=2.8pt}, torange!85!black, line width=.7pt]
  (2.64,1.95) -- (2.64,2.9)
  node[midway, left=3.5pt, align=right, font=\scriptsize\bfseries, torange!80!black,
       fill=white, fill opacity=.82, text opacity=1, inner sep=1.5pt]
  {articulability\\[-2.5pt]gap};
\draw[black!30, densely dotted, line width=.5pt] (1.20,1.05) -- (1.54,1.05);
\draw[black!30, densely dotted, line width=.5pt] (2.30,1.95) -- (2.64,1.95);
\node[panel] at (1.9,3.52) {(b)\; the gap anatomy};
\end{scope}

\end{tikzpicture}

%% file: figures/fig_infilling.tex
\begin{tikzpicture}[
  fieldbox/.style={rectangle, rounded corners=5pt, draw=black!35, line width=.6pt,
    fill=black!3, inner sep=7pt},
  pill/.style={rectangle, rounded corners=7pt, inner sep=4.5pt, font=\scriptsize,
    minimum height=5.5mm},
  shared/.style={pill, fill=ylabc!85, text=white},
  fieldonly/.style={pill, fill=ylabc!35, text=black!70},
  newpill/.style={pill, fill=torange!85, text=white, font=\scriptsize\bfseries},
  inherit/.style={-{Stealth[length=2mm]}, ylabc!60, line width=.9pt},
  lab/.style={font=\small\bfseries, text=black!75}]
\definecolor{ylabc}{HTML}{5A6B8C}

\node[fieldbox, minimum width=9.2cm, minimum height=1.6cm] (F) at (0,0) {};
\node[lab, anchor=west] at ($(F.north west)+(.15,-.32)$) {Field: mathematical proofs};
\node[shared]   (m1) at ($(F.center)+(-3.15,-.22)$) {$m_1$ logical rigor};
\node[shared]   (m2) at ($(F.center)+(-.05,-.22)$)  {$m_2$ clarity of exposition};
\node[fieldonly](m3) at ($(F.center)+(3.05,-.22)$)  {$m_3$ generality};

\node[fieldbox, minimum width=6.2cm, minimum height=1.6cm] (A) at (-3.6,-3.05) {};
\node[lab, anchor=west, font=\small] at ($(A.north west)+(.15,-.32)$)
  {Subfield: linear algebra};
\node[shared, fill=ylabc!70] (a1) at ($(A.center)+(-2.35,-.22)$) {$m_1$};
\node[shared, fill=ylabc!70] (a2) at ($(A.center)+(-1.6,-.22)$) {$m_2$};
\node[newpill] (a4) at ($(A.center)+(.95,-.22)$) {$m_4$ coordinate-free elegance};

\node[fieldbox, minimum width=6.2cm, minimum height=1.6cm] (B) at (3.6,-3.05) {};
\node[lab, anchor=west, font=\small] at ($(B.north west)+(.15,-.32)$)
  {Subfield: combinatorics};
\node[shared, fill=ylabc!70] (b1) at ($(B.center)+(-2.4,-.22)$) {$m_1$};
\node[shared, fill=ylabc!70] (b2) at ($(B.center)+(-1.65,-.22)$) {$m_2$};
\node[newpill] (b5) at ($(B.center)+(.75,-.22)$) {$m_5$ bijective insight};

\draw[inherit] (m1.south) to[bend right=22] (a1.north);
\draw[inherit] (m2.south) to[bend right=24] (a2.north);
\draw[inherit] (m1.south) to[bend left=8]  (b1.north);
\draw[inherit] (m2.south) to[bend left=22] (b2.north);

\node[font=\scriptsize\itshape, text=torange!80!black, align=center]
  at (-3.6,-4.25) {discovered by metric infilling:\\held-out lift within subfield only};
\node[font=\scriptsize\itshape, text=torange!80!black, align=center]
  at (3.6,-4.25) {discovered by metric infilling:\\held-out lift within subfield only};
\node[font=\scriptsize\itshape, text=ylabc!80!black, align=center,
      fill=white, inner sep=1.5pt] at (0,-1.78) {shared core:\\transfers across subfields};
\end{tikzpicture}